%% file: main.tex
\documentclass[10pt,twocolumn,letterpaper]{article}

\usepackage[pagenumbers]{cvpr} 

\input{preamble}

\definecolor{cvprblue}{rgb}{0.21,0.49,0.74}
\usepackage[pagebackref,breaklinks,colorlinks,citecolor=cvprblue]{hyperref}

\usepackage{multirow}
\usepackage{graphicx}
\usepackage{booktabs}
\usepackage{bm}
\usepackage{bbm}
\usepackage{makecell}
\usepackage{pbox}
\usepackage{float}
\usepackage[export]{adjustbox}

\usepackage{color}

\def\confName{CVPR}
\def\confYear{2024}

\title{Improving Image Restoration through \\ Removing Degradations in Textual Representations}

\author{Jingbo Lin$^{1}$, Zhilu Zhang$^{1}$, Yuxiang Wei$^{1}$, Dongwei Ren$^{1}$, Dongsheng Jiang$^{2}$, Wangmeng Zuo$^{1,}$\thanks{Correspondence author.} \\
\textsuperscript{1}Harbin Institute of Technology \ \textsuperscript{2}Huawei Cloud Computing Co., Ltd. \\
{\tt \small jblincs1996@gmail.com, cszlzhang@outlook.com, yuxiang.wei.cs@gmail.com, }
\\ 
{\tt \small rendongweihit@gmail.com, dongsheng\_jiang@outlook.com, cswmzuo@gmail.com}
}

\begin{document}

\maketitle
\input{main_sec/0_abstract}    
\input{main_sec/1_intro}

\input{main_sec/2_related_works}
\input{main_sec/3_methods}
\input{main_sec/4_experiments}
\input{main_sec/5_conclusion}

{
    \small
    \bibliographystyle{ieeenat_fullname}
    \bibliography{main}
}


\input{main_sec/suppl}

\end{document}

%% file: preamble.tex
\usepackage[dvipsnames]{xcolor}


%% file: main_sec/0_abstract.tex
\begin{abstract}
In this paper, we introduce a new perspective for improving image restoration by removing degradation in the textual representations of a given degraded image.
Intuitively, restoration is much easier on text modality than image one. 
For example, it can be easily conducted by removing degradation-related words while keeping the content-aware words.
Hence, we combine the advantages of images in detail description and ones of text in degradation removal to perform restoration.
To address the cross-modal assistance, we propose to map the degraded images into textual representations for removing the degradations, and then convert the restored textual representations into a guidance image for assisting image restoration.
In particular, We ingeniously embed an image-to-text mapper and text restoration module into CLIP-equipped text-to-image models to generate the guidance.
Then, we adopt a simple coarse-to-fine approach to dynamically inject multi-scale information from guidance to image restoration networks.
Extensive experiments are conducted on various image restoration tasks, including deblurring, dehazing, deraining, and denoising, and all-in-one image restoration. 
The results showcase that our method outperforms state-of-the-art ones across all these tasks.
The codes and models are available at \url{https://github.com/mrluin/TextualDegRemoval}.
\end{abstract}

%% file: main_sec/1_intro.tex
\section{Introduction}
\label{sec:introduction}
%
Image restoration aims to reconstruct a high-quality clean image from its degraded observations.
Most existing methods design deep networks for specific restoration tasks, including image denoising~\cite{DnCNN, FFDNet, DRUNet, SwinIR, Restormer}, deblurring~\cite{MIMO-Unet+, MPRNet, HINet, Uformer, NAFNet},  deraining~\cite{MPRNet, Uformer, IDT, DRSformer},  dehazing~\cite{DehazeNet, Dehamer, Maxim, Dehazeformer, SFNet}, \etc.
Recent works~\cite{AirNet, ADMS, IDR, PromptIR, WGWS, teacher_student} expect to explore a unified model for multiple degradations.
However, the severely ill-posed nature of the task makes it non-trivial to separate degradations and desired image content.
Especially for unified models, the potential conflicts in dealing with various degradations bring more uncertainty.

From a broader perspective, the purpose of the restoration is to enhance the clarity of scenes for human perception and recognition, while performing it on image modality is just one option.
Moreover, degradations in image modality are tightly coupled with desirable content, making their removal challenging.
When recording the scenes in some other modalities, this problem can be alleviated.
Let us take the text modality as an example. 
Assume the textual description of a clean scene is represented by `a scene of $\ast$'. 
When this scene is subjected to rainfall, the description can be converted into `a rainy scene of $\ast$'. 
Thus, deraining can be readily achieved by simply removing the rain-related text `rainy'. Furthermore, it is also convenient to remove multiple degradations in textual space with one unified model.
Note that text modality may only describe rough aspects and ignore some details.
We can further introduce image modality to combine their complementary potential in degradation removal and image restoration.

\input{figures/001_introduction_figure.tex}

Motivated by the above, we propose to perform restoration first in a modality where degradations and content are loosely coupled, and then utilize the corrected content to guide image restoration.
In particular, we adopt the commonly used textual modality.
Note that the text corresponding to the scene is not always available and the cross-modal assistance is difficult to achieve directly.
The gap between text and image should be bridged.
First, degraded images should be mapped into textual space for removing degradations. 
Second, the restored text should be mapped into a guidance image to assist restoration of the degraded image.
As shown in \cref{fig:introduction_figure}, for the former, we have tried to convert degraded images to textual captions by image-to-text (I2T) models (\eg, BLIP~\cite{BLIP, BLIPv2}), and find the captions can explicitly express both degradation types (\eg, blur, rain, haze, and noise) and content information.
For the latter, we can utilize the text-to-image (T2I) models (\eg, Imagen~\cite{saharia2022photorealistic}), which have demonstrated impressive image generation capability from the given texts.

Although such a scheme is conceptually feasible, converting to explicit text may lose a lot of information from images, and additional degraded-clean text pairs need to be collected to train the text restoration module.
Fortunately, Contrastive Language-Image Pre-training (CLIP)~\cite{CLIP} implicitly aligns concepts of images and text.
And we can leverage the CLIP-equipped T2I models (\eg, Stable Diffusion~\cite{stablediffusion}) to build an end-to-end framework for generating clear guidance from degraded images gracefully.
Specifically, CLIP has been demonstrated as an effective I2T converter~\cite{ding2023clip}.
For converting degraded images into degraded textual representations, we adopt the image encoder of CLIP, and add an I2T mapper after the encoder. 
For restoring degraded representations, we further append a textual restoration module after I2T mapper.
Then, I2T mapper and textual restoration module can be sequentially trained by employing paired data from different image restoration datasets, without additional text data.
Due to the ease of removing degradations on textual level, we train the I2T mapper and textual restoration module with multiple degradations at once. 
When training is done, it can serve multiple tasks, generating content-related and degradation-free guidance images from degraded images.
Finally, we adopt a simple coarse-to-fine approach to dynamically inject multi-scale information from guidance to classic image restoration networks.
Extensive experiments are conducted on multiple tasks, including all-in-one restoration, image deburring, dehazing, deraining, and denoising.
The results show our method achieves better universally than corresponding state-of-the-art methods.
Especially for all-in-one restoration, 0.5 dB PSNR improvement is obtained in comparison with PromptIR~\cite{PromptIR}.

The main contributions can be summarized as follows:
\begin{itemize}
    \item We introduce a new perspective for image restoration, \ie, performing restoration first in textual space where degradations and content are loosely coupled, and then utilizing the restored content to guide image restoration.
    \item To address the cross-modal assistance, we propose to embed an image-to-text mapper and textual restoration module into CLIP-equipped text-to-image models to generate clear guidance from degraded images.
    \item Extensive experiments on multiple tasks demonstrate that our method improves the performance of state-of-the-art image restoration networks.
\end{itemize}

%% file: figures/001_introduction_figure.tex
\begin{figure*}[t]
    \centering
    \setlength{\abovecaptionskip}{0.1cm}
    \setlength{\belowcaptionskip}{-0.3cm}
    \includegraphics[width=0.97\linewidth]{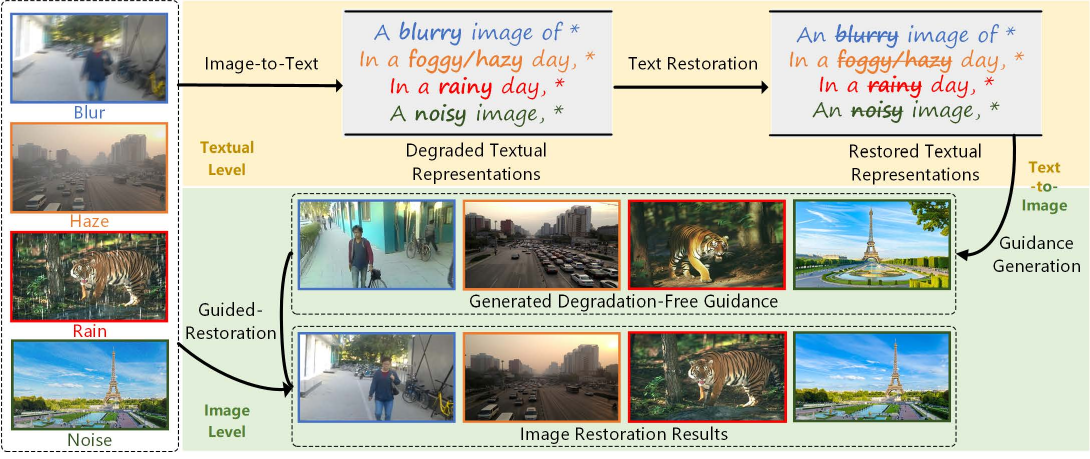}
    \caption{\textbf{Overview of the proposed method}. We propose to improve image restoration by performing restoration on the textual level, in which content and degradation information are loosely coupled. We first encode image concepts into textual space and then remove degradation-related information. To achieve cross-modal assistance, we employ pre-trained T2I models to generate clean guidance for the image restoration process.}
    \label{fig:introduction_figure}
    \vspace{-1mm}
\end{figure*}

%% file: main_sec/2_related_works.tex
\input{figures/002_method_figure.tex}

\section{Related Work}
\label{sec:related_work}
\subsection{Image Restoration}
\textbf{Image Restoration for Specific Tasks}.
Image restoration is a fundamental computer vision problem, which aims to reconstruct high-quality images from corresponding degraded inputs.
In recent years, deep learning has achieved great progress in image restoration.
Starting from some simple convolutional neural networks (CNNs)~\cite{SRCNN, DnCNN}, the introduction of channel-attention~\cite{RCAN}, spatial-attention~\cite{self-guided, ZamirAKHKYS20}, non-local operation~\cite{RNAN, NLRN}, skip-connection architectures~\cite{recurrent, ZamirAKHKYS20} and multi-stage scheme~\cite{Uformer, MPRNet} enables image restoration performance continuously improve.
With the emergence of vision transformers~\cite{liu2021swin,dosovitskiy2020image}, the capability of capturing long-range dependencies in the image allows transformer-based methods to achieve better performance, gradually replacing previous CNN-based methods. To balance computational cost, window-based attention~\cite{SwinIR, HAT} and transposed attention~\cite{Restormer} are also introduced into image restoration tasks.
Albeit image restoration performance has benefited from various advanced architecture designs, most of the works only focus on one specific degradation, due to the difficulty of learning to remove multiple degradations.

\noindent\textbf{All-in-One Image Restoration}.
Promoting by the development of unified model, some works focus on the solution to the all-in-one image restoration~\cite{AirNet, ADMS, IDR, PromptIR, WGWS, teacher_student}.
The first all-in-one image restoration work, \ie, IPT~\cite{IPT}, employs ViT-based backbone with multi-heads and multi-tails for inputs with different degradations. However, the method is limited in handling specific synthesized degradations and cannot directly generalize to unknown tasks.
AirNet~\cite{AirNet} develops the unified model for denoising-deraining-dehazing, which utilizes contrastive learning to capture degradation representations of different tasks and adaptively injects the degradation priors into backbone restoration framework for aiding in learning better results.
ADMS~\cite{ADMS} exploits FAIG~\cite{FAIG} in all-in-one image restoration, by learning specific filters and degradation classifiers, achieving better performance on deraining-denoising-deblurring and deraining-desnowing-dehazing tasks.
IDR~\cite{IDR} proposes a two-stage training strategy, which first learns separate task-oriented hubs for each degradation. Then, in the second stage, it reformulates the learned hubs into a single ingredient-oriented hub by learnable PCA and adopts the reformulated hub as prior to adaptively aid in restoring corrupted images.
PromptIR~\cite{PromptIR} learns to encode and adopt degradation information as prompts for dehazing-deraining-denoising task.
Although existing works can unify a set of image restoration tasks into one unified model, the performance is still limited caused by the large gap among different degradations.

\noindent\textbf{Prior-Based Image Restoration}.
Prior-based restoration aims at improving performance by introducing external priors, including structures, images, pre-trained models, \etc.
For instance, some works~\cite{c2matching, MASA-SR, DATSR} adopt information from the high-resolution reference image to help improve the performance of super-resolution.
Recent works~\cite{StableSR, DiffBIR, PASD} utilize pre-trained generative priors to restore more realistic and natural results.
TextIR~\cite{textir} develops a text-driven image restoration framework by incorporating information on textual representation.
In this work, we provide a new perspective, \ie, utilizing restoration in textual space to assist image restoration.
We also combine the advantages of both text-based prior (by learning degradations in textual level) and image-based prior ( by giving clean guidance) for better performance.

\subsection{Text-to-Image Generation}
Text-to-image generation models~\cite{ramesh2022hierarchical,ding2022cogview2,gafni2022make,nichol2021glide,ramesh2021zero,xia2021tedigan,sauer2023stylegan} have attracted intensive attention in recent years due to its ability to generate high-quality and diverse images based on given text descriptions.
A variety of techniques, including generative adversarial networks (GAN)~\cite{goodfellow2014generative}, autoregressive models, and diffusion models have been investigated.
Initial studies~\cite{li2019controllable,xia2021tedigan,sauer2023stylegan} mainly rely on GAN-based architectures, and train a conditional model from given paired image-caption datasets to generate samples.
Some efforts focus on autoregressive models~\cite{ding2021cogview,gafni2022make,ramesh2021zero,chang2023muse,yu2022scaling} have also shown exciting results. 
These models, such as CogView~\cite{ding2021cogview} and Muse~\cite{chang2023muse} first learn a discrete codebook through training an autoencoder, and then adopt an autoregressive transformer to predict the tokens sequentially.
With the development of diffusion models~\cite{ho2020denoising,song2020denoising}, text conditioned image synthesis has shown remarkable improvement.
By training with huge corpora, large diffusion models, such as DALLE-2~\cite{ramesh2022hierarchical}, Imagen~\cite{saharia2022photorealistic}, Stable Diffusion~\cite{rombach2022high}, and DALLE-3~\cite{BetkerImprovingIG} have demonstrated excellent semantic understanding, and can generate diverse and photo-realistic images.

%% file: figures/002_method_figure.tex
\begin{figure*}[t]
	\centering
	\setlength{\abovecaptionskip}{0.1cm}
	\setlength{\belowcaptionskip}{-0.3cm}
	\includegraphics[width=0.95\textwidth]{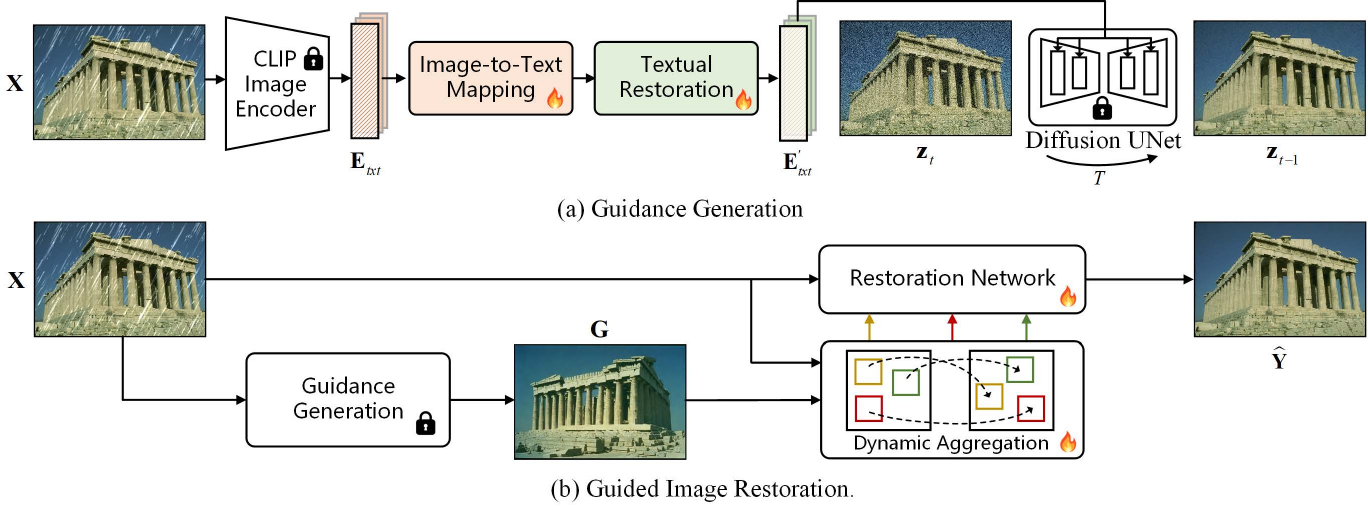}
   \vspace{-2mm}
	\caption{\textbf{Illustration of the proposed pipeline.}  \textbf{(a)} We sequentially train image-to-text mapper  $\mathcal{M}_{i2t}$ and textual restoration module $\mathcal{M}_{clean}$ to convert image concepts into textual representations and remove textual degradation information, respectively. 
    \textbf{(b)} The guidance image is used to assist the image restoration process.}
	\label{fig:main_framework}
   \vspace{-1mm}
\end{figure*}

%% file: main_sec/3_methods.tex
\section{Proposed Method}
\label{sec:method}
In this section, we present our proposed method for image restoration.
As illustrated in Fig.~\ref{fig:main_framework}, we suggest first conducting restoration in the textual modality, in which degradation is loosely coupled with the content and can be easily removed.
Then we in turn utilize the restored results as guidance to improve image restoration.
Specifically, in Sec.~\ref{sec:degradation-free_guidance_generation}, we propose an image-to-text mapper $\mathcal{M}_{i2t}$ and a textual restoration module $\mathcal{M}_{clean}$ to extract the degradation-free textual representations $\mathbf{E}_{txt}'$ from degraded images $\mathbf{X}\in\mathbbm{R}^{H\times W\times 3}$.
Then with a pre-trained T2I model, we can generate a guidance image $\mathbf{G}\in\mathbbm{R}^{H\times W\times 3}$ that is related to the content of $\mathbf{X}$ but free of degradation.
In Sec.~\ref{sec:guided_restoration}, to leverage clean content information from $\mathbf{G}$ for enhancing image restoration performance, we introduce a coarse-to-fine approach that dynamically incorporates multi-scale information from guidance $\mathbf{G}$ into restoration frameworks.
Besides, we first give a preliminary knowledge of the T2I model in Sec.~\ref{sec:preliminary}.

\subsection{Preliminary}
\label{sec:preliminary}

In this work, we employ Stable Diffusion~\cite{stablediffusion} as our text-to-image model. 
Stable Diffusion pretrains an autoencoder $(\mathcal{E}(\cdot), \mathcal{D}(\cdot) )$ to map the input image $\mathbf{X}$ into a lower dimensional latent space by $\mathbf{z}=\mathcal{E}(\mathbf{X})$.
The decoder $\mathcal{D}(\cdot)$ learns to map the latent code back to the image as $\mathcal{D}(\mathcal{E}(\mathbf{X})) \approx \mathbf{X}$. 
Then, the conditional diffusion model $\epsilon_\theta(\cdot)$ is trained on the latent space to generate latent codes based on text condition $\mathbf{p}$. 
To train the diffusion model, simple mean-squared loss is adopted as,
\begin{equation}
\begin{small}
L_{LDM}  \!
=  \!
 \mathbb{E}_{\mathbf{z}\sim\mathcal{E}(\mathbf{X}),\mathbf{p},\epsilon\sim\mathcal{N}(0,1),t} \Big[\Vert\epsilon \! -  \! \epsilon_\theta(\mathbf{z}_t, t, \tau_\theta^t(\mathbf{p})) \Vert_2^2\Big],
\label{equ:ldm_loss}
\end{small}
\end{equation}
where $\epsilon$ denotes the unscaled noise, $t$ is the timestep, $\mathbf{z}_t$ is the latent noised to time $t$, and $\tau_\theta^t(\cdot)$ represents the pre-trained CLIP~\cite{CLIP} text encoder. During inference, a random Gaussian noise $\mathbf{z}_T$ is iteratively denoised to $\mathbf{z}_0$, and the final image is obtained through the decoder $\mathbf{X}^\prime=\mathcal{D}(\mathbf{z}_0)$.

\subsection{Degradation-Free Guidance Generation}
\label{sec:degradation-free_guidance_generation}
Image captioning models (\eg, BLIP2~\cite{BLIPv2}) can generate text descriptions of degraded images, which can be taken into the text restoration module and text-to-image models to reconstruct clean guidance images.
However, such descriptions may lose too many details, resulting in severely inconsistent content between the guidance and the input images.
To address these problems, we propose an image-to-text mapper $\mathcal{M}_{i2t}$ to encode the degraded image $\mathbf{X}$ as implicit textual representations $\mathbf{E}_{txt}$ rather than explicit texts, which can be used to reconstruct the content more faithfully.
Then, we utilize a textual restoration module $\mathcal{M}_{clean}$ to remove the degradation from $\mathbf{E}_{txt}$ and obtain degradation-free textual representations $\mathbf{E}'_{txt}$. 
\noindent\textbf{Image-to-Text Mapping}.
Following~\cite{ELITE}, we adopt the textual word embedding space of CLIP~\cite{CLIP} as the target space in image-to-text mapping. 
Benefiting from the aligned vision-language feature space of CLIP, we first adopt pre-trained CLIP image encoder $\tau_\theta^i(\cdot)$ to encode input $\mathbf{X}$ into CLIP image embedding $\mathbf{E}_{img}$. 
Then, we propose an image-to-text mapper $\mathcal{M}_{i2t}$ to project CLIP image embedding into textual word embedding $\mathbf{E}_{txt}$,
\begin{equation}
    \mathbf{E}_{txt} = \mathcal{M}_{i2t} (\tau_\theta^i(\mathbf{X})) ,
\label{eqn:global_mapping}
\end{equation}
where $\mathbf{E}_{txt} \in \mathbbm{R}^{N\times D}$ and $D$ is the dimension of textual word embedding.
$N$ is the number of learned words.
To make the obtained word embedding describe more details of the input image, we set $N$ to a large number (\eg, $N$$=$$20$).
%

\noindent\textbf{Textual Restoration}.
Although $\mathcal{M}_{i2t}$ can project images into the textual word embedding space, the projected textual word embedding will include corresponding degradation information if input $\mathbf{X}$ is degraded. 
Condition on the degraded textual word embedding, the synthesized guidance inevitably reflects the corresponding degradation pattern.
Therefore, we further propose a textual restoration module $\mathcal{M}_{clean}$ to remove the degradation information from the textual word embedding $\mathbf{E}_{txt}$,
\begin{equation}
    \mathbf{E}_{txt}' = \mathcal{M}_{clean}(\mathbf{E}_{txt}),
\label{equ:clean_mapping}
\end{equation}
where $\mathbf{E}_{txt}'$ denote the restored textual representations. 
When feeding $\mathbf{E}_{txt}'$ into Stable Diffusion, we can obtain a degradation-free image $\mathbf{G}$, which has similar content as $\mathbf{X}$ but is free of degradation, as shown in Fig.~\ref{fig:introduction_figure}.

\noindent\textbf{Model Optimization}.
During training, the image-to-text mapper $\mathcal{M}_{i2t}$ and textual restoration module $\mathcal{M}_{clean}$ are optimized sequentially. 
In the first training stage,
we collect both clean and degraded images from different restoration tasks as training data.
We adopt ~\cref{equ:ldm_loss} as a loss function, which constrains the diffusion model to reconstruct clean and degraded images conditioning on their own projected embedding $\mathbf{E}_{txt}$. 
In the second training stage, 
we use pairs of degraded-clean images from different image restoration datasets as training data.
Based on pre-trained $\mathcal{M}_{i2t}$, we further deploy $\mathcal{M}_{clean}$ to remove degradation-related concepts from degraded embedding $\mathbf{E}_{txt}$, obtaining $\mathbf{E}'_{txt}$.
We still adopt ~\cref{equ:ldm_loss}, which constrains the diffusion model to reconstruct clean images conditioning on restored embedding $\mathbf{E}'_{txt}$. 
Note that, due to the ease of removing degradations in the textual space, it is feasible to train $\mathcal{M}_{i2t}$ and $\mathcal{M}_{clean}$ with multiple degradations simultaneously. 
When training is done, it can serve multiple image restoration tasks, generating content-related and degradation-free guidance images $\mathbf{G}$ from degraded images with one unified model (as shown in Fig.~\ref{fig:introduction_figure}).
More training details can be seen in the \textit{Suppl}.

\input{tables/001_all_in_one_restoration_table.tex}

\subsection{Guided Restoration}
\label{sec:guided_restoration}
Although guidance image $\mathbf{G}$ is clean, it is inevitable that there are content differences between $\mathbf{G}$ and degraded image $\mathbf{X}$. 
Simply fusing the features of $\mathbf{G}$ and $\mathbf{X}$ is not enough to improve image restoration.
Instead, we suggest a dynamic aggregation module to extract and exploit helpful information from $\mathbf{G}$.
\noindent \textbf{Dynamic Aggregation}.
First, we perform feature matching~\cite{MASA-SR} between the guidance $\mathbf{G}$ and given degraded image $\mathbf{X}$.
Specifically, we use a shared image encoder to extract multi-scale features for $\mathbf{X}$ and $\mathbf{G}$, named  $\mathbf{F}_x$ and $\mathbf{F}_g$, respectively. 
According to the similarity score between $\mathbf{F}_x$ and $\mathbf{F}_g$, we search for the most useful feature from $\mathbf{F}_g$ for each small patch in $\mathbf{F}_x$ in a coarse-to-fine manner.
After searching, we get a set of guidance features $\hat{\mathbf{F}}_g$, whose content is aligned spatially with the input content.
Second, we integrate the matched feature $\hat{\mathbf{F}}_g$ into the image restoration backbone.
Both CNN-based and transformer-based restoration networks can be adopted.
Without bells and whistles, we modulate the original input features as,
\begin{equation}
\mathbf{F}_x = \mathbf{F}_x + \alpha \cdot \mathcal{B}([\mathbf{F}_x, \hat{\mathbf{F}}_g]),
\label{eqn:fusion}
\end{equation}
where $[\cdot, \cdot]$ represents concatenation operation, $\mathcal{B}$ represents one CNN-based  block or transformer-based block, $\alpha$ is the hyper-parameter to trade-off feature integration.
Moreover, we perform it over multiple layers or levels.
More details can be seen in the \textit{Suppl}.

\input{figures/003_all_in_one_visual_comparison.tex}

\input{tables/002_deblurring_tables.tex}
\input{figures/005_motion_deblurring_visual_comparison.tex}
\input{figures/004_defocus_deblurring_visual_comparison.tex}

%
\noindent\textbf{Loss Function}.
To train guided restoration framework, we simply adopt $\ell_1$ loss as the reconstruction loss between predicted results $\hat{\mathbf{Y}}$ and target $\mathbf{Y}$, \ie,
\begin{equation}
\ell_1 = ||\hat{\mathbf{Y}} - \mathbf{Y} ||_1.
\end{equation}

%% file: tables/001_all_in_one_restoration_table.tex
%
\begin{table}[t]
\begin{center}
\caption{\small \underline{\textbf{All-in-one image restoration results}}. Following PromptIR~\cite{PromptIR}, we train and evaluate the proposed method in all-in-one image restoration task, our method outperforms PromptIR across all the benchmark datasets.}
\label{table:all_in_one_restoration_results}
\vspace{-3mm}
\setlength{\tabcolsep}{1.9pt}
\scalebox{0.63}{
\begin{tabular}{l | c | c | c c c | c }
\toprule[0.15em]
   & \textbf{Dehazing} & \textbf{Derain} & \multicolumn{3}{c|}{\textbf{Denoise} on BSD68} & \textbf{Average} \\

\textbf{Method} & on SOTS & on Rain100L & $\sigma=15$ & $\sigma=25$ & $\sigma=50$ &  \\
\midrule[0.1em]
BRDNet~\cite{BRDNet} & 23.23/0.895 & 27.42/0.895 & 32.26/0.898 & 29.74/0.836 & 26.34/0.836 & 27.80/0.843  \\
LPNet~\cite{LPNet}   & 20.84/0.828 & 24.88/0.784 & 26.47/0.778 & 24.77/0.748 & 21.26/0.552 & 23.64/0.738 \\
FDGAN~\cite{FDGAN}   & 24.71/0.924 & 29.89/0.933 & 30.25/0.910 & 28.81/0.868 & 26.43/0.776 & 28.02/0.883 \\
MPRNet~\cite{MPRNet} & 25.28/0.954 & 33.57/0.954 & 33.54/0.927 & 30.89/0.880 & 27.56/0.779 & 30.17/0.899 \\
DL~\cite{DL} & 26.92/0.391 & 32.62/0.931 & 33.05/0.914 & 30.41/0.861 & 26.90/0.740 & 29.98/0.875 \\
AirNet~\cite{AirNet} & 27.94/0.962 & 34.90/0.967 & 33.92/0.933 & 31.26/0.888 & 28.00/0.797  & 31.20/0.910 \\
\midrule[0.1em]
PromptIR~\cite{PromptIR} & \underline{30.58}/\underline{0.974} & \underline{36.37}/\underline{0.972} & \underline{33.98}/\underline{0.933} & \underline{31.31}/\underline{0.888} & \underline{28.06}/\underline{0.799} & \underline{32.06}/\underline{0.913} \\
\textbf{Ours} & \textbf{31.63}/\textbf{0.980} & \textbf{37.58}/\textbf{0.979} & \textbf{34.01}/\textbf{0.933} & \textbf{31.39}/\textbf{0.890} & \textbf{28.18}/\textbf{0.802} & \textbf{32.56}/\textbf{0.916} \\
\bottomrule[0.15em]
\end{tabular}
}
\end{center}
\vspace{-4mm}
\end{table}
%

%% file: figures/003_all_in_one_visual_comparison.tex
\begin{figure}[t]
\begin{center}
\setlength{\tabcolsep}{1.5pt}
\scalebox{0.88}{
\begin{tabular}[b]{c c c c}
\vspace{0.2em}
\includegraphics[width=.13\textwidth,valign=t]{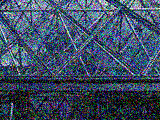} &   
\includegraphics[width=.13\textwidth,valign=t]{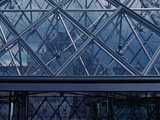} &   
\includegraphics[width=.13\textwidth,valign=t]{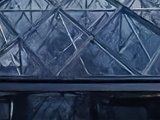} &   
\includegraphics[width=.13\textwidth,valign=t]{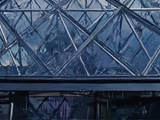}   \\
\vspace{0.2em}
\includegraphics[width=.13\textwidth,valign=t]{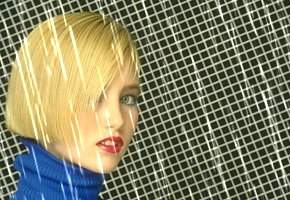}    &   
\includegraphics[width=.13\textwidth,valign=t]{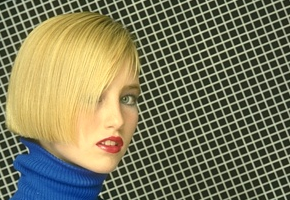}   &   
\includegraphics[width=.13\textwidth,valign=t]{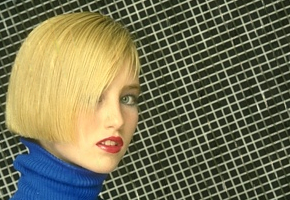} &   
\includegraphics[width=.13\textwidth,valign=t]{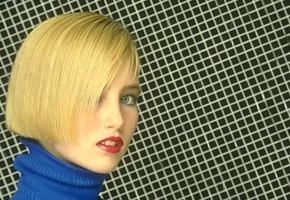}     \\
\includegraphics[width=.13\textwidth,valign=t]{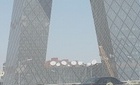} &   
\includegraphics[width=.13\textwidth,valign=t]{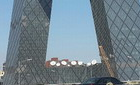} &   
\includegraphics[width=.13\textwidth,valign=t]{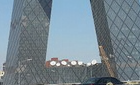} &   
\includegraphics[width=.13\textwidth,valign=t]{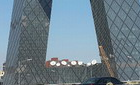}   \\
\small~Degraded  & \small~Label & \small~PromptIR~\cite{PromptIR} & \small~\textbf{Ours}
\\
\end{tabular}
}
\vspace*{-6mm}
\caption{\small \underline{\textbf{All-in-one image restoration results}}. Top: image denoising, mid: image deraining, bottom: image dehazing.}
\label{fig:all_in_one_visual_comparison}
\end{center}
\vspace{-7mm}
\end{figure}

%% file: tables/002_deblurring_tables.tex
\begin{table*}[!t]
\parbox{.5\linewidth}{
\centering
\caption{\small \underline{\textbf{Motion image deblurring results}}. We train models with GoPro training data. We evaluate our method on GoPro, HIDE, RealBlur benchmark datasets. PSNR and SSIM scores are calculated on RGB-channels.} 
\label{table:motion_deblurring_results}
\vspace{-2mm}
\setlength{\tabcolsep}{1.9pt}
\scalebox{0.71}{
\begin{tabular}{l  | c  c | c  c | c  c | c  c }
\toprule[0.15em]
\multirow{2}{*}{\textbf{Method}} & \multicolumn{2}{c|}{\textbf{GoPro}~\cite{GoPro}} & \multicolumn{2}{c|}{\textbf{HIDE}~\cite{HIDE}} & \multicolumn{2}{c|}{\textbf{RealBlur-R}~\cite{RealBlur}} & \multicolumn{2}{c}{\textbf{RealBlur-J}~\cite{RealBlur}} \\ 
                        & PSNR$\textcolor{black}{\uparrow}$ & SSIM$\textcolor{black}{\uparrow}$ & PSNR$\textcolor{black}{\uparrow}$ & SSIM$\textcolor{black}{\uparrow}$ & PSNR$\textcolor{black}{\uparrow}$ & SSIM$\textcolor{black}{\uparrow}$  & PSNR$\textcolor{black}{\uparrow}$ & SSIM$\textcolor{black}{\uparrow}$ \\
\midrule[0.1em]           
DBGAN~\cite{DBGAN}           & 31.10 & 0.942 & 28.94 & 0.915 & 33.78 & 0.909 & 24.93 & 0.745 \\
MT-RNN~\cite{MT-RNN}          & 31.15 & 0.945 & 29.15 & 0.918 & 35.79 & 0.951 & 28.44 & 0.862 \\
DMPHN~\cite{DMPHN}           & 31.20 & 0.940 & 29.09 & 0.924 & 35.70 & 0.948 & 28.42 & 0.860 \\
SPAIR~\cite{SPAIR}           & 32.06 & 0.953 & 30.29 & 0.931 & -     &  -    & 28.81 & 0.875 \\
MIMO-Unet+~\cite{MIMO-Unet+}      & 32.45 & 0.957 & 29.99 & 0.930 & 35.54 & 0.947 & 27.63 & 0.837 \\ 
IPT~\cite{IPT}             & 32.52 & -     &  -    & -    & -      & -     & -     & -     \\
MPRNet~\cite{MPRNet}          & 32.66 & 0.959 & 30.96 & 0.939 & 35.99 & 0.952 & 28.70 & 0.873 \\
HINet~\cite{HINet}           & 32.71 & 0.959 & 30.32 & 0.932 & -     & -     & -     & -     \\
Uformer~\cite{Uformer}         & 32.97 & 0.967 & -     & -     & -     & -     & -     & -     \\
\midrule[0.1em]
Restormer~\cite{Restormer}       & 32.92 & 0.961 & 31.22 & 0.942 & \underline{36.19} & \underline{0.957} & \underline{28.96} & \textbf{0.879} \\
\textbf{Ours-Restormer}                    & 33.11 & 0.962 & 31.26 & 0.943 & \textbf{36.47} & \textbf{0.959} & \textbf{29.17} & \underline{0.875}\\
\midrule[0.1em]
NAFNet~\cite{NAFNet}          & \underline{33.69} & \underline{0.966} & \underline{31.32} & \underline{0.943} & 33.62 & 0.944 & 26.33 & 0.856 \\
\textbf{Ours-NAFNet}                    & \textbf{33.97} & \textbf{0.968} & \textbf{31.57} & \textbf{0.946} & 33.87 & 0.950 & 26.76 & 0.861 \\
\bottomrule[0.15em]
\end{tabular}
}
}
\hfill
\parbox{.47\linewidth}{
\centering
\caption{\underline{\textbf{Defocus image deblurring results}}. We train and evaluate methods on DPDD dataset~\cite{DPDNet}. $S$ denotes single-image defocus deblurring model. $D$ denotes dual-pixel defocus deblurring. PSNR and SSIM scores are calculated on RGB channels.}
\label{table:defocus_deblurring_results}
\vspace{-2mm}
\setlength{\tabcolsep}{3pt}
\scalebox{0.66}{
\begin{tabular}{l | c c| c c | c c}
\toprule[0.15em]
   & \multicolumn{2}{c|}{\textbf{Indoor Scenes}} & \multicolumn{2}{c|}{\textbf{Outdoor Scenes}} & \multicolumn{2}{c}{\textbf{Combined}} \\
   \textbf{Method} & PSNR~$\textcolor{black}{\uparrow}$ & SSIM~$\textcolor{black}{\uparrow}$
                   & PSNR~$\textcolor{black}{\uparrow}$ & SSIM~$\textcolor{black}{\uparrow}$
                   & PSNR~$\textcolor{black}{\uparrow}$ & SSIM~$\textcolor{black}{\uparrow}$ \\
\midrule[0.15em]
EBDB$_S$~\cite{EBDB}      & 25.77 & 0.772 & 21.25 & 0.599 & 23.45 & 0.683 \\
DMENet$_S$~\cite{DMENet}  & 25.50 & 0.788 & 21.43 & 0.644 & 23.41 & 0.714 \\
JNB$_S$~\cite{JNB}        & 26.73 & 0.828 & 21.10 & 0.608 & 23.84 & 0.715 \\
DPDNet$_S$~\cite{DPDNet}  & 26.54 & 0.816 & 22.25 & 0.682 & 24.34 & 0.747 \\

KPAC$_S$~\cite{KPAC}      & 27.97 & 0.852 & 22.62 & 0.701 & 25.22 & 0.774 \\
IFAN$_S$~\cite{IFAN}      & 28.11 & 0.861 & 22.76 & 0.720 & 25.37 & 0.789 \\
\midrule[0.1em]
Restormer$_S$~\cite{Restormer} & \underline{28.87} & \underline{0.882} & \underline{23.24} & \underline{0.743} & \underline{25.98} & \underline{0.811}  \\
\textbf{Ours}$_S$ & \textbf{29.11} & \textbf{0.889}  & \textbf{23.35} & \textbf{0.748} & \textbf{26.15} & \textbf{0.817}  \\ 
\midrule[0.1em]
\midrule[0.1em]
DPDNet$_D$~\cite{DPDNet}   & 27.48 & 0.849 & 22.90 & 0.726 & 25.13 & 0.786  \\
RDPD$_D$~\cite{RDPD}       & 28.10 & 0.843 & 22.82 & 0.704 & 25.39 & 0.772  \\
Uformer$_D$~\cite{Uformer} & 28.23 & 0.860 & 23.10 & 0.728 & 25.65 & 0.795  \\
IFAN$_D$~\cite{IFAN}       & 28.66 & 0.868 & 23.46 & 0.743 & 25.99 & 0.804  \\
\midrule[0.1em]
Restormer$_D$~\cite{Restormer} & \underline{29.48} & \underline{0.895} & \underline{23.97} & \underline{0.773}  & \underline{26.66} & \underline{0.833} \\
\textbf{Ours}$_D$ & \textbf{29.62} & \textbf{0.899} & \textbf{24.16} & \textbf{0.775} & \textbf{26.82} & \textbf{0.835}\\ 
\bottomrule[0.15em]
\end{tabular}
}
}
\vspace{0mm}
\end{table*}

%% file: figures/005_motion_deblurring_visual_comparison.tex
\begin{figure*}[!t]
\begin{center}
\scalebox{0.97}{
\begin{tabular}[b]{c@{ } c@{ }  c@{ } c@{ } c@{ } }
\hspace{-4mm}
    \multirow{3}{*}{\includegraphics[width=.31\textwidth,valign=t]{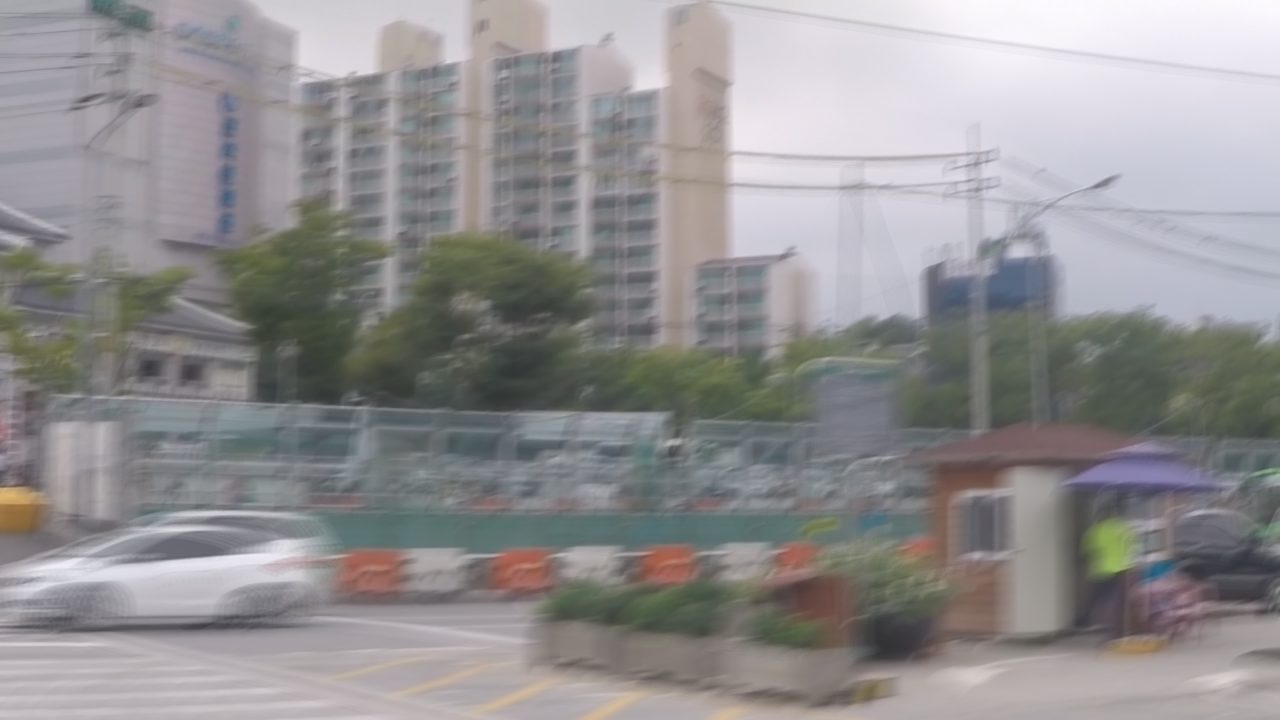}} &   
    \includegraphics[width=.173\textwidth,valign=t]{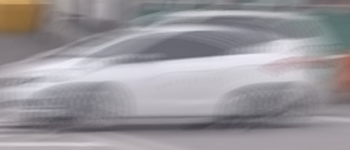}         &
    \includegraphics[width=.173\textwidth,valign=t]{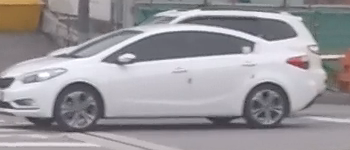}        &   
    \includegraphics[width=.173\textwidth,valign=t]{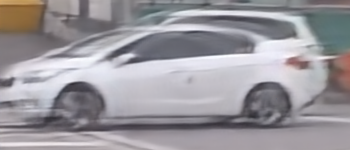}         &
    \includegraphics[width=.173\textwidth,valign=t]{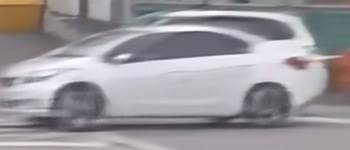} 
 \\
    & \small~Degraded &\small~Label & \small~DMPHN~\cite{DMPHN} & \small~MT-RNN~\cite{MT-RNN}          \\

  & \includegraphics[width=.173\textwidth,valign=t]{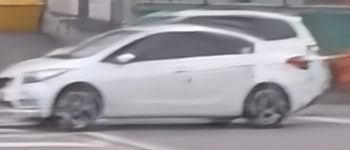}   &  
    \includegraphics[width=.173\textwidth,valign=t]{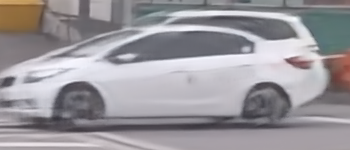}        &
    \includegraphics[width=.173\textwidth,valign=t]{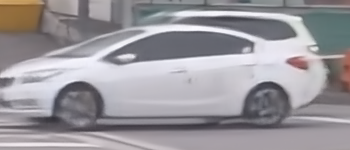}       &
    \includegraphics[width=.173\textwidth,valign=t]{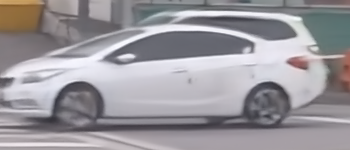}         \\

     Input image & \small~MIMOUnet~\cite{MIMO-Unet+} & \small~HINet~\cite{HINet} & \small~NAFNet~\cite{NAFNet}  & \small~\textbf{Ours} 
\\
\end{tabular}
}
\end{center}
\vspace{-6mm}
\caption{ \small  \underline{\textbf{Motion image deblurring results}}. Compared with others, our method can predict clearer results with clearer boundaries.
}
\label{fig:motion_deblurring_visual_comparison}
\vspace{0mm}
\end{figure*}

%% file: figures/004_defocus_deblurring_visual_comparison.tex
\begin{figure*}[!t]
\begin{center}
\scalebox{0.97}{
\begin{tabular}[b]{c@{ } c@{ }  c@{ } c@{ } c@{ } }
\hspace{-4mm}
    \multirow{3}{*}{\includegraphics[width=.285  \textwidth,valign=t]{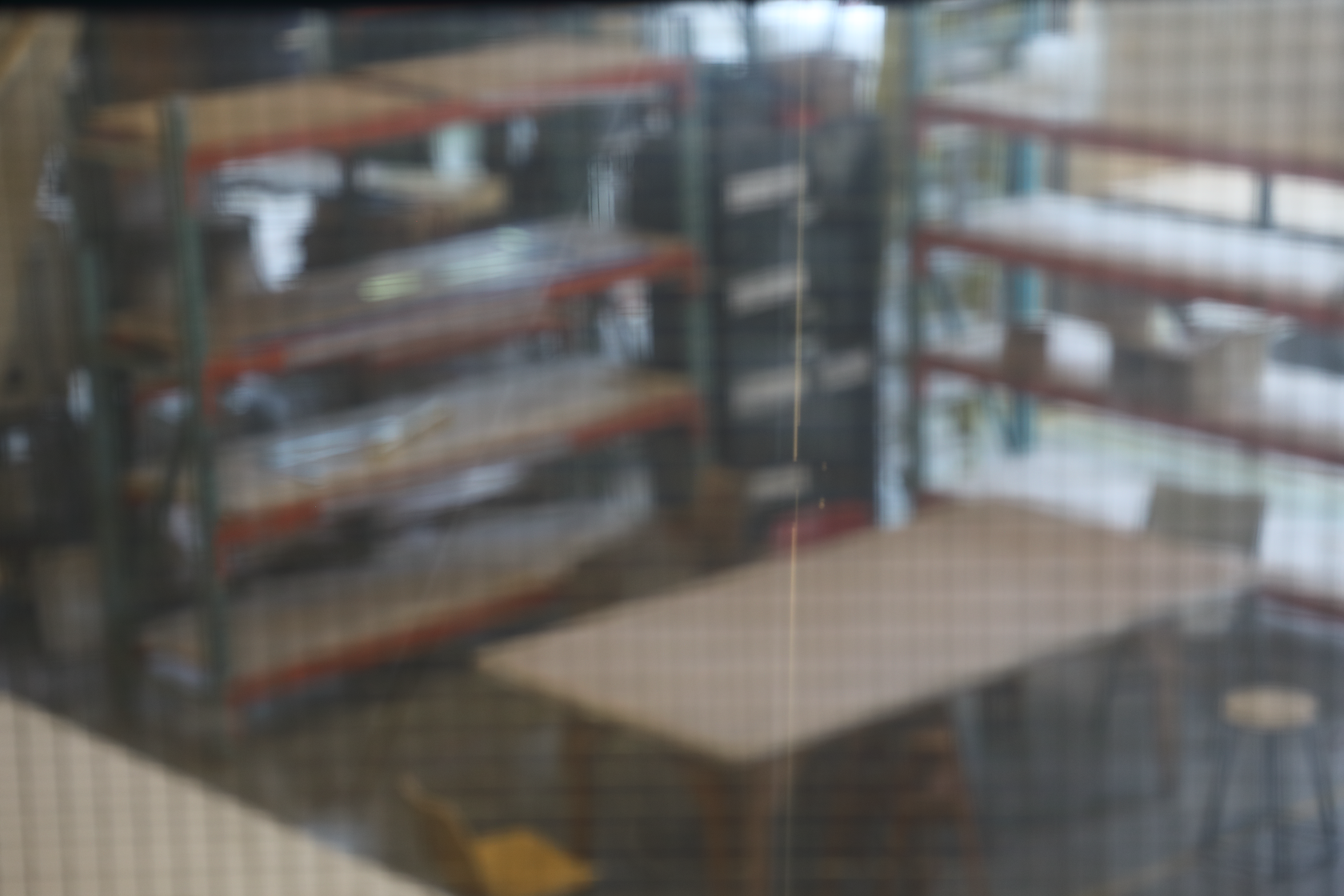}} &   
    \includegraphics[width=.17\textwidth,valign=t]{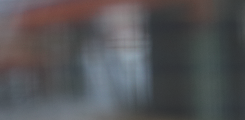}         &
    \includegraphics[width=.17\textwidth,valign=t]{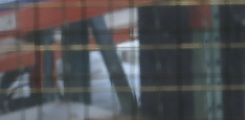}        &   
    \includegraphics[width=.17\textwidth,valign=t]{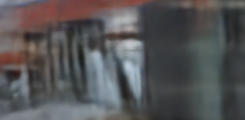}         &
    \includegraphics[width=.17\textwidth,valign=t]{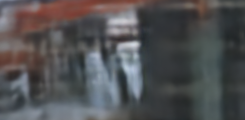} 
 \\
    & \small~Degraded &\small~Label & \small~DMPHN~\cite{DMPHN} & \small~MPRNet~\cite{MPRNet}          \\

  & \includegraphics[width=.17\textwidth,valign=t]{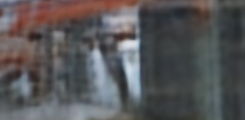}   &  
    \includegraphics[width=.17\textwidth,valign=t]{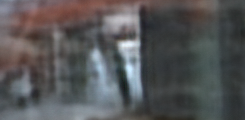}        &
    \includegraphics[width=.17\textwidth,valign=t]{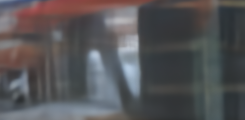}       &
    \includegraphics[width=.17\textwidth,valign=t]{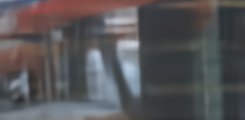}         \\

     Input image & \small~DPDNet~\cite{DPDNet} & \small~RDPD~\cite{RDPD} & \small~Restormer~\cite{Restormer}  & \small~\textbf{Ours} 
\\
\end{tabular}
}
\end{center}
\vspace{-6mm}
\caption{ \small  \underline{\textbf{Defocus image deblurring results}}. Compared with others, our method can predict clearer results with clearer boundaries.
}
\label{fig:defocus_deblurring_visual_comparison}
\vspace{-0.5em}
\end{figure*}

%% file: main_sec/4_experiments.tex
\section{Experiments}
\label{sec:experiments}
We evaluate the proposed method on five image restoration tasks, including
\textbf{(1)} all-in-one image restoration~\cite{PromptIR},
\textbf{(2)} image deblurring~\cite{NAFNet, Restormer} (\ie, motion image deblurring and defocus image deblurring),
\textbf{(3)} image dehazing~\cite{SFNet},   
\textbf{(4)} image deraining~\cite{DRSformer},
and \textbf{(5)} image denoising~\cite{Restormer} (\ie, grayscale and color image denoising on Gaussian noise, and real-world image denoising).
\noindent \textbf{Implementation Details}.
We adopt publicly released stable-diffusion-v2.1-base as our T2I generative model.
$N$ is set to $20$ in image-to-text mapper.
%
%
Moreover, we employ the state-of-the-art methods as our image restoration backbone in different image restoration tasks: 
\textbf{(1)} we adopt PromptIR~\cite{PromptIR} in all-in-one image restoration, 
\textbf{(2)} we adopt NAFNet~\cite{NAFNet} and Restormer~\cite{Restormer} in single-image motion deblurring, and Restormer~\cite{Restormer} in defocus deblurring,
\textbf{(3)} we adopt SFNet~\cite{SFNet} in image dehazing, 
\textbf{(4)} we adopt DRSformer~\cite{DRSformer} in image deraining, 
\textbf{(5)} we adopt Restormer~\cite{Restormer} in image denoising.
The training settings of our guided restoration network are consistent with those of the corresponding backbone methods. 
%
Details of datasets, network architecture, and training hyper-parameters are provided in \textit{Suppl}.

\subsection{All-in-One Restoration Results}
\label{sec:all_in_one_restoration}
Following~\cite{PromptIR}, we adopt BSD400~\cite{BSD} and WED~\cite{WED} for color image denoising, Rain100L~\cite{DL} for image deraining, and SOTS~\cite{reside} for image dehazing. 
We train and evaluate the proposed method on these three tasks in all-in-one image restoration setting.
The large difference among different degradations makes unifying multiple image restoration tasks into one unified network difficult, and thus the performance of existing all-in-one image restoration methods is limited.
Compared with performing all-in-one restoration in the image level, it is much easier for us to conduct all-in-one restoration in the textual level.
Once trained, our method can generate clean textual representation for different image restoration tasks, and thus we can synthesize guidance images for various types of degradation.
Comparison results in Table~\ref{table:all_in_one_restoration_results} show that our method outperforms PromptIR~\cite{PromptIR} across all the benchmark datasets, achieving +1.05 dB PSNR on dehazing task, +1.21 dB PSNR on deraining task, and +0.5 dB PSNR on average improvement.
As shown in Fig.~\ref{fig:all_in_one_visual_comparison}, the prediction of the proposed method is finer and more detailed compared with PromptIR.

\subsection{Image Deblurring Results}
\label{sec:image_deblurring_results}
\noindent \textbf{Motion Deblurring}.
%
Following~\cite{Restormer, NAFNet}, we train the proposed method on GoPro training data and evaluate our method on GoPro~\cite{GoPro}, HIDE~\cite{HIDE}, and real-world datasets (RealBlur-R~\cite{RealBlur} and RealBlur-J~\cite{RealBlur}).
As shown in Table~\ref{table:motion_deblurring_results}, benefiting from synthesized high-quality guidance, Ours-Restormer and Ours-NAFNet outperform the state-of-the-art methods on all four benchmark datasets. 
Our method achieves +0.28 dB PSNR improvement on GoPro testing data.
The visual results in Fig.~\ref{fig:motion_deblurring_visual_comparison} show that our method can restore images with sharper boundaries and details, demonstrating its effectiveness.

\input{tables/005_deraining_dehazing_table.tex}
\input{tables/003_denoising_gaussian_table.tex}
\input{figures/006_gaussian_color_denoising_visual_comparison.tex}
\input{figures/007_gaussian_grayscale_denoising_visual_comparison.tex}

\noindent \textbf{Defocus Deblurring}.
Following~\cite{Restormer}, we train and evaluate the proposed method on DPDD~\cite{DPDNet} dataset for single-image defocus deblurring and dual-pixel defocus deblurring. 
From Table~\ref{table:defocus_deblurring_results}, compared with Restormer~\cite{Restormer}, our method achieves +0.17 dB PSNR and +0.16 dB PSNR gains on single-image defocus deblurring and dual-pixel defocus deblurring, respectively.
As illustrated in Fig.~\ref{fig:defocus_deblurring_visual_comparison}, the prediction of our method shows a clearer structure.


\subsection{Image Dehazing Results}
\label{sec:image_dehazing_results}
For image dehazing task, we conduct experiments on the synthetic benchmark RESIDE~\cite{reside} dataset. 
We train the proposed method on indoor scene and outdoor scene data, and evaluate the performance on SOTS-indoor and SOTS-outdoor testing dataset~\cite{reside} separately. 
Table~\ref{table:dehazing_results} reports the quantitative comparison results.
One can see that our method outperforms existing methods on both indoor and outdoor scenes, achieving +0.24 dB PSNR average improvement.
Visual comparisons can be found in the \textit{Suppl}.
%


\input{tables/004_denoising_realworld_table.tex}
\input{tables/006_ablation_studies_arxiv.tex}

\subsection{Image Deraining Results}
\label{sec:image_deraining_results}
Following~\cite{DRSformer}, we train and evaluate the proposed method separately on Rain200L~\cite{Rain200H}, Rain200H~\cite{Rain200H}, DID-Data~\cite{DID-Data}, and DDN-Data~\cite{DDN-Data} datasets.
From Table~\ref{table:deraining_results}, our method achieves +0.31 dB PSNR average improvement on Rain200L, DID-Data, and DDN-Data against DRSformer~\cite{DRSformer}. 
On the Rain200H dataset, we only have comparable performance, as its severe rain streak may interfere with the dynamic aggregation process between the guidance information and the degraded images.
Visual comparisons of deraining results will be provided in the \textit{Suppl}.
%

\subsection{Image Denoising Results}
\label{sec:image_denoising_results}
\noindent \textbf{Image Denoising on Gaussian Noise}. 
Following~\cite{Restormer}, we train the blind denoising model ($\sigma$ with range of $[0, 50]$) and non-blind denoising models ($\sigma$$=$$15,25,50$) for grayscale and color image denoising on Gaussian noise.
We train the proposed method on DFBW dataset with synthetic noise degradation.
For grayscale image denoising, we evaluate our method on Set12~\cite{DnCNN}, BSD68~\cite{CBSD68}, and Urban100~\cite{Urban100}.
For color image denoising, we evaluate our method on 
CBSD68~\cite{CBSD68}, Kodak24, McMaster~\cite{McMaster}, and Urban100~\cite{Urban100}.
As shown in Table~\ref{table:gaussian_grayscale_denoising_results} and Table~\ref{table:gaussian_color_denoising_results}, our method achieves comparable performance with baseline on low noise-level ($\sigma$$=$$15$), while our method outperforms baseline when noise becomes heavier, \eg, achieving +0.25 dB PSNR gain on Urban100~\cite{Urban100} with Gaussian color noise $\sigma$$=$$50$).
Qualitative comparisons in Fig.~\ref{fig:gaussian_color_denoising_visual_comparison} and Fig.~\ref{fig:gaussian_grayscale_denoising_visual_comparison} show that our method can restore finer texture and sharper boundaries while other methods tend to generate smoother results.

\noindent\textbf{Real-World Image Denoising}.
Furthermore, We train and evaluate the proposed method on real-world denoising dataset SIDD~\cite{sidd}.
From Table~\ref{table:realworld_denoising_results}, although Restormer~\cite{Restormer} has already achieved superior performance on real-world denoising task, our method can also bring +0.07 dB PSNR improvement, demonstrating its effectiveness.
Visual comparison results can be found in \textit{Suppl}.
%

\subsection{Ablation Studies}
\label{sec:ablation_studies}

In ablation studies, 
we mainly conduct discussion from three perspectives:
\textbf{(1)} the effect of the textural representation size,
\textbf{(2)} the effect of guidance integration strategy
and \textbf{(3)} the effect of generated guidance.
During experiments, we adopt NAFNet with 32 channels as our baseline. We train each method on GoPro datasets for 30k iterations and evaluate the performance on GoPro testing data. 
%
%
%

%
\noindent\textbf{Effect of Textural Representation Size}.
In our method, we use the number of words $N$ to control the representation ability of $\mathbf{E}_{txt}$, which is used as condition information to generate guidance. 
%
%
We conduct the experiments on five settings, \ie, $N=5,10,20,30,40$. 
As shown in Table~\ref{table:ablation_condition_information}, the restoration performance generally improves from $N$$=$$5$ to $N$$=$$40$, as more informative textual descriptions can help T2I model generate more faithful guidance.
%
Since the performance only improves slightly when $N$ is larger than $20$, we finally set $N$ to $20$ to make a trade-off between performance and computational cost.
\noindent \textbf{Effect of Integrating Strategy}.
To evaluate the effect of different guidance integrating strategies, we conduct experiments by integrating guidance information into different modules of image restoration networks, \ie, Enc. (encoder only), Dec. (decoder only), and Enc. \& Dec. (both encoder and decoder).
As shown in Table~\ref{table:ablation_injection}, the experimental results show that integrating guidance information into both encoder and decoder achieves better performance.

\noindent\textbf{Effect of Generated Guidance}.
To assess the effect of generated guidance, we conduct an experiment by replacing it with degraded input (named `Degra.').
From Table~\ref{table:ablation_ref}, when replacing the generated guidance with degraded input, it cannot obtain better results compared with baseline, which shows dynamically aggregating the degraded image itself cannot help the image restoration. In contrast, the performance can improve effectively with our guidance.

%% file: tables/005_deraining_dehazing_table.tex
\begin{table*}[t]
\parbox{.45\linewidth}{
\centering
\caption{\underline{\textbf{Image dehazing results}}. We separately train and evaluate our method indoor scene and outdoor scene. PSNR and SSIM scores are calculated on RGB-channels.} 
\label{table:dehazing_results}
\vspace{-1mm}
\setlength{\tabcolsep}{1.5pt}
\scalebox{0.74}{
\begin{tabular}{l | c  c | c  c }
\toprule[0.15em]
\multirow{2}{*}{\textbf{Method}} & \multicolumn{2}{c|}{\textbf{SOTS-Indoor}~\cite{reside}} & \multicolumn{2}{c}{\textbf{SOTS-Outdoor}~\cite{reside}}   \\ 
                         & PSNR$\textcolor{black}{\uparrow}$ & SSIM$\textcolor{black}{\uparrow}$  & PSNR$\textcolor{black}{\uparrow}$ & SSIM$\textcolor{black}{\uparrow}$  \\
\midrule[0.1em]
DehazeNet~\cite{DehazeNet}            & 19.82 & 0.821 & 24.75 & 0.927 \\
AOD-Net~\cite{AOD-Net}                & 20.51 & 0.861 & 24.14 & 0.920 \\
GridDehazeNet~\cite{GridDehazeNet}    & 32.16 & 0.984 & 30.86 & 0.982 \\
MSBDN~\cite{MSBDN}                    & 33.67 & 0.985 & 33.48 & 0.982 \\
FFA-Net~\cite{FFA-Net}                & 36.39 & 0.989 & 33.57 & 0.984 \\
ACER-Net~\cite{ACER-Net}              & 37.17 & 0.990 & -     & -     \\
DeHamer~\cite{Dehamer}                & 36.63 & 0.988 & 35.18 & 0.986 \\
MAXIM-2S~\cite{Maxim}                 & 38.11 & 0.991 & 34.19 & 0.985 \\
PMNet~\cite{PMNet}                    & 38.41 & 0.990 & 34.74 & 0.985 \\
DehazeFormer-L~\cite{Dehazeformer}    & 40.05 & 0.996 & -     & -     \\
\midrule[0.1em]
SFNet~\cite{SFNet}                    & \underline{41.24} & \underline{0.996} & \underline{40.05} & \underline{0.996} \\
\textbf{Ours}                            & \textbf{41.48} & \textbf{0.996} & \textbf{40.29} & \textbf{0.996} \\
\bottomrule[0.15em]
\end{tabular}}
}
\hfill
\parbox{.53\linewidth}{
\centering
\caption{\small \underline{\textbf{Image deraining results}}. We separately train and evaluate our method on Rain200H, Rain200L, DID-Data, and DDN-Data. PSNR and SSIM scores are calculated on Y channel in YCbCr color space.}
\label{table:deraining_results}
\vspace{-3mm}
\setlength{\tabcolsep}{1.9pt}
\scalebox{0.75}{
\begin{tabular}{l | c  c | c  c | c  c| c  c}
\toprule[0.15em]
\multirow{2}{*}{\textbf{Method}} & \multicolumn{2}{c|}{\textbf{Rain200L}~\cite{Rain200H}} & \multicolumn{2}{c|}{\textbf{Rain200H}~\cite{Rain200H}} & \multicolumn{2}{c|}{\textbf{DID-Data}~\cite{DID-Data}} & \multicolumn{2}{c}{\textbf{DDN-Data}~\cite{DDN-Data}} \\ 
                        & PSNR$\textcolor{black}{\uparrow}$ & SSIM$\textcolor{black}{\uparrow}$  & PSNR$\textcolor{black}{\uparrow}$ & SSIM$\textcolor{black}{\uparrow}$ & PSNR$\textcolor{black}{\uparrow}$ & SSIM$\textcolor{black}{\uparrow}$  & PSNR$\textcolor{black}{\uparrow}$ & SSIM$\textcolor{black}{\uparrow}$  \\
\midrule[0.1em]
DDN~\cite{DDN}                  & 34.68 & 0.967 & 26.05 & 0.805 & 30.97 & 0.911 & 30.00 & 0.904 \\
RESCAN~\cite{RESCAN}            & 36.09 & 0.967 & 26.75 & 0.835 & 33.38 & 0.941 & 31.94 & 0.935 \\
PReNet~\cite{PReNet}            & 37.80 & 0.981 & 29.04 & 0.899 & 33.17 & 0.948 & 32.60 & 0.946 \\
MSPFN~\cite{MSPFN}              & 38.53 & 0.983 & 29.36 & 0.903 & 33.72 & 0.955 & 32.99 & 0.933 \\
RCDNet~\cite{RCDNet}            & 39.17 & 0.989 & 30.24 & 0.904 & 34.08 & 0.953 & 33.04 & 0.947 \\
MPRNet~\cite{MPRNet}            & 39.47 & 0.982 & 30.67 & 0.911 & 33.99 & 0.959 & 33.10 & 0.935 \\
DualGCN~\cite{DualGCN}          & 40.73 & 0.989 & 31.15 & 0.912 & 34.37 & 0.962 & 33.01 & 0.949 \\
SPDNet~\cite{SPDNet}            & 40.50 & 0.988 & 31.28 & 0.920 & 34.57 & 0.956 & 33.15 & 0.946 \\
Uformer~\cite{Uformer}          & 40.20 & 0.986 & 30.80 & 0.910 & 35.02 & 0.962 & 33.95 & 0.955 \\
Restormer~\cite{Restormer}      & 40.99 & 0.989 & 32.00 & 0.932 & 35.29 & \underline{0.964} & 34.20 & \underline{0.957} \\
IDT~\cite{IDT}                  & 40.74 & 0.988 & \underline{32.10} & \textbf{0.934} & 34.89 & 0.962 & 33.84 & 0.955 \\
\midrule[0.1em]
DRSformer~\cite{DRSformer}       & \underline{41.21} & \underline{0.989} & \textbf{32.16} & \underline{0.933} & \underline{35.24} & 0.962 & \underline{34.23} & 0.955 \\
\textbf{Ours}                   & \textbf{41.59}    & \textbf{0.990}    & 31.97 & 0.931 & \textbf{35.46} & \textbf{0.964} & \textbf{34.57} & \textbf{0.958} \\
\bottomrule[0.15em]
\end{tabular}
}}
\vspace*{-1mm}
\end{table*}

%% file: tables/003_denoising_gaussian_table.tex
\begin{table*}[!t]
\parbox{.435\linewidth}{
\centering
\caption{\small \underline{\textbf{Grayscale image denoising on Gaussian noise}}. Upper-bracket: models are trained on a range of noise levels. Lower-bracket: models are trained on the fixed noise level.} 
\label{table:gaussian_grayscale_denoising_results}
\vspace{-2mm}
\setlength{\tabcolsep}{1.5pt}
\scalebox{0.7}{
\begin{tabular}{l | c c c | c c c | c c c}
\toprule[0.15em]
   & \multicolumn{3}{c|}{\textbf{Set12}~\cite{DnCNN}} & \multicolumn{3}{c|}{\textbf{BSD68}~\cite{BSD68}} & \multicolumn{3}{c}{\textbf{Urban100}~\cite{Urban100}} \\
   \textbf{Method} & $\sigma$$=$$15$ & $\sigma$$=$$25$ & $\sigma$$=$$50$ & $\sigma$$=$$15$ & $\sigma$$=$$25$ & $\sigma$$=$$50$ & $\sigma$$=$$15$ & $\sigma$$=$$25$ & $\sigma$$=$$50$ \\
\midrule[0.1em]
DnCNN~\cite{DnCNN}        & 32.67 & 30.35 & 27.18 & 31.62 & 29.16 & 26.23 & 32.28 & 29.80 & 26.35 \\
FFDNet~\cite{FFDNet}  & 32.75 & 30.43 & 27.32 & 31.63 & 29.19 & 26.29 & 32.40 & 29.90 & 26.50 \\ 
IRCNN~\cite{IRCNN}        & 32.76 & 30.37 & 27.12 & 31.63 & 29.15 & 26.19 & 32.46 & 29.80 & 26.22 \\ 
DRUNet~\cite{DRUNet}      & 33.25 & 30.94 & 27.90 & 31.91 & 29.48 & 26.59 & 33.44 & 31.11 & 27.96 \\ 
\midrule[0.1em]
Restormer\cite{Restormer} & \underline{33.35} & \underline{31.04} & \underline{28.01} & \underline{31.95} & \underline{29.51} & \underline{26.62} & \textbf{33.67} & \underline{31.39} & \underline{28.33} \\ 
\textbf{Ours}             & \textbf{33.35} & \textbf{31.30} & \textbf{28.13} & \textbf{31.98} & \textbf{29.58} & \textbf{26.77} & \underline{33.62} & \textbf{31.47} & \textbf{28.46}  \\
\midrule[0.1em]
\midrule[0.1em]
FOCNet~\cite{FOCNet}      & 33.07 & 30.73 & 27.68 & 31.83 & 29.38 & 26.50 & 33.15 & 30.64 & 27.40 \\
MWCNN~\cite{MWCNN}        & 33.15 & 30.79 & 27.74 & 31.86 & 29.41 & 26.53 & 33.17 & 30.66 & 27.42 \\
NLRN~\cite{NLRN}          & 33.16 & 30.80 & 27.64 & 31.88 & 29.41 & 26.47 & 33.45 & 30.94 & 27.49 \\
RNAN~\cite{RNAN}          & - & - & 27.70 & - & - & 26.48 & - & - & 27.65\\
DeamNet~\cite{DeamNet}    & 33.19 & 30.81 & 27.74 & 31.91 & 29.44 & 26.54 & 33.37 & 30.85 & 27.53 \\
DAGL~\cite{DAGL}          & 33.28 & 30.93 & 27.81 & 31.93 & 29.46 & 26.51 & 33.79 & 31.39 & 27.97 \\
SwinIR~\cite{SwinIR}      & 33.36 & 31.01 & 27.91 & \textbf{31.97} & 29.50 & 26.58 & 33.70 & 31.30 & 27.98 \\ 
\midrule[0.1em]
Restormer~\cite{Restormer} & \underline{33.42} & \underline{31.08} & \underline{28.00} & \underline{31.96} & \underline{29.52} & \underline{26.62} & \textbf{33.79} & \underline{31.46} & \underline{28.29} \\ 
\textbf{Ours}              & \textbf{33.47} & \textbf{31.15} & \textbf{28.12} & 31.92 & \textbf{29.67} & \textbf{26.78} & \underline{33.78} & \textbf{31.58} & \textbf{28.38} \\
\bottomrule[0.15em]
\end{tabular}}
}
\hfill
\parbox{.54\linewidth}{
\centering
\caption{\small \underline{\textbf{Color image denoising on Gaussian noise}}. Upper-bracket: models are trained on a range of noise levels. Lower-bracket: models are trained on the fixed noise level. PSNR is calculated on RGB channels.}
\label{table:gaussian_color_denoising_results}
\vspace{-2mm}
\setlength{\tabcolsep}{1.5pt}
\scalebox{0.7}{
\begin{tabular}{l | c c c | c c c | c c c | c c c}
\toprule[0.15em]
   & \multicolumn{3}{c|}{\textbf{CBSD68}~\cite{CBSD68}} & \multicolumn{3}{c|}{\textbf{Kodak24}} & \multicolumn{3}{c|}{\textbf{McMaster}~\cite{McMaster}} & \multicolumn{3}{c}{\textbf{Urban100}~\cite{Urban100}} \\
   \textbf{Method} & $\sigma$$=$$15$ & $\sigma$$=$$25$ & $\sigma$$=$$50$ & $\sigma$$=$$15$ & $\sigma$$=$$25$ & $\sigma$$=$$50$ & $\sigma$$=$$15$ & $\sigma$$=$$25$ & $\sigma$$=$$50$ & $\sigma$$=$$15$ & $\sigma$$=$$25$ & $\sigma$$=$$50$ \\
\midrule[0.1em]
IRCNN~\cite{IRCNN}    & 33.86 & 31.16 & 27.86 & 34.69 & 32.18 & 28.93 & 34.58 & 32.18 & 28.91 & 33.78 & 31.20 & 27.70 \\
FFDNet~\cite{FFDNet}  & 33.87 & 31.21 & 27.96 & 34.63 & 32.13 & 28.98 & 34.66 & 32.35 & 29.18 & 33.83 & 31.40 & 28.05 \\
DnCNN~\cite{DnCNN}    & 33.90 & 31.24 & 27.95 & 34.60 & 32.14 & 28.95 & 33.45 & 31.52 & 28.62 & 32.98 & 30.81 & 27.59 \\
DSNet~\cite{DSNet}    & 33.91 & 31.28 & 28.05 & 34.63 & 32.16 & 29.05 & 34.67 & 32.40 & 29.28 & - & - & -             \\
DRUNet~\cite{DRUNet}  & 34.30 & 31.69 & 28.51 & 35.31 & 32.89 & 29.86 & 35.40 & 33.14 & 30.08 & 34.81 & 32.60 & 29.61 \\
\midrule[0.1em]
Restormer~\cite{Restormer} & \textbf{34.39} & \underline{31.78} & \underline{28.59} & \underline{35.44} & \underline{33.02} & \underline{30.00} & \underline{35.55} & \underline{33.31} & \underline{30.29} & \textbf{35.06} & \underline{32.91} & \underline{30.02} \\
\textbf{Ours}              & \underline{34.37} & \textbf{31.87} & \textbf{28.68} & \textbf{35.52} & \textbf{33.13} & \textbf{30.15} & \textbf{35.62} & \textbf{33.38} & \textbf{30.40} & \underline{35.03} & \textbf{32.97} & \textbf{30.19} \\
\midrule[0.1em]
\midrule[0.1em]
RPCNN~\cite{RPCNN}         & - & 31.24 & 28.06 & - & 32.34 & 29.25 & - & 32.33 & 29.33 & - & 31.81 & 28.62\\
BRDNet~\cite{BRDNet}       & 34.10 & 31.43 & 28.16 & 34.88 & 32.41 & 29.22 & 35.08 & 32.75 & 29.52 & 34.42 & 31.99 & 28.56 \\
RNAN~\cite{RNAN}           & - & - & 28.27 & - & - & 29.58 & - & - & 29.72 & - & - & 29.08\\
RDN~\cite{RDN}             & - & - & 28.31 & - & - & 29.66 & - & - & - & - & - & 29.38\\
IPT~\cite{IPT}             & - & - & 28.39 & - & - & 29.64 & - & - & 29.98 & - & - & 29.71\\
SwinIR~\cite{SwinIR}       & 34.42 & 31.78 & 28.56 & 35.34 & 32.89 & 29.79 & 35.61 & 33.20 & 30.22 & \underline{35.13} & 32.90 & 29.82  \\
\midrule[0.1em]
Restormer~\cite{Restormer} & \underline{34.40} & \underline{31.79} & \underline{28.60} & \underline{35.47} & \underline{33.04} & \underline{30.01} & \underline{35.61} & \underline{33.34} & \underline{30.30} & \textbf{35.13} & \underline{32.96} & \underline{30.02}  \\ 
\textbf{Ours}  & \textbf{34.48} & \textbf{31.97} & \textbf{28.83}  & \textbf{35.58} & \textbf{33.21} & \textbf{30.23}
                           & \textbf{35.75} & \textbf{33.56} & \textbf{30.46} & 35.11 & \textbf{33.13} & \textbf{30.27} \\             
\bottomrule[0.15em]
\end{tabular}}
}
\vspace*{-1mm}
\end{table*}

%% file: figures/006_gaussian_color_denoising_visual_comparison.tex
\begin{figure*}[!t]
\begin{center}
\scalebox{0.97}{
\begin{tabular}[b]{c@{ } c@{ }  c@{ } c@{ } c@{ } c@{ } c@{ } c@{ } }
\hspace{-4mm}
    \includegraphics[width=.174\textwidth,valign=t]{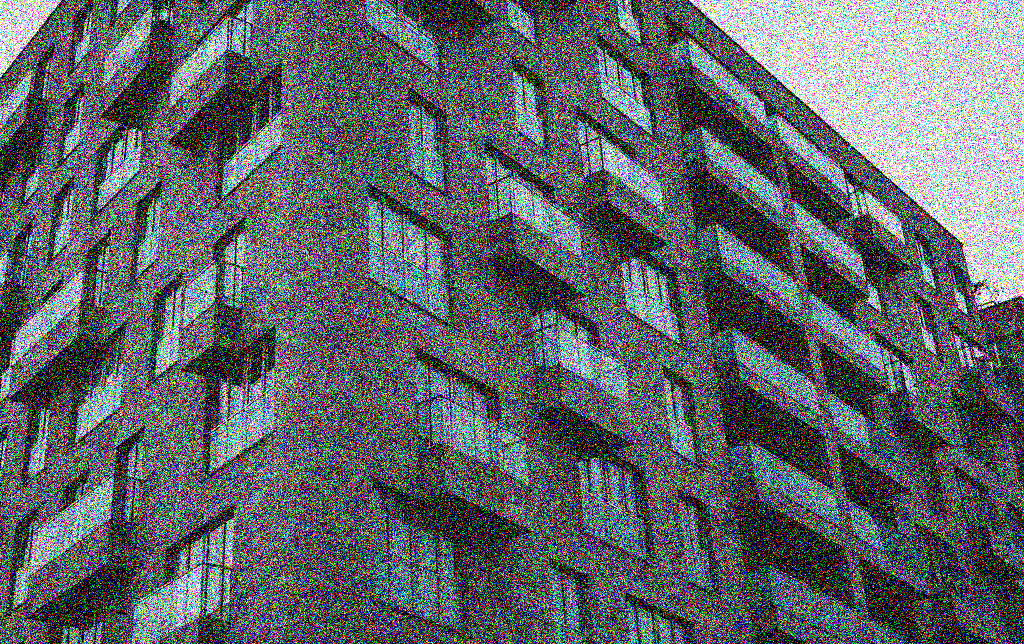} &   
    \includegraphics[width=.115\textwidth,valign=t]{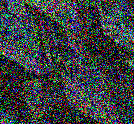}  &
    \includegraphics[width=.115\textwidth,valign=t]{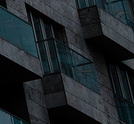} &   
    \includegraphics[width=.115\textwidth,valign=t]{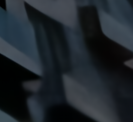}    &
    \includegraphics[width=.115\textwidth,valign=t]{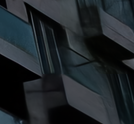}   &
    \includegraphics[width=.115\textwidth,valign=t]{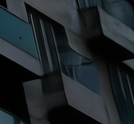} &
    \includegraphics[width=.115\textwidth,valign=t]{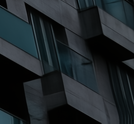} &
    \includegraphics[width=.115\textwidth,valign=t]{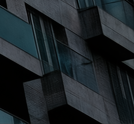}      \\
    
    \small~Input image & \small~Degraded &\small~Label & \small~RDN~\cite{RDN} & \small~RNAN~\cite{RNAN} & \small~SwinIR~\cite{SwinIR} & \small~Restormer~\cite{Restormer}  & \small~\textbf{Ours} \\

\end{tabular}
}
\end{center}
\vspace{-7mm}
\caption{ \small  \underline{\textbf{Color image denoising results on Gaussian noise}}. Compared with others, the proposed method can obtain finer results. 
}
\label{fig:gaussian_color_denoising_visual_comparison}
\vspace{-1mm}
\end{figure*}

%% file: figures/007_gaussian_grayscale_denoising_visual_comparison.tex
\begin{figure*}[!t]
\begin{center}
\scalebox{0.97}{
\begin{tabular}[b]{c@{ } c@{ }  c@{ } c@{ } c@{ } c@{ } c@{ } c@{ } }
\hspace{-4mm}
    \includegraphics[width=.133\textwidth,valign=t]{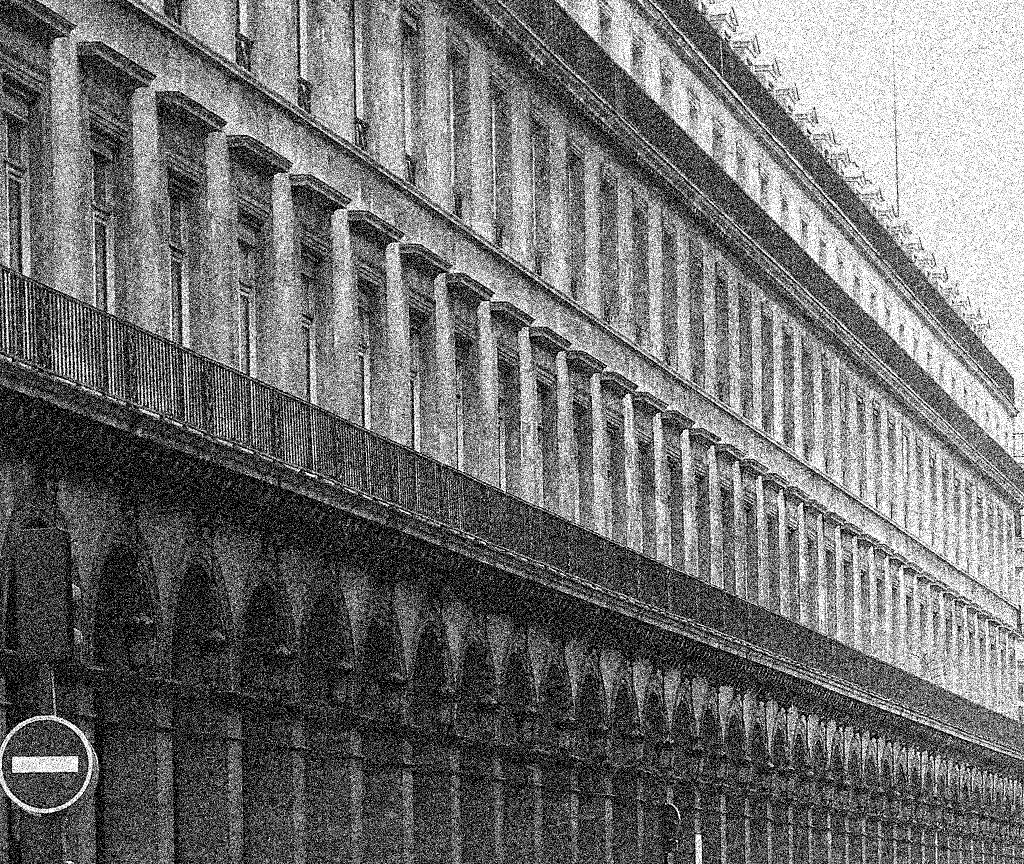} &   
    \includegraphics[width=.122\textwidth,valign=t]{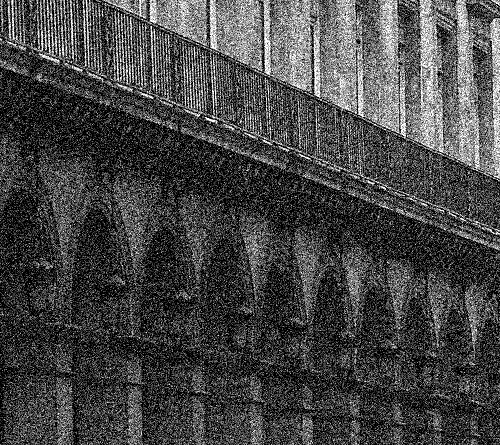}  &
    \includegraphics[width=.122\textwidth,valign=t]{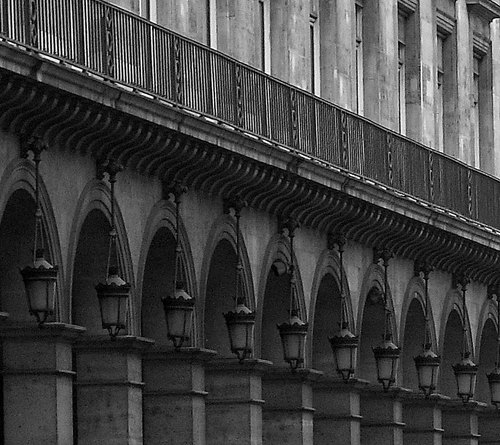} &   
    \includegraphics[width=.122\textwidth,valign=t]{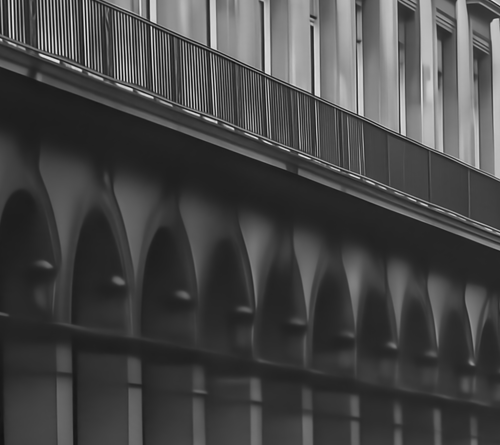}    &
    \includegraphics[width=.122\textwidth,valign=t]{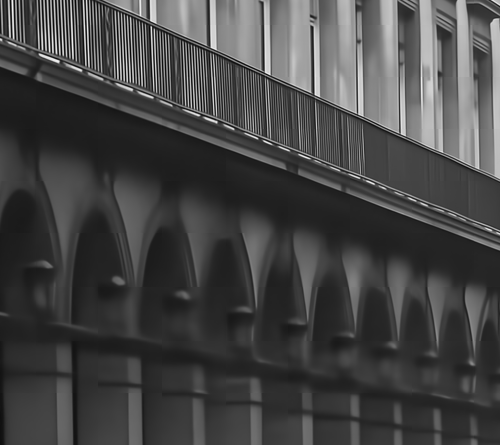}   &
    \includegraphics[width=.122\textwidth,valign=t]{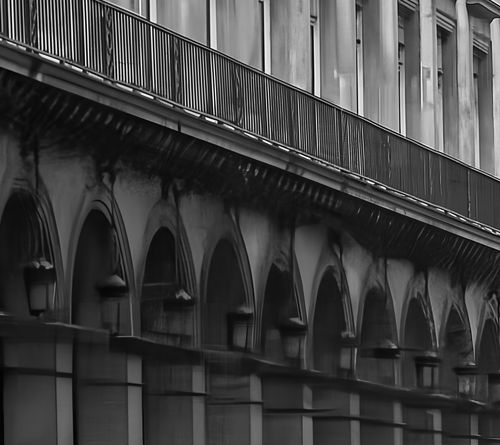} &
    \includegraphics[width=.122\textwidth,valign=t]{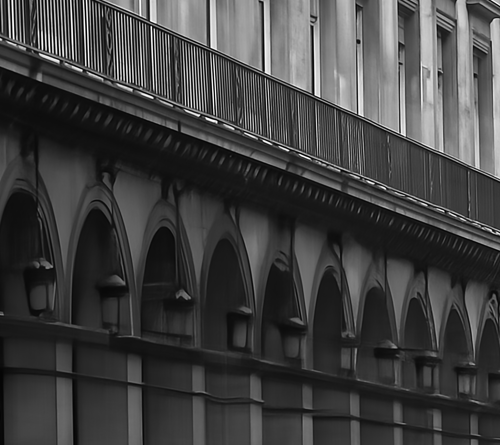} &
    \includegraphics[width=.122\textwidth,valign=t]{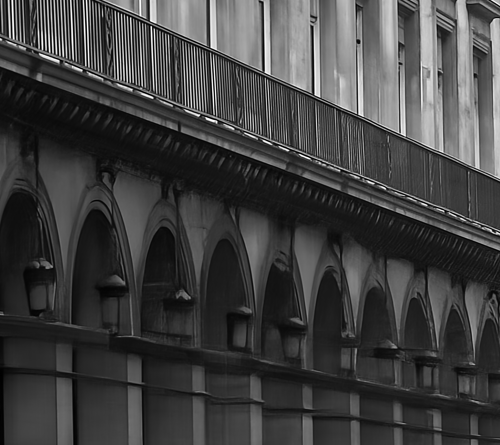}      \\
    
    \small~Input image & \small~Degraded &\small~Label & \small~DeamNet~\cite{DeamNet} & \small~RNAN~\cite{RNAN} & \small~SwinIR~\cite{SwinIR} & \small~Restormer~\cite{Restormer}  & \small~\textbf{Ours} \\

\end{tabular}
}
\end{center}
\vspace{-7mm}
\caption{\small \underline{\textbf{Grayscale image denoising results on Gaussian noise}}. Our method can restore detailed results especially in texture regions.}
\label{fig:gaussian_grayscale_denoising_visual_comparison}
\vspace{-0.5em}
\end{figure*}

%% file: tables/004_denoising_realworld_table.tex
\begin{table*}[!t]
\begin{center}
\caption{\small \small \underline{\textbf{Real-world image denoising results}}. We train and evaluate our mehtod on SIDD~\cite{sidd} datasets. \textcolor{red}{$\ast$} denotes methods using additional training data. PSNR and SSIM scores are calculated on RGB channels.}
\label{table:realworld_denoising_results}
\vspace{-2mm}
\setlength{\tabcolsep}{2.5pt}
\scalebox{0.68}{
\begin{tabular}{c| c| c c c c c c c c c c c c c c c | c c }
\toprule[0.15em]
& \textbf{Method} & DnCNN   & BM3D & CBDNet\textcolor{red}{*}   & RIDNet\textcolor{red}{*}  & AINDNet\textcolor{red}{*}  & VDN & SADNet\textcolor{red}{*} & DANet+\textcolor{red}{*} & CycleISP\textcolor{red}{*} & MIRNet & DeamNet\textcolor{red}{*} & MPRNet & DAGL &  Uformer & Restormer & \textbf{Ours} \\
\textbf{Dataset} & & \cite{DnCNN} & \cite{BM3D} & \cite{CBDNet} & \cite{RIDNet} & \cite{AINDNet} & \cite{VDN} &  \cite{SADNet}	 & \cite{DANet} &  \cite{CycleISP} & \cite{MIRNet} & \cite{DeamNet}& \cite{MPRNet} & \cite{DAGL} & \cite{Uformer} & \cite{Restormer} &  \\
\midrule[0.1em]
\textbf{SIDD} & PSNR~$\textcolor{black}{\uparrow}$ &  23.66 &  25.65 & 30.78  &  38.71  &  39.08  & 39.28  & 39.46  & 39.47 & 39.52  & 39.72  & 39.47  & 39.71 & 38.94 & 39.77 & \underline{40.02} & \textbf{40.09}\\
~\cite{sidd}  & SSIM~$\textcolor{black}{\uparrow}$ &  0.583 &  0.685 & 0.801  &  0.951  &  0.954  & 0.956  & 0.957  & 0.957 & 0.957  & 0.959  & 0.957  & 0.958 & 0.953 & 0.959 & \underline{0.960} & \textbf{0.960}\\
\bottomrule[0.15em]
\end{tabular}
}
\end{center}
\vspace{-4mm}
\end{table*}

%% file: tables/006_ablation_studies_arxiv.tex
\begin{table*}[t]
\parbox{.38\linewidth}{
\centering
\caption{\small Effect of condition information. } 
\label{table:ablation_condition_information}
\vspace{-3mm}
\setlength{\tabcolsep}{2pt}
\scalebox{0.75}{
\begin{tabular}{l | c | c | c | c | c | c}
\toprule[0.15em]
Method & baseline & $N$$=$$5$ & $N$$=$$10$ & $N$$=$$20$ & $N$$=$$30$ & $N$$=$$40$ \\
\midrule[0.1em]
%
PSNR$\uparrow$  & 30.16  & 31.13 & 31.36 & 31.57 & 31.51 & 31.60 \\
SSIM$\uparrow$  & 0.932  & 0.941 & 0.945 & 0.947 & 0.947 & 0.948 \\
\bottomrule[0.15em]
\end{tabular}
}
}
\hfill
\parbox{.30\linewidth}{
\centering
\caption{\small Effect of integration strategy. }
\label{table:ablation_injection}
\vspace{-3mm}
\setlength{\tabcolsep}{2pt}
\scalebox{0.75}{
\begin{tabular}{l | c | c | c | c}
\toprule[0.15em]
 Method                 & baseline & Enc. & Dec. & Enc. \&  Dec. \\
\midrule[0.1em]
PSNR$\uparrow$ &  30.16  & 31.37   & 30.31   & 31.57 \\
SSIM$\uparrow$ &  0.932 & 0.946  & 0.934  & 0.947 \\
\bottomrule[0.15em]
\end{tabular}
}}
\hfill
\parbox{.30\linewidth}{
\centering
\caption{\small Effect of generated guidance. }
\label{table:ablation_ref}
\vspace{-3mm}
\setlength{\tabcolsep}{7pt}
\scalebox{0.75}{
\begin{tabular}{l | c | c | c}
\toprule[0.15em]
 Method                 & baseline & Degra. & Ours \\
\midrule[0.1em]
PSNR$\uparrow$ & 30.16  &  30.13  & 31.57 \\
SSIM$\uparrow$ & 0.932  &  0.931  & 0.947 \\
\bottomrule[0.15em]
\end{tabular}
}}
\vspace*{-1mm}
\end{table*}

%% file: main_sec/5_conclusion.tex
\vspace{-0.3em}
\section{Conclusion}
\label{sec:conclusion}
In this paper, we provide a new perspective for image restoration.
Considering that content and degradation are tightly coupled in image representation, while decoupled in textual representation.
Rather than direct restoration on image level, we suggest conducting restoration on textual level and in turn utilizing the restored textual content to assist image restoration.
To this end, we propose a plug-and-play approach that first encodes degraded images into degraded textual representations and then removes textual degradation information to obtain restored textual representations. 
Conditioned on the representations, we employ a pre-trained T2I model to synthesize clean guidance images for improving image restoration.
We evaluate our method on various image restoration tasks. The experimental results demonstrate it outperforms the state-of-the-art ones.

%% file: main_sec/suppl.tex
\maketitlesupplementary

\renewcommand{\thesection}{\Alph{section}}
\renewcommand{\thetable}{\Alph{table}}
\renewcommand{\thefigure}{\Alph{figure}}
\renewcommand{\theequation}{\Alph{equation}}
\setcounter{section}{0}
\setcounter{figure}{0}
\setcounter{table}{0}
\setcounter{equation}{0}


The content of the supplementary material involves: 
\begin{itemize}
\vspace{2mm}
\item Experimental details  in \cref{sec:experimental_details}. 
\vspace{2mm}
\item Effect of textual restoration  in \cref{sec:effect_textual_restoration}.
\vspace{2mm}
\item Explicit textual representation v.s. implicit textual representation in \cref{sec:explicit_vs_implicit}.
\vspace{2mm}
\item Guidance visualization in \cref{sec:guidance_visualization}.
\vspace{2mm}
\item More visual comparisons in \cref{sec:more_visual_comparisons}.
\vspace{2mm}
\end{itemize}

\section{Experimental Details}
\label{sec:experimental_details}
In this section, we list experimental details for different image restoration task (\ie, all-in-one image restoration~\cite{PromptIR}, image deblurring~\cite{NAFNet, Restormer}, image dehazing~\cite{SFNet}, image deraining~\cite{DRSformer}, and image denoising~\cite{Restormer}), the proposed degradation-free guidance generation process (\ie, image-to-text mapping and textual restoration), and guided-restoration.

\noindent\textbf{All-in-One Image Restoration}.
We adopt PromptIR~\cite{PromptIR} as our backbone in all-in-one image restoration.
Following~\cite{PromptIR}, network has 8 stages (the first 7 stages as main network, the last stage as refinement), the number of blocks for each stages is [4, 6, 6, 8, 6, 6, 4, 4], network width is 48, the number of heads for each stages is [1, 2, 4, 8, 4, 2, 1, 1]. 
In perspective of training data, we adopt concatenation of 400 images from BSD~\cite{BSD} and 4,744 images from WED~\cite{WED} dataset as denoising training data, 200 images from Rain100L~\cite{Rain200H} for deraining task, 72,135 images from SOTS for dehazing task. Considering dataset size gap among different tasks, we properly enlarge deraining data and denoising data as~\cite{PromptIR}.
To train, we adopt AdamW optimizer with CosineAnnealing learning rate scheduler, the initial learning rate of the main restoration network and dynamic aggregation is set to 2e-4 and 1e-4, respectively. We train all-in-one image restoration on 4 Tesla-V100 GPUs with training patch size 128, batch size 48.
Performance reported on Table {\color{red}1} is referred to~\cite{PromptIR}. PSNR and SSIM scores are calculated on RGB channels, except which of deraining task are calculated on Y-channel in YCbCr color space.

\noindent\textbf{Image Deblurring}.
We adopt NAFNet~\cite{NAFNet} as our backbone in single-image motion deblurring, Restormer~\cite{Restormer} as backbone in defocus deblurring.
In single-image motion deblurring task, we follow~\cite{NAFNet}, main restoration network has 9 stages, and the number of blocks for each stage is [1, 1, 1, 28, 1, 1, 1, 1, 1], network width is 64.
We adopt GoPro~\cite{GoPro} as our training data and directly evaluated the trained model on GoPro validation set, HIDE~\cite{HIDE} testing set, and Realblur~\cite{RealBlur} dataset. GoPro dataset has 4,214 blur-sharp paris of data (2,103 for training and 1,111 for validation), testset of HIDE dataset consist of 2,025 images, and two subsets of Realblur both have 980 images.
To train single-image motion deblurring, we adopt AdamW optimizer with CosineAnnealing learning rate scheduler, the initial learning rate of the main restoration network and dynamic aggregation module is 1e-4 and 5e-5, respectively. We train single-image motion deblurring on 8 Tesla-V100 GPUs with training patch size of 256, batch size of 16.
In defocus deblurring task, we follow~\cite{Restormer}, main restoration network has 8 stages (the last stage as refinement stage), the number of blocks of each stage is [4, 6, 6, 8, 6, 6, 4, 4], network width is 48, the number of heads of each stage is [1, 2, 4, 8, 4, 2, 1, 1]. Note that input channels is 6 in dual-pixel defocus deblurring, the output channels is 3 in both single-image defocus deblurring and dual-pixel defocus deblurring.
We adopt DPDD~\cite{DPDNet} dataset as our training data. DPDD dataset contains 500 indoor \& outdoor scenes captured by DSLR camera. Each scene includes three defocus input images and a corresponding all-in-focus ground-truth image. Three input images are labeled as left, right and center views. The left and right defocused sub-aperture views are acquired with a wide camera aperture setting, and the corresponding all-in-focus ground-truth image captured with a narrow aperture. Following Restormer~\cite{Restormer}, we use sub-aperture data to train dual-pixel defocus deblurring, and we use center input image to train single-image defocus deblurring. To perform evaluation, we separately evaluate trained model in indoor \& outdoor scene testing data and the average performance is calculated by weighted combination.
To train defocus deblurring, we maintain progressive learning in official implementation, we adopt AdamW optimizer with CosineAnnealing learning rate scheduler, the initial learning rate of main restoration network and dynamic aggregation module is set to 3e-4 and 1e-4, respectively.
Performance reported in Table {\color{red}2} and Table {\color{red}3} is referred to ~\cite{NAFNet} and ~\cite{Restormer}. PSNR and SSIM scores are calculated on RGB channels.

\noindent\textbf{Image Dehazing}.
We adopt SFNet~\cite{SFNet} as our backbone in image dehazing.
Following~\cite{SFNet}, network width is set to 32, the number of resblocks is 16.
We train image dehazing on RESIDE~\cite{reside} dataset, we train and evaluate method separately on indoor scene and outdoor scene data.
To train image dehazing, we use Adam optimizer with CosineAnnealing learning rate scheduler, the initial learning rate of main network and dynamic aggregation is set to 1e-4 and 5e-5, respectively.
Performance in Table {\color{red}4} is referred to ~\cite{SFNet}. PSNR and SSIM scores are calculated on RGB channels.

\noindent\textbf{Image Deraining}.
We adopt DRSformer~\cite{DRSformer} as our backbone in image deraining.
Following~\cite{DRSformer}, we adopt 7 stages for main restoration network, the number of blocks for each stage is [4, 6, 6 ,8, 6, 6, 4], network width is set to 48, the number of heads for each stage is [1, 2, 4, 8, 4, 2, 1]. 
For training, we separately train and evaluate the proposed method on four datasets with synthetic rainstreak degradation, including Rain200L~\cite{Rain200H}, Rain200H~\cite{Rain200H}, DID-Data~\cite{DID-Data}, and DDN-Data~\cite{DDN-Data}. Rain200H and Rain200L dataset contain 1,800 pairs of rainy-clean images for training and 200 pairs images for testing. In DID-Data and DDN-Data, the synthetic rainstreak has different directions and different levels. DID-Data contains 12,000 pairs of images for training and 1,200 pairs of images for testing. DDN-Data contains 12,600 pairs of images for training and 1,400 images for testing. 
We employ MEFC~\cite{DRSformer} module for Rain200H, DID-Data, and DDN-Data.
During training, we adopt AdamW optimizer with CosineAnnealing learning rate scheduler, patch size and batch size is set to 128 and 16 with 4 Tesla-V100 GPUs, the initial learning rate of main network and dynamic aggregation is set to 1e-4 and 5e-5, respectively.
Performance in Table  {\color{red}5} is referred to ~\cite{DRSformer}. PSNR and SSIM scores are calculated on Y channel in YCbCr color space.

\noindent\textbf{Image Denoising}.
We adopt Restormer~\cite{Restormer} as our backbone in image denoising.
Following~\cite{Restormer}, we employ the bias-free network with 8 stages, the number of blocks for each stage is [4, 6, 6, 8, 6, 6, 4, 4], network width is set to 48, the number of heads of each stage is [1, 2, 4, 8, 4, 2, 1, 1].
We adopt concatenation data of Div2k~\cite{Div2k} (800 images for training), Flickr2k (2,650 images for training), BSD~\cite{BSD} (400 images for training), and WED~\cite{WED} (4,744 images for training) to train Gaussian grayscale denoising and Gaussian color denoising.
We adopt 320 high-resolution images in SIDD~\cite{sidd} dataset for real-world denoising.
During training, we also maintain the progressive learning strategy, we adopt AdamW optimizer with CosinAnnealing learning rate scheduler, the initial learning rate of main restoration network and dynamic aggregation is set to 3e-4 and 1e-4, respectively.
Performance in Table {\color{red}6}, Table {\color{red}7}, and Table {\color{red}8} is referred to ~\cite{Restormer}. PSNR and SSIM scores are calculated on RGB channels.

\noindent\textbf{Image-to-Text Mapping}.
To enable our image-to-text mapping network can project both clean images and degraded images into textual space, we use the collection of high-quality data, degraded data from different image restoration tasks as our training data. High-quality data includes LSDIR~\cite{LSDIR} dataset and HQ-50K~\cite{HQ50K} dataset, LSDIR dataset contains 84,991 high-quality images for training, HQ-50K dataset contains 50,000 high-quality images for training. Degraded data includes GoPro, RESIDE, Rain200H, Rain200L, DID-Data, DDN-Data, and DFBW data with synthetic Gaussian noise.
During training, we crop high-quality high-resolution data (LSDIR and HQ-50K) into $512$$\times$$512$ as input, for others we centerly crop images along shorter side and resize them to $512$$\times$$512$ as input.
The mapping network is implemented as four-layer MLP network, and we adopt $N$$=$$20$ words to control representation capability of textual word embedding.
To encode image concepts into textual space, feature from the last layer of CLIP image encoder is selected as input to image-to-text mapping network. 
The learning rate is set to 1e-6 and batch size is set to 4.

\noindent\textbf{Textual Restoration}.
To train textual restoration network, we use concatenation of training dataset used in different image restoration tasks as our training data.
We adopt the same strategy to preprocess pairs of degraded-clean data to $512$$\times$$512$ patches as input.
The same with image-to-text mapping network, the textual restoration network is also implemented by four-layer MLP network.
The learning rate is set to 1e-6 and training batch size is set to 4.
During guidance generation, we use 200 steps of DDIM scheduler with scale of 5.

\noindent\textbf{Guided-Restoration}.
Following~\cite{MASA-SR}, the dynamic aggregation includes two steps: feature matching and feature aggregation.
In feature matching, we adopt a shared $n$-stages encoder to extract multi-scale feature from degraded input and clean guidance, $n$ depends on total downsampling ratio of the main restoration network, each stage is with 4 residual blocks, the width of encoder is the same to the width of main network.
We then adopt a coarse-to-fine manner to match useful information for each patch of degraded input, \eg, we first match in coarse block level then match in fine patch level.
In coarse matching, feature block size is set to 8, dilation ratio is set to [1, 2, 3]. In fine matching, patch size is set to 3.
For feature aggregation, we employ a more general way. We simply use concatenation \& residual/self-attention blocks with adaptive scaling factor $\alpha$ to fuse guidance information to main restoration network, \ie, Eq. ({\color{red}4}).

\section{Effect of Textual Restoration}
\label{sec:effect_textual_restoration}
In this section, we demonstrate the effectiveness of our textual restoration. We discard textual restoration and directly use the output of image-to-text mapping network as conditional input for diffusion model, and the visual results of the synthetic guidance images are shown in Fig.~\ref{fig:suppl_textual_restoration}, denoted as \textbf{w/o. textual restoration}.
%

\input{suppl_figures/suppl_017_single_i2tmapping.tex}

\section{Explicit Textual Representation \\ v.s. Implicit Textual Representation}
\label{sec:explicit_vs_implicit}
In this section, we compare synthetic guidance images conditioned on explicit text representation and implicit textual representation,
1) explicit text representation: we first convert degraded images into image caption by BLIPv2~\cite{BLIPv2}, then we manually discarding degradation-related text in image caption, finally we use the processed image caption as text prompt input to StableDiffusion to get synthetic guidance images. Denoted as \textbf{Explicit}.
2) implicit textual representation: our method, which is denoted as \textbf{Ours}.
As shown in Fig.~\ref{fig:suppl_explicit_implicit_deblur}, Fig.~\ref{fig:suppl_explicit_implicit_derain}, Fig.~\ref{fig:suppl_explicit_implicit_dehaze}, and Fig.~\ref{fig:suppl_explicit_implicit_denoise}, we illustrate visual comparison for image deblurring, image deraining, image dehazing, and image denoising tasks. 
We can found though explicit text representation can describe content of degraded image properly, the synthetic results cannot maintain style, details and texture of original content. And in image denoising task, explicitly converting degraded noise image into image caption usually leads to wrong captions and thus cannot provide useful guidance image for restoration.
\section{Guidance Visualization}
\label{sec:guidance_visualization}

We illustrate synthesized guidance images for each image restoration:image deblurring shows in Fig.~\ref{fig:suppl_guidance_images_deblur}, image deraining shows in Fig.~\ref{fig:suppl_guidance_images_derain}, image dehazing shows in Fig.~\ref{fig:suppl_guidance_images_dehaze}, image denoising shows in Fig.~\ref{fig:suppl_guidance_images_denoise}.

\section{More Visual Comparisons}
\label{sec:more_visual_comparisons}
We provide visual comparison for different image restoration tasks:
\begin{itemize}
    \item All-in-one image restoration: image deraining results show in Fig.~\ref{fig:suppl_all_in_one_deraining_visual_results_1} and Fig.~\ref{fig:suppl_all_in_one_deraining_visual_results_2}, image dehazing results show in Fig.~\ref{fig:suppl_all_in_one_dehazing_visual_results}, image denoising results show in Fig.~\ref{fig:suppl_all_in_one_denoising_visual_results}.
    \item Image deblurring results: single-image motion deblurring results show in Fig.~\ref{fig:suppl_single_image_motion_deblurring_visual_results_1} and Fig.~\ref{fig:suppl_single_image_motion_deblurring_visual_results_2}, defocus deblurring results show in Fig.~\ref{fig:suppl_defocus_deblurring_visual_results_1} and Fig.~\ref{fig:suppl_defocus_deblurring_visual_results_2}.
    \item Image dehazing results: Fig.~\ref{fig:suppl_dehazing_visual_results}.
    \item Image deraining results: Fig.~\ref{fig:suppl_deraining_visual_results_1}, Fig.~\ref{fig:suppl_deraining_visual_results_2}, and Fig.~\ref{fig:suppl_deraining_visual_results_3}.
    \item Image denoising results: Fig.~\ref{fig:suppl_denoising_visual_results}.
\end{itemize}

\input{suppl_figures/suppl_013_explicit_implicit_comparison_deblur.tex}
\input{suppl_figures/suppl_014_explicit_implicit_comparison_derain.tex}
\input{suppl_figures/suppl_015_explicit_implicit_comparison_dehaze.tex}
\input{suppl_figures/suppl_016_explicit_implicit_comparison_denoise.tex}

\input{suppl_figures/suppl_009_guidance_images_deblur.tex}
\input{suppl_figures/suppl_010_guidance_images_derain.tex}
\input{suppl_figures/suppl_011_guidance_images_dehaze.tex}
\input{suppl_figures/suppl_012_guidance_images_denoise.tex}

\input{suppl_figures/suppl_001_01_all_in_one_deraining_visual_results.tex}
\input{suppl_figures/suppl_001_02_all_in_one_deraining_visual_results.tex}

\input{suppl_figures/suppl_002_all_in_one_dehazing_visual_results.tex}

\input{suppl_figures/suppl_003_all_in_one_denoising_visual_results.tex}

\input{suppl_figures/suppl_004_01_motion_deblurring_visual_results.tex}
\input{suppl_figures/suppl_004_02_motion_deblurring_visual_results.tex}

\input{suppl_figures/suppl_005_01_defocus_deblurring_visual_results.tex}
\input{suppl_figures/suppl_005_02_defocus_deblurring_visual_results.tex}

\input{suppl_figures/suppl_007_dehazing_visual_results.tex}

\input{suppl_figures/suppl_006_01_deraining_visual_results.tex}
\input{suppl_figures/suppl_006_02_deraining_visual_results.tex}
\input{suppl_figures/suppl_006_03_deraining_visual_results.tex}

\input{suppl_figures/suppl_008_denoising_visual_results.tex}

%% file: suppl_figures/suppl_017_single_i2tmapping.tex
\begin{figure}[t]
\begin{center}
\setlength{\tabcolsep}{1.5pt}
\scalebox{0.85}{
\begin{tabular}[b]{c c c}
\vspace{0.2em}
\includegraphics[width=.18\textwidth,valign=t]{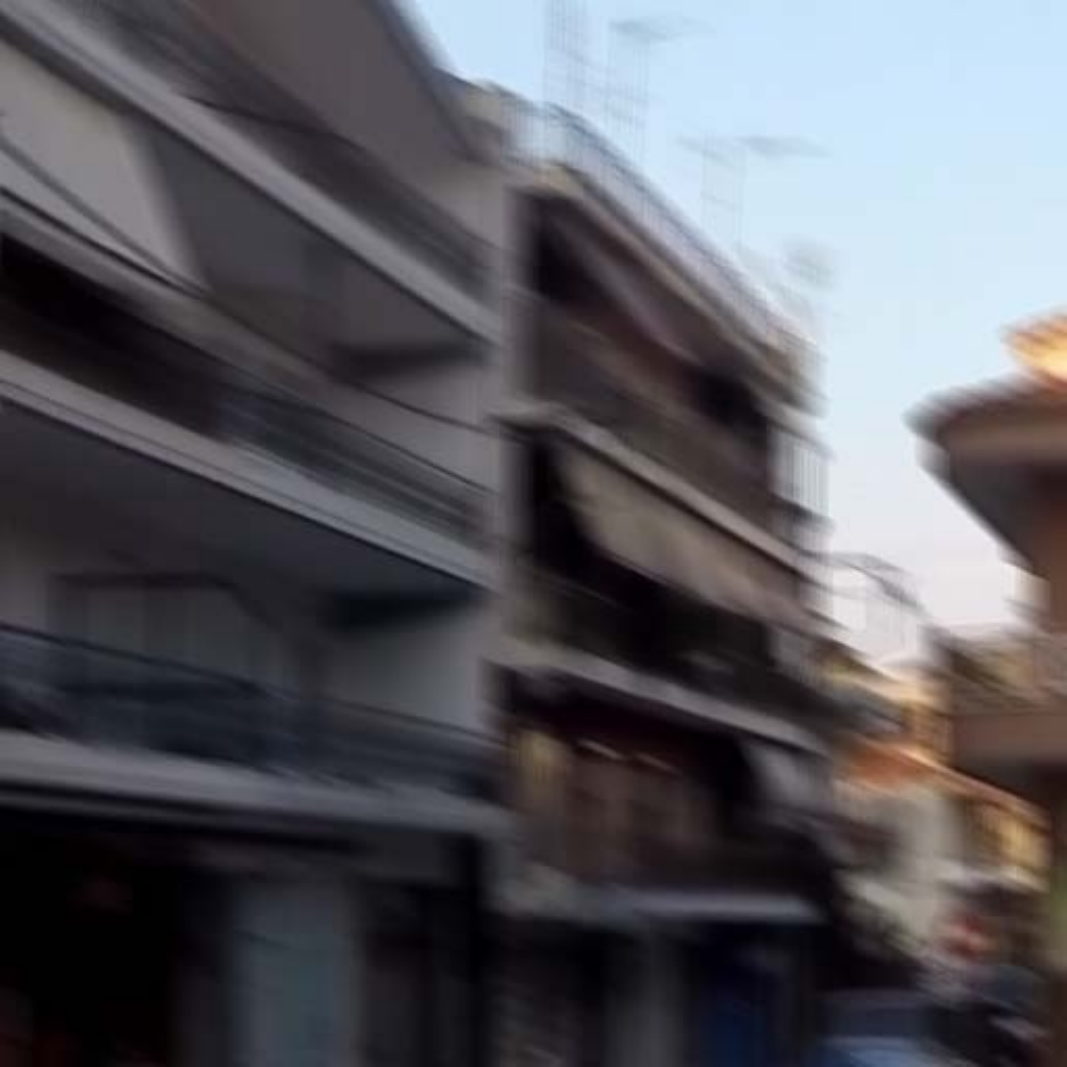} &   
\includegraphics[width=.18\textwidth,valign=t]{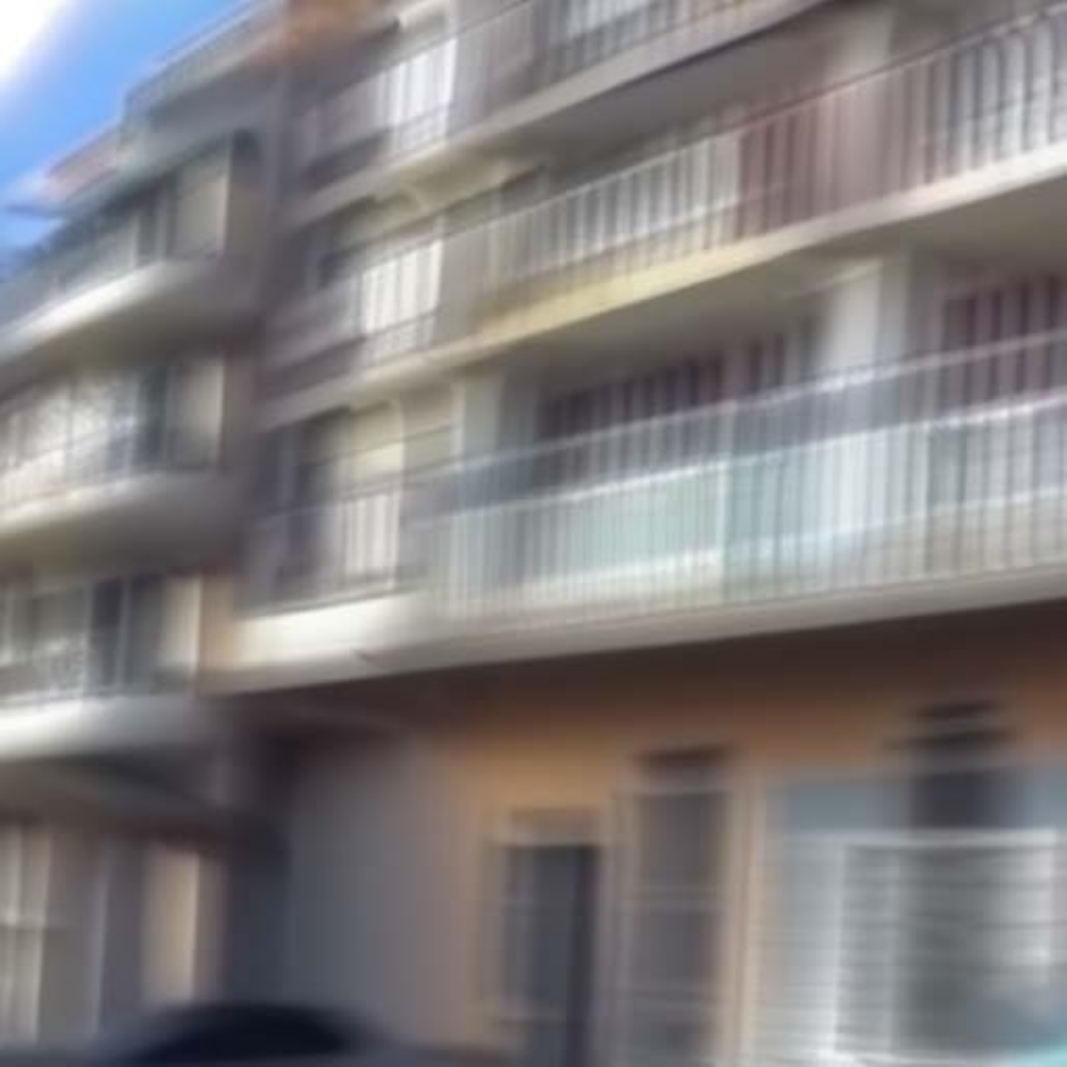} &
\includegraphics[width=.18\textwidth,valign=t]{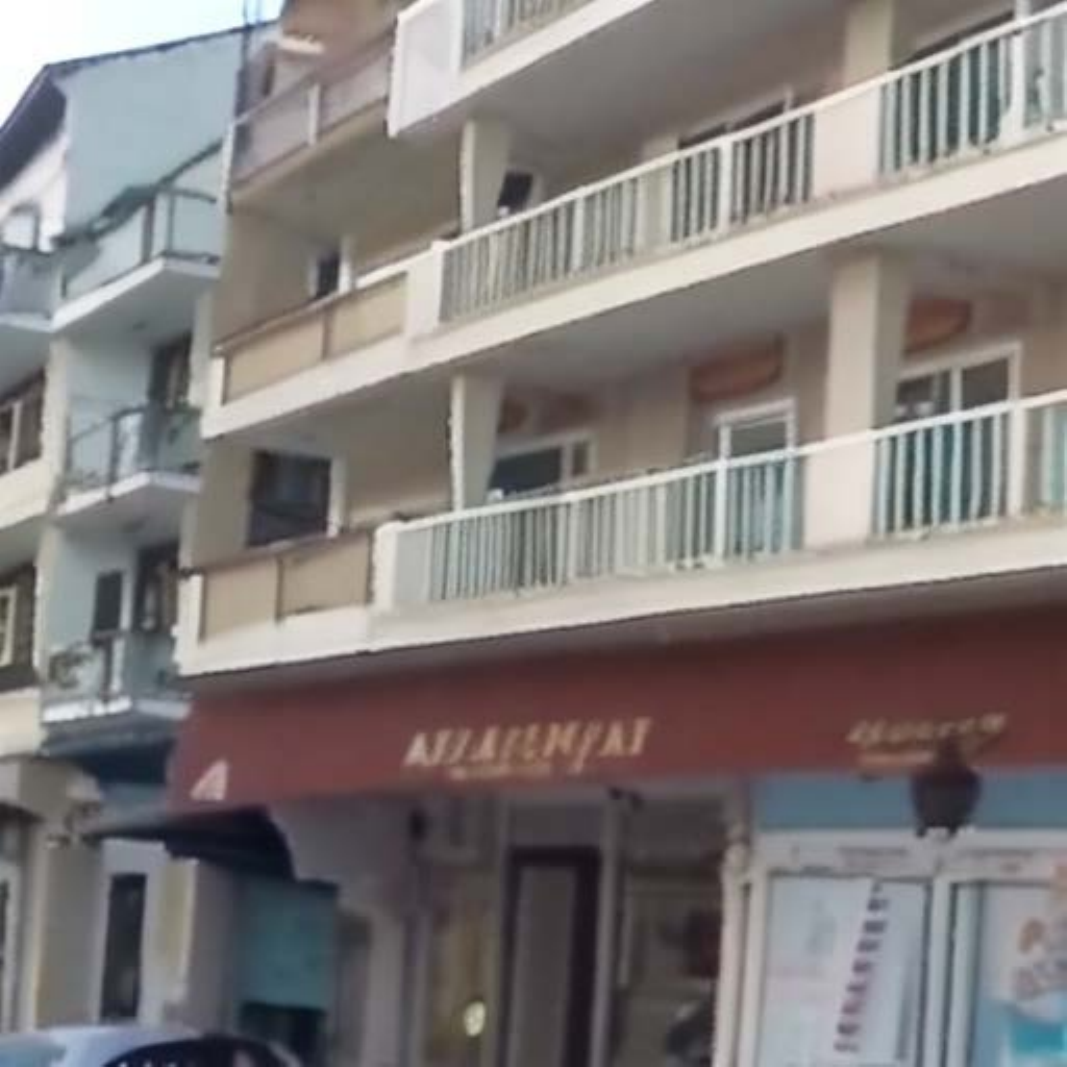}     \\
\vspace{0.2em}
\includegraphics[width=.18\textwidth,valign=t]{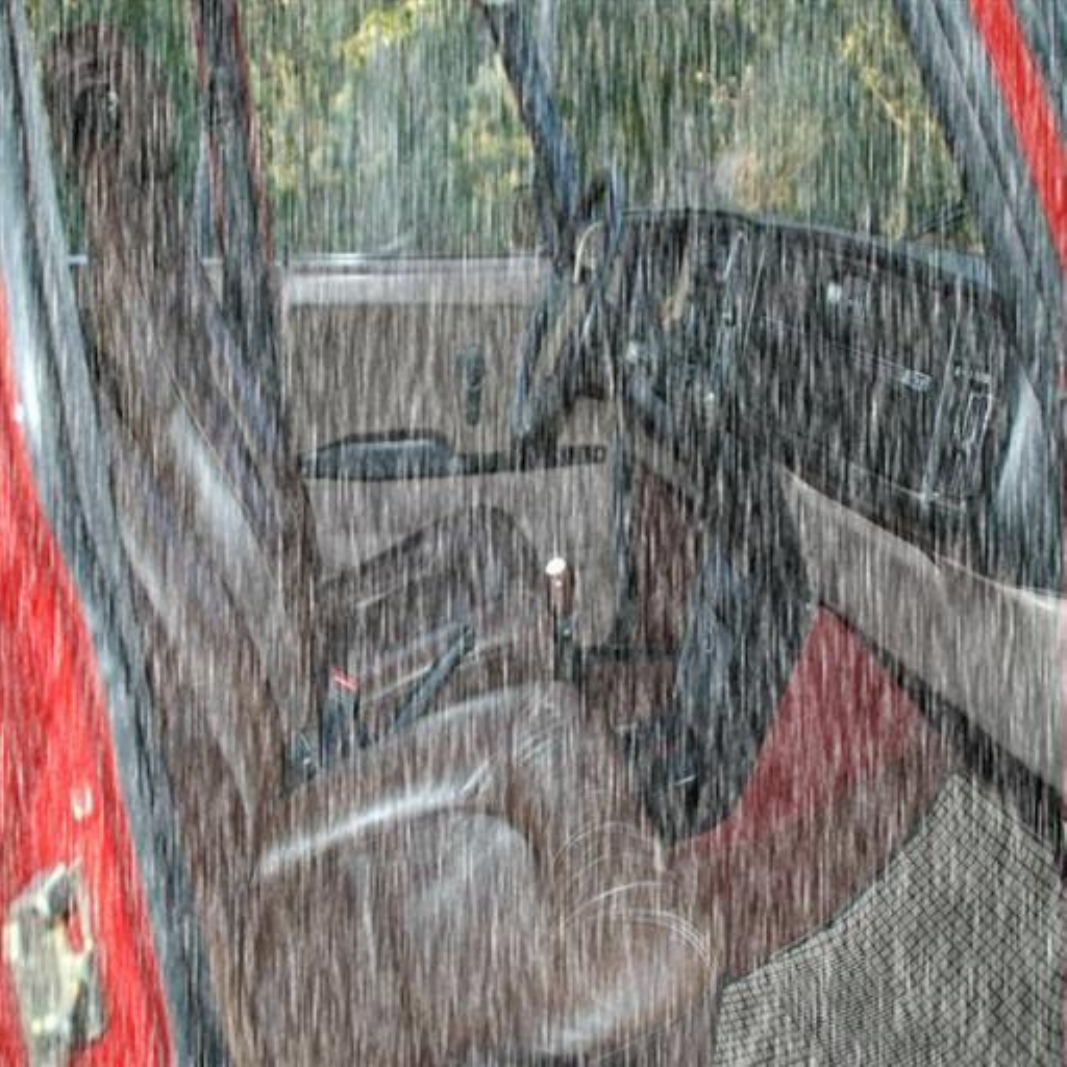}    &   
\includegraphics[width=.18\textwidth,valign=t]{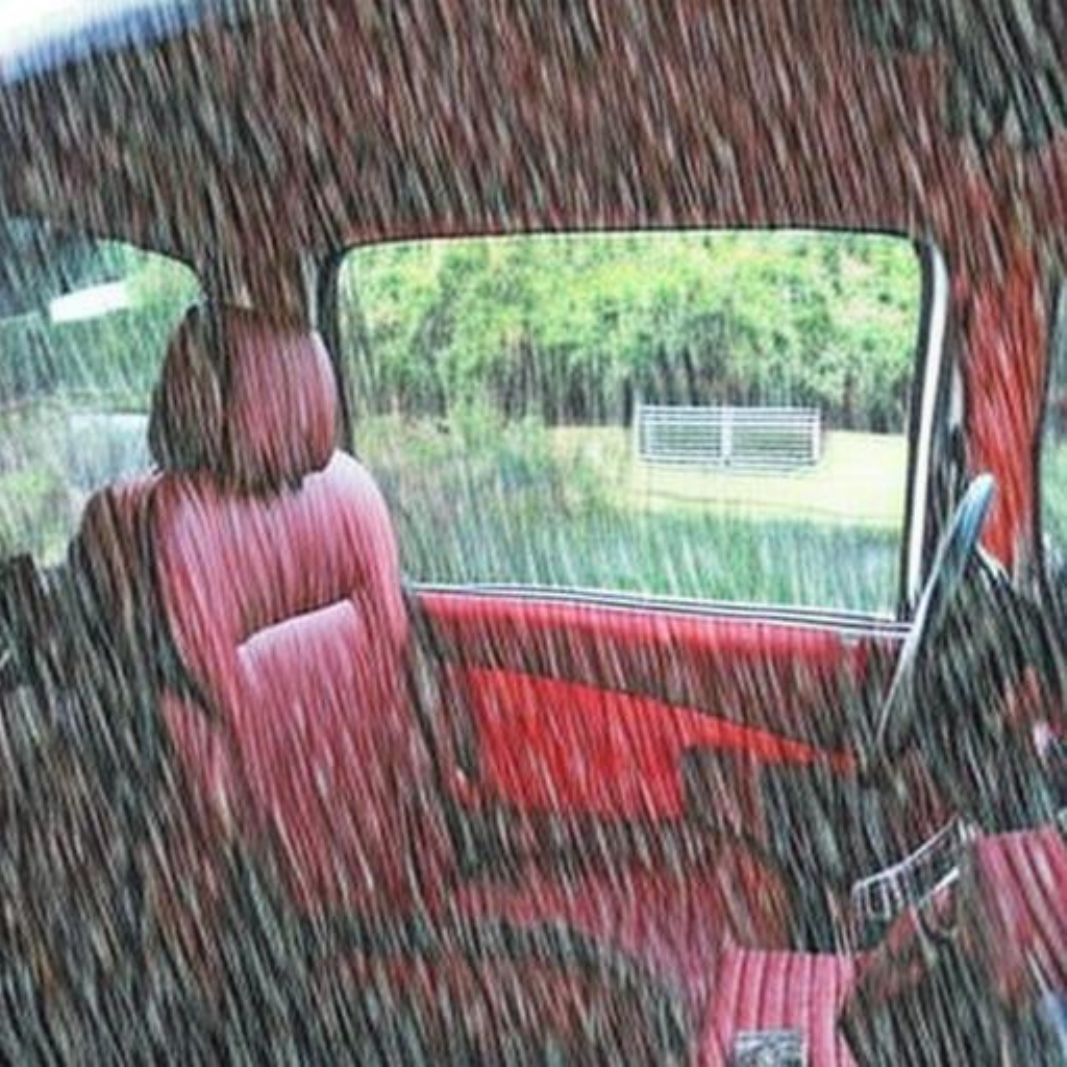}   &   
\includegraphics[width=.18\textwidth,valign=t]{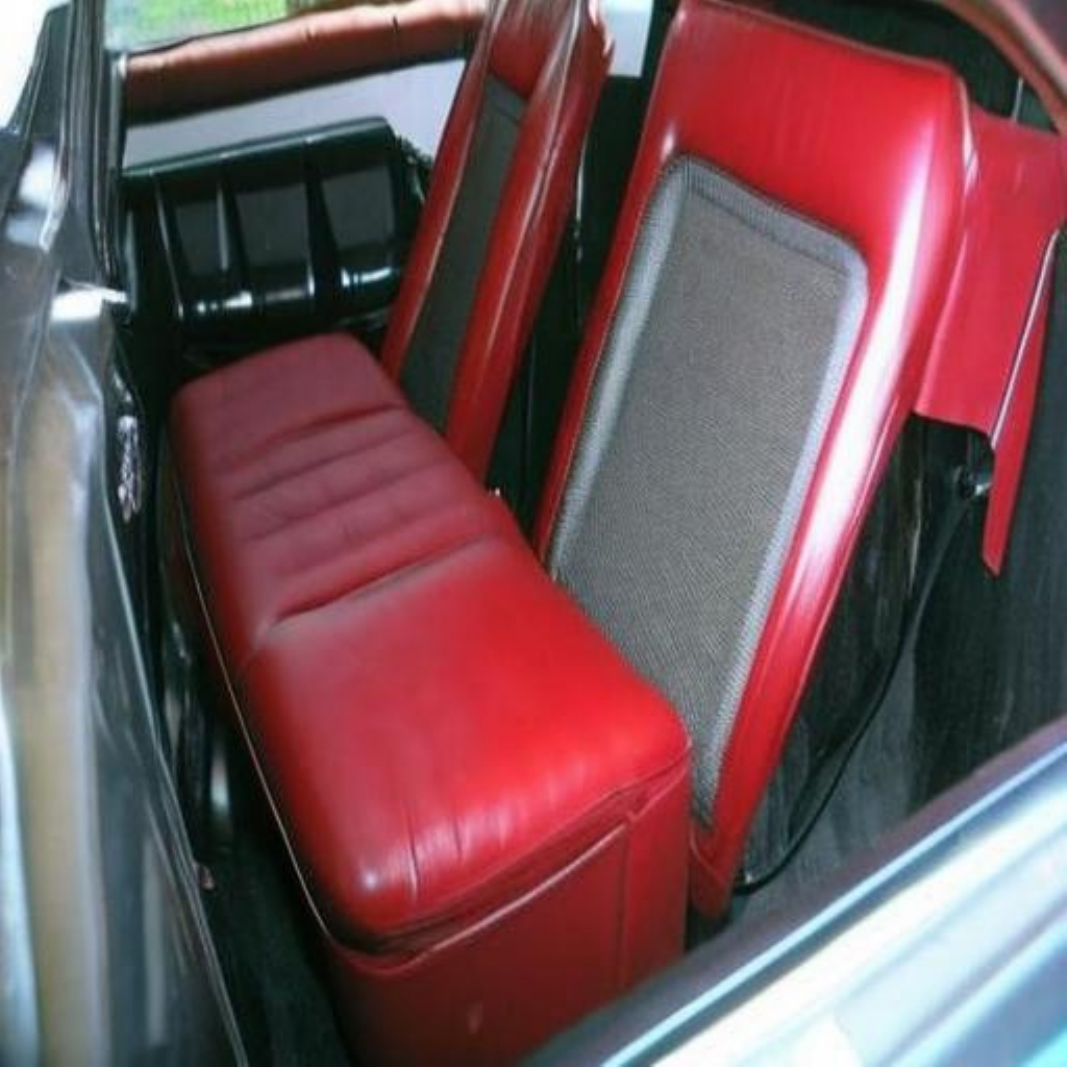}     \\
\vspace{0.2em}
\includegraphics[width=.18\textwidth,valign=t]{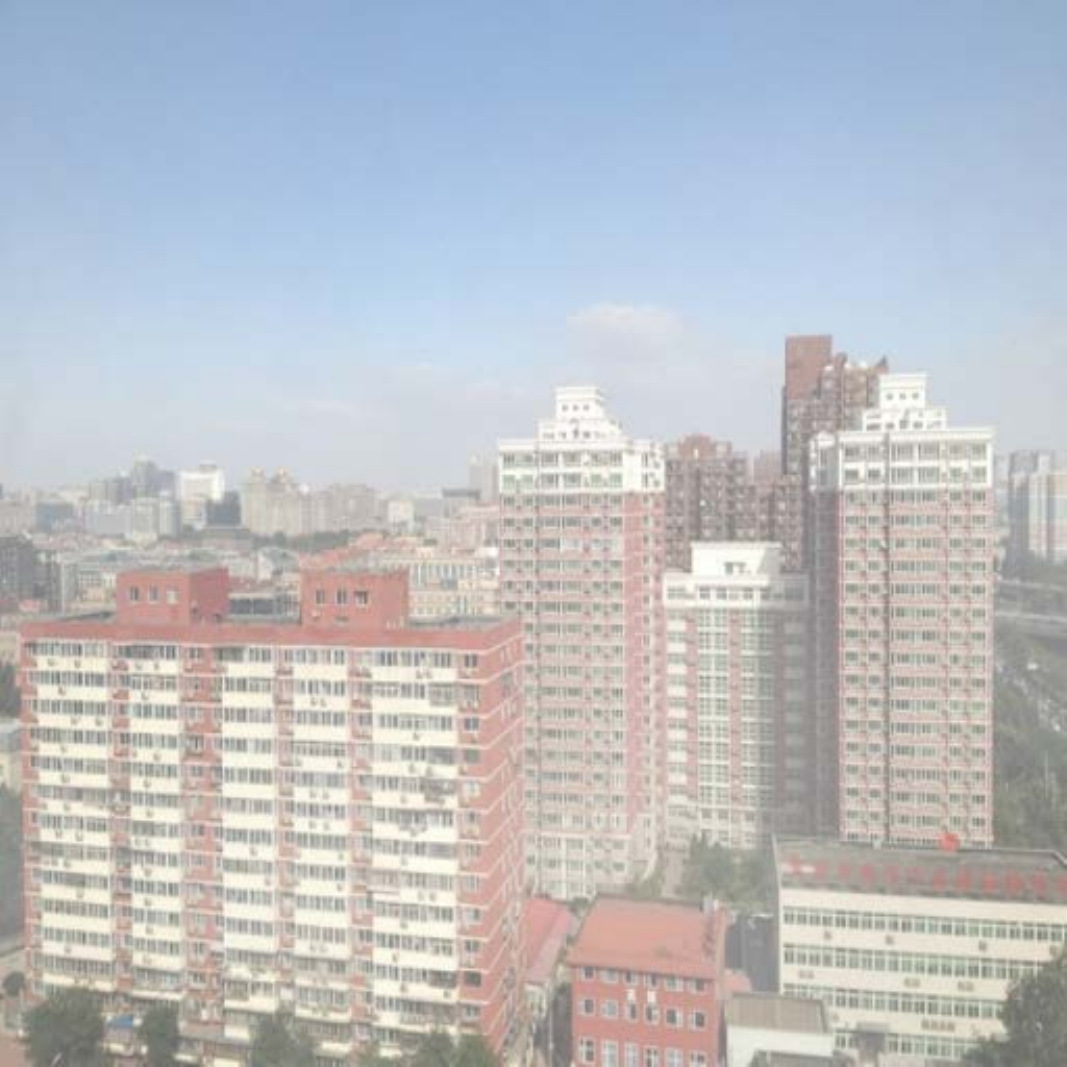} &   
\includegraphics[width=.18\textwidth,valign=t]{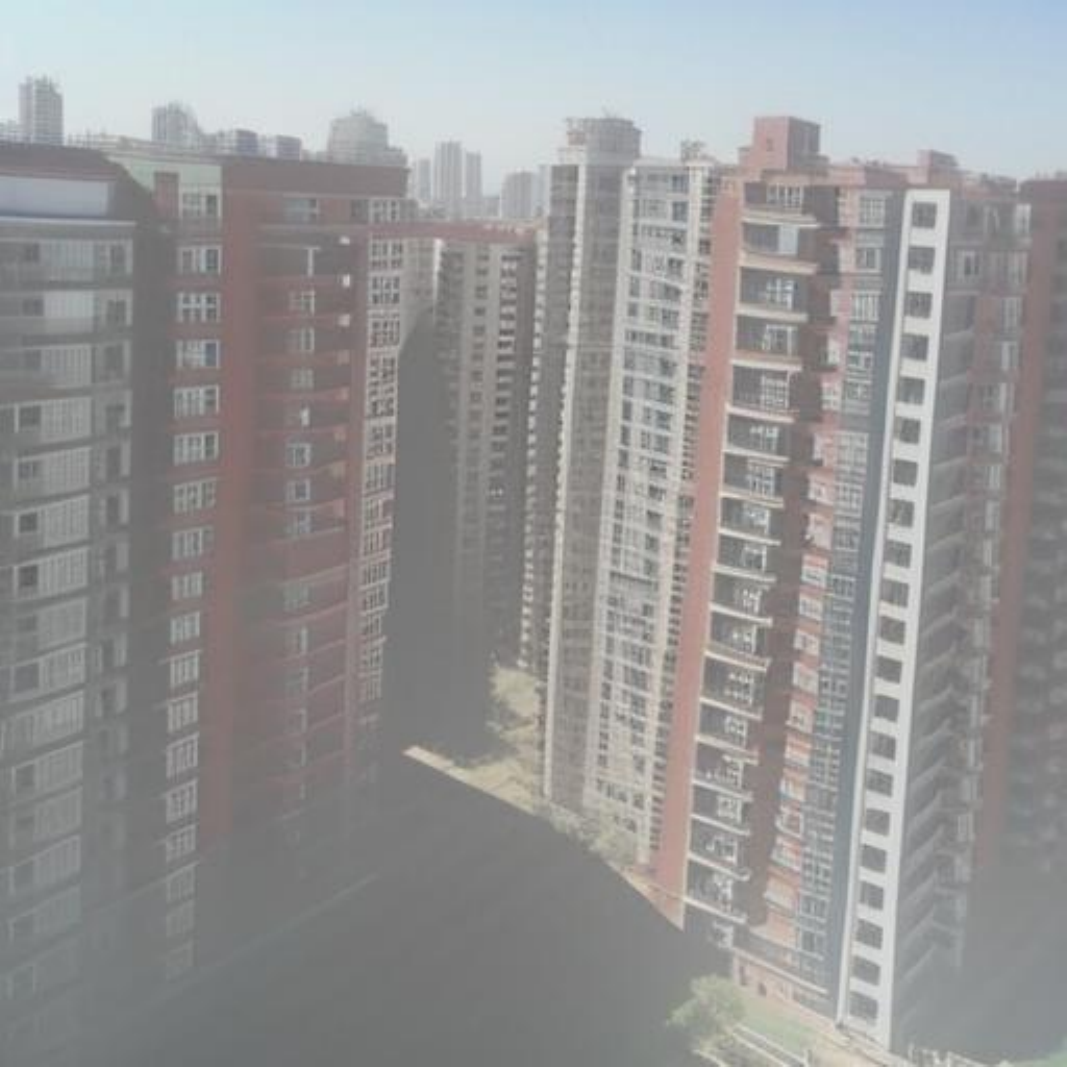} &    
\includegraphics[width=.18\textwidth,valign=t]{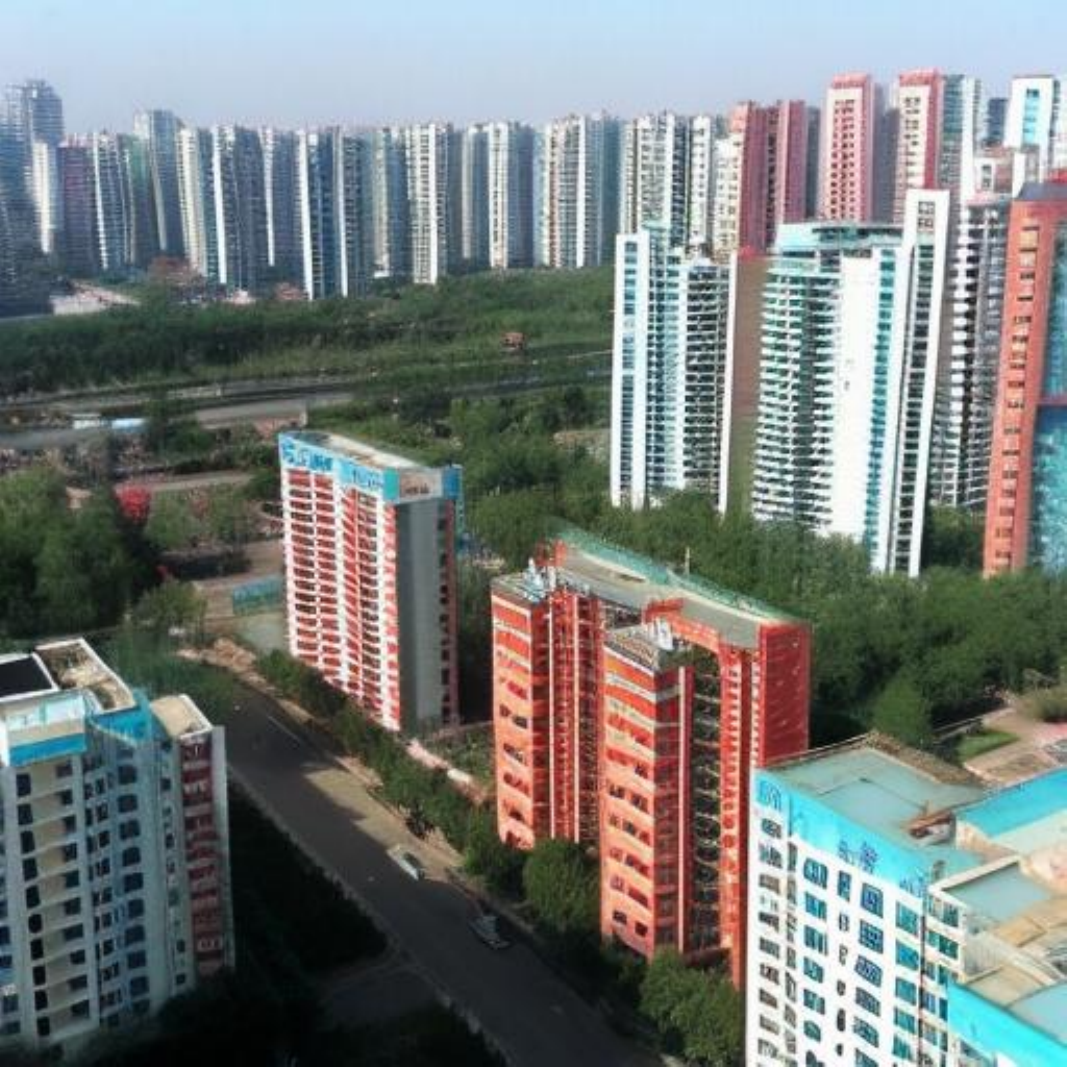}   \\
\vspace{0.2em}
\includegraphics[width=.18\textwidth,valign=t]{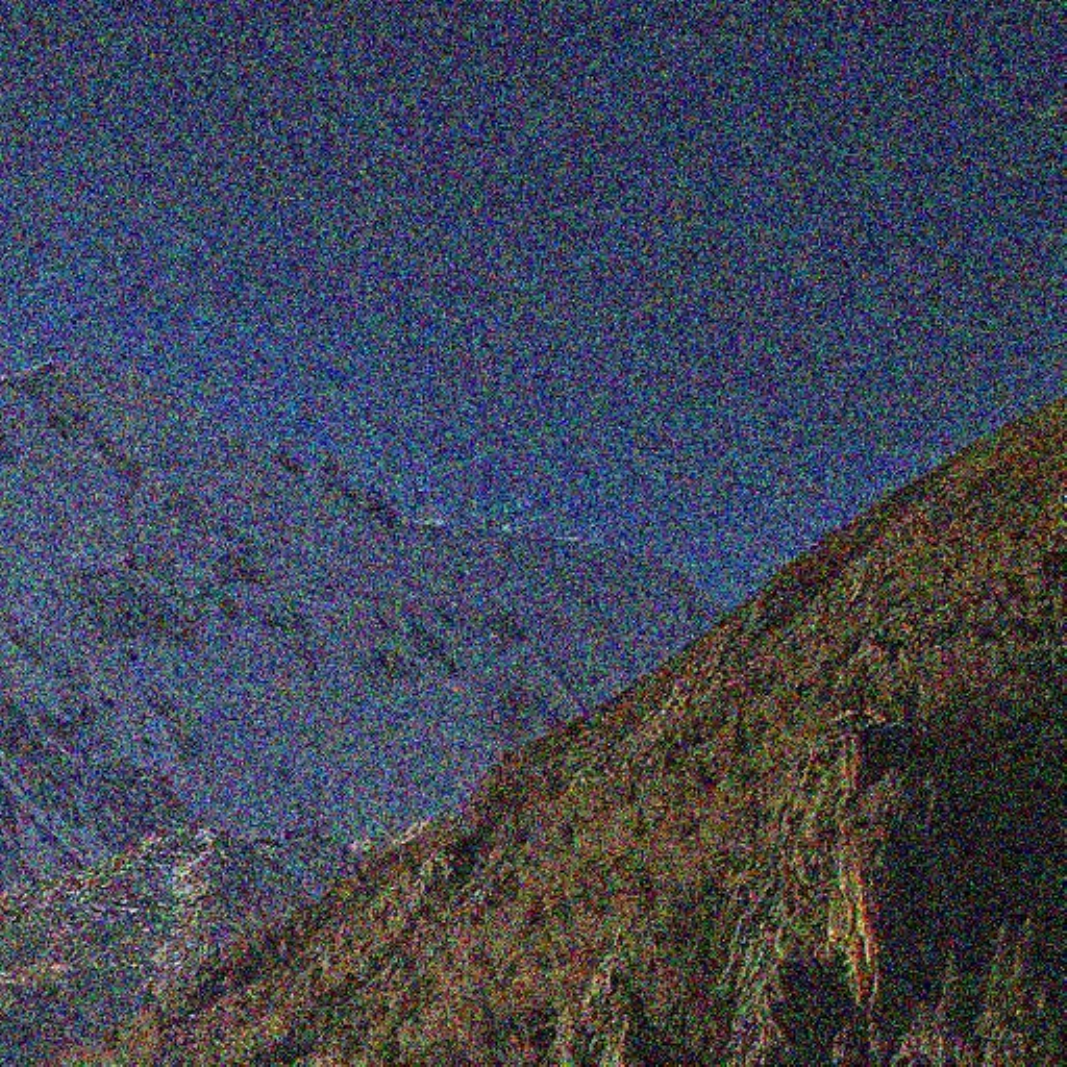} &   
\includegraphics[width=.18\textwidth,valign=t]{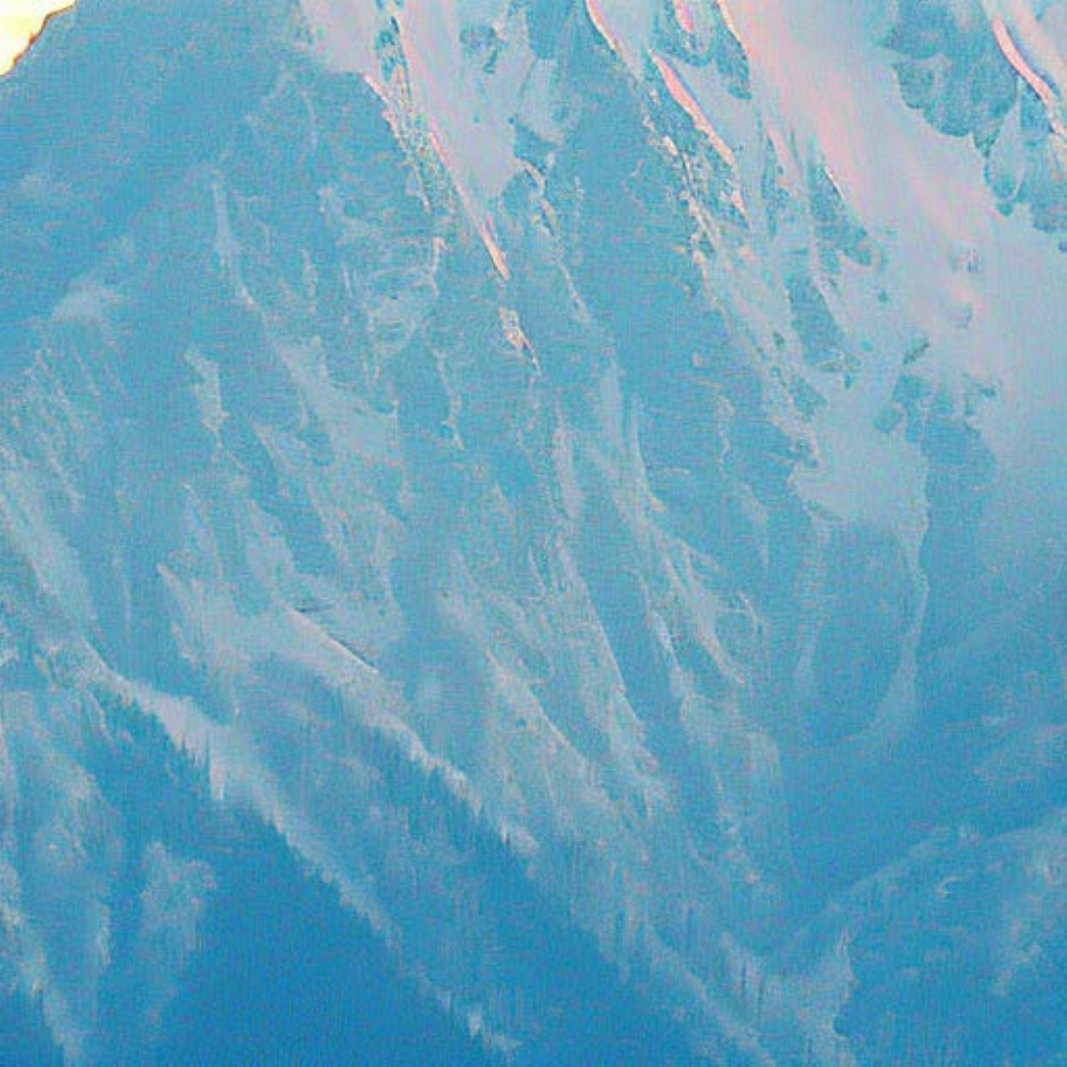} &    
\includegraphics[width=.18\textwidth,valign=t]{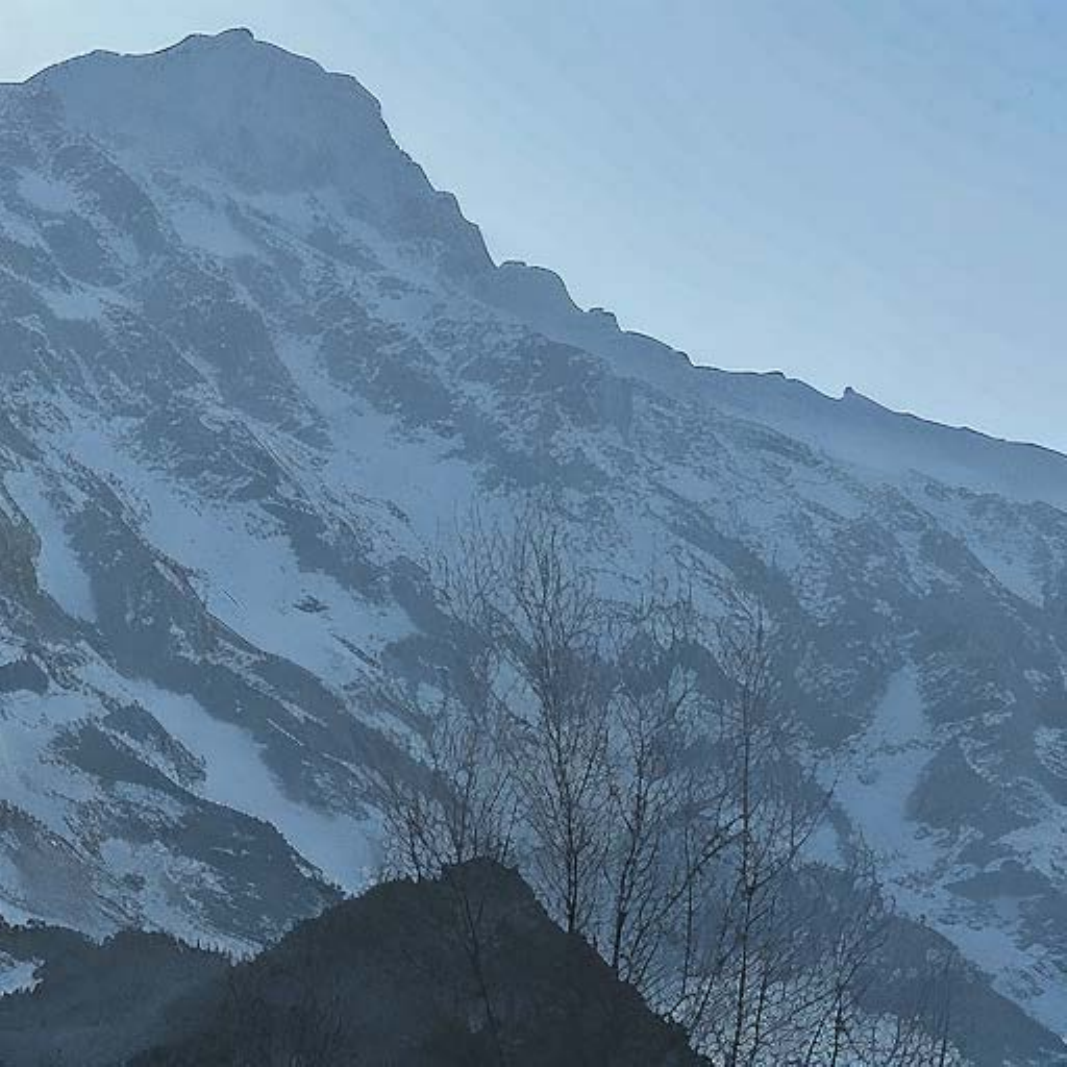}   \\
\vspace{0.2em}
\small~Degraded  & \small~w/o. textual restoration & \small~w/. textual restoration \\
\end{tabular}
}
\vspace*{-6mm}
\caption{\small Visual comparison of w/o. textual restoration and w/. textual restoration.}
\label{fig:suppl_textual_restoration}
\end{center}
\vspace{-2mm}
\end{figure}

%% file: suppl_figures/suppl_013_explicit_implicit_comparison_deblur.tex
\begin{figure*}[t]
    \centering
    \setlength{\abovecaptionskip}{0.1cm}
    \setlength{\belowcaptionskip}{-0.3cm}
    \includegraphics[width=0.97\linewidth]{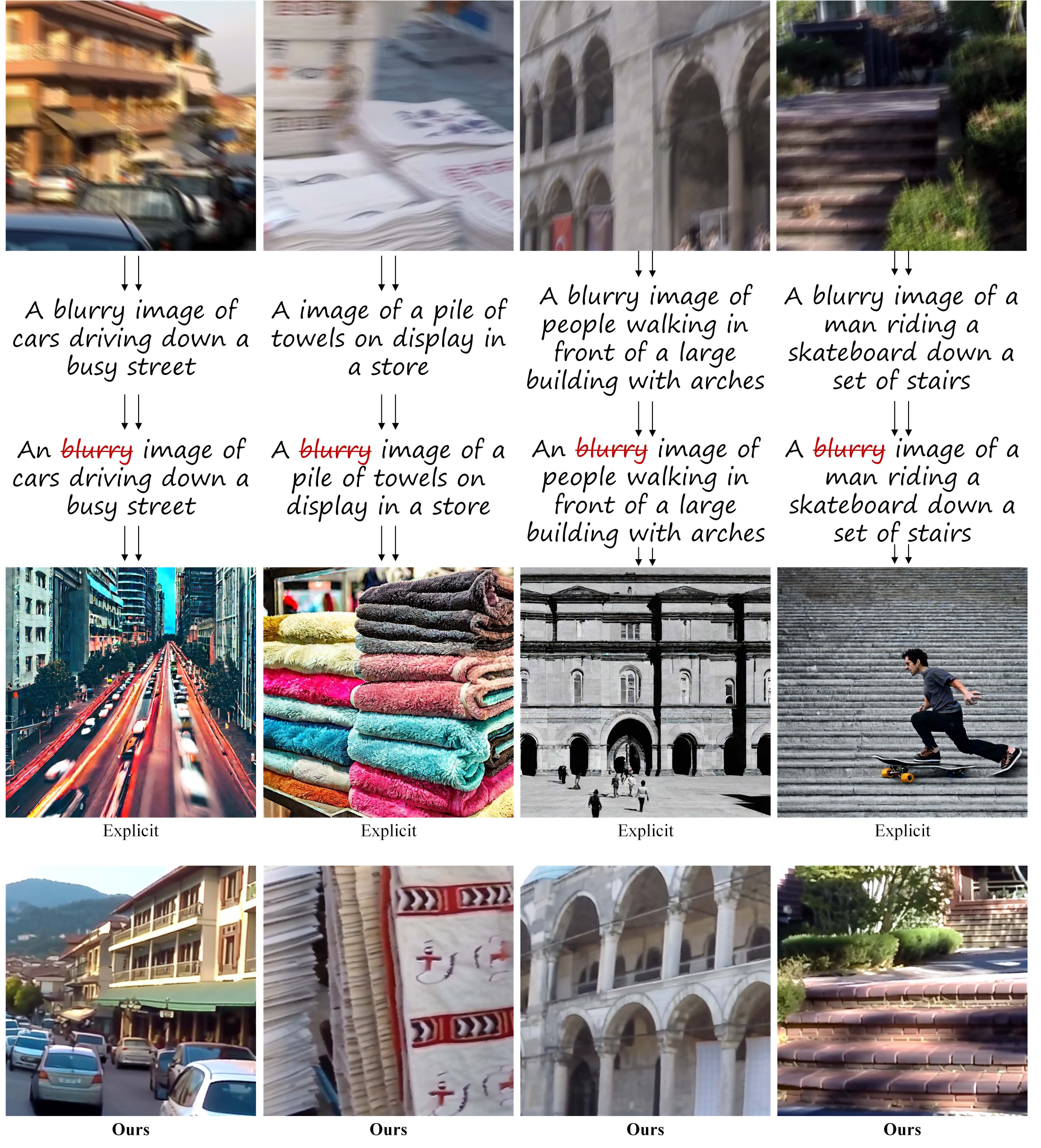}
    \caption{\small Visual comparison of synthetic guidance by explicit and implicit textual representation on image deblurring task.}
    \label{fig:suppl_explicit_implicit_deblur}
    \vspace{-1mm}
\end{figure*}

%% file: suppl_figures/suppl_014_explicit_implicit_comparison_derain.tex
\begin{figure*}[t]
    \centering
    \setlength{\abovecaptionskip}{0.1cm}
    \setlength{\belowcaptionskip}{-0.3cm}
    \includegraphics[width=0.97\linewidth]{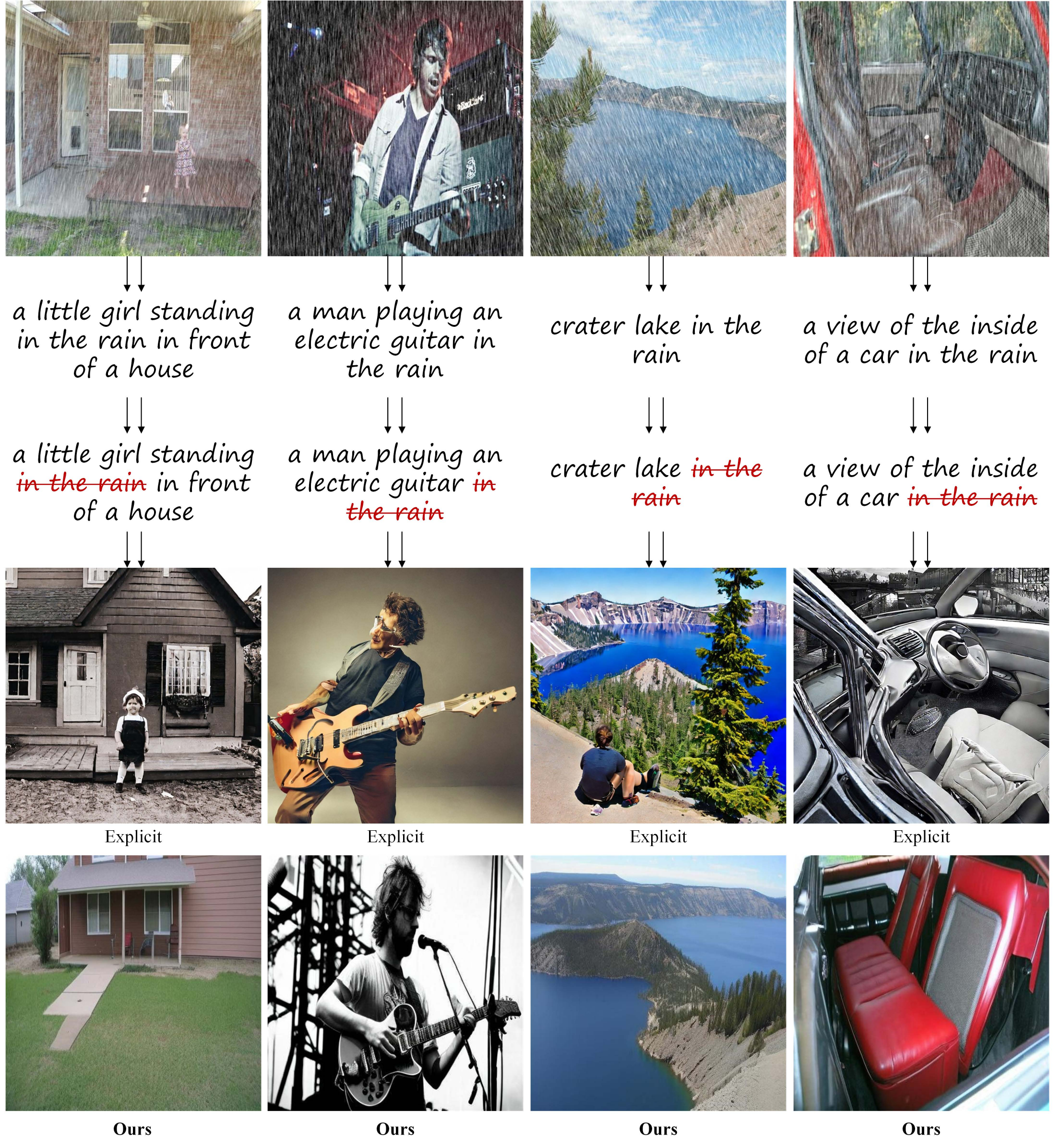}
    \caption{\small Visual comparison of synthetic guidance by explicit and implicit textual representation on image deraining task.}
    \label{fig:suppl_explicit_implicit_derain}
    \vspace{-1mm}
\end{figure*}

%% file: suppl_figures/suppl_015_explicit_implicit_comparison_dehaze.tex
\begin{figure*}[t]
    \centering
    \setlength{\abovecaptionskip}{0.1cm}
    \setlength{\belowcaptionskip}{-0.3cm}
    \includegraphics[width=0.97\linewidth]{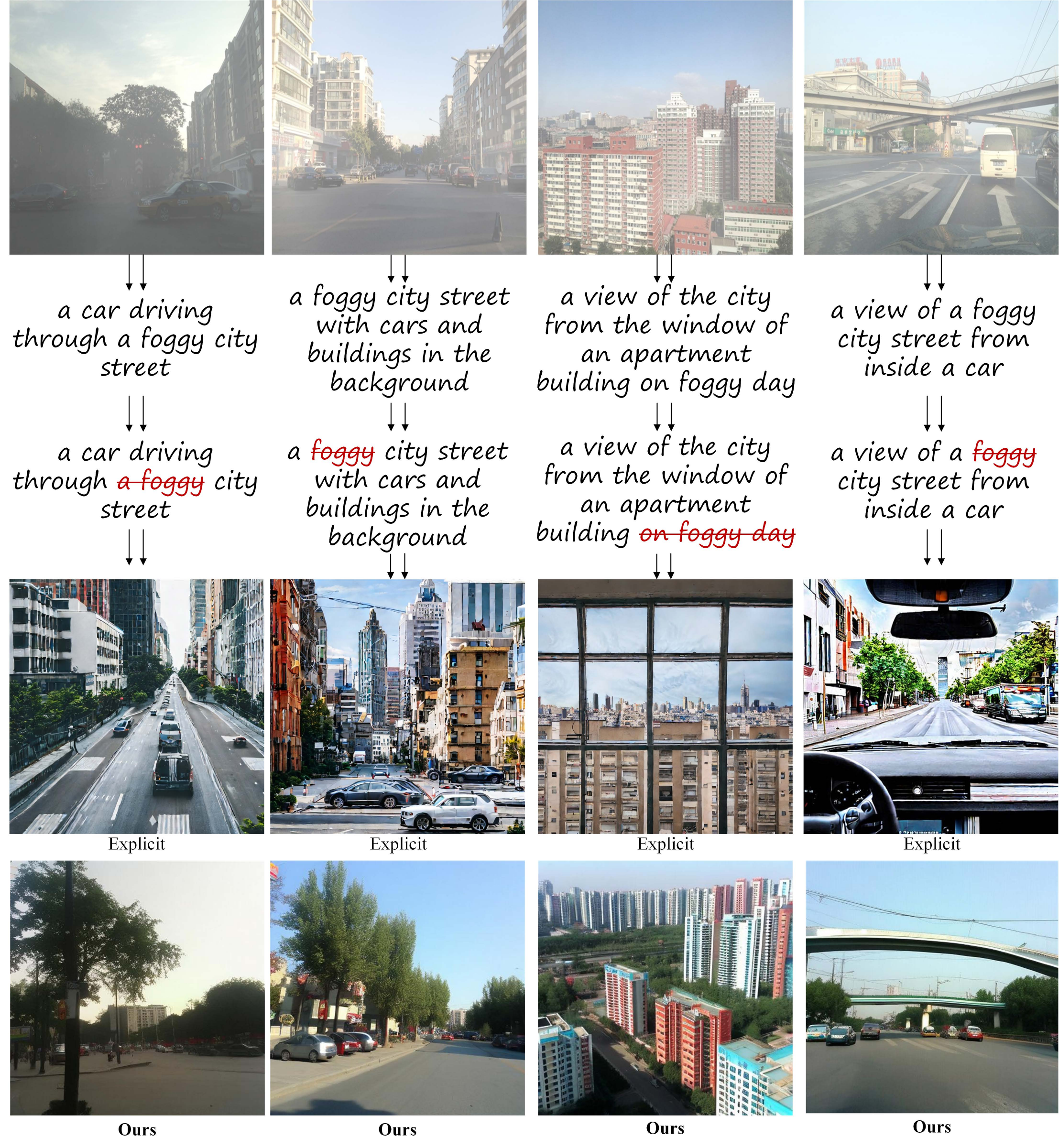}
    \caption{\small Visual comparison of synthetic guidance by explicit and implicit textual representation on image dehazing task.}
    \label{fig:suppl_explicit_implicit_dehaze}
    \vspace{-1mm}
\end{figure*}

%% file: suppl_figures/suppl_016_explicit_implicit_comparison_denoise.tex
\begin{figure*}[t]
    \centering
    \setlength{\abovecaptionskip}{0.1cm}
    \setlength{\belowcaptionskip}{-0.3cm}
    \includegraphics[width=0.97\linewidth]{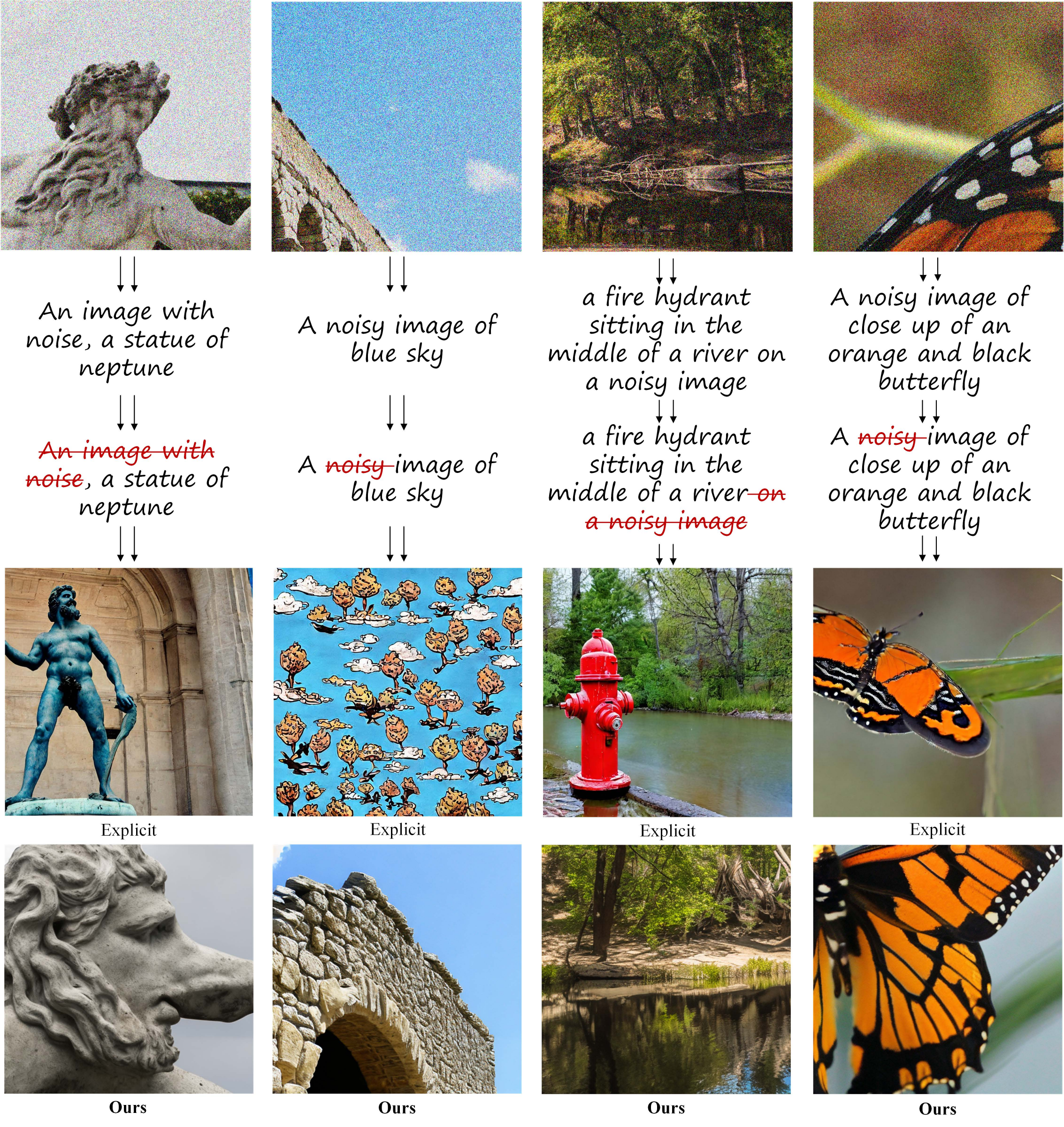}
    \caption{\small Visual comparison of synthetic guidance by explicit and implicit textual representation on image denoising task.}
    \label{fig:suppl_explicit_implicit_denoise}
    \vspace{-1mm}
\end{figure*}

%% file: suppl_figures/suppl_009_guidance_images_deblur.tex
\begin{figure*}[t]
\begin{center}
\setlength{\tabcolsep}{1.5pt}
\scalebox{0.88}{
\begin{tabular}[b]{c c c c}
\includegraphics[width=.25\textwidth,valign=t]{suppl_figures/guidance_images/GOPR0372_07_00-000047-1_s001_input.pdf}  &   
\includegraphics[width=.25\textwidth,valign=t]{suppl_figures/guidance_images/GOPR0372_07_00-000047-1_s001_guidance.pdf} & 
\includegraphics[width=.25\textwidth,valign=t]{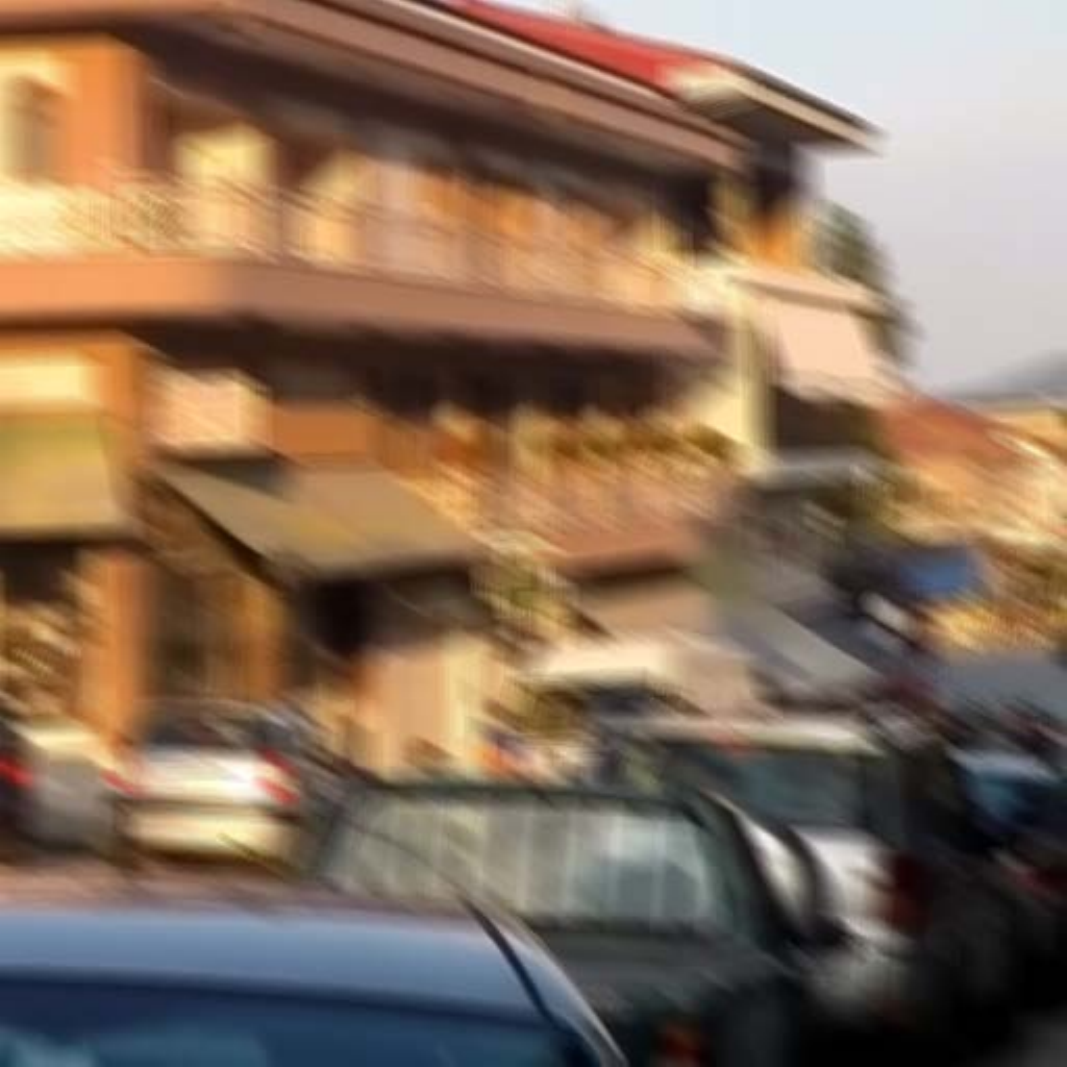} &   
\includegraphics[width=.25\textwidth,valign=t]{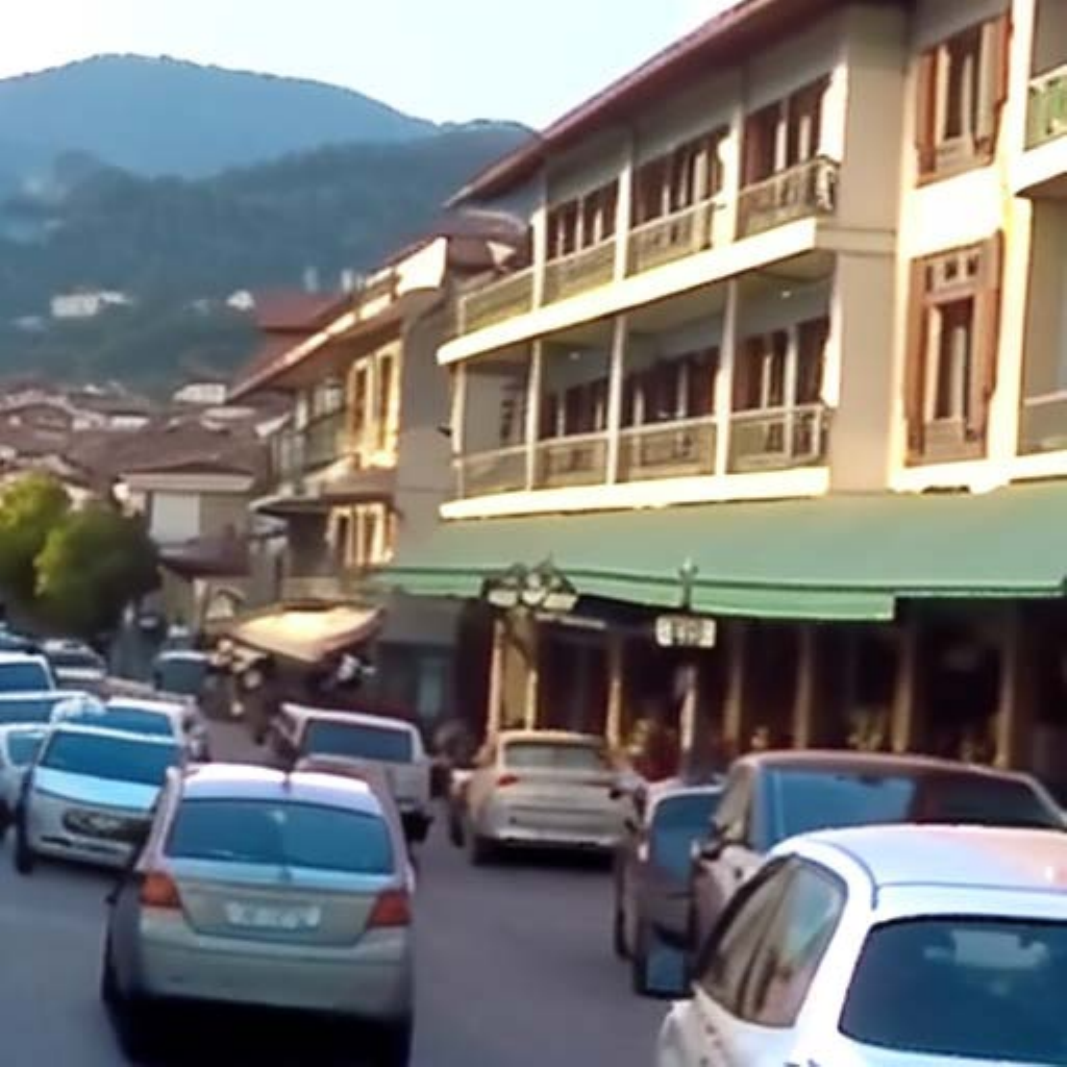} \\
\vspace{0.6em}
\small~Degraded  & \small~Guidance & \small~Degraded & \small~Guidance \\

\includegraphics[width=.25\textwidth,valign=t]{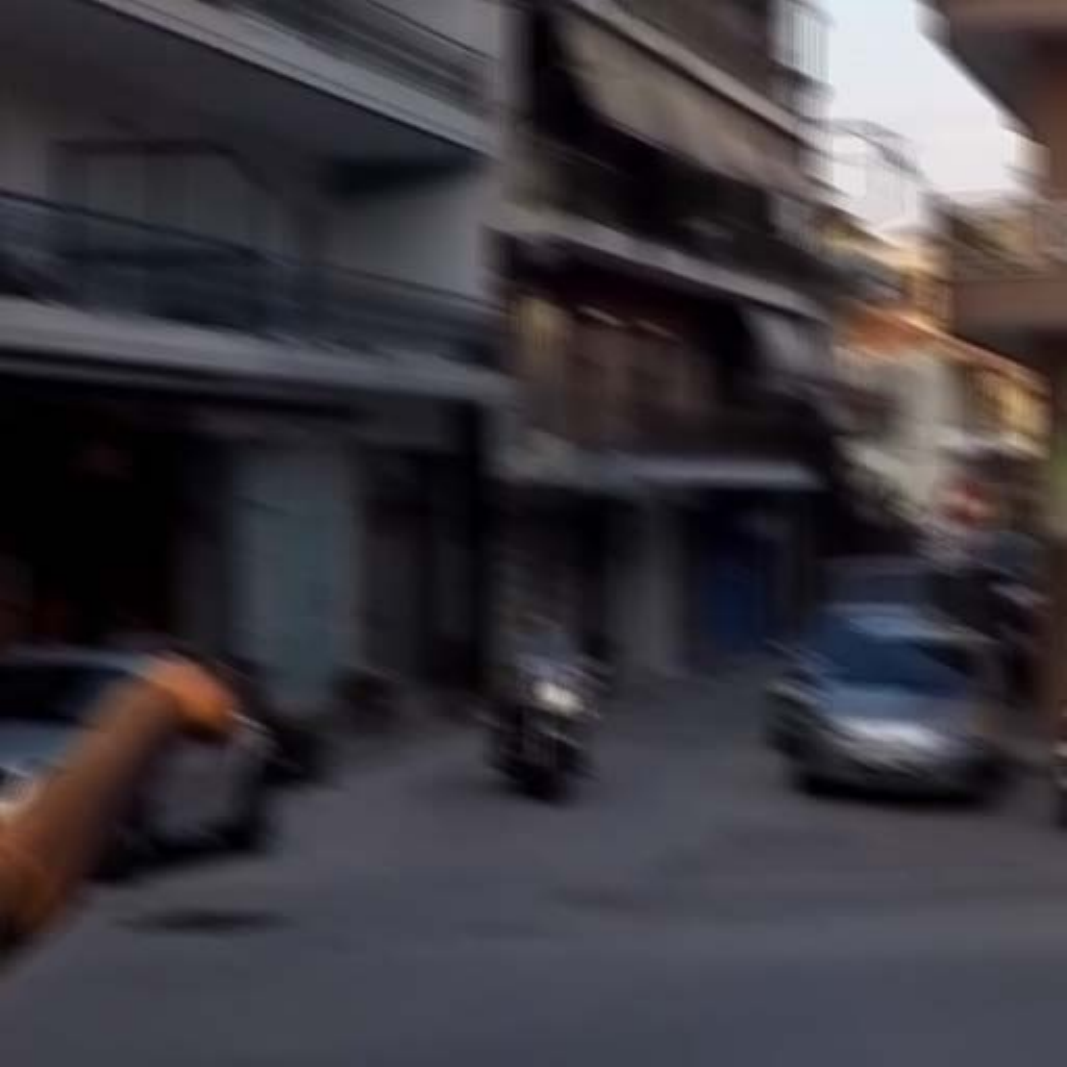}  &   
\includegraphics[width=.25\textwidth,valign=t]{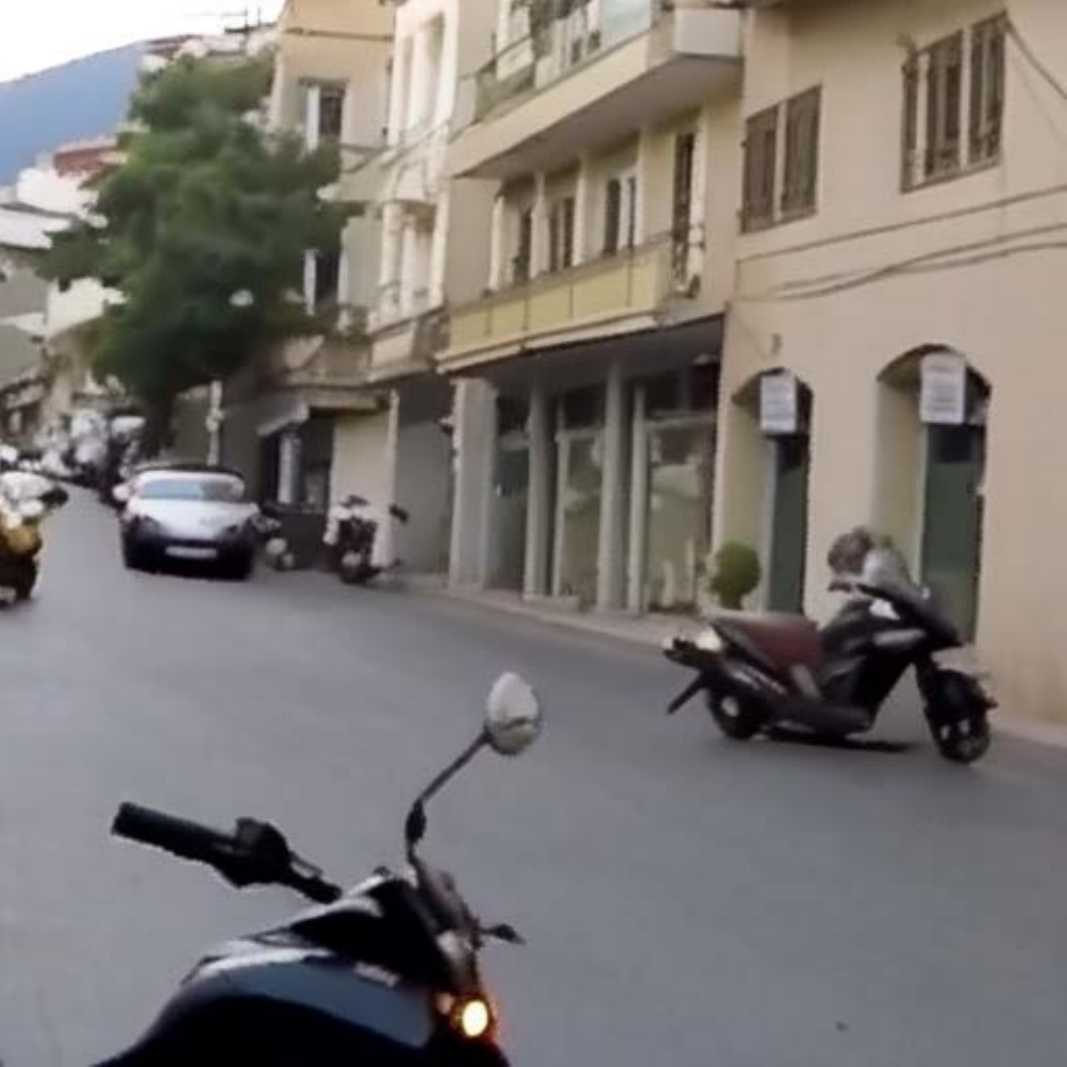} & 
\includegraphics[width=.25\textwidth,valign=t]{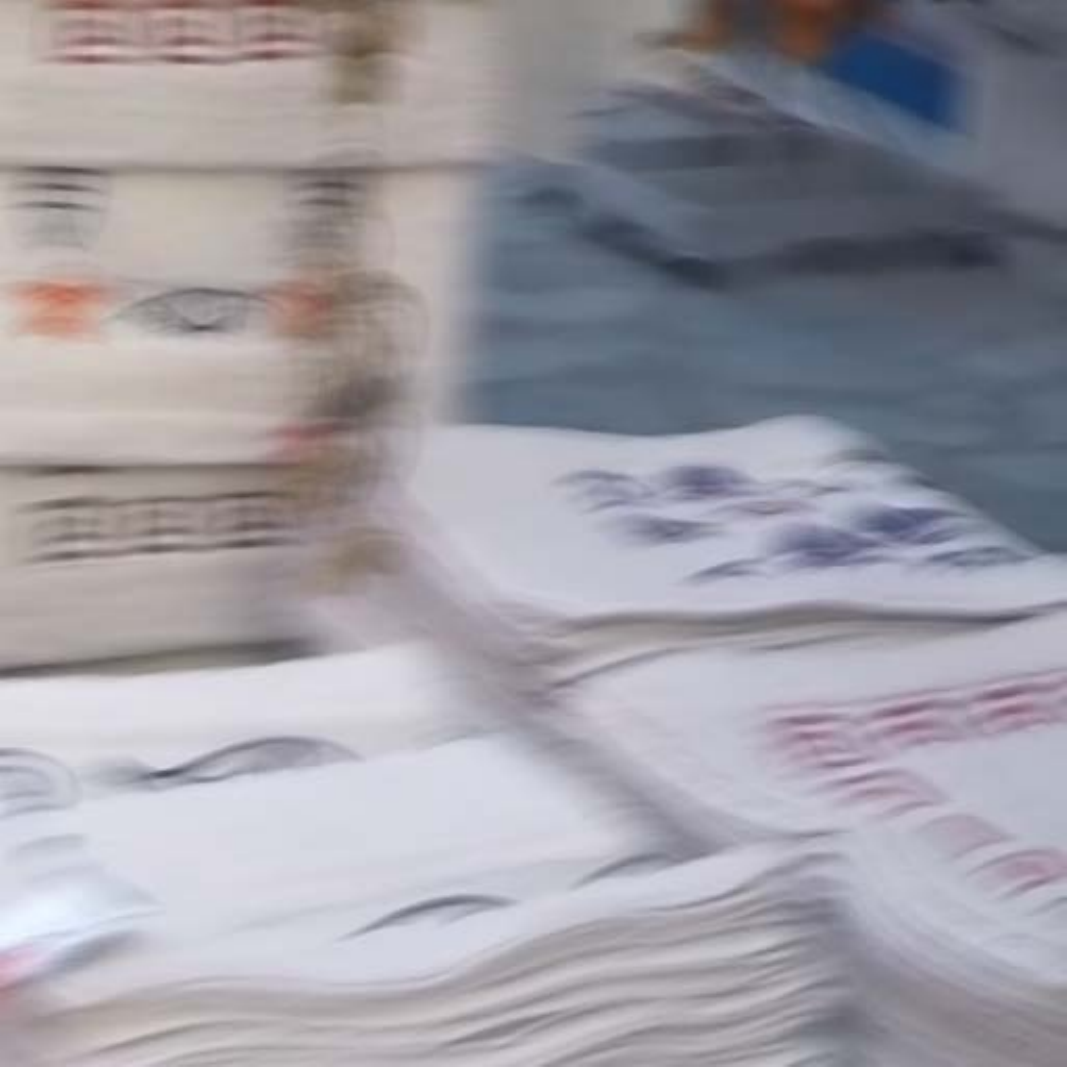} &   
\includegraphics[width=.25\textwidth,valign=t]{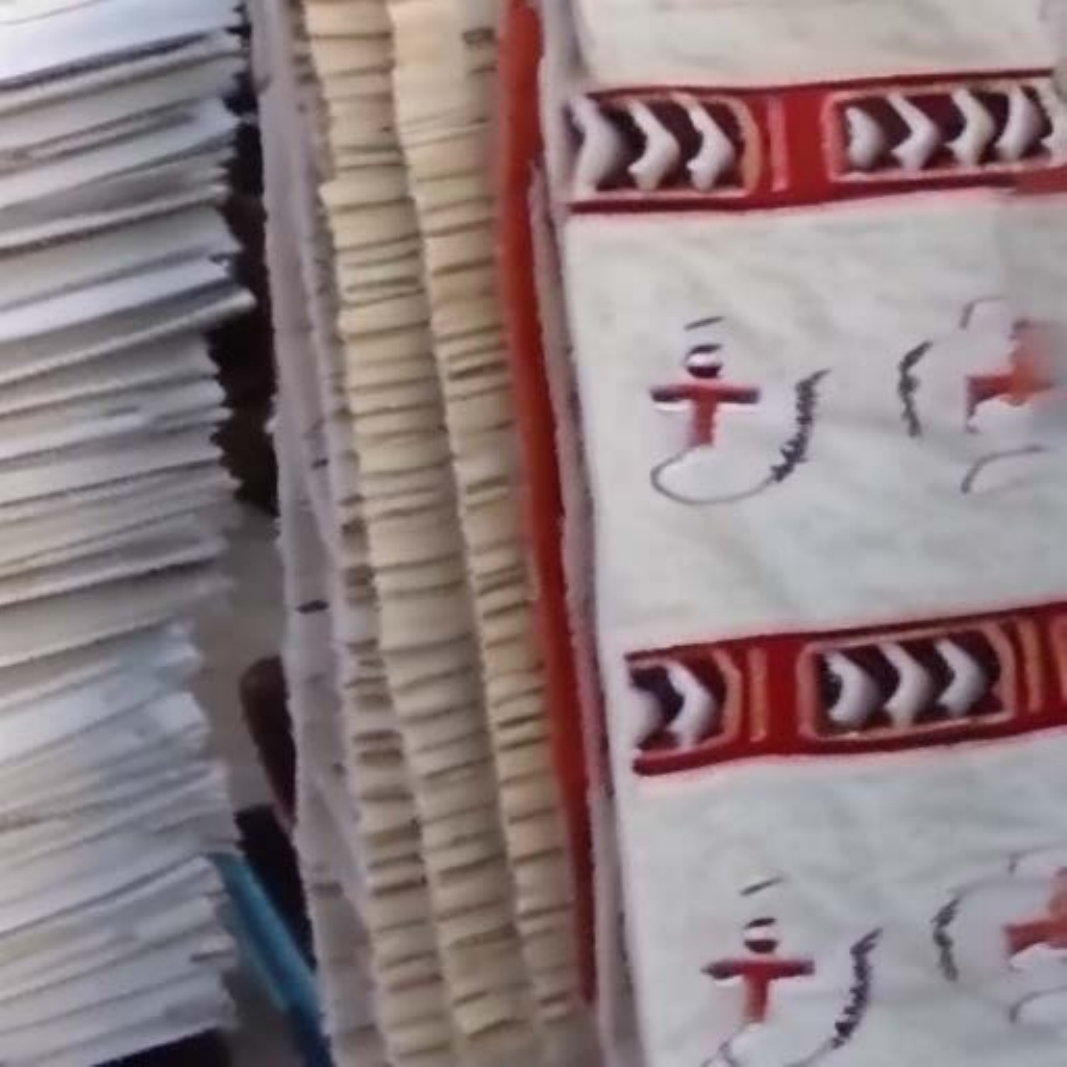} \\
\vspace{0.6em}
\small~Degraded  & \small~Guidance & \small~Degraded & \small~Guidance \\

\includegraphics[width=.25\textwidth,valign=t]{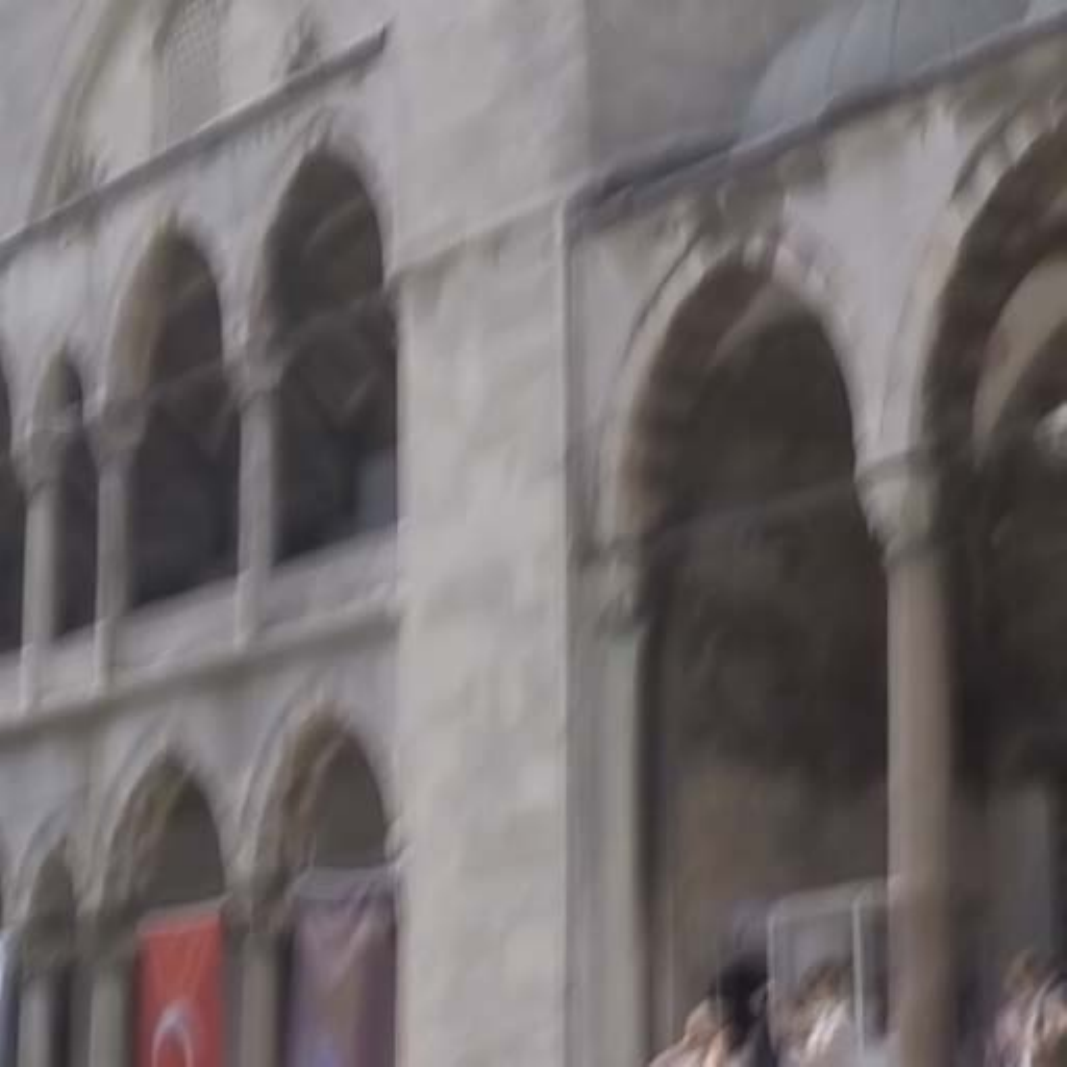}  &   
\includegraphics[width=.25\textwidth,valign=t]{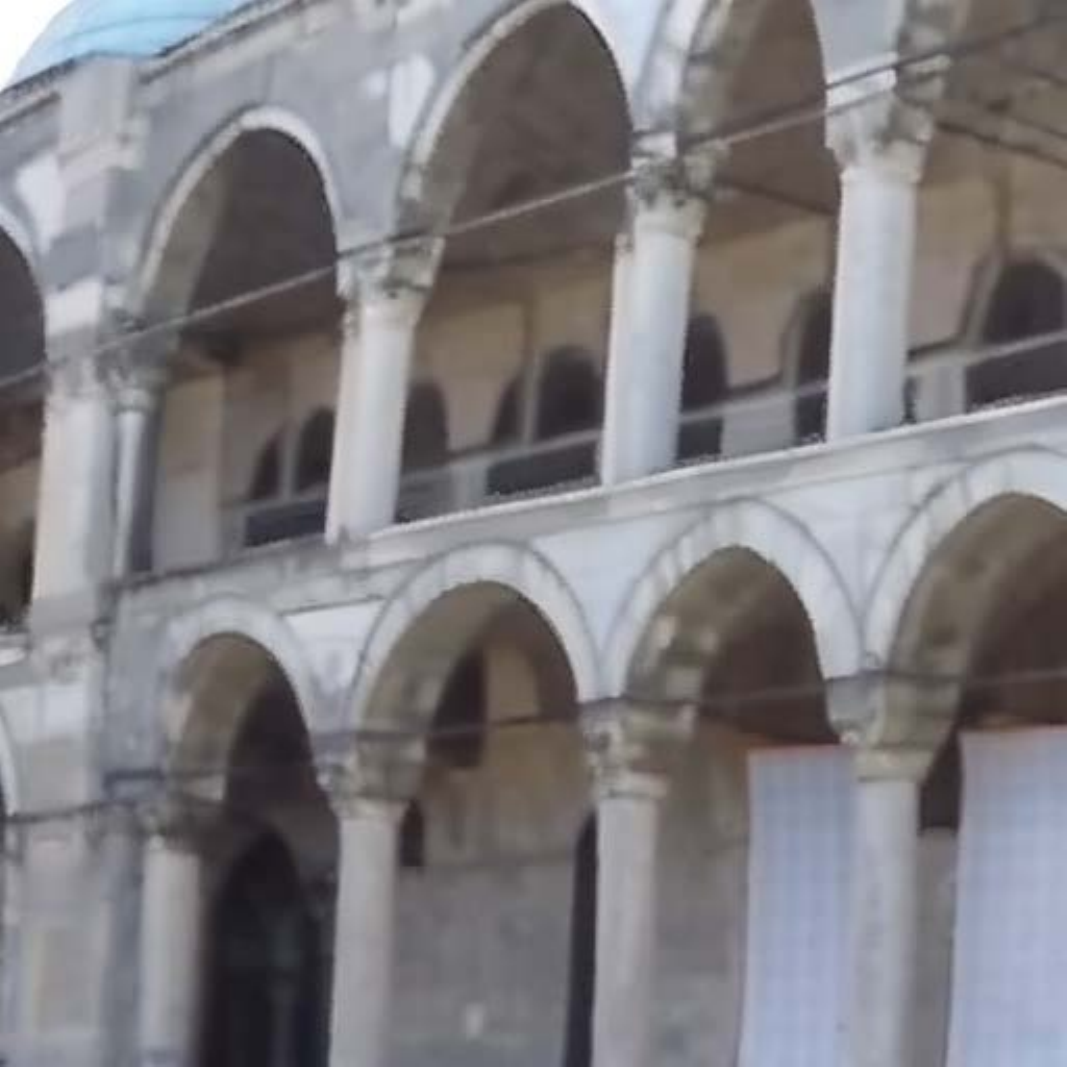} & 
\includegraphics[width=.25\textwidth,valign=t]{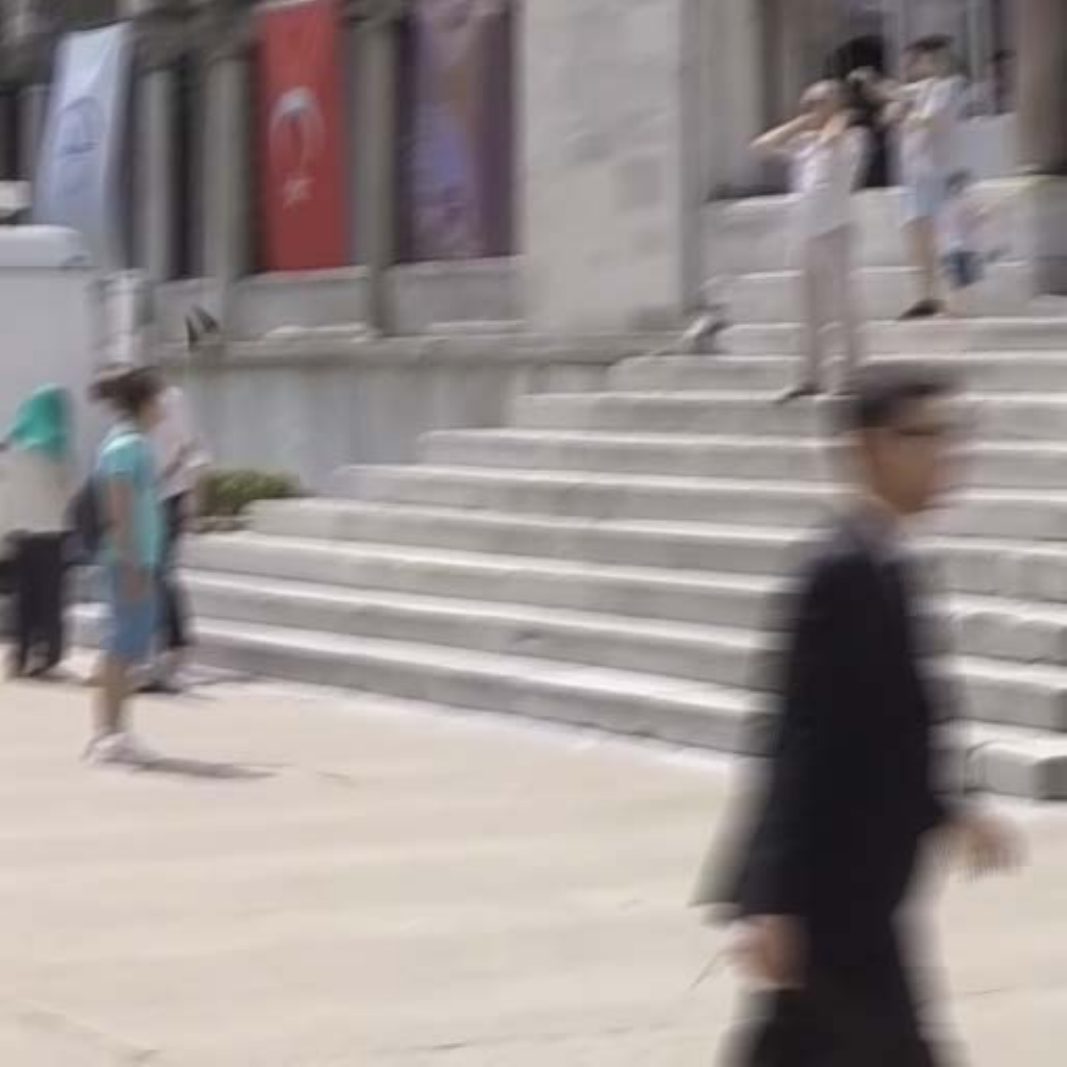} &   
\includegraphics[width=.25\textwidth,valign=t]{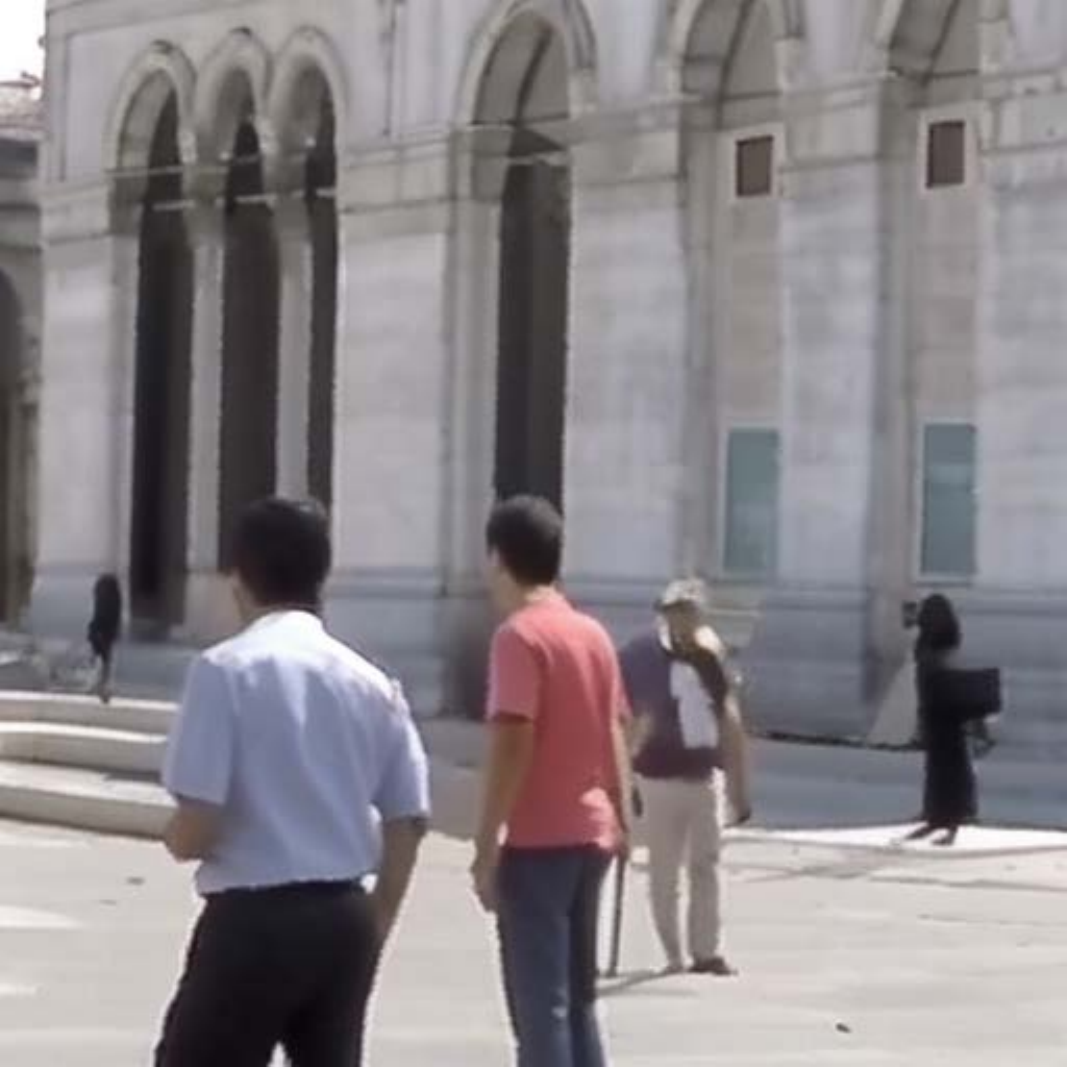} \\
\vspace{0.6em}
\small~Degraded  & \small~Guidance & \small~Degraded & \small~Guidance \\
\end{tabular}
}
\caption{\small Illustration of guidance images for image deblurring task.}
\label{fig:suppl_guidance_images_deblur}
\end{center}
\vspace{-7mm}
\end{figure*}

%% file: suppl_figures/suppl_010_guidance_images_derain.tex
\begin{figure*}[t]
\begin{center}
\setlength{\tabcolsep}{1.5pt}
\scalebox{0.88}{
\begin{tabular}[b]{c c c c}
\includegraphics[width=.25\textwidth,valign=t]{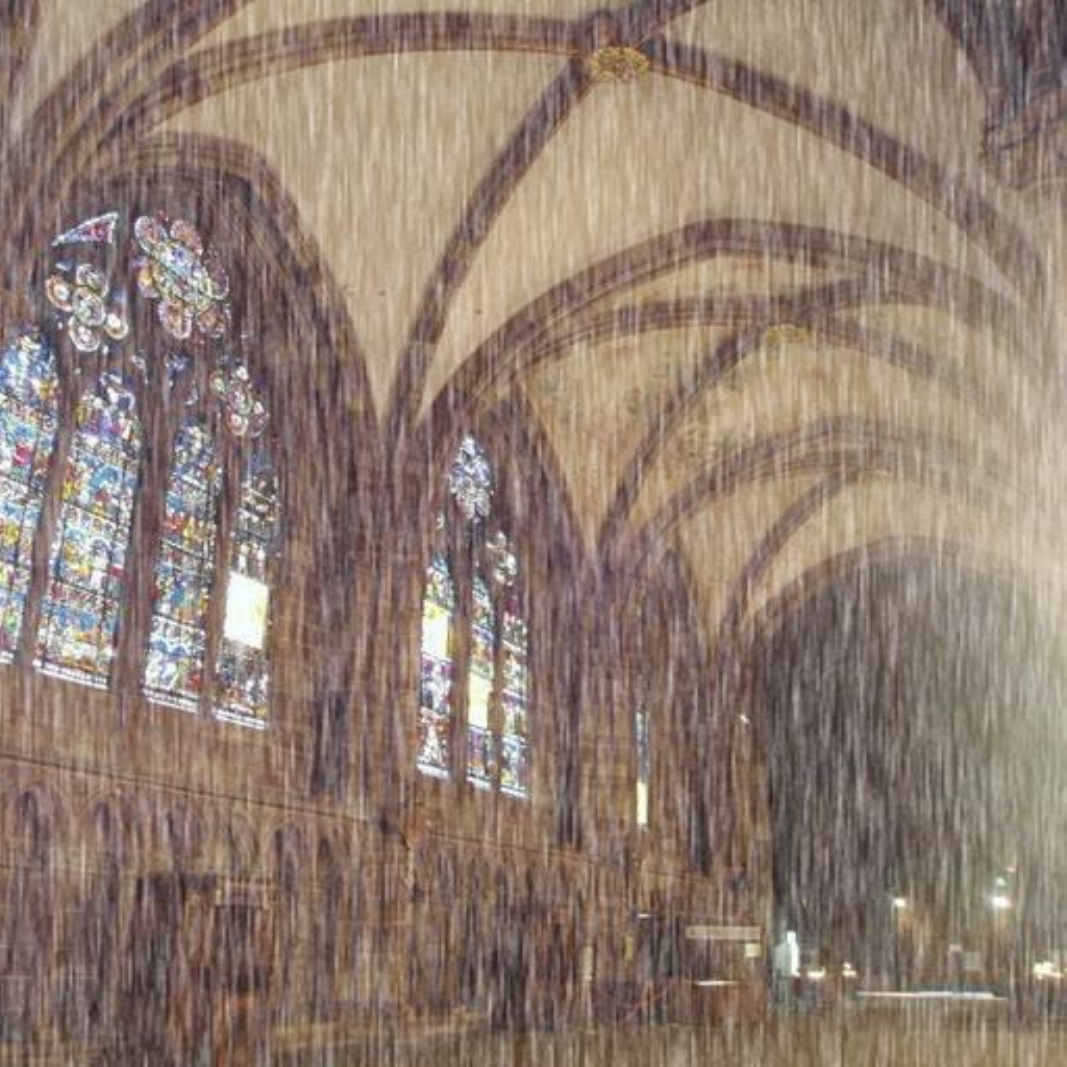}  &   
\includegraphics[width=.25\textwidth,valign=t]{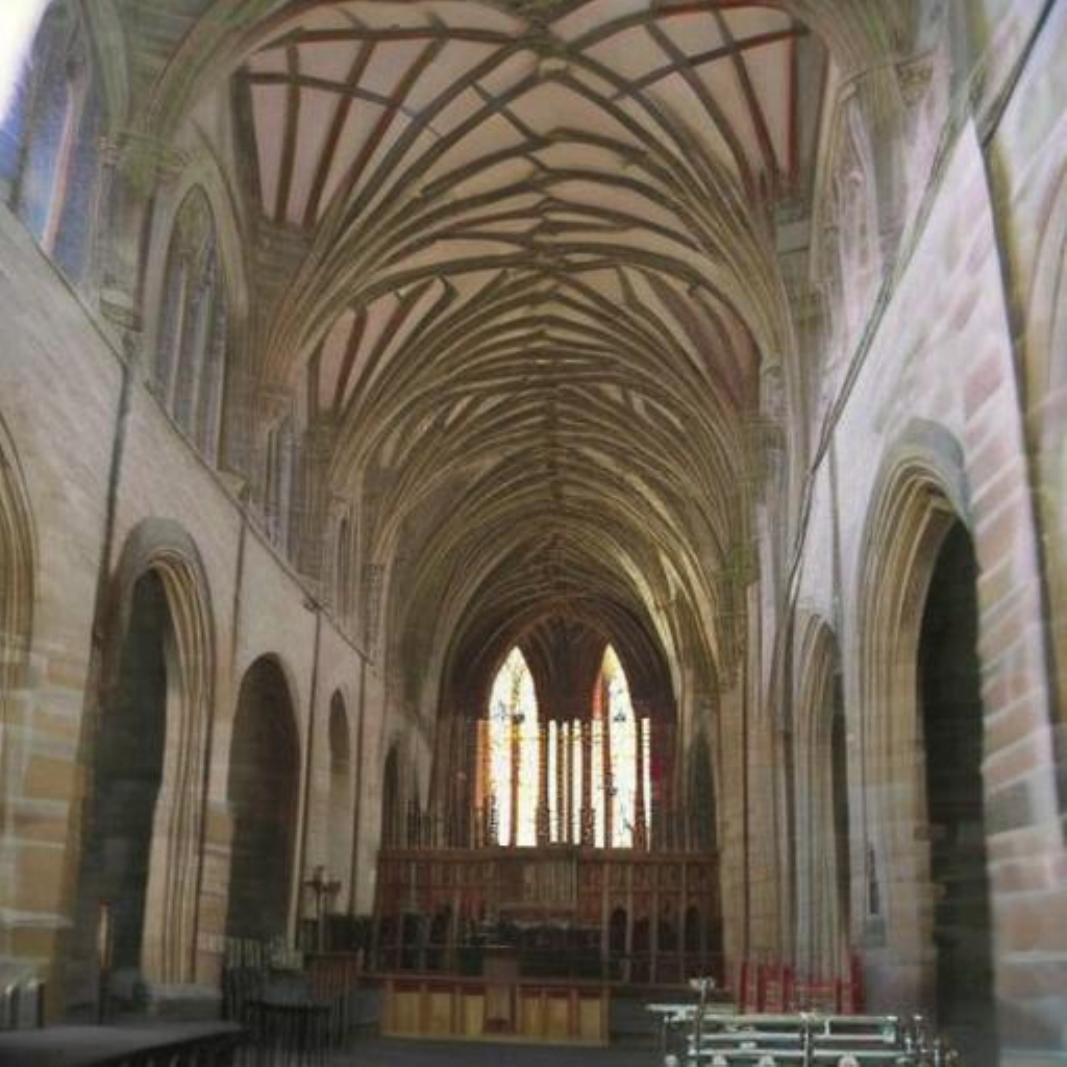} & 
\includegraphics[width=.25\textwidth,valign=t]{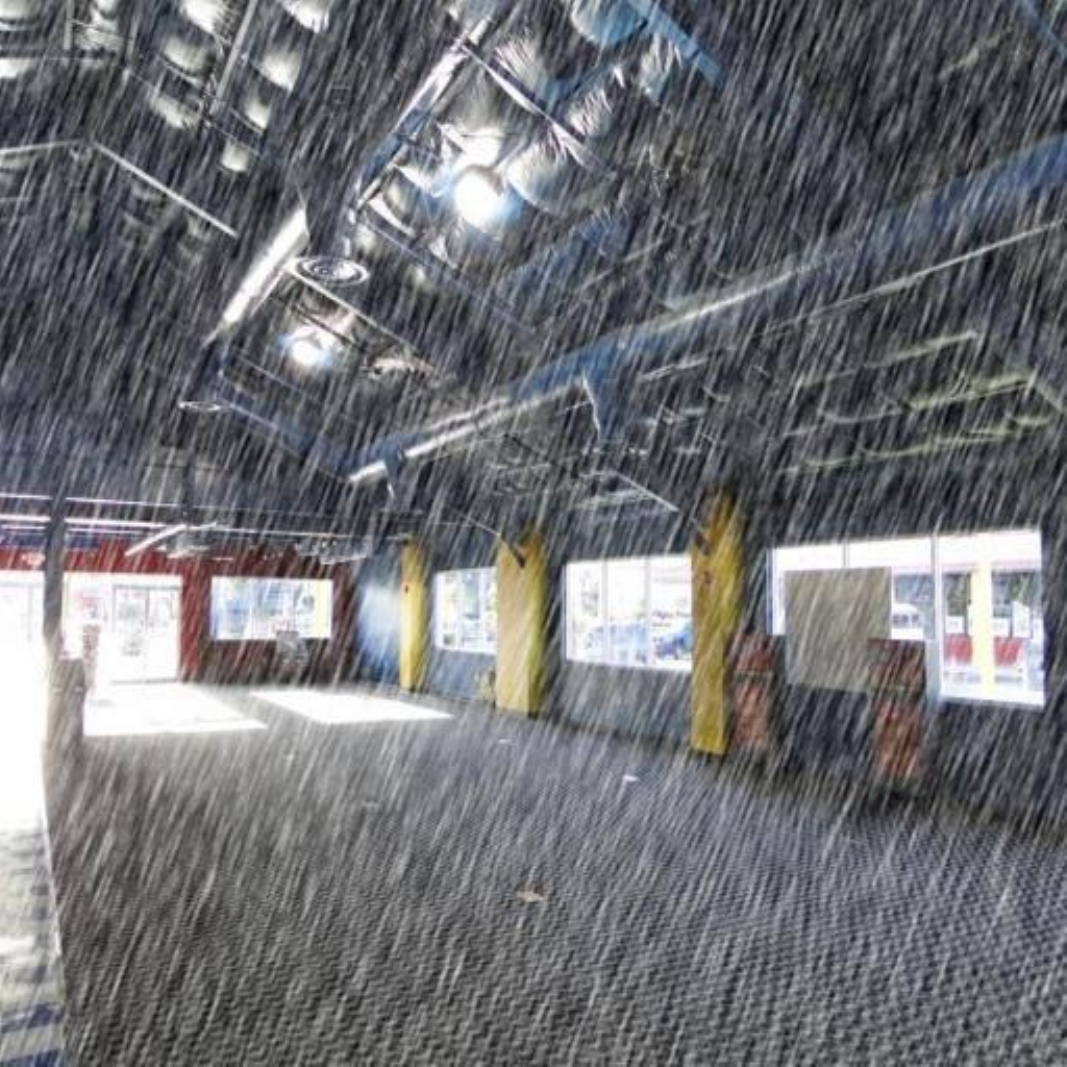} &   
\includegraphics[width=.25\textwidth,valign=t]{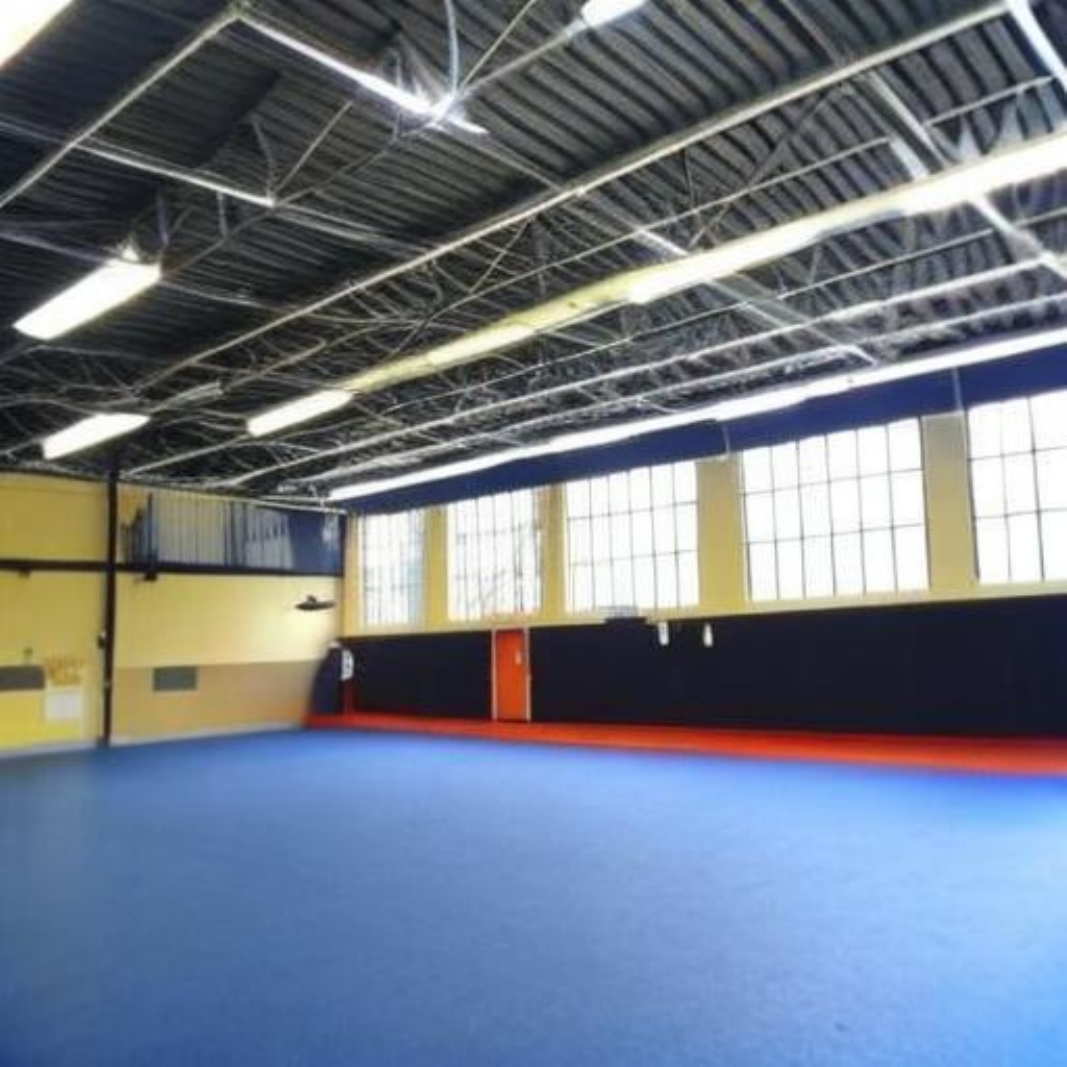} \\
\vspace{0.6em}
\small~Degraded  & \small~Guidance & \small~Degraded & \small~Guidance \\

\includegraphics[width=.25\textwidth,valign=t]{suppl_figures/guidance_images/10003_input.pdf}  &   
\includegraphics[width=.25\textwidth,valign=t]{suppl_figures/guidance_images/10003_guidance.pdf} & 
\includegraphics[width=.25\textwidth,valign=t]{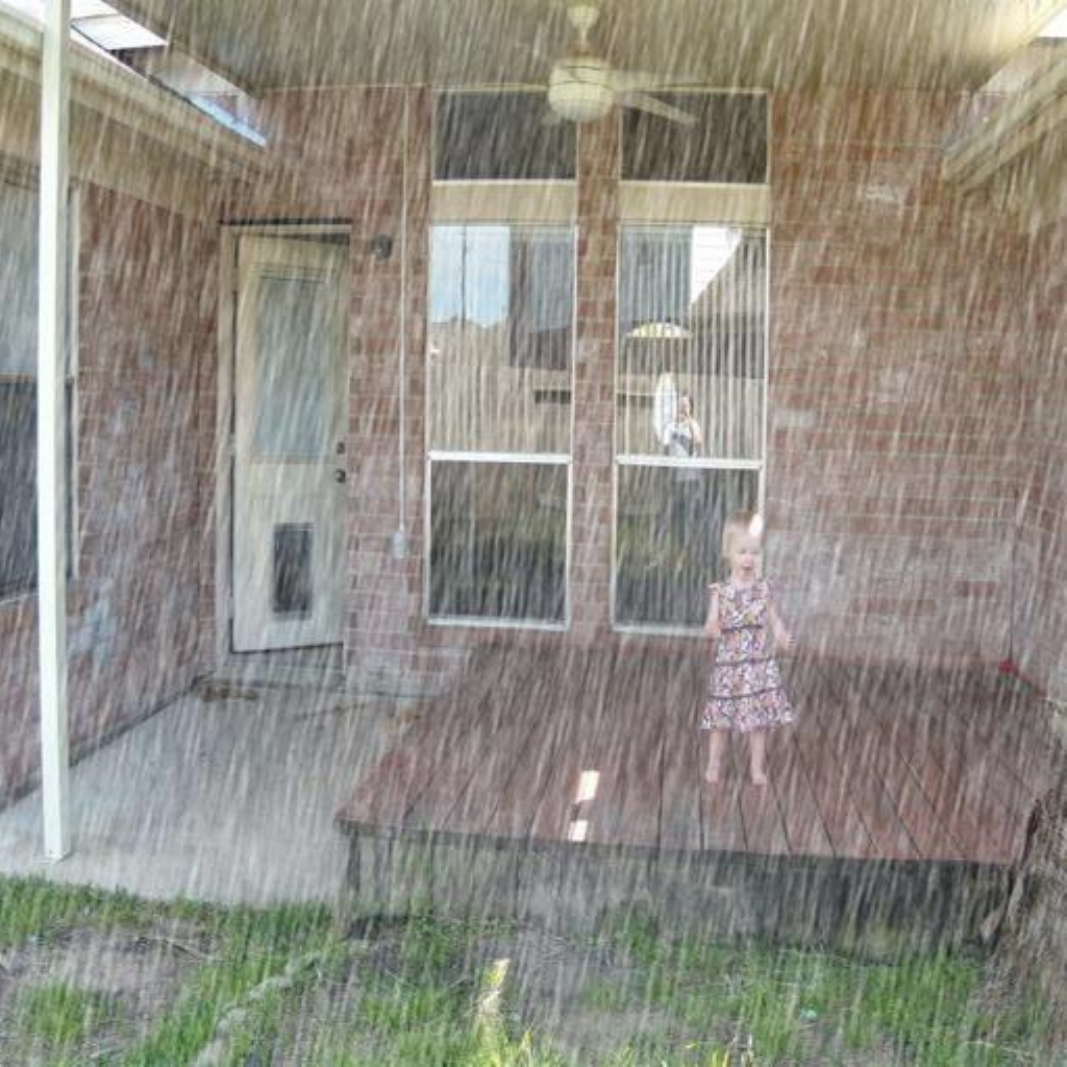} &   
\includegraphics[width=.25\textwidth,valign=t]{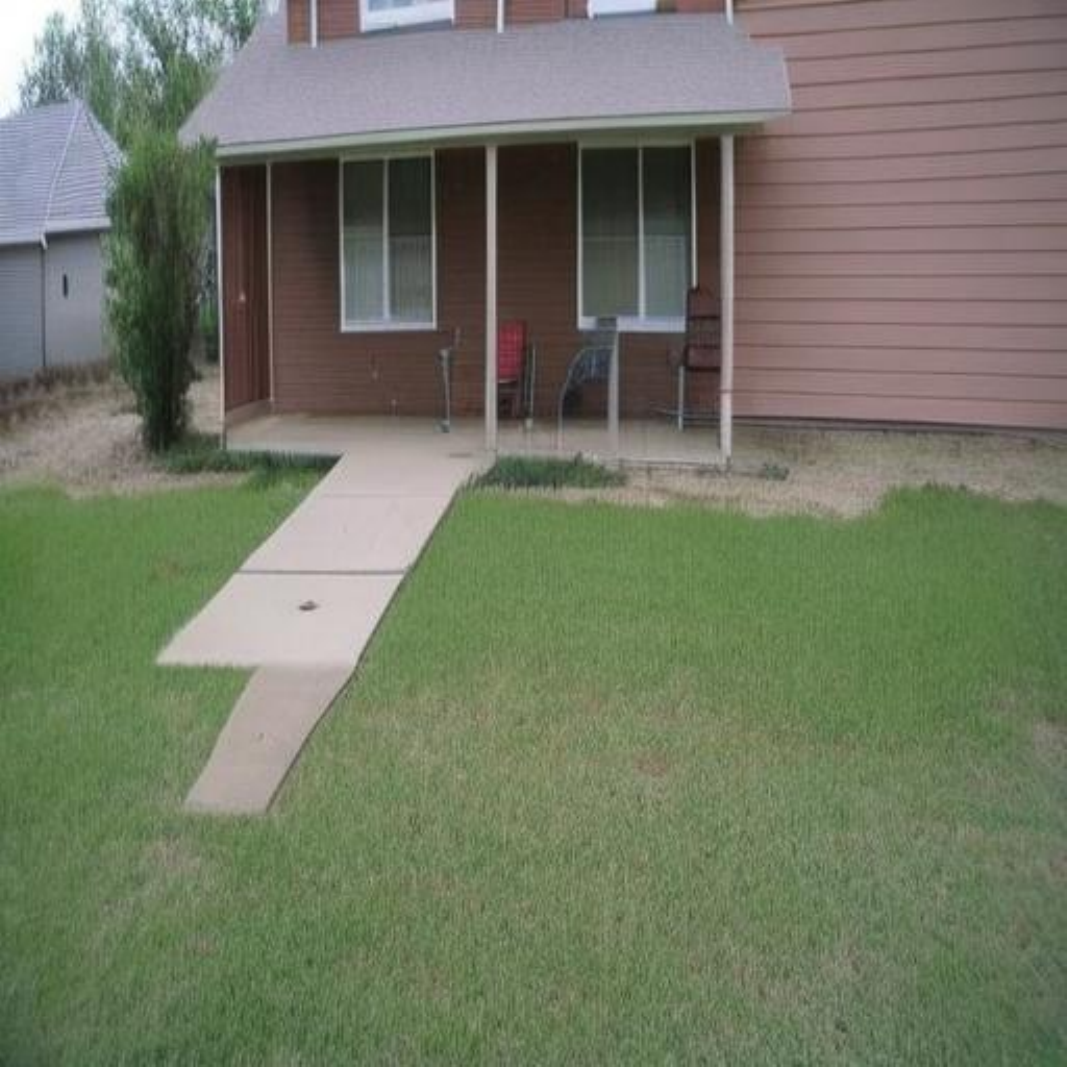} \\
\vspace{0.6em}
\small~Degraded  & \small~Guidance & \small~Degraded & \small~Guidance \\

\includegraphics[width=.25\textwidth,valign=t]{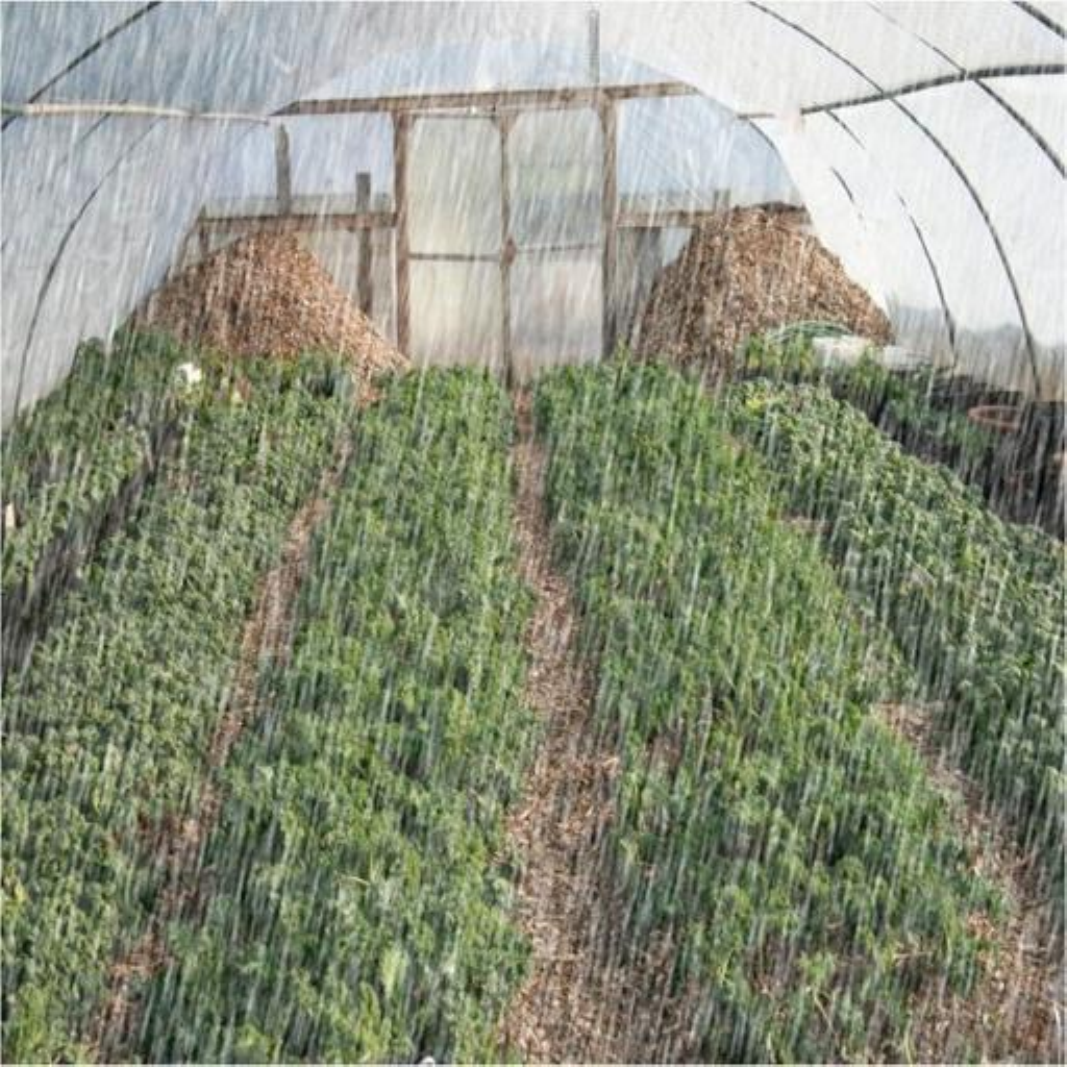}  &   
\includegraphics[width=.25\textwidth,valign=t]{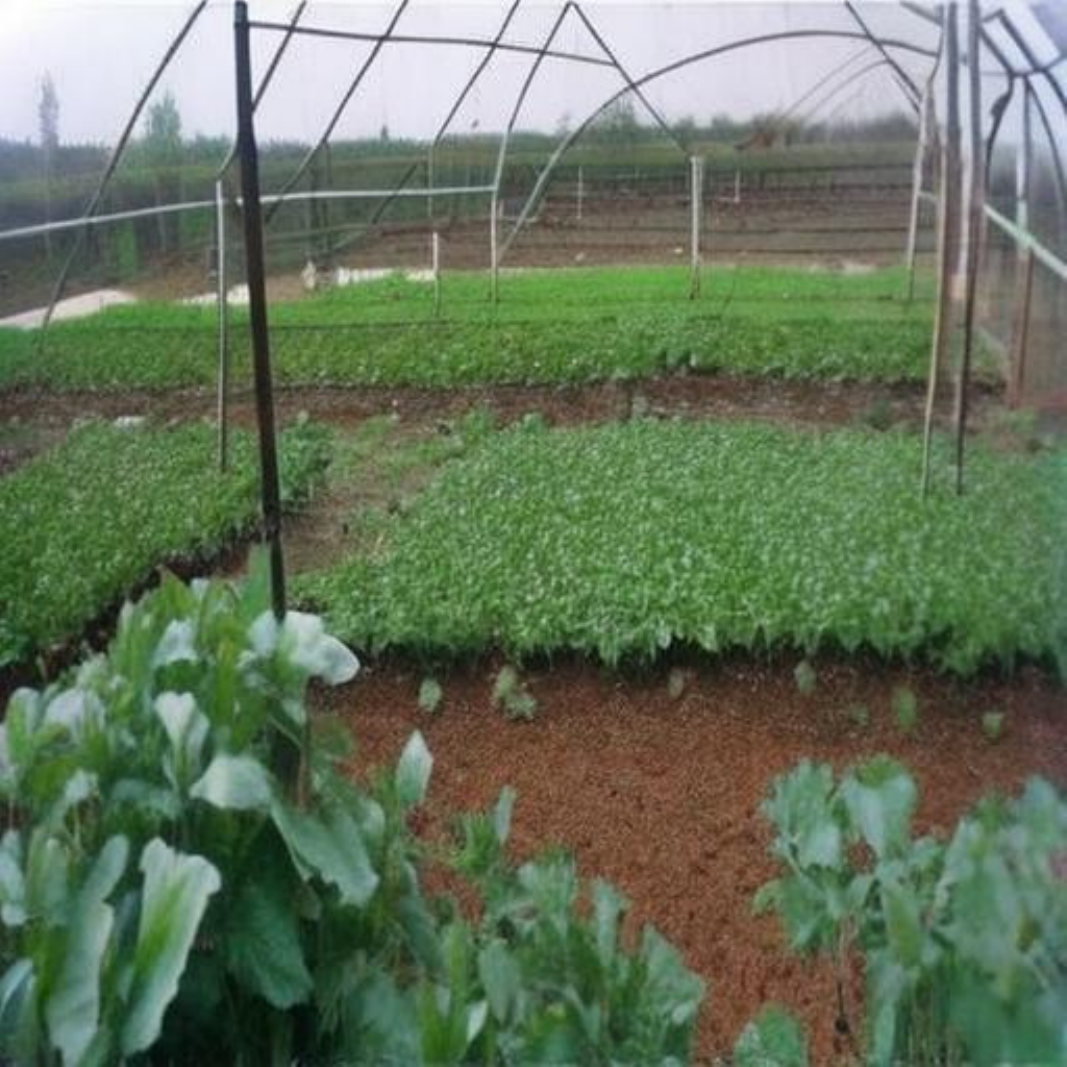} & 
\includegraphics[width=.25\textwidth,valign=t]{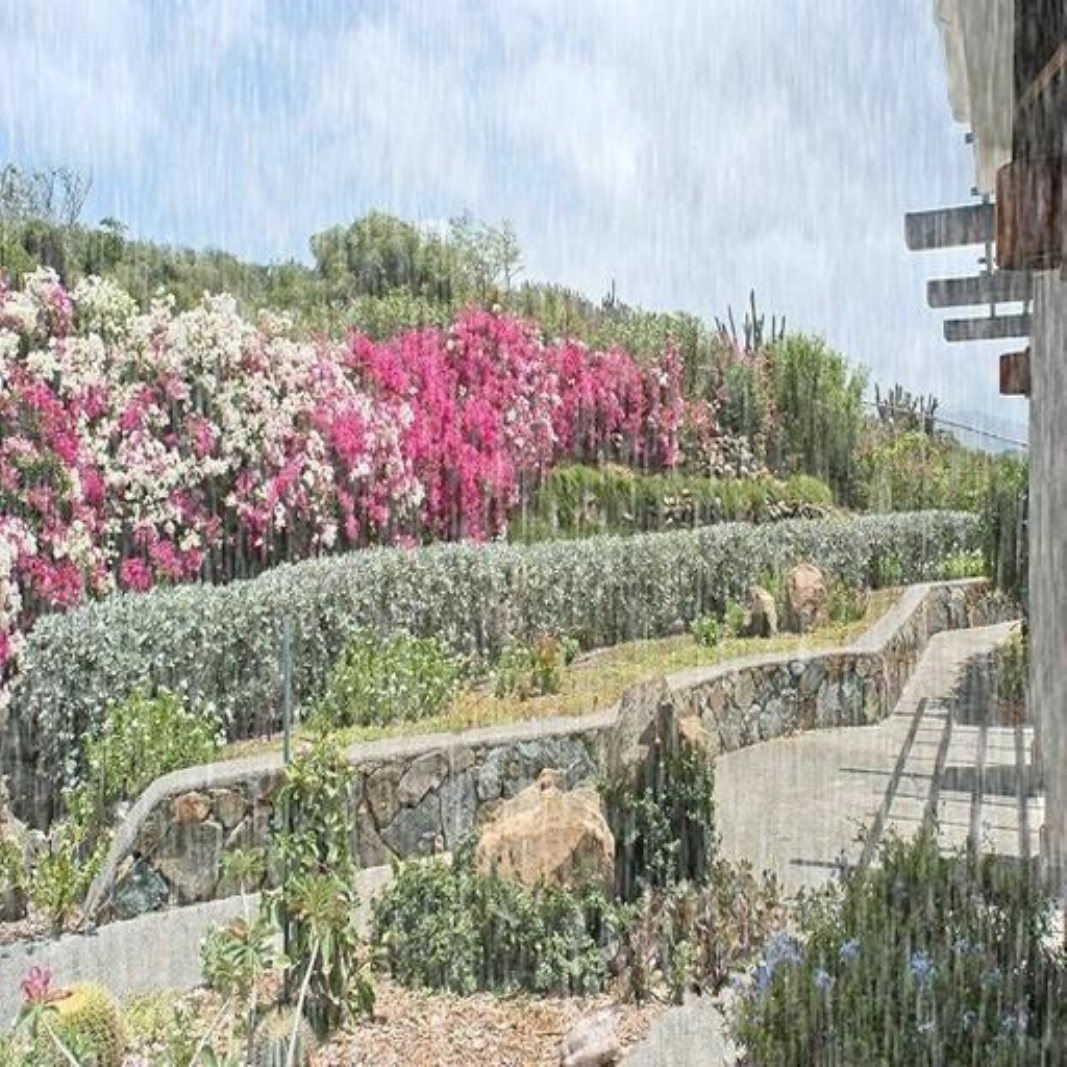} &   
\includegraphics[width=.25\textwidth,valign=t]{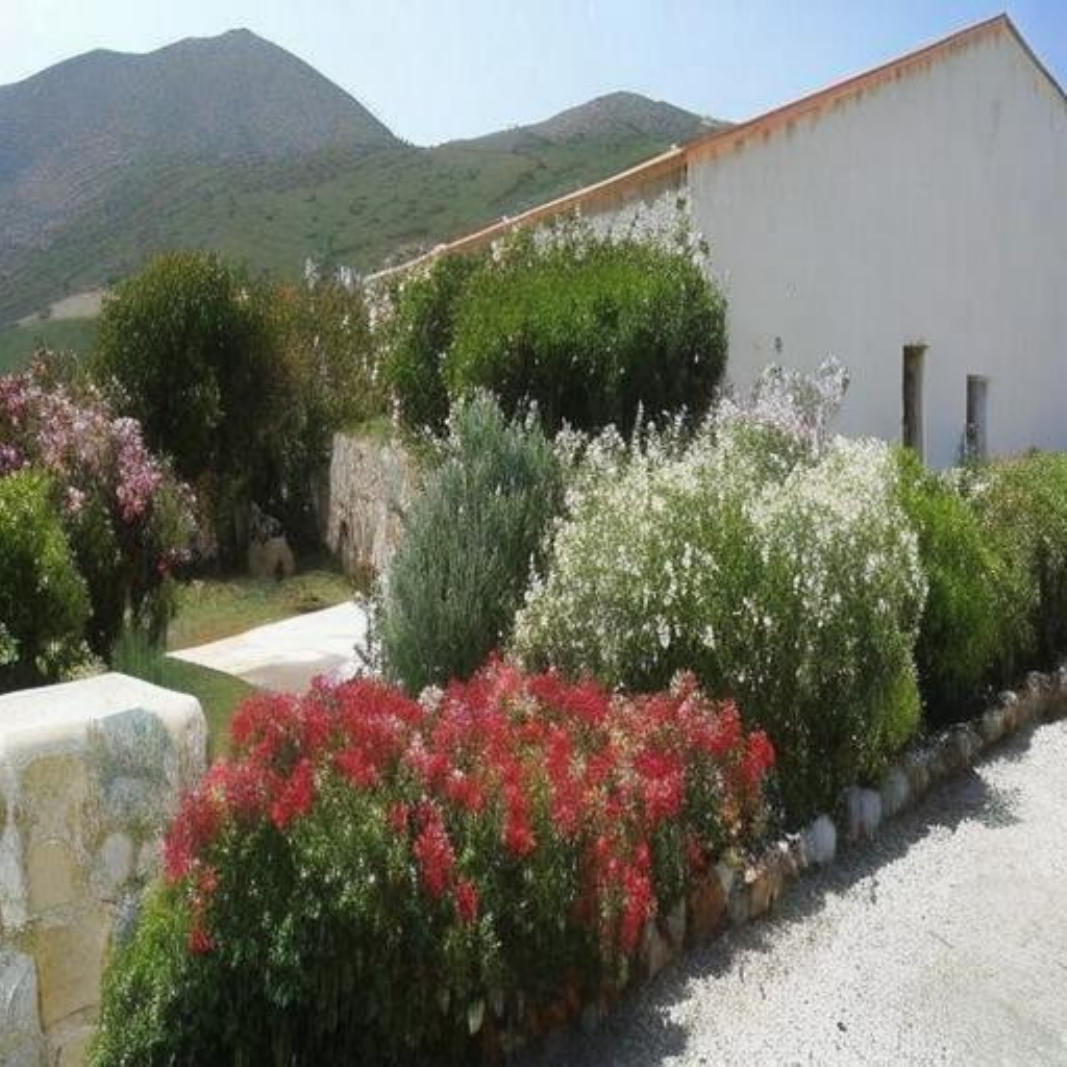} \\
\vspace{0.6em}
\small~Degraded  & \small~Guidance & \small~Degraded & \small~Guidance \\

\includegraphics[width=.25\textwidth,valign=t]{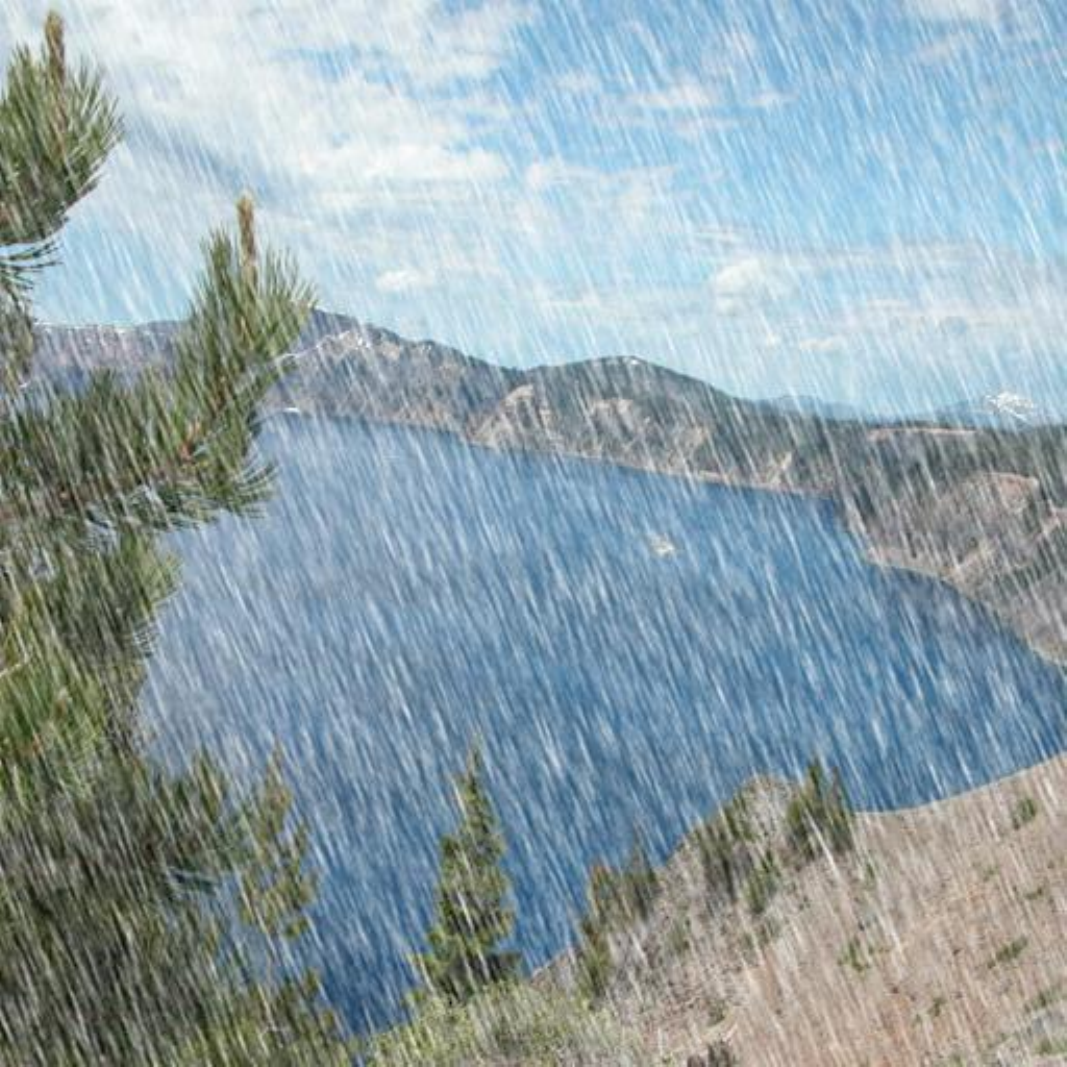}  &   
\includegraphics[width=.25\textwidth,valign=t]{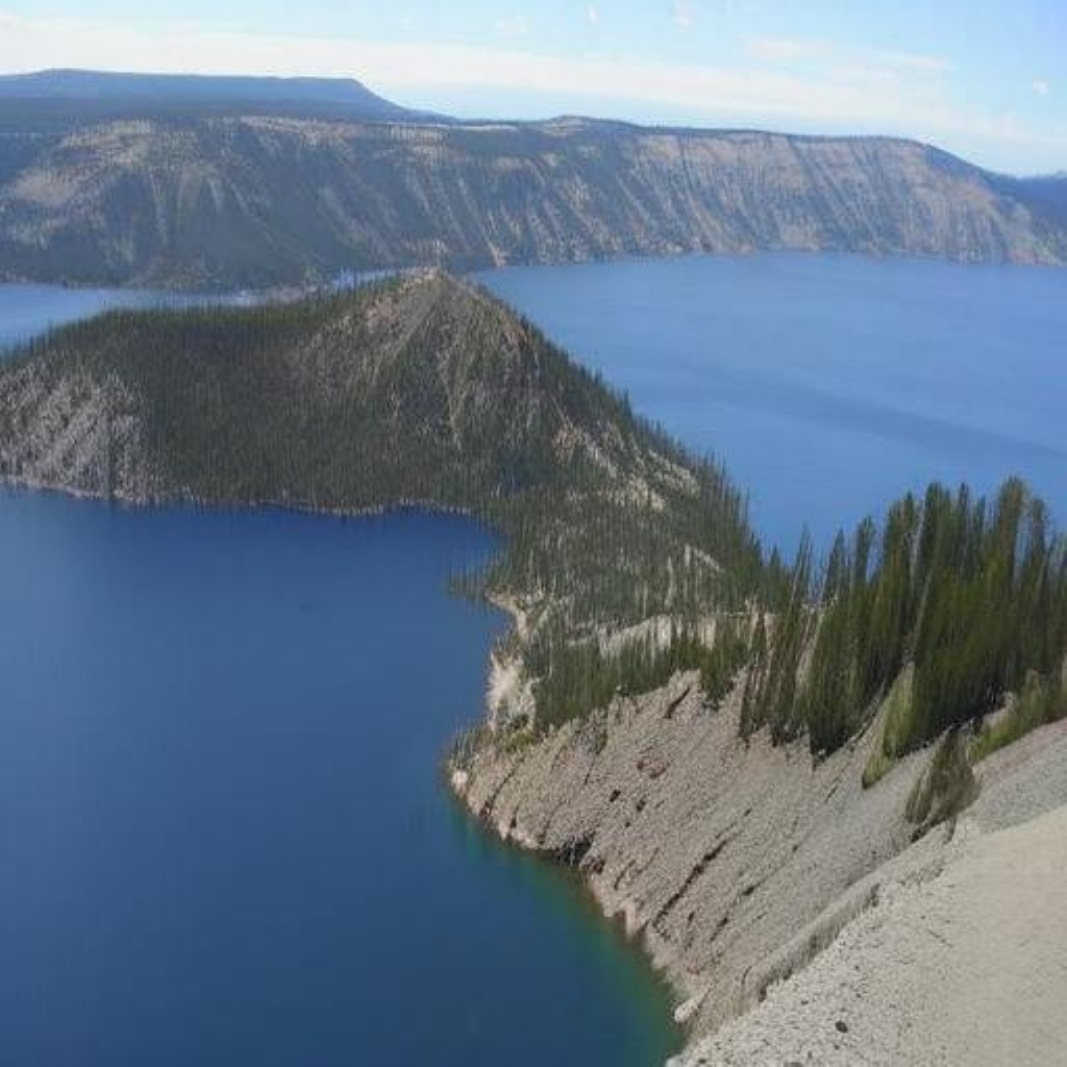} & 
\includegraphics[width=.25\textwidth,valign=t]{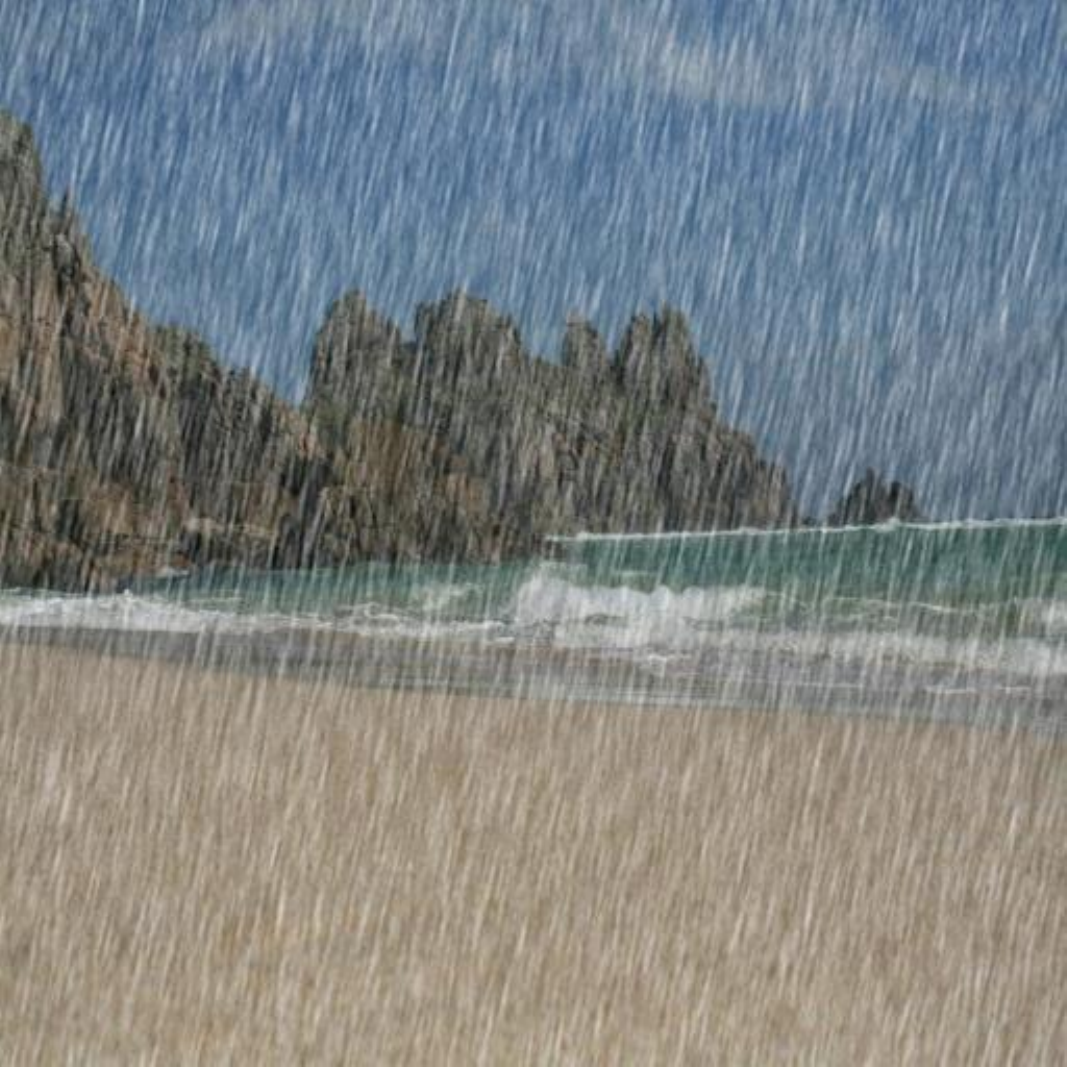} &   
\includegraphics[width=.25\textwidth,valign=t]{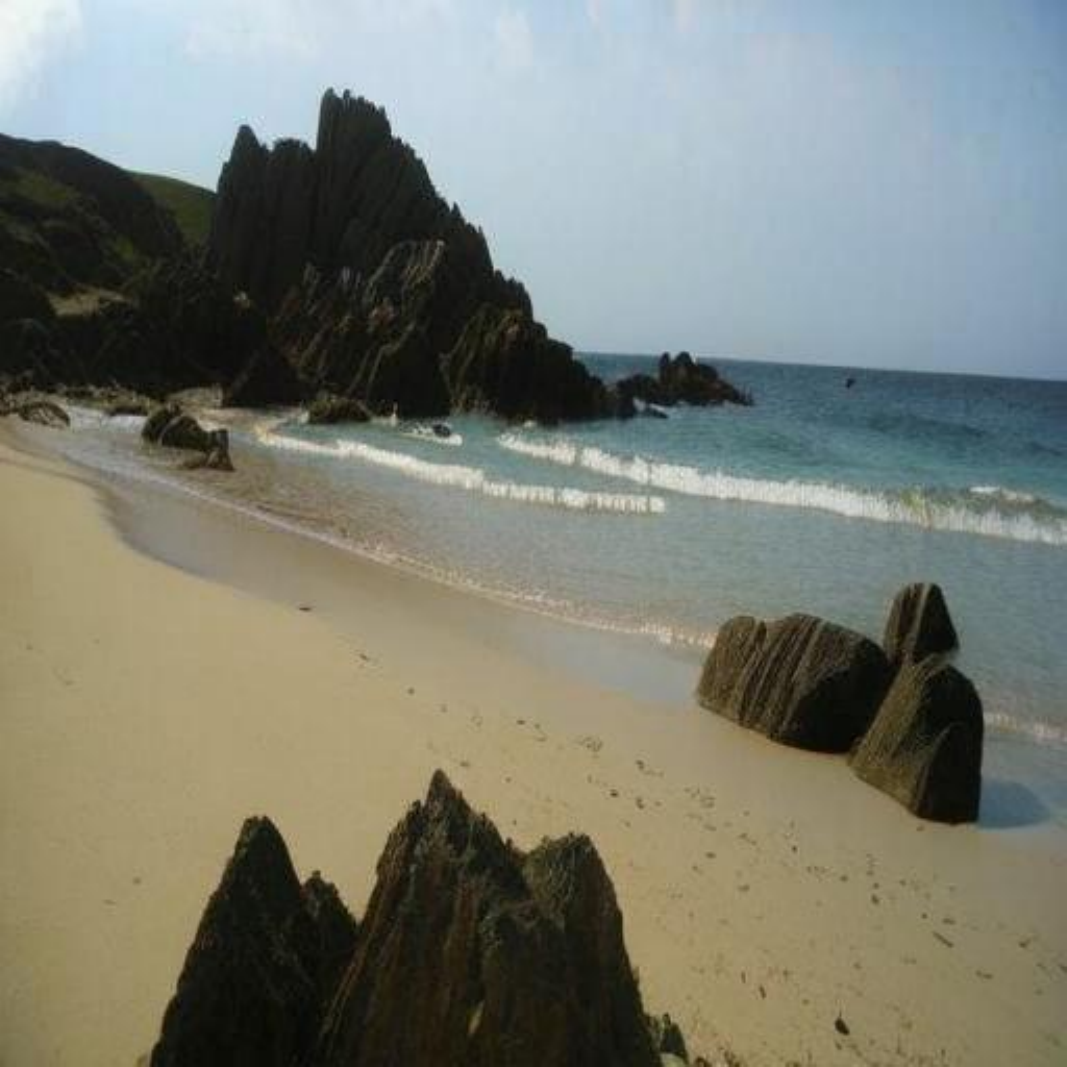} \\
\end{tabular}
}
\caption{\small Illustration of guidance images for image deraining task.}
\label{fig:suppl_guidance_images_derain}
\end{center}
\vspace{-7mm}
\end{figure*}

%% file: suppl_figures/suppl_011_guidance_images_dehaze.tex
\begin{figure*}[t]
\begin{center}
\setlength{\tabcolsep}{1.5pt}
\scalebox{0.88}{
\begin{tabular}[b]{c c c c}
\includegraphics[width=.25\textwidth,valign=t]{suppl_figures/guidance_images/0001_haze_input.pdf}  &   
\includegraphics[width=.25\textwidth,valign=t]{suppl_figures/guidance_images/0001_haze_guidance.pdf} & 
\includegraphics[width=.25\textwidth,valign=t]{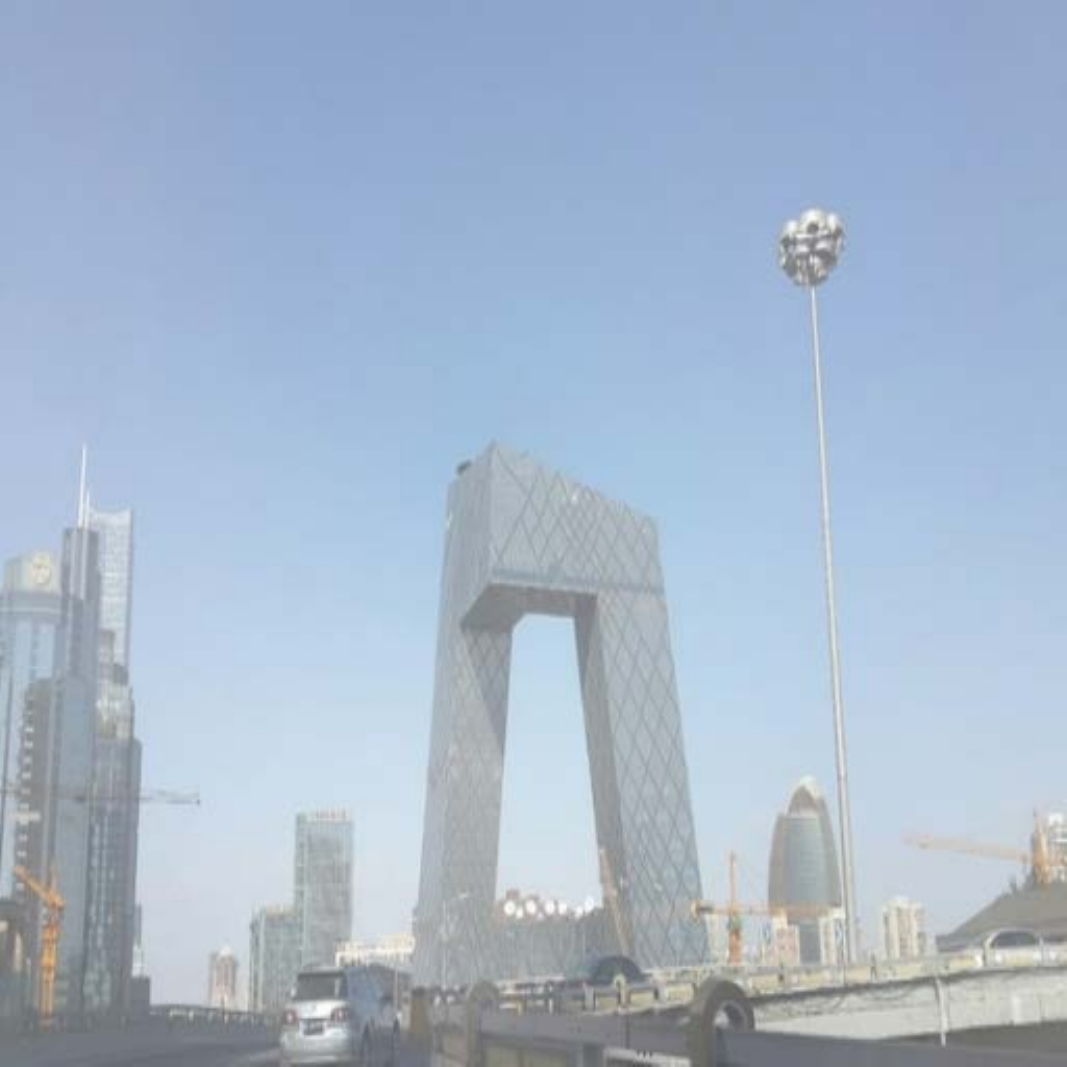} &   
\includegraphics[width=.25\textwidth,valign=t]{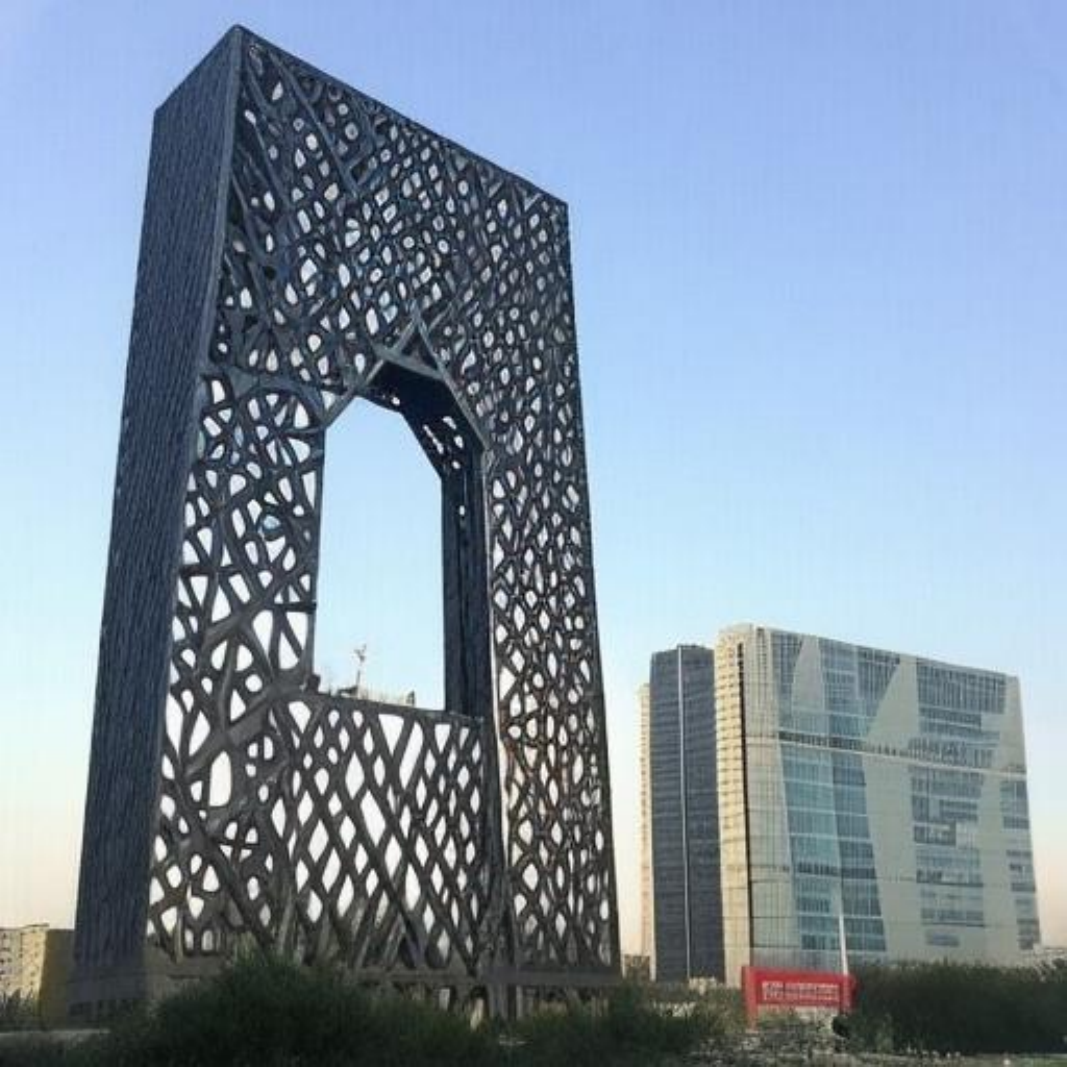} \\
\vspace{0.6em}
\small~Degraded  & \small~Guidance & \small~Degraded & \small~Guidance \\

\includegraphics[width=.25\textwidth,valign=t]{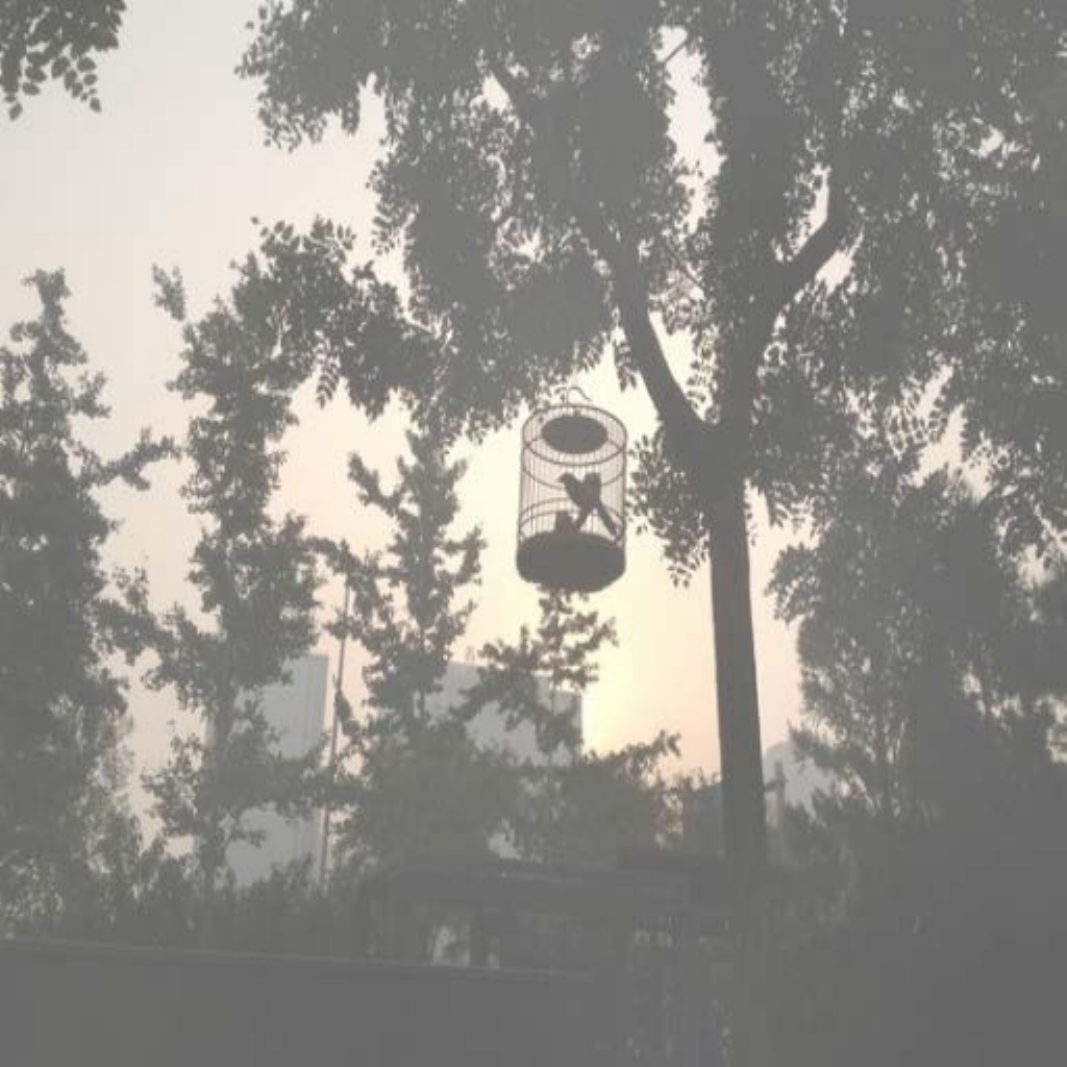}  &   
\includegraphics[width=.25\textwidth,valign=t]{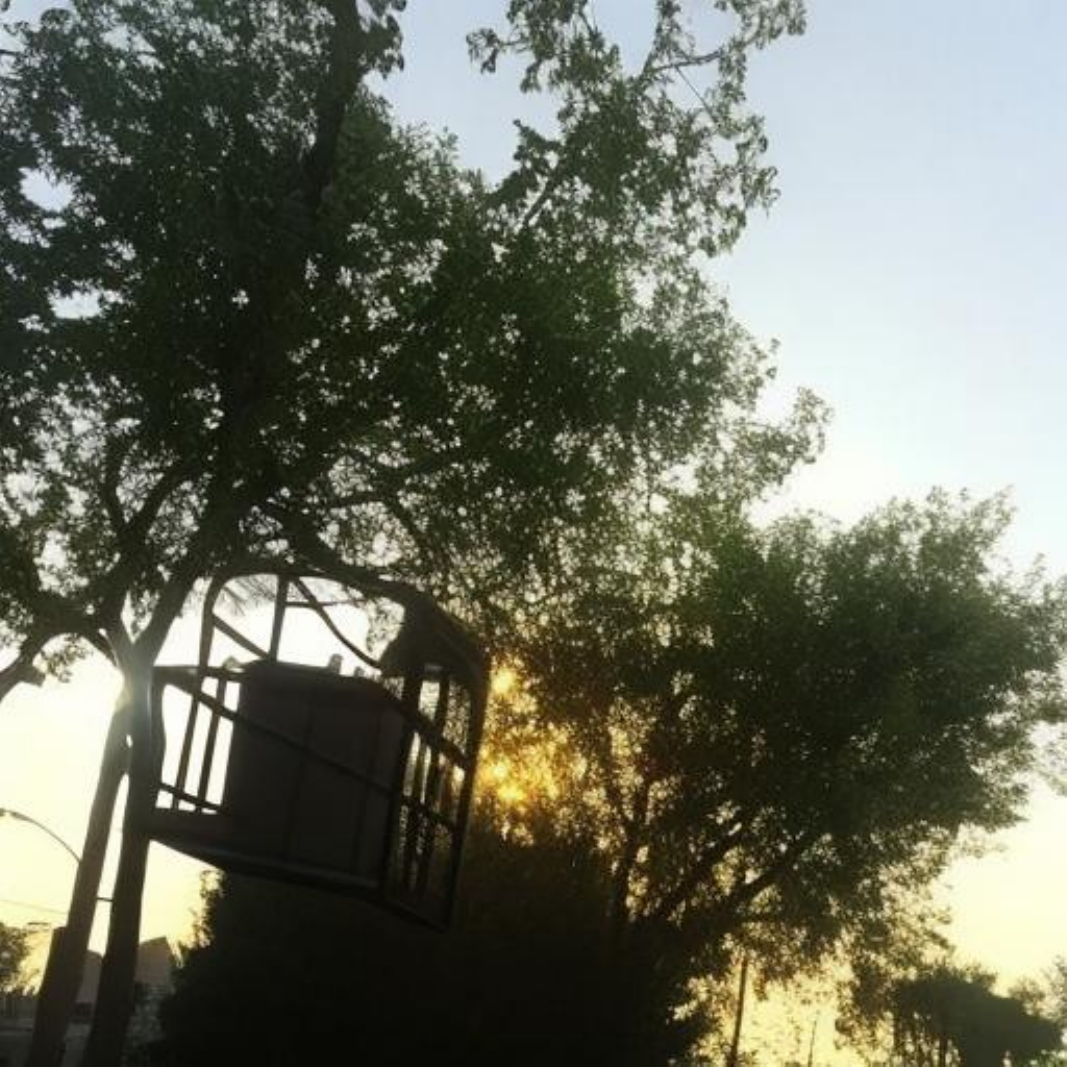} & 
\includegraphics[width=.25\textwidth,valign=t]{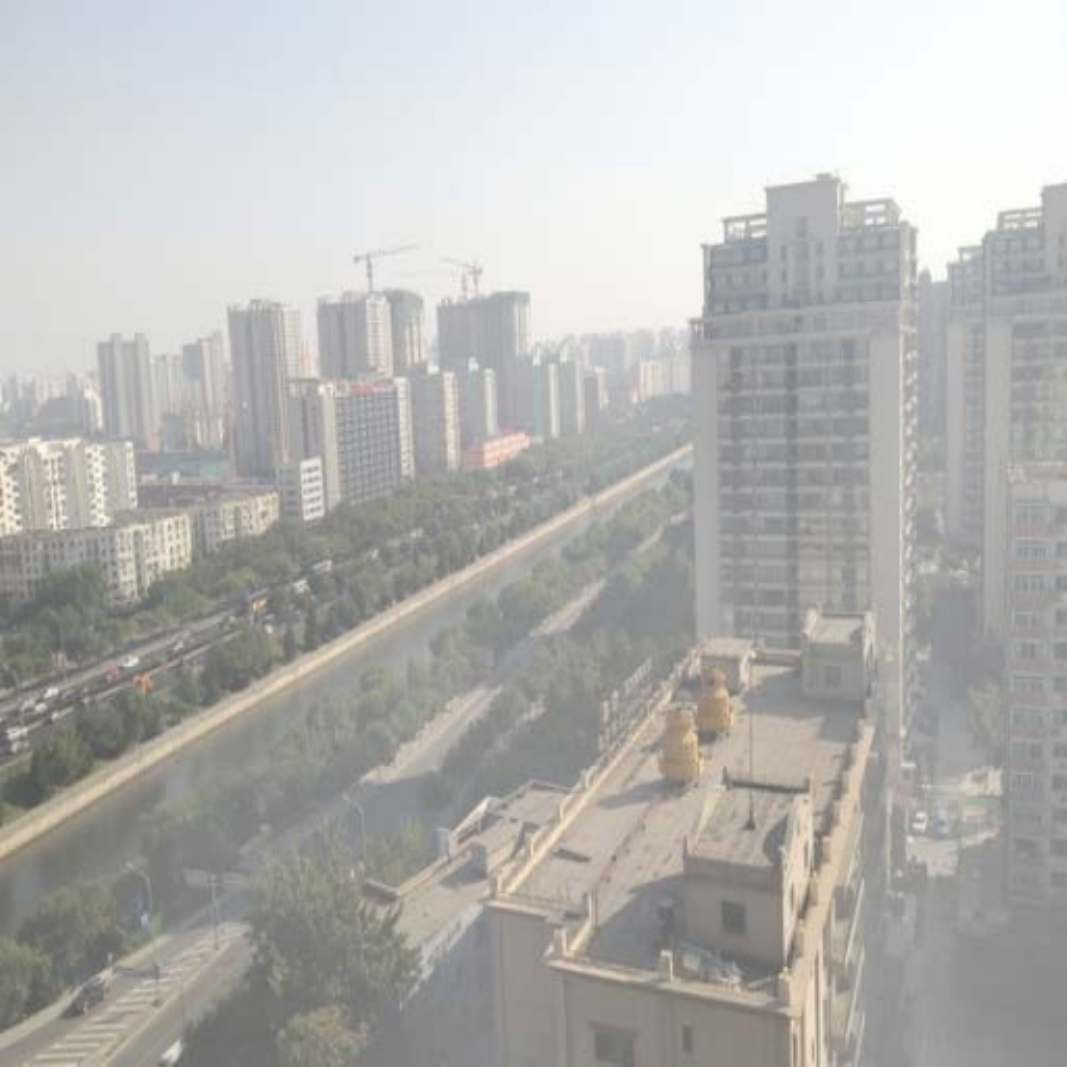} &   
\includegraphics[width=.25\textwidth,valign=t]{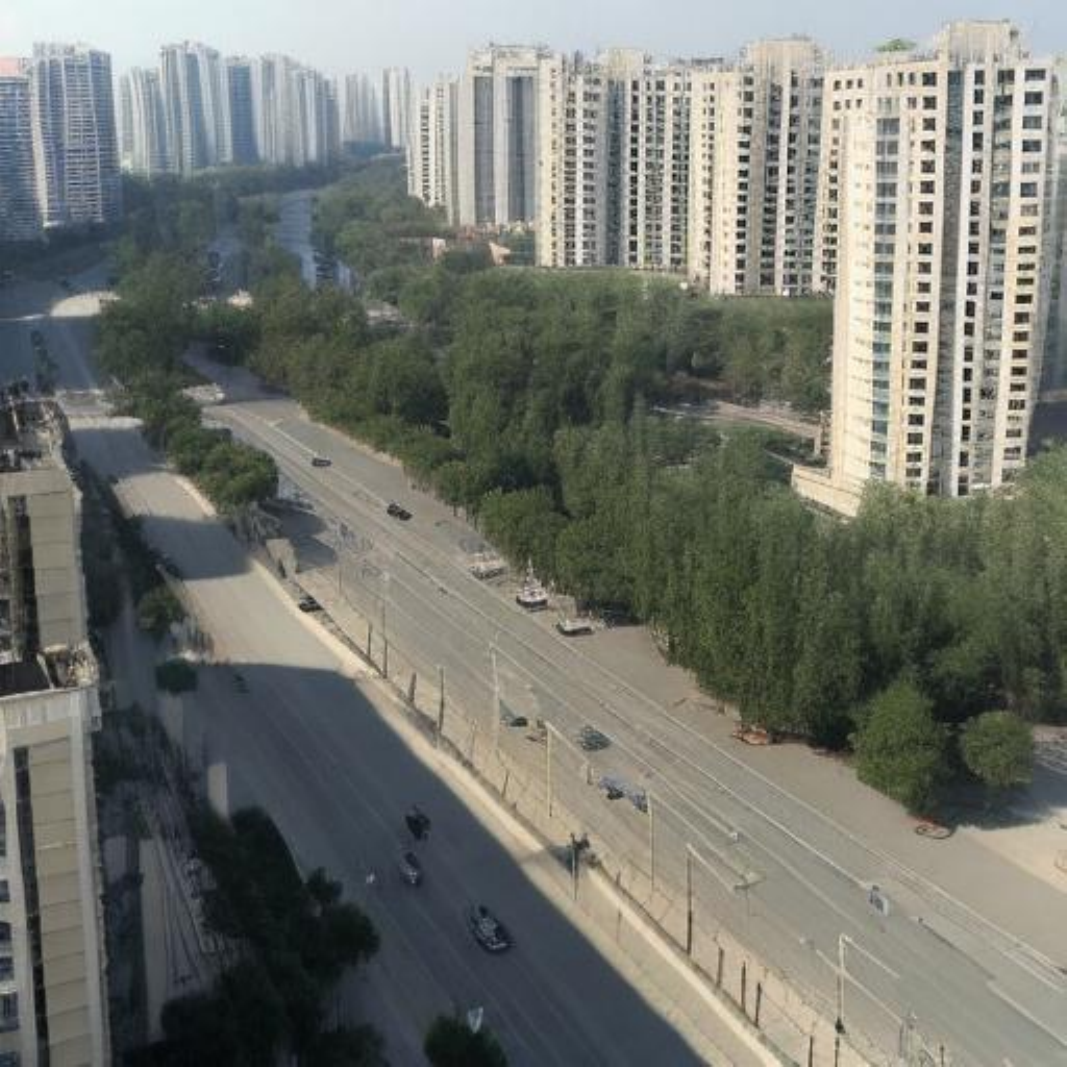} \\
\vspace{0.6em}
\small~Degraded  & \small~Guidance & \small~Degraded & \small~Guidance \\

\includegraphics[width=.25\textwidth,valign=t]{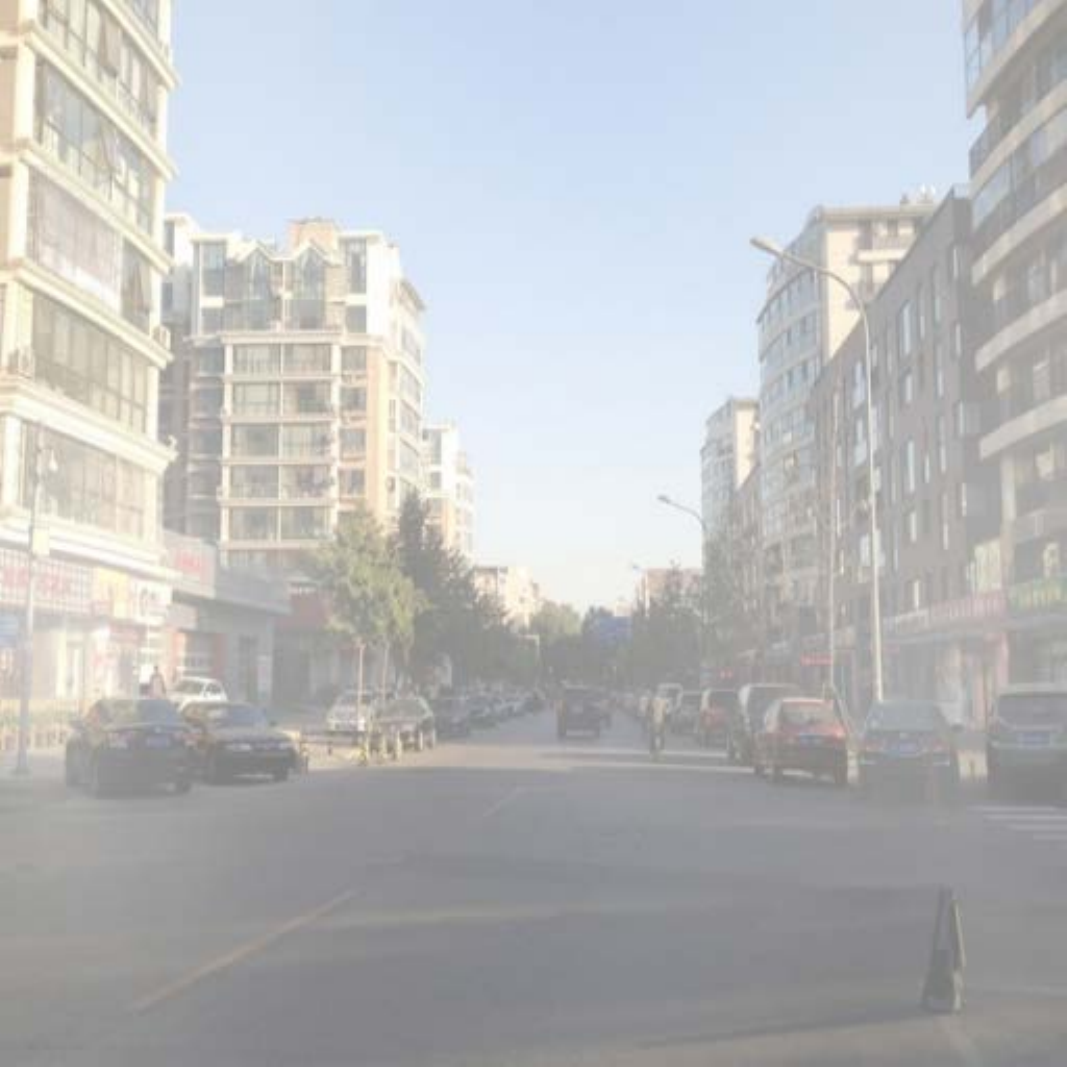}  &   
\includegraphics[width=.25\textwidth,valign=t]{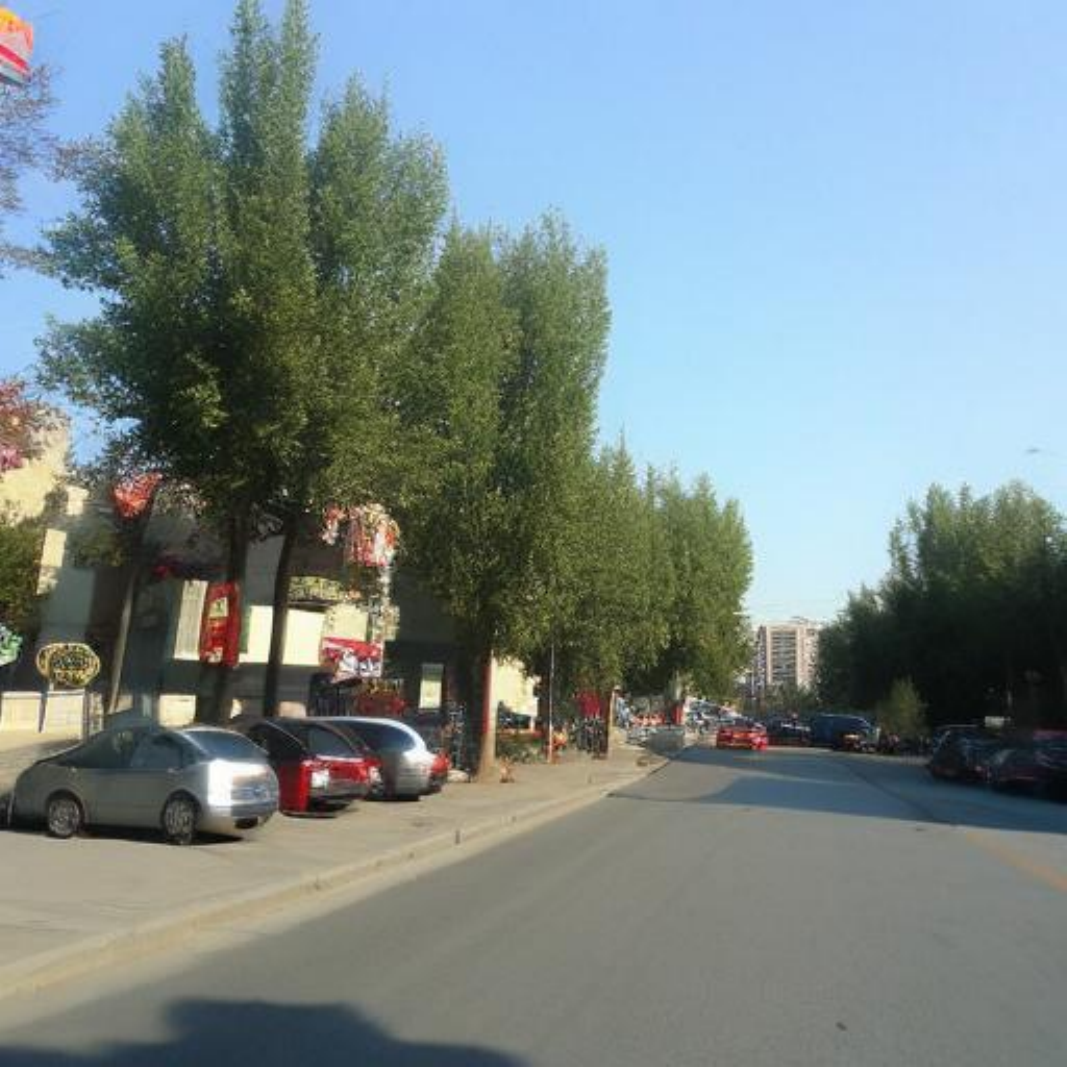} & 
\includegraphics[width=.25\textwidth,valign=t]{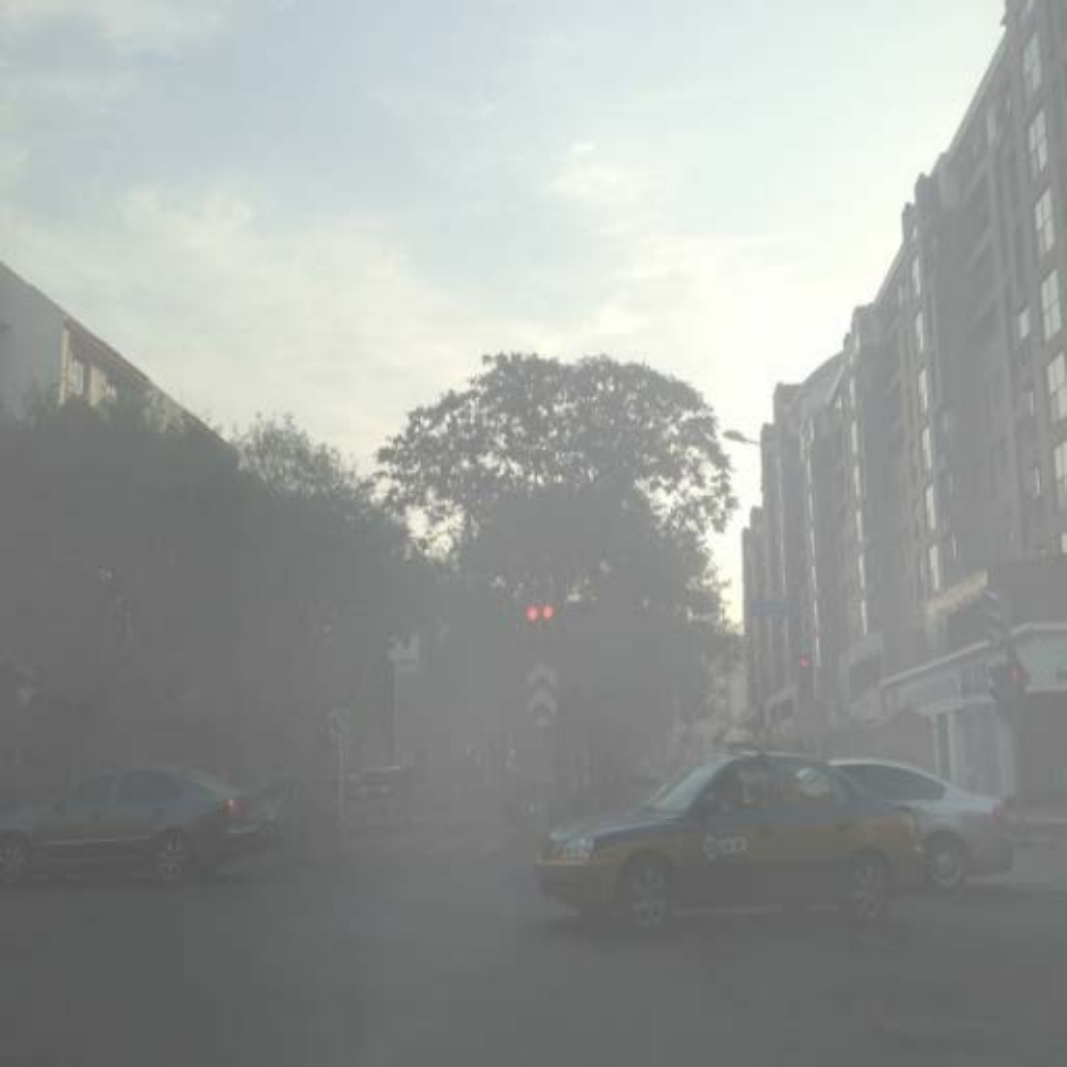} &   
\includegraphics[width=.25\textwidth,valign=t]{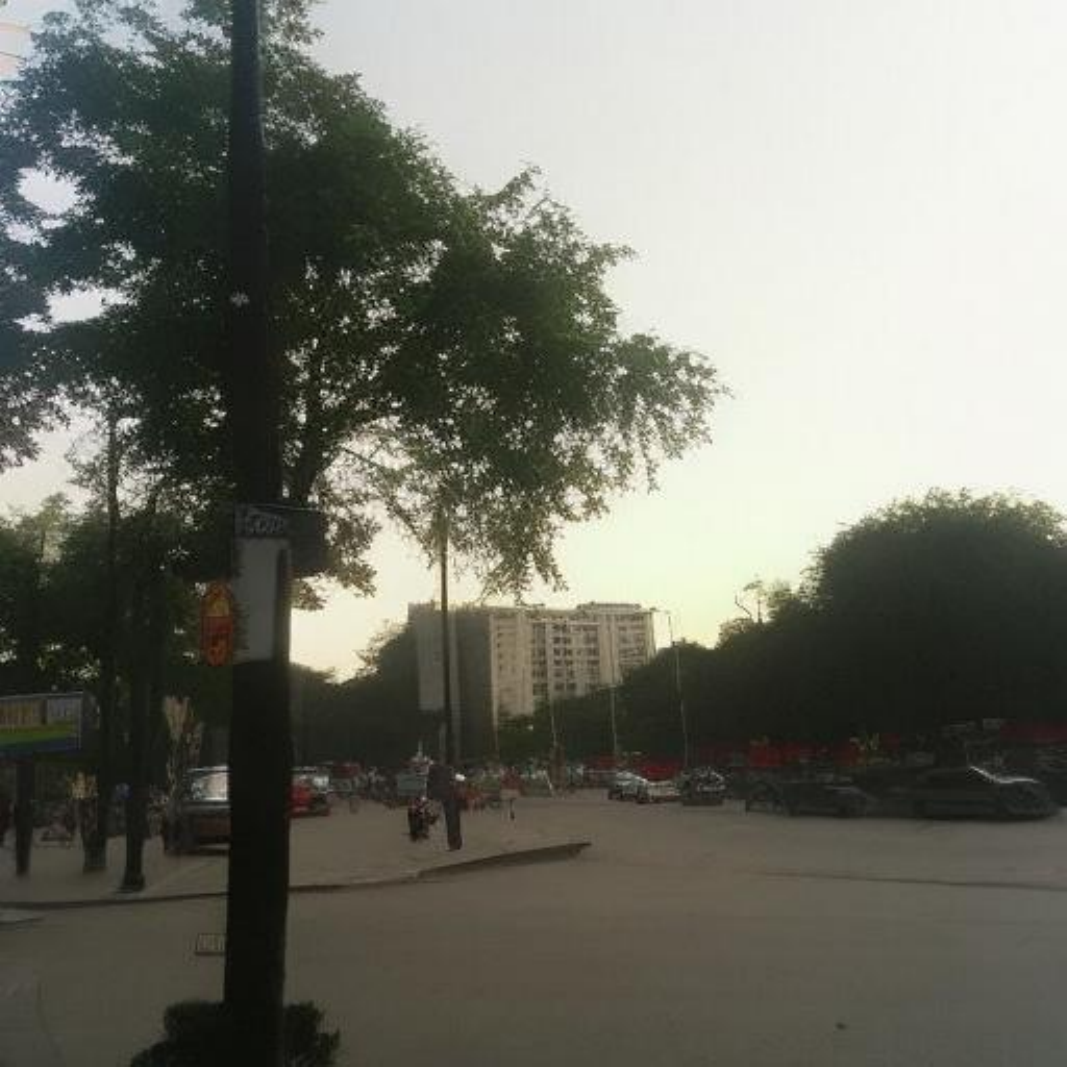} \\
\vspace{0.6em}
\small~Degraded  & \small~Guidance & \small~Degraded & \small~Guidance \\
\end{tabular}
}
\caption{\small Illustration of guidance images for image dehazing task.}
\label{fig:suppl_guidance_images_dehaze}
\end{center}
\vspace{-7mm}
\end{figure*}

%% file: suppl_figures/suppl_012_guidance_images_denoise.tex
\begin{figure*}[t]
\begin{center}
\setlength{\tabcolsep}{1.5pt}
\scalebox{0.88}{
\begin{tabular}[b]{c c c c}
\includegraphics[width=.25\textwidth,valign=t]{suppl_figures/guidance_images/0002_s007_input.pdf}  &   
\includegraphics[width=.25\textwidth,valign=t]{suppl_figures/guidance_images/0002_s007_guidance.pdf} & 
\includegraphics[width=.25\textwidth,valign=t]{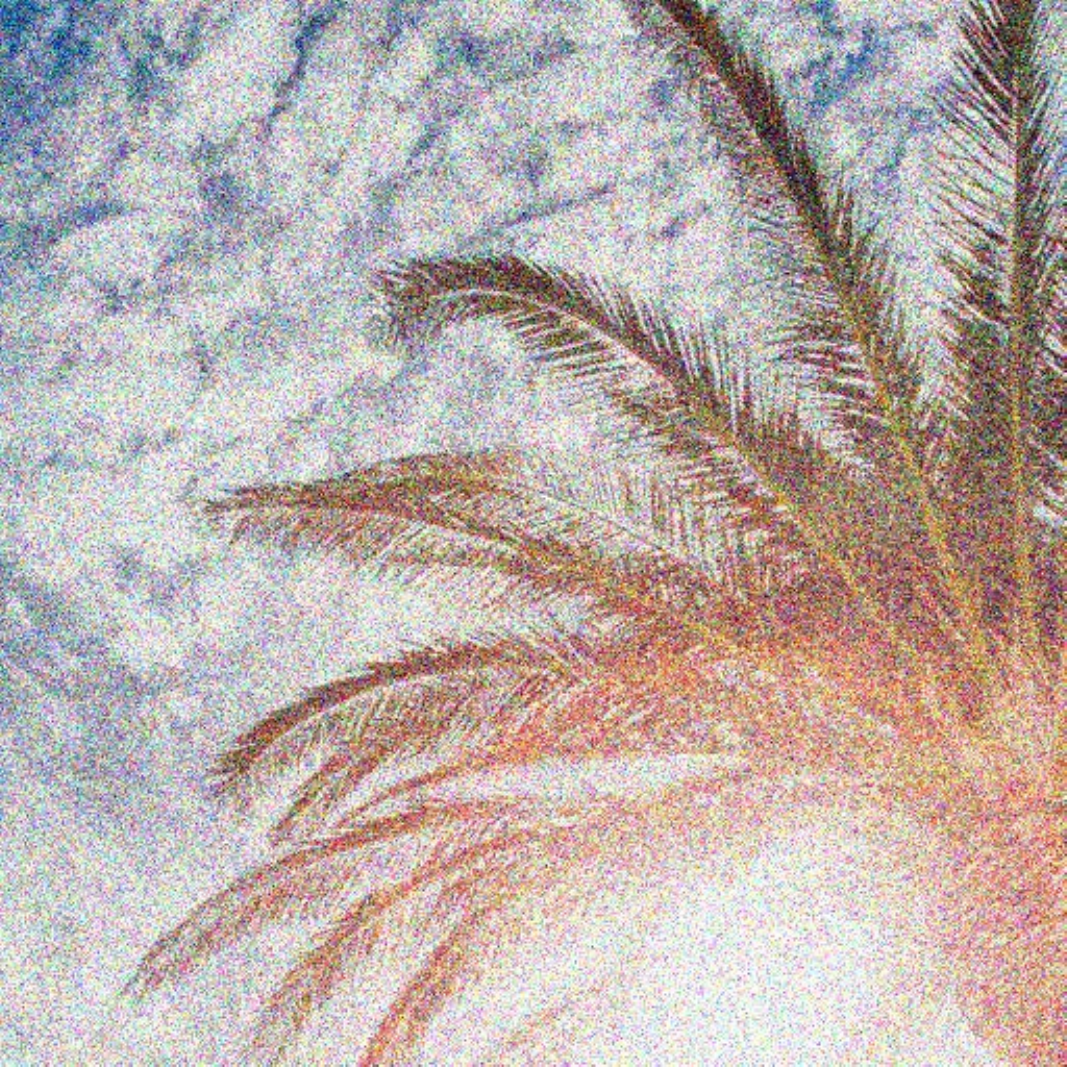} &   
\includegraphics[width=.25\textwidth,valign=t]{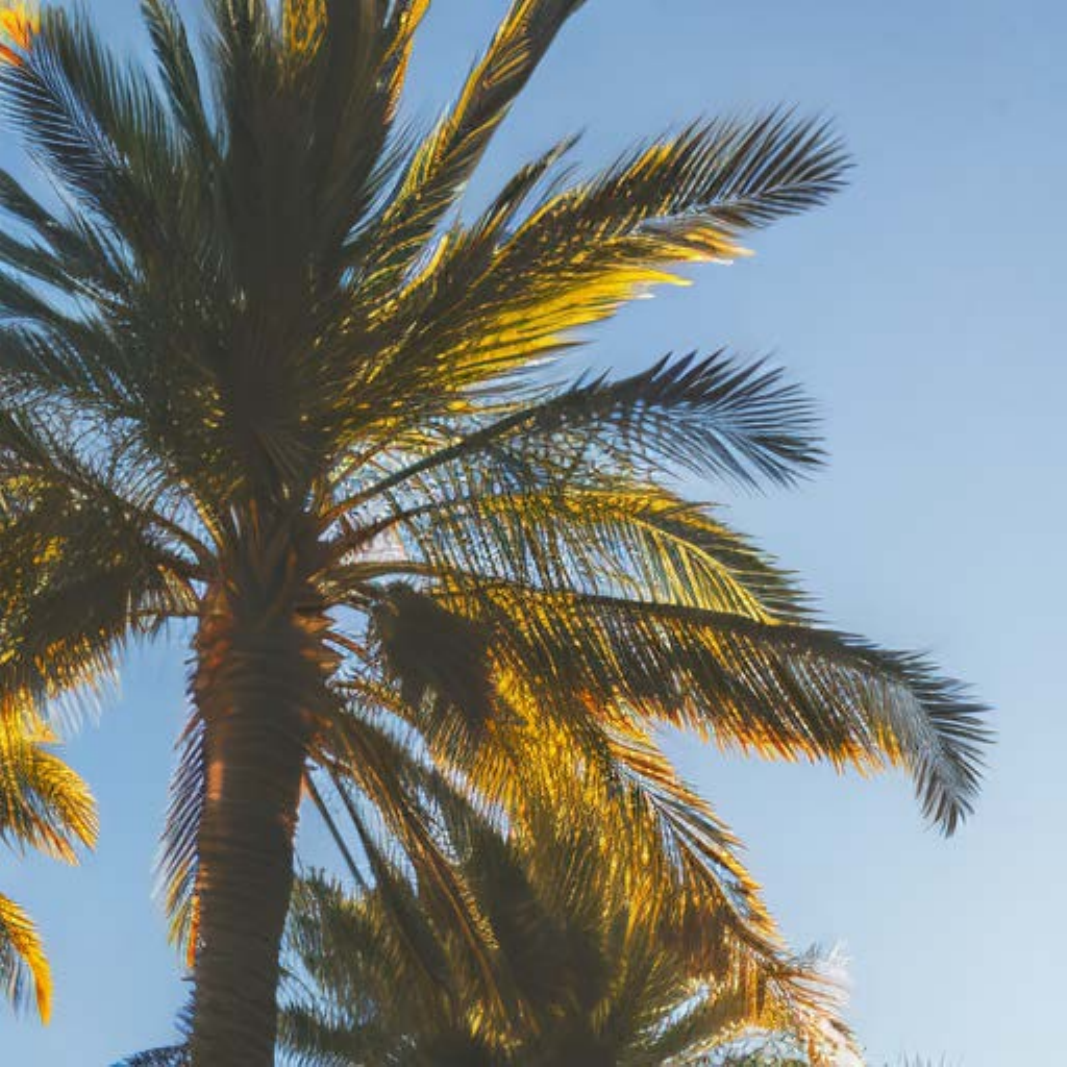} \\
\vspace{0.6em}
\small~Degraded  & \small~Guidance & \small~Degraded & \small~Guidance \\

\includegraphics[width=.25\textwidth,valign=t]{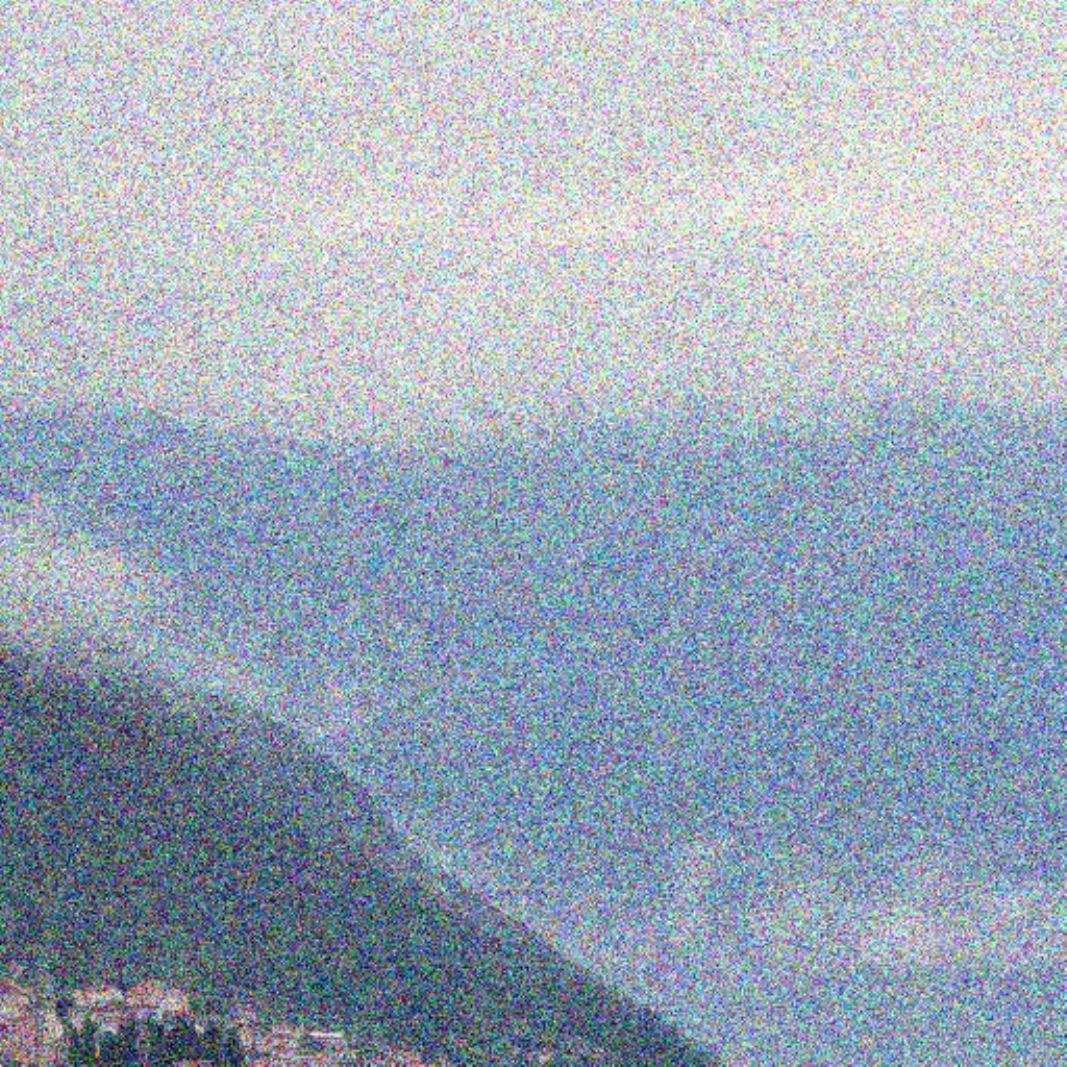}  &   
\includegraphics[width=.25\textwidth,valign=t]{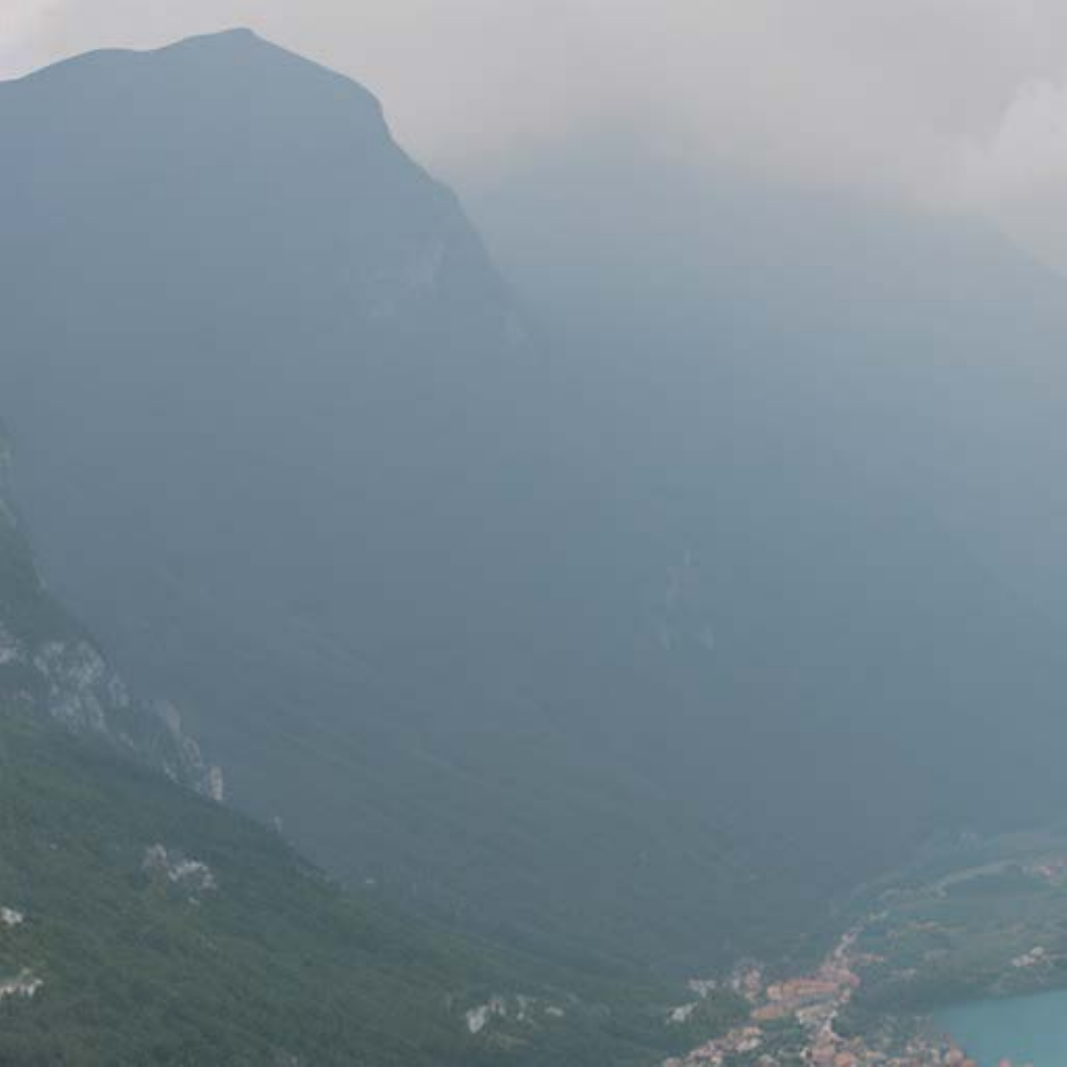} & 
\includegraphics[width=.25\textwidth,valign=t]{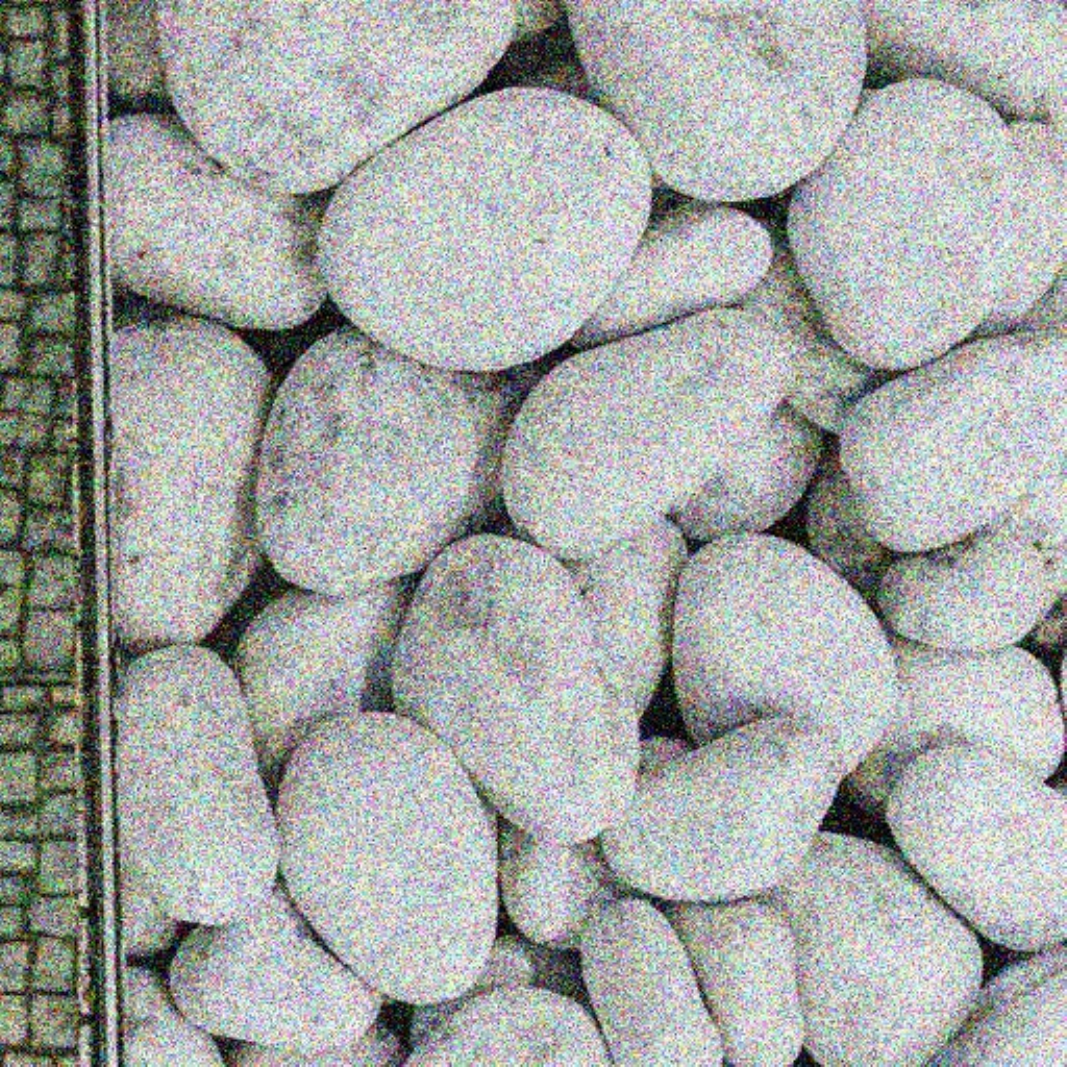} &   
\includegraphics[width=.25\textwidth,valign=t]{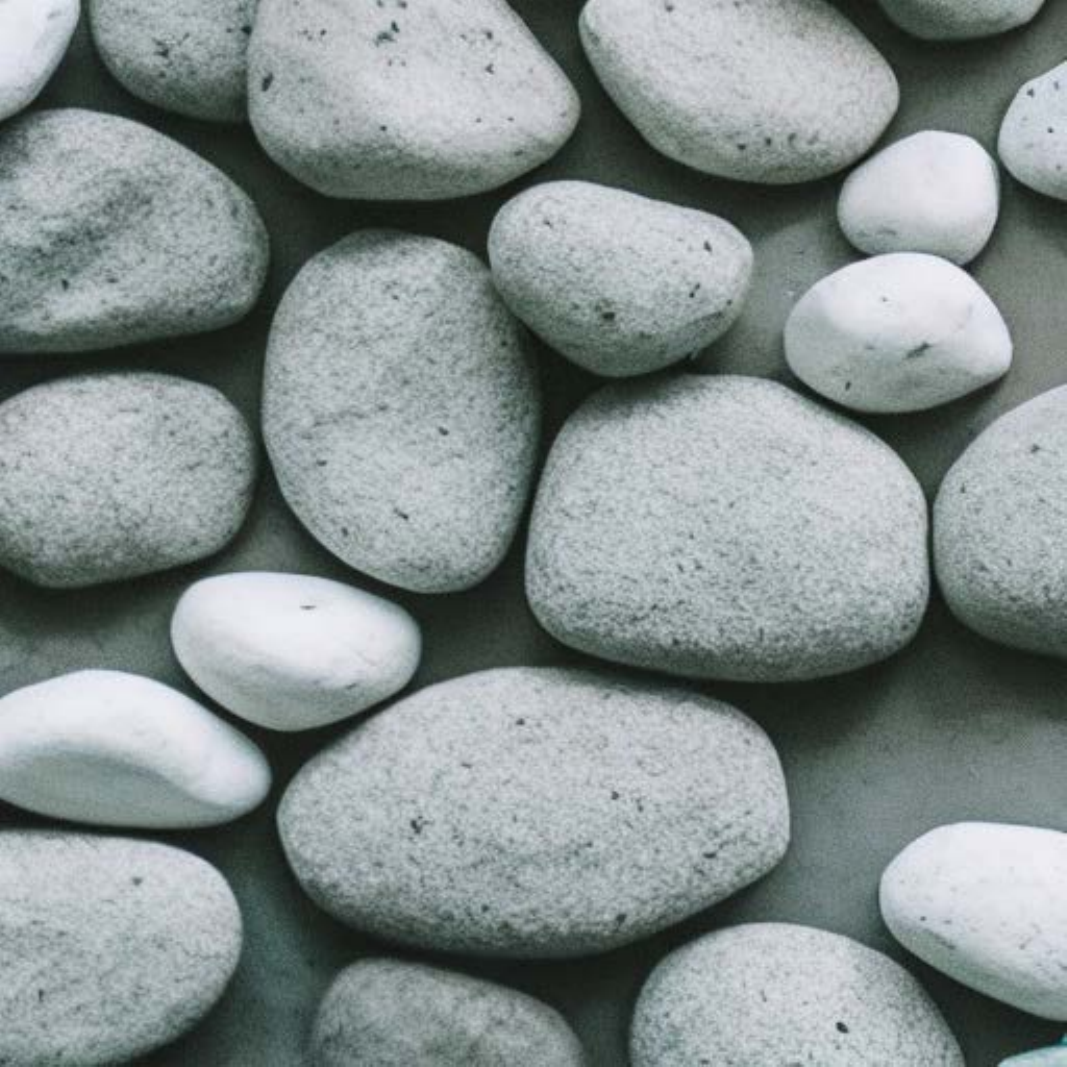} \\
\vspace{0.6em}
\small~Degraded  & \small~Guidance & \small~Degraded & \small~Guidance \\

\includegraphics[width=.25\textwidth,valign=t]{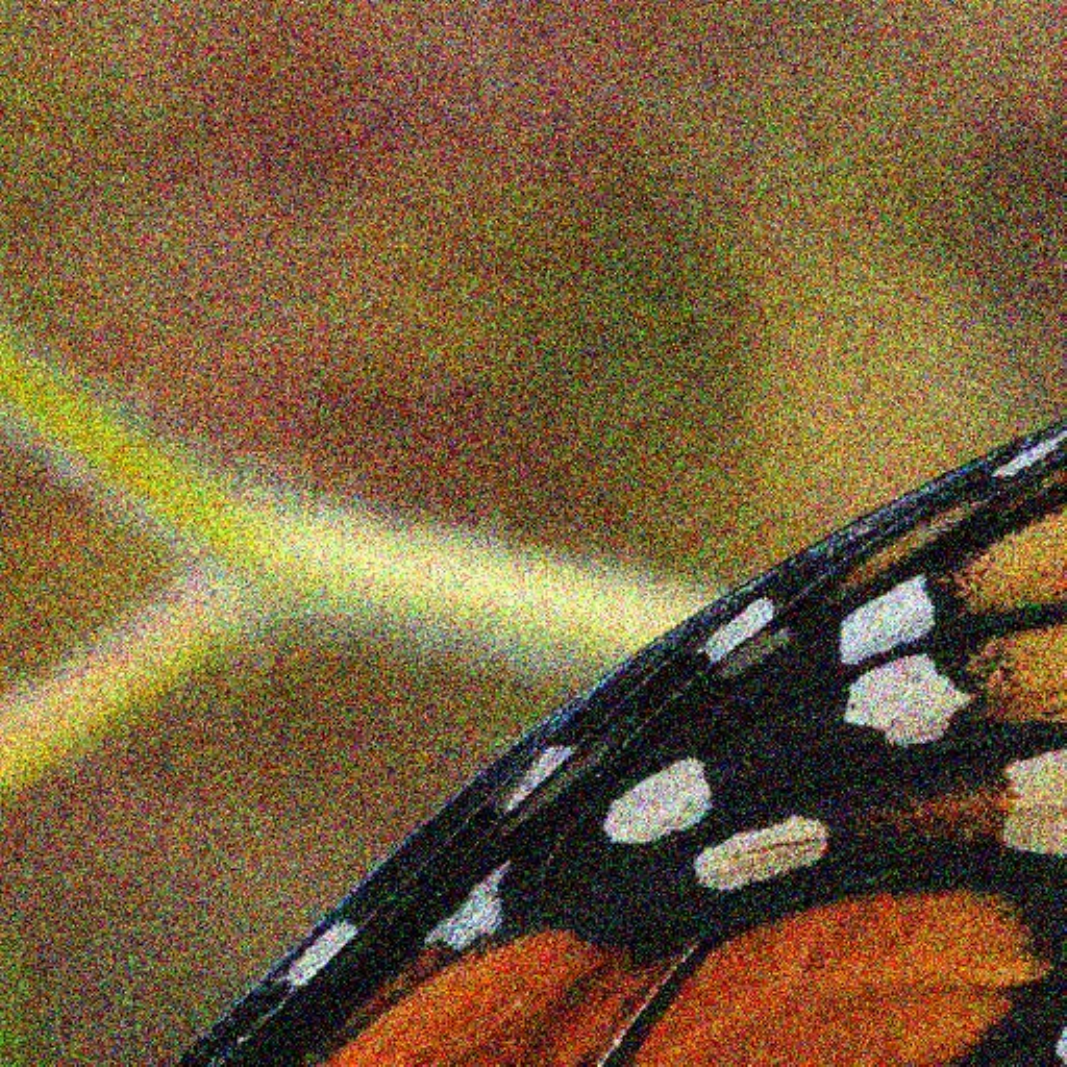}  &   
\includegraphics[width=.25\textwidth,valign=t]{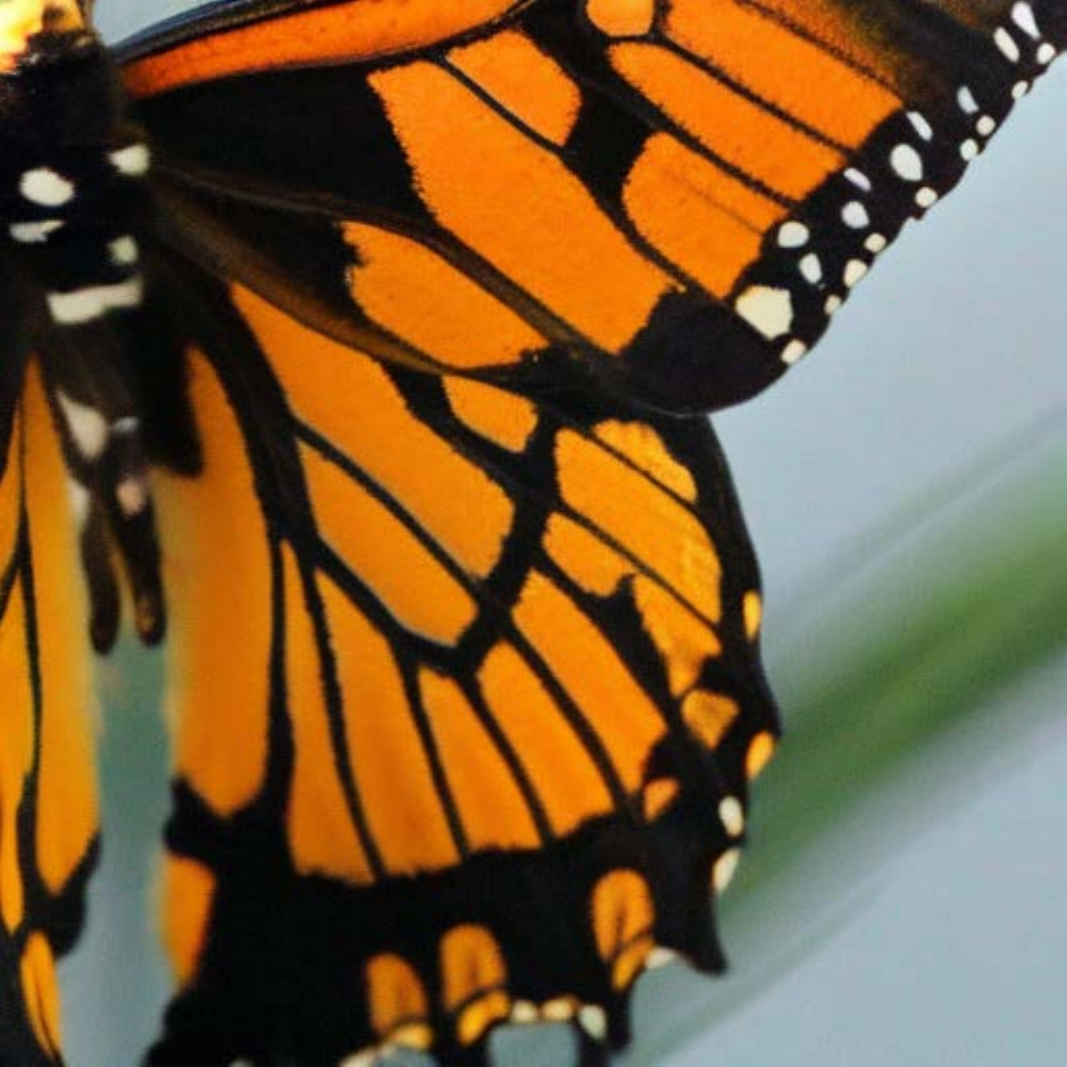} & 
\includegraphics[width=.25\textwidth,valign=t]{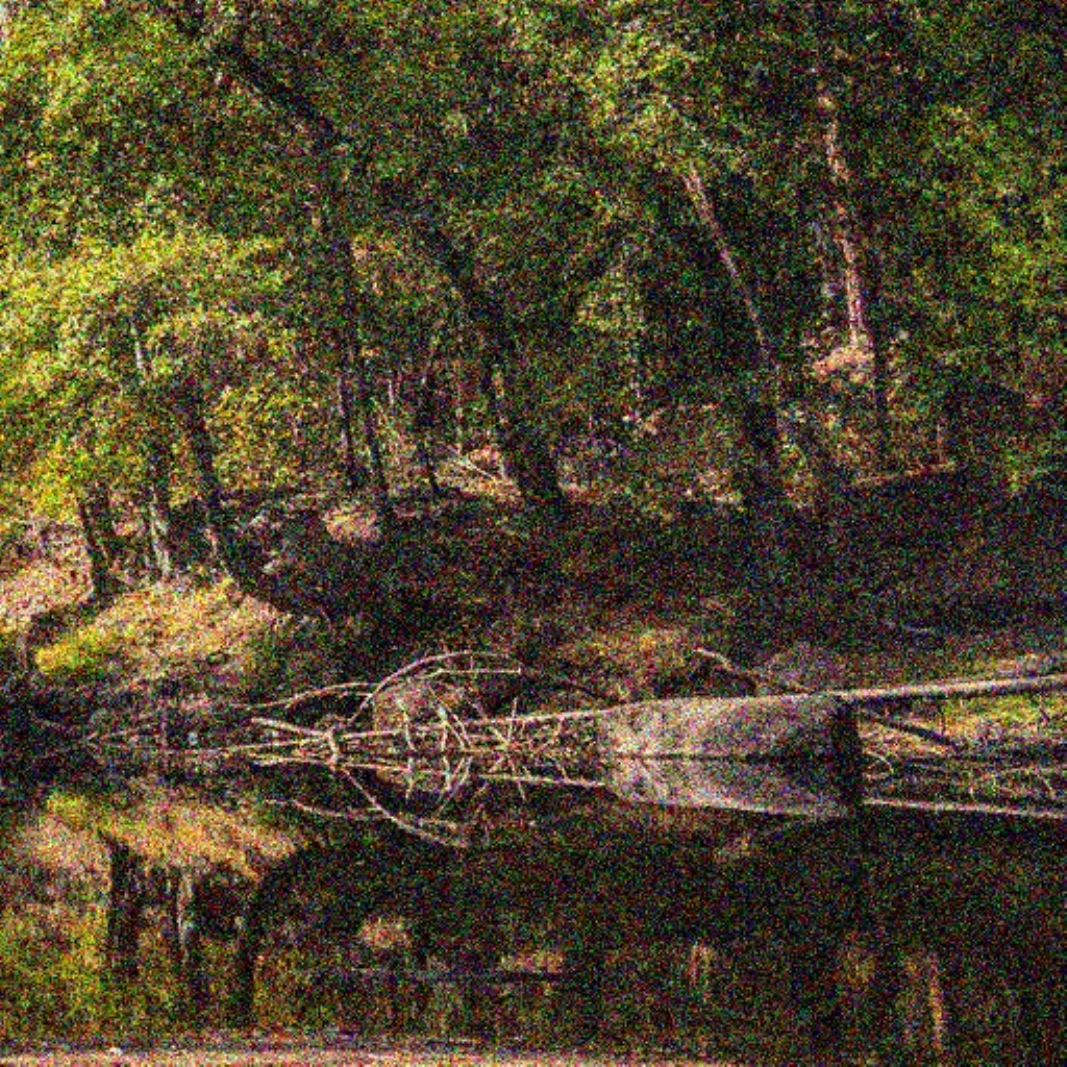} &   
\includegraphics[width=.25\textwidth,valign=t]{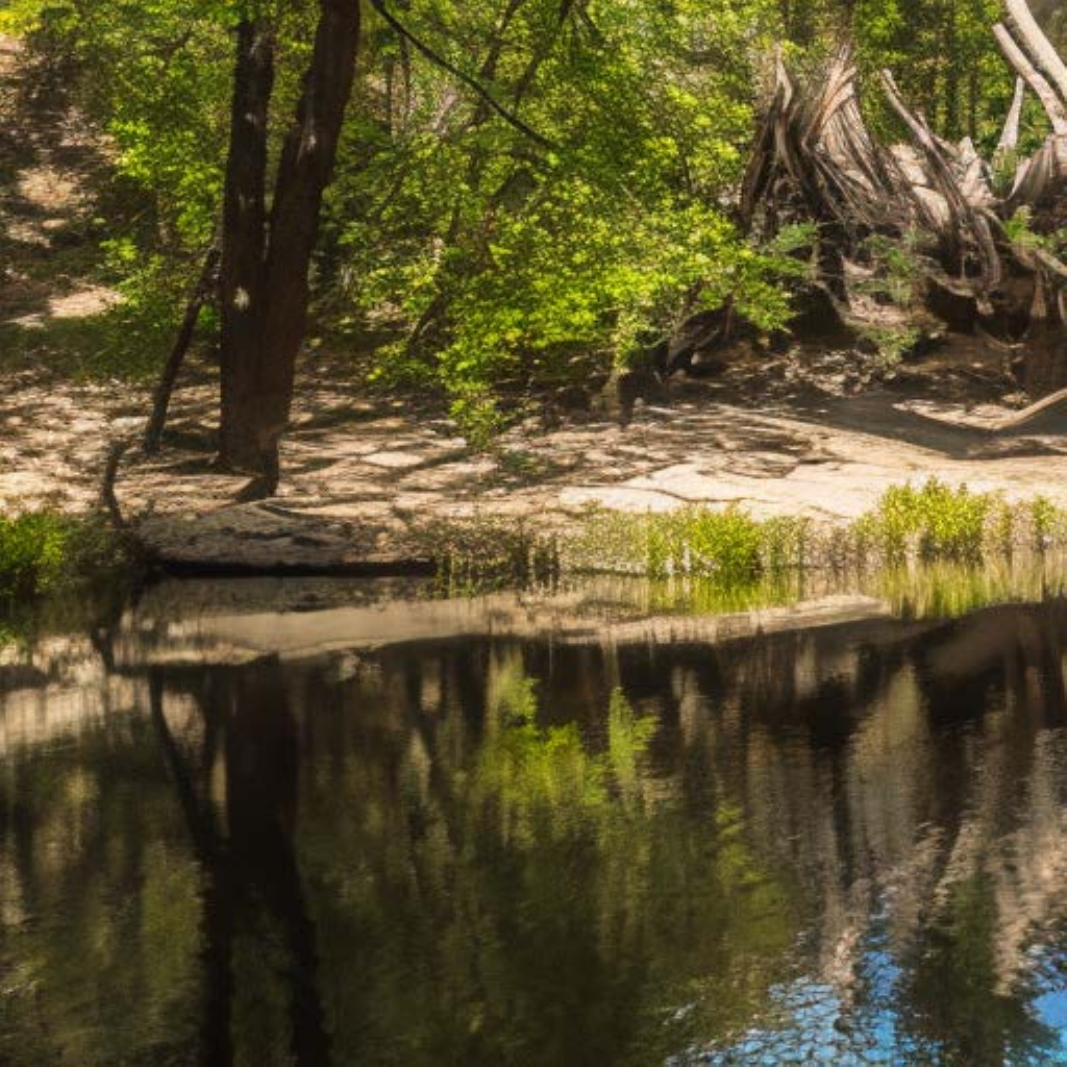} \\
\vspace{0.6em}
\small~Degraded  & \small~Guidance & \small~Degraded & \small~Guidance \\
\end{tabular}
}
\caption{\small Illustration of guidance images for image denoising task.}
\label{fig:suppl_guidance_images_denoise}
\end{center}
\vspace{-7mm}
\end{figure*}

%% file: suppl_figures/suppl_001_01_all_in_one_deraining_visual_results.tex
\begin{figure*}[t]
\begin{center}
\setlength{\tabcolsep}{1.5pt}
\scalebox{0.88}{
\begin{tabular}[b]{c c c c}
\vspace{0.4em}
\includegraphics[width=.2\textwidth,valign=t]{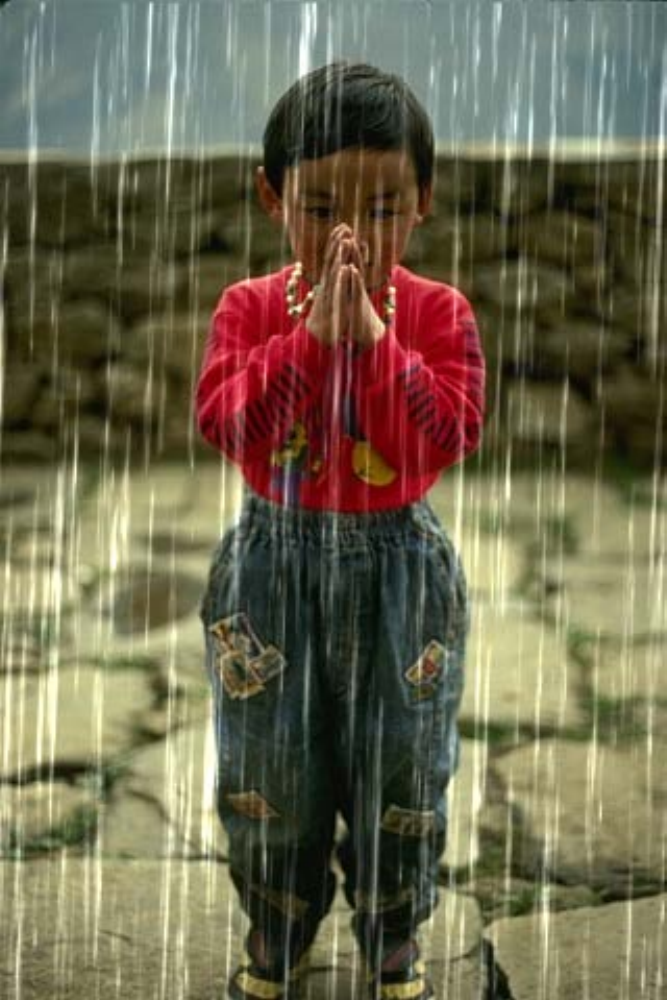} &   
\includegraphics[width=.2\textwidth,valign=t]{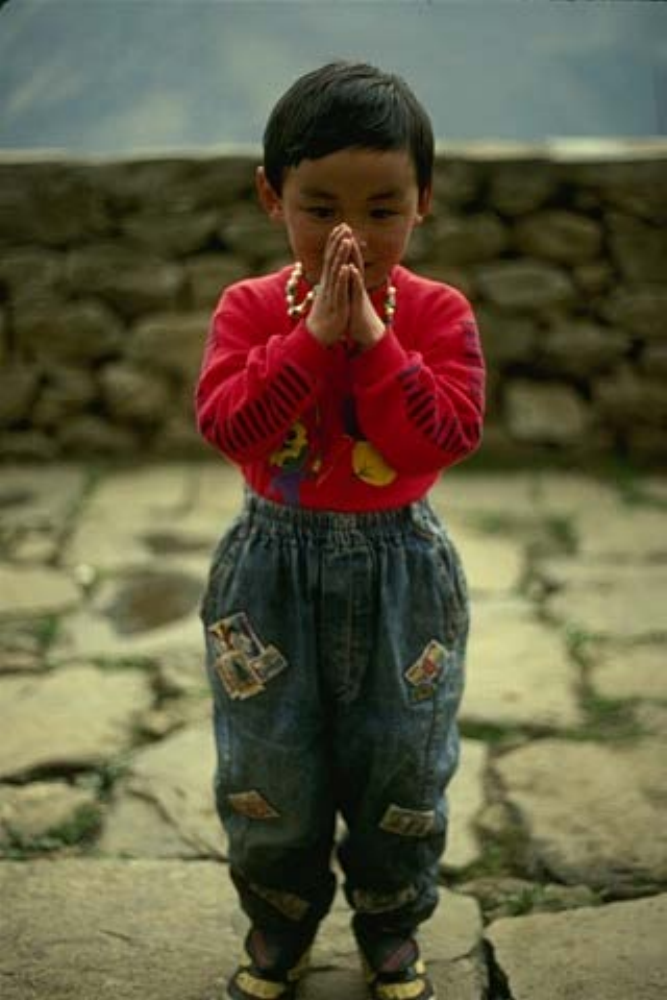} &   
\includegraphics[width=.2\textwidth,valign=t]{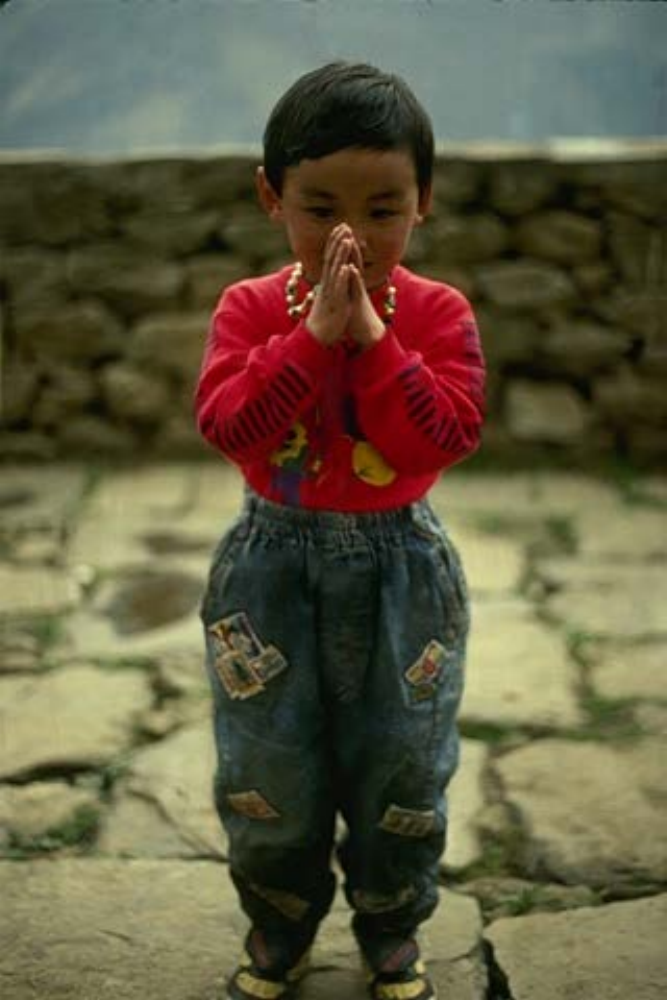} &   
\includegraphics[width=.2\textwidth,valign=t]{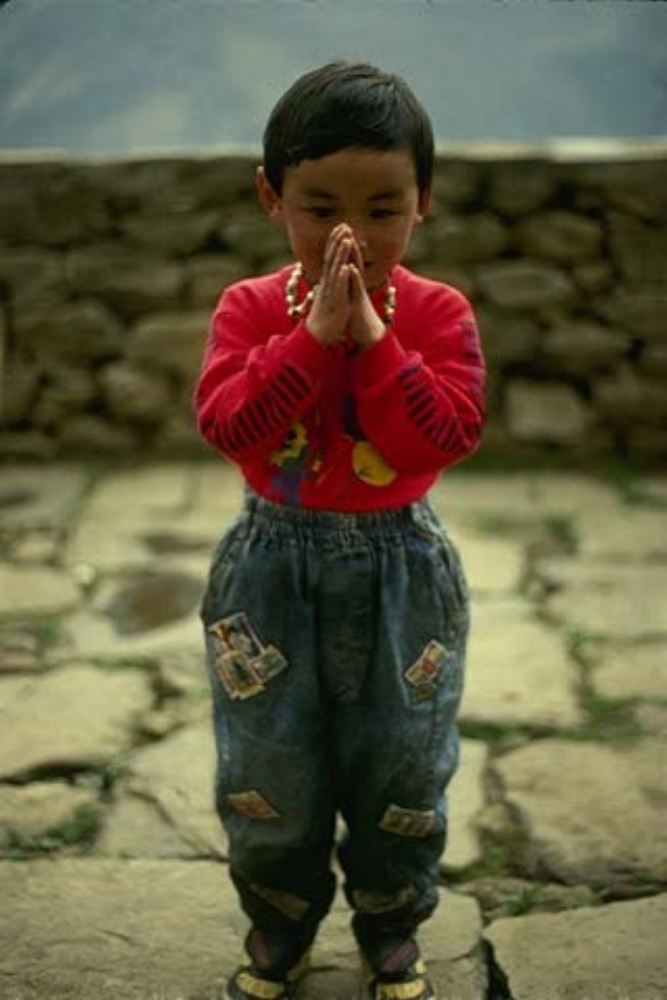}   \\
\vspace{0.4em}
\small~Degraded  & \small~Label & \small~PromptIR~\cite{PromptIR}, 37.68 dB  & \small~\textbf{Ours}, \textbf{39.06 dB}\\
\vspace{0.4em}
\includegraphics[width=.2\textwidth,valign=t]{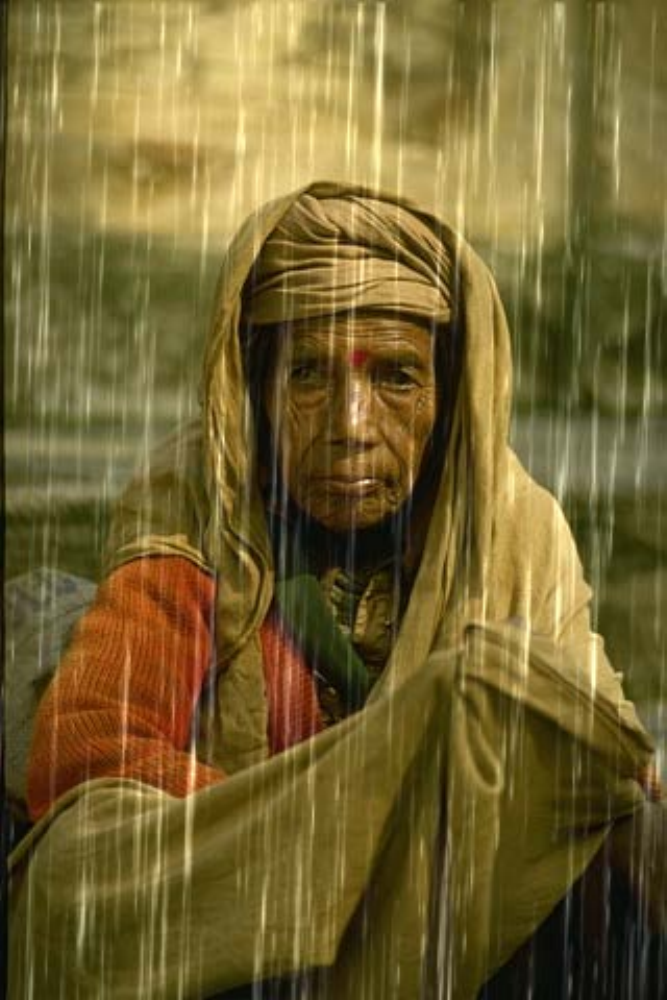} &   
\includegraphics[width=.2\textwidth,valign=t]{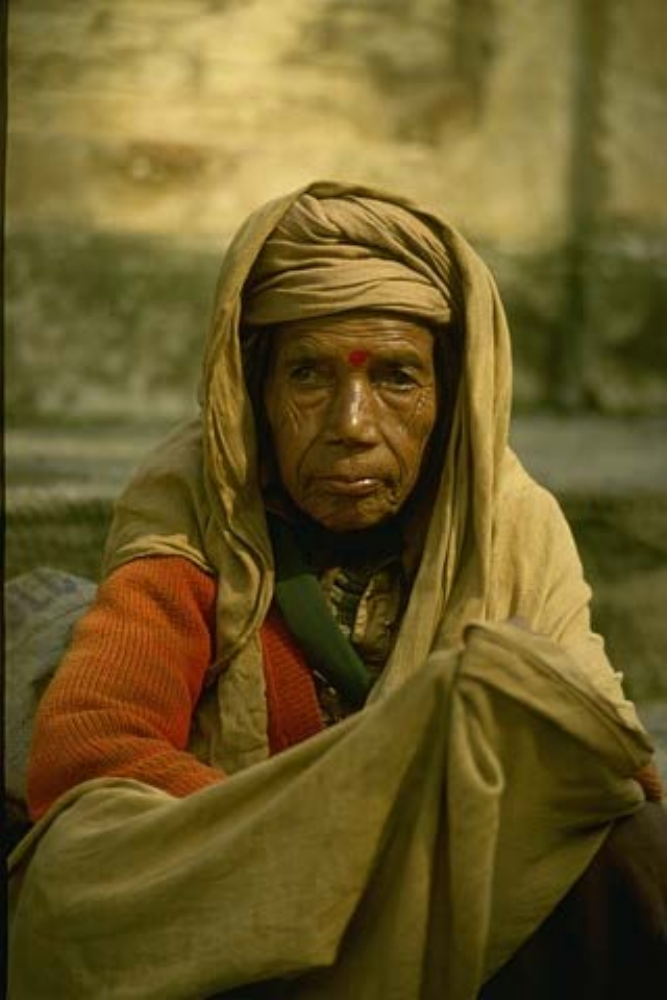} &   
\includegraphics[width=.2\textwidth,valign=t]{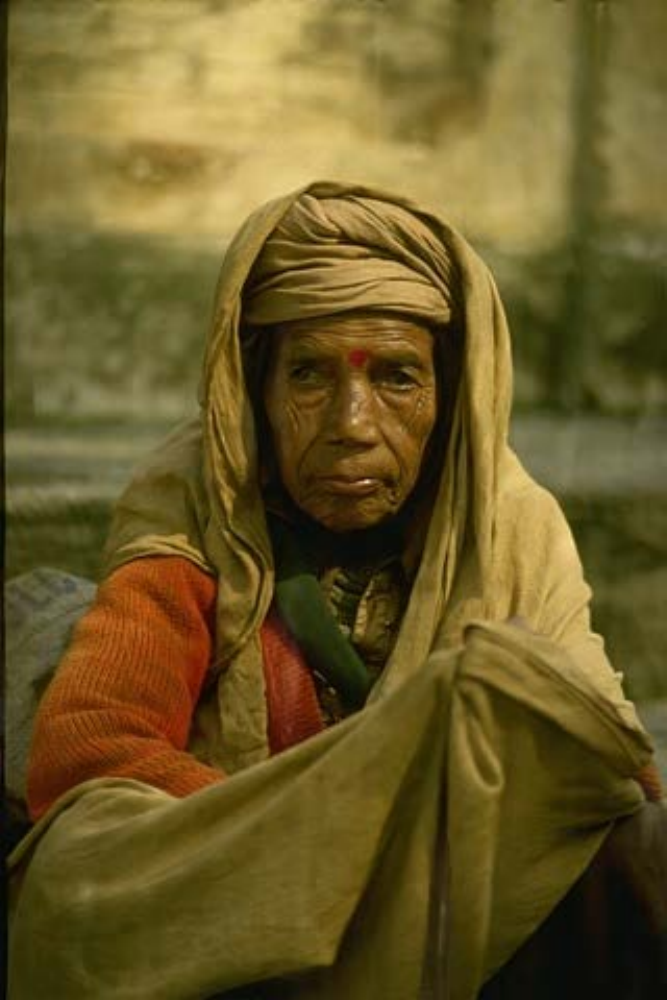} &   
\includegraphics[width=.2\textwidth,valign=t]{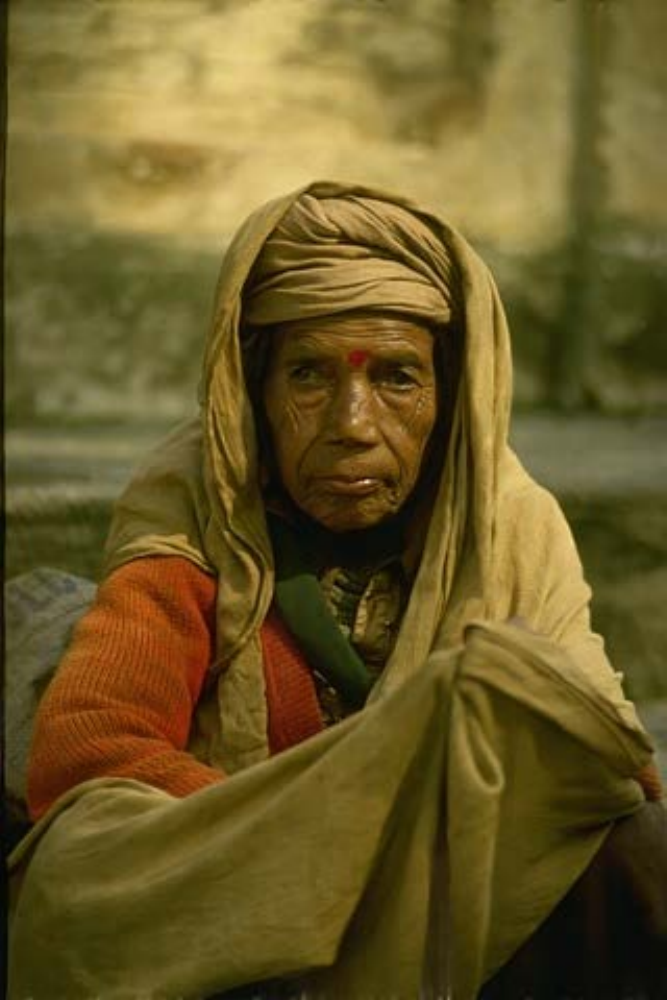}   \\
\vspace{0.4em}
\small~Degraded  & \small~Label & \small~PromptIR~\cite{PromptIR}, 38.50 dB  & \small~\textbf{Ours}, \textbf{40.50 dB} \\
\vspace{0.4em}
\includegraphics[width=.2\textwidth,valign=t]{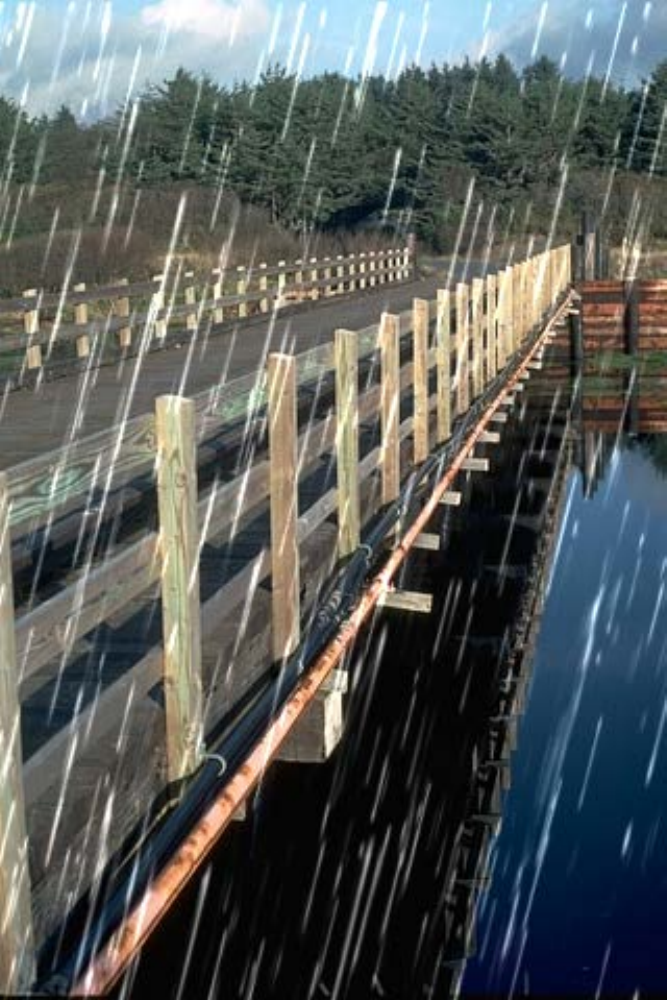} &   
\includegraphics[width=.2\textwidth,valign=t]{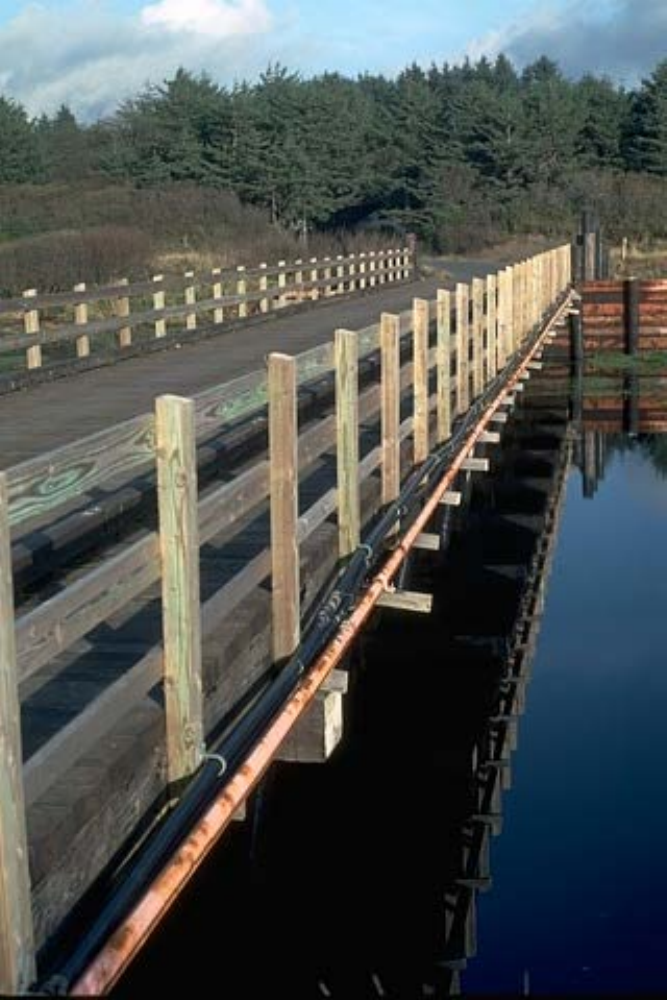} &   
\includegraphics[width=.2\textwidth,valign=t]{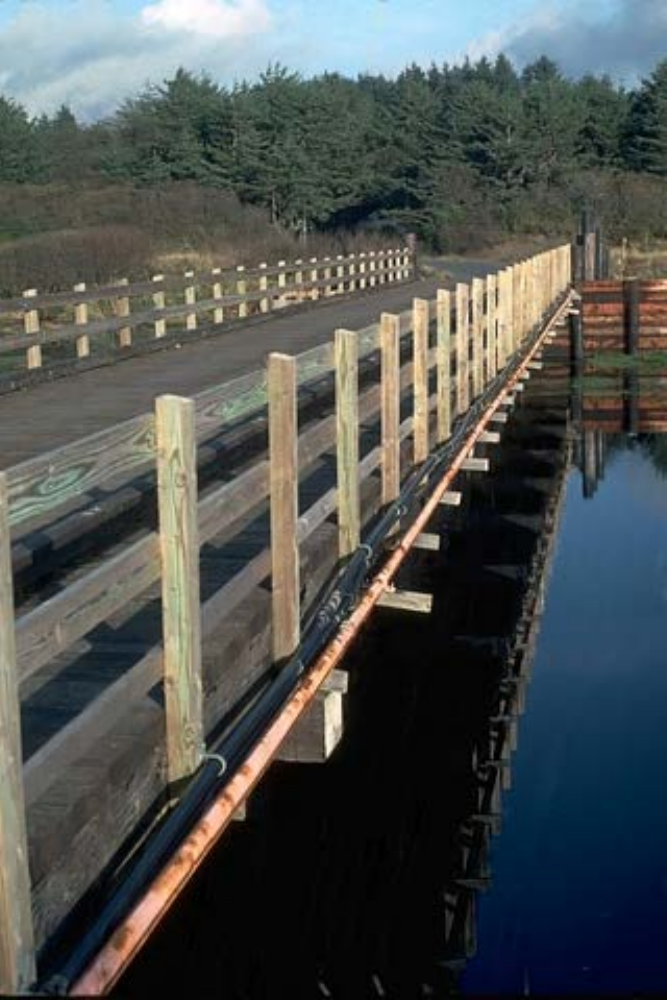} &   
\includegraphics[width=.2\textwidth,valign=t]{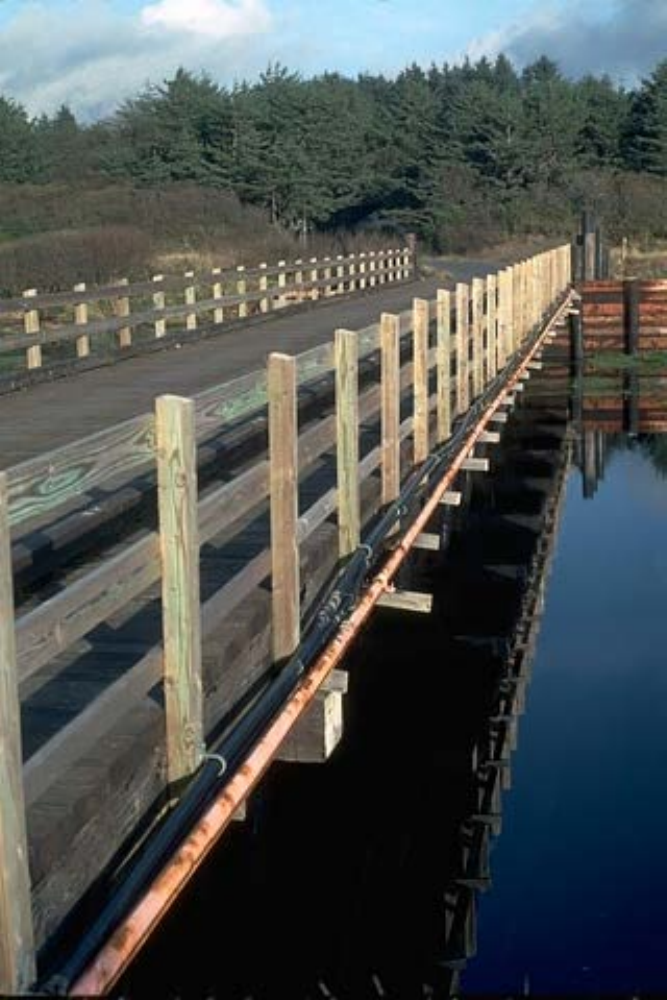}   \\
\vspace{0.4em}
\small~Degraded  & \small~Label & \small~PromptIR~\cite{PromptIR}, 36.59 dB  & \small~\textbf{Ours}, \textbf{38.13 dB} \\
\end{tabular}
}
\caption{\small Image Deraining on Rain100L~\cite{Rain200H}.}
\label{fig:suppl_all_in_one_deraining_visual_results_1}
\end{center}
\vspace{-7mm}
\end{figure*}

%% file: suppl_figures/suppl_001_02_all_in_one_deraining_visual_results.tex
\begin{figure*}[t]
\begin{center}
\setlength{\tabcolsep}{1.5pt}
\scalebox{0.88}{
\begin{tabular}[b]{c c c c}
\vspace{0.4em}
\includegraphics[width=.25\textwidth,valign=t]{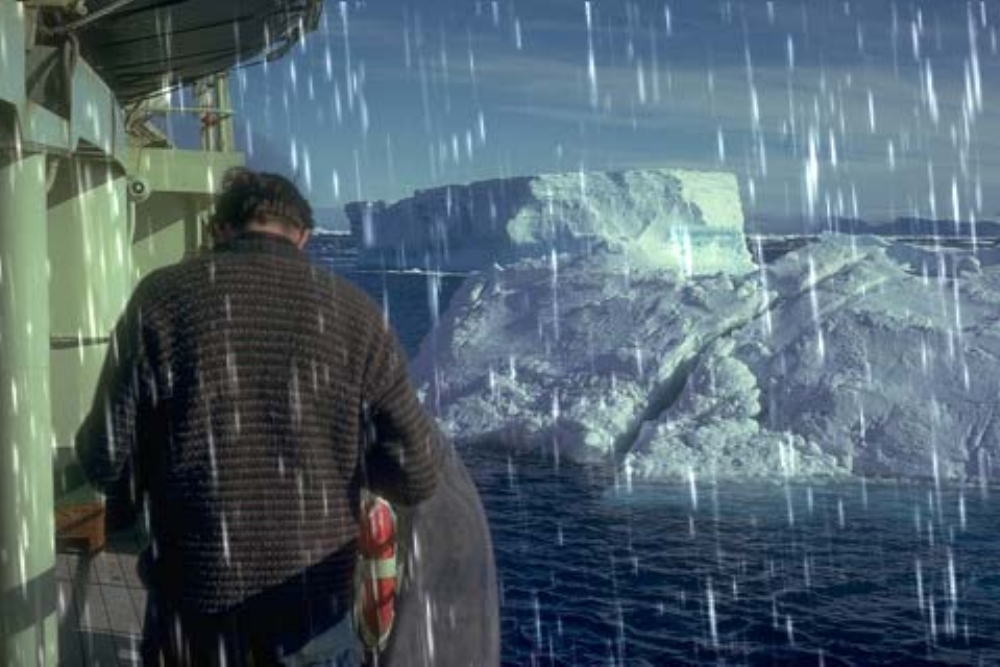} &   
\includegraphics[width=.25\textwidth,valign=t]{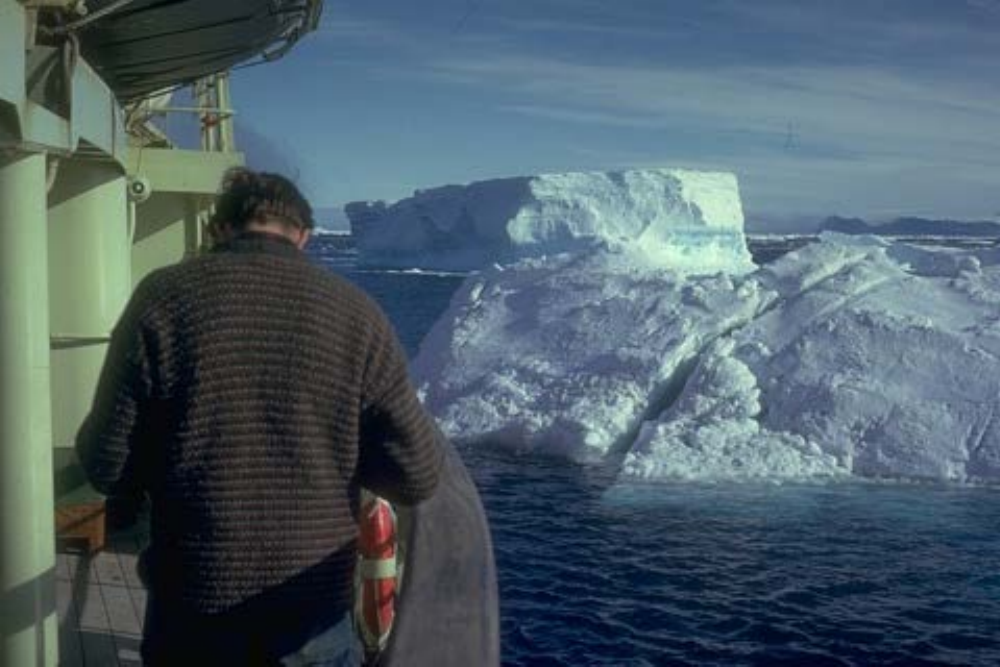} &   
\includegraphics[width=.25\textwidth,valign=t]{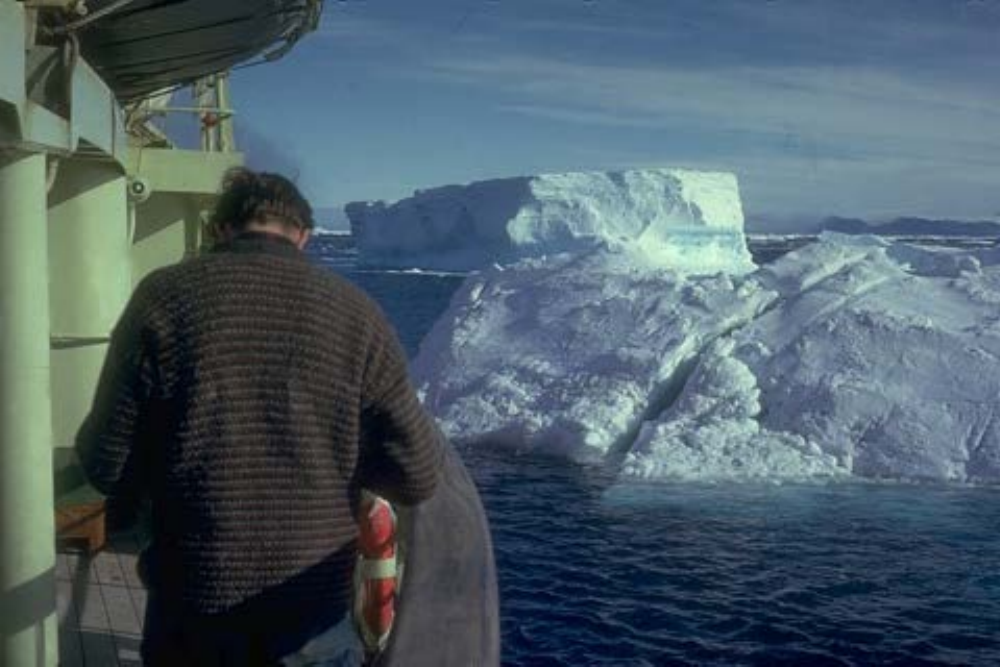} &   
\includegraphics[width=.25\textwidth,valign=t]{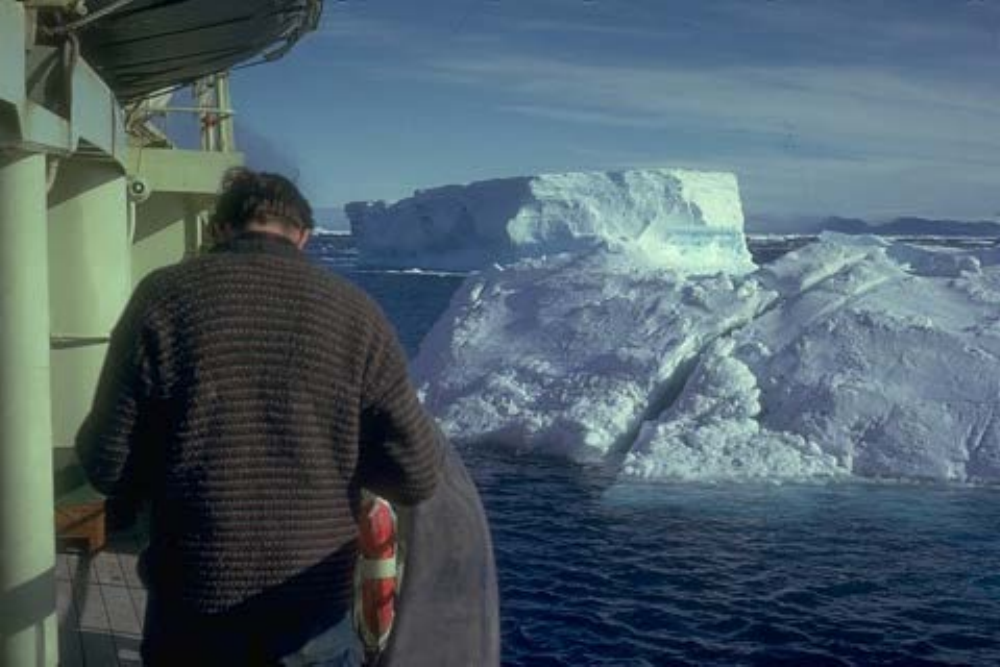}   \\
\vspace{0.4em}
\small~Degraded  & \small~Label & \small~PromptIR~\cite{PromptIR}, 39.70 dB & \small~\textbf{Ours}, \textbf{41.57 dB} \\
\vspace{0.4em}
\includegraphics[width=.25\textwidth,valign=t]{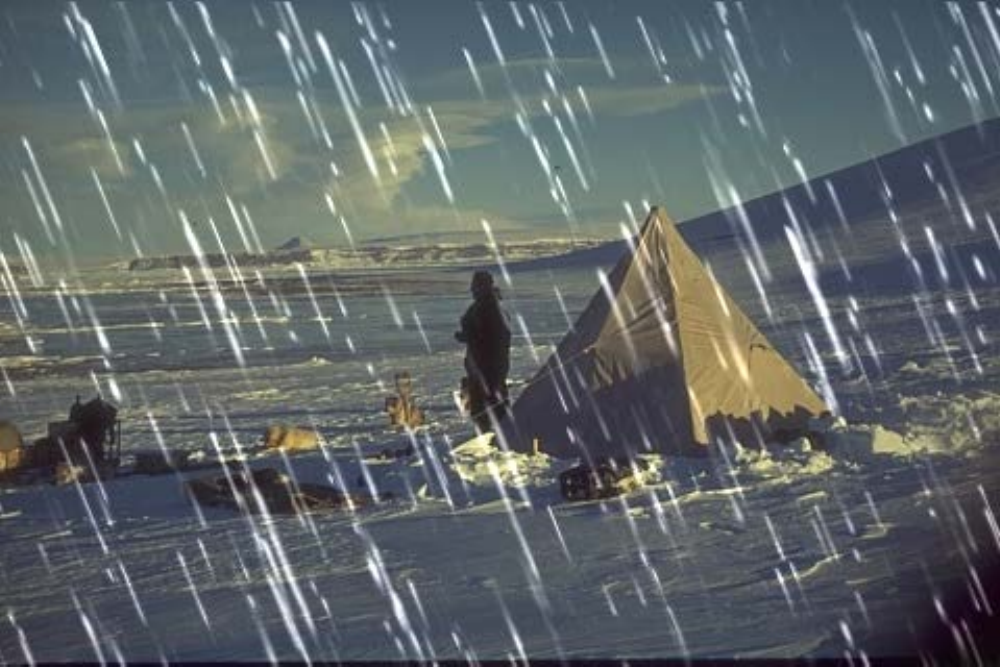} &   
\includegraphics[width=.25\textwidth,valign=t]{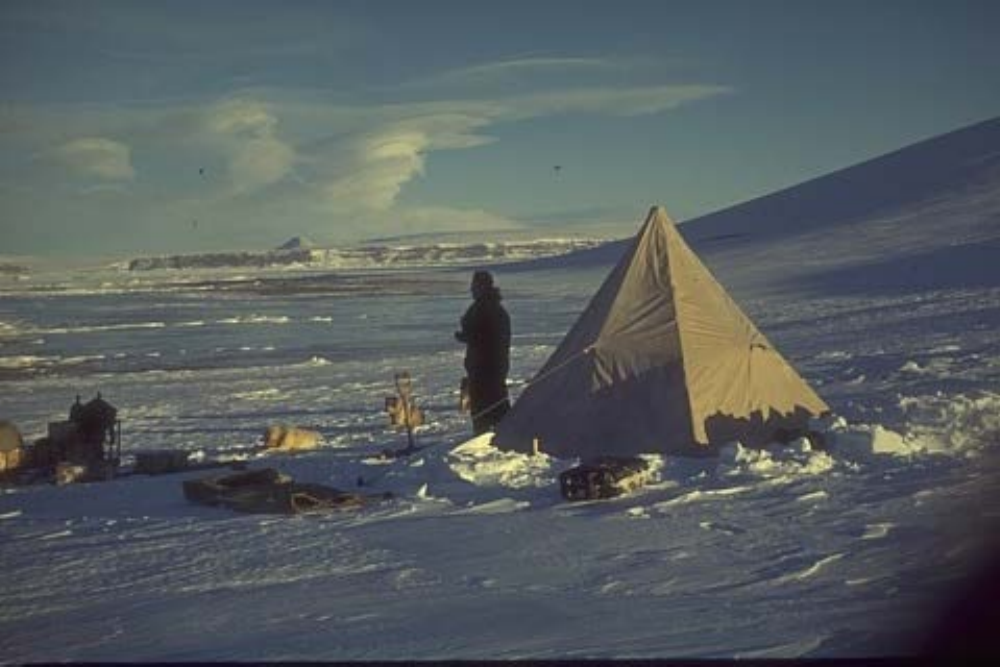} &   
\includegraphics[width=.25\textwidth,valign=t]{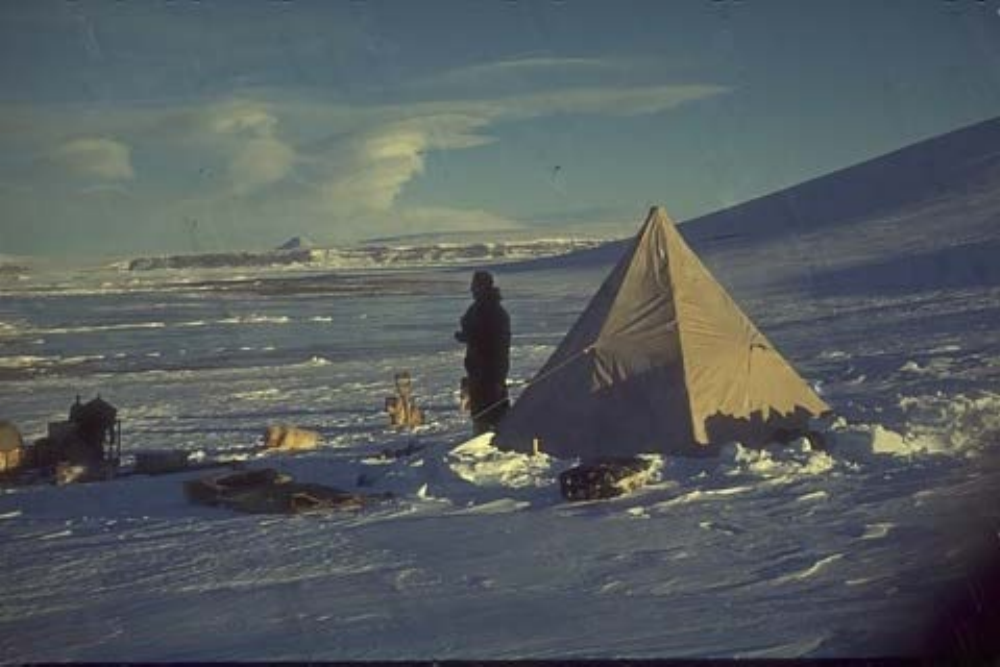} &   
\includegraphics[width=.25\textwidth,valign=t]{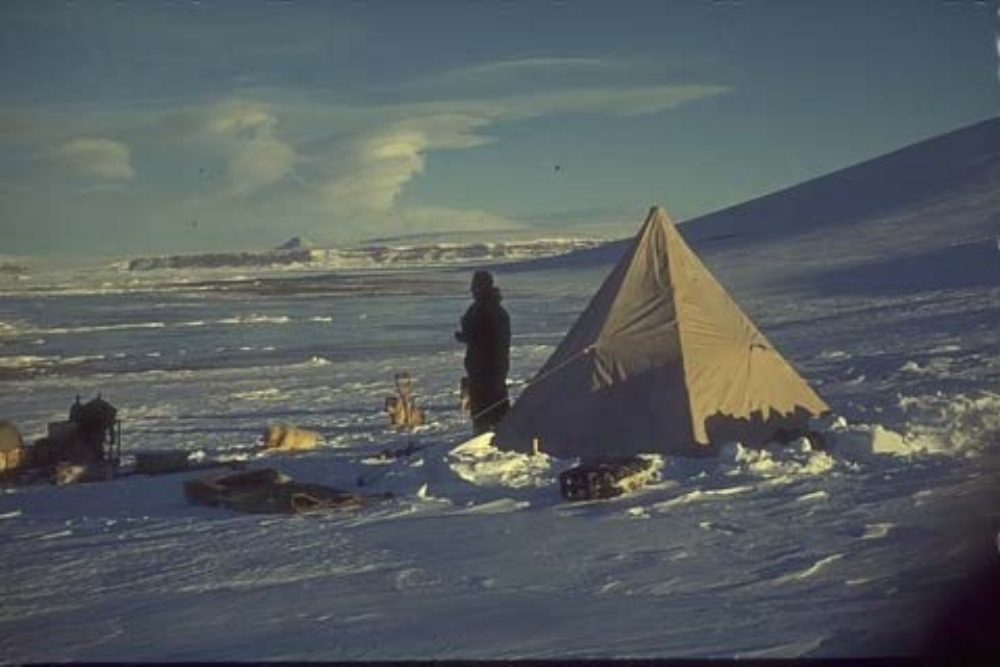}   \\
\vspace{0.4em}
\small~Degraded  & \small~Label & \small~PromptIR~\cite{PromptIR}, 40.99 dB & \small~\textbf{Ours}, \textbf{42.92 dB} \\
\vspace{0.4em}
\includegraphics[width=.25\textwidth,valign=t]{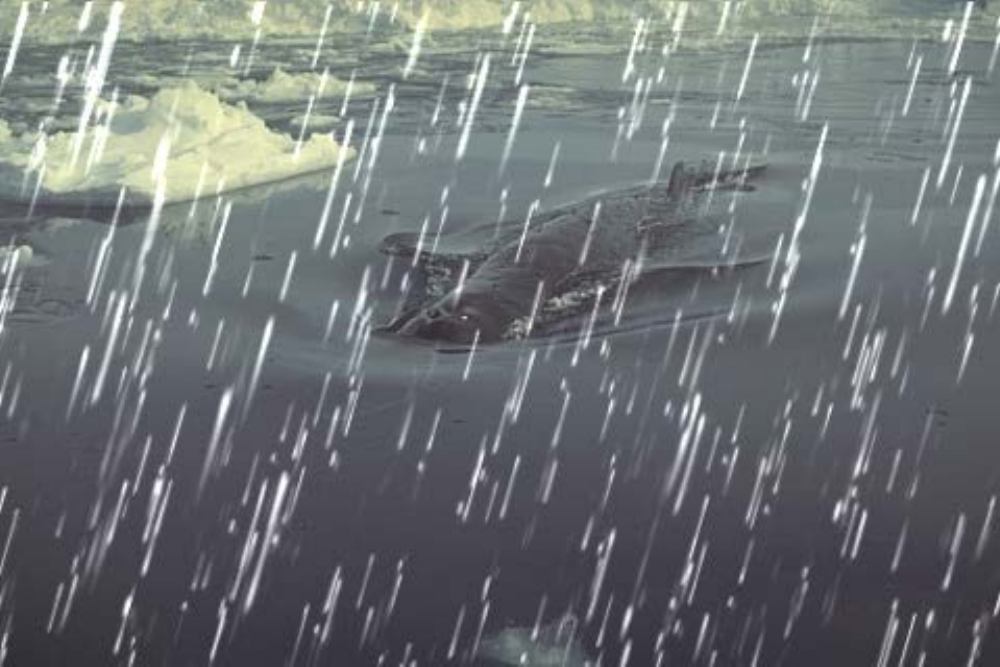} &   
\includegraphics[width=.25\textwidth,valign=t]{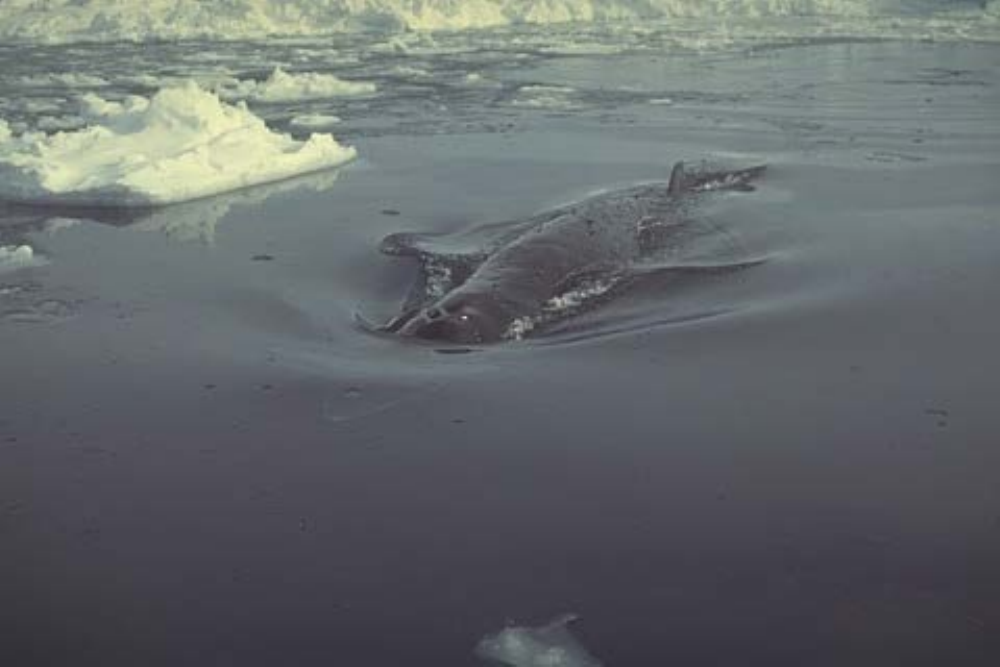} &   
\includegraphics[width=.25\textwidth,valign=t]{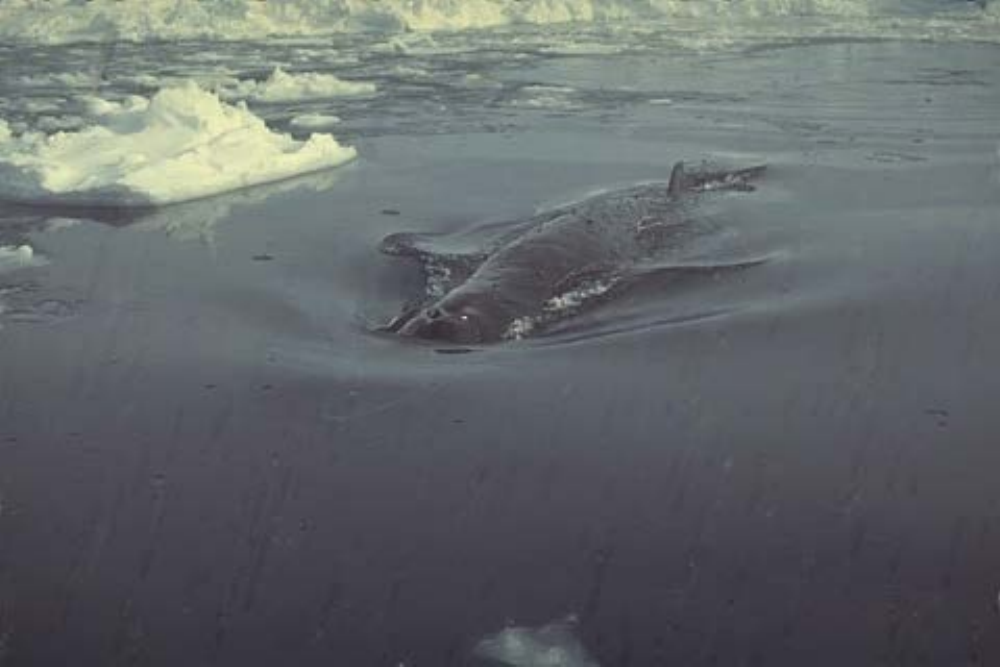} &   
\includegraphics[width=.25\textwidth,valign=t]{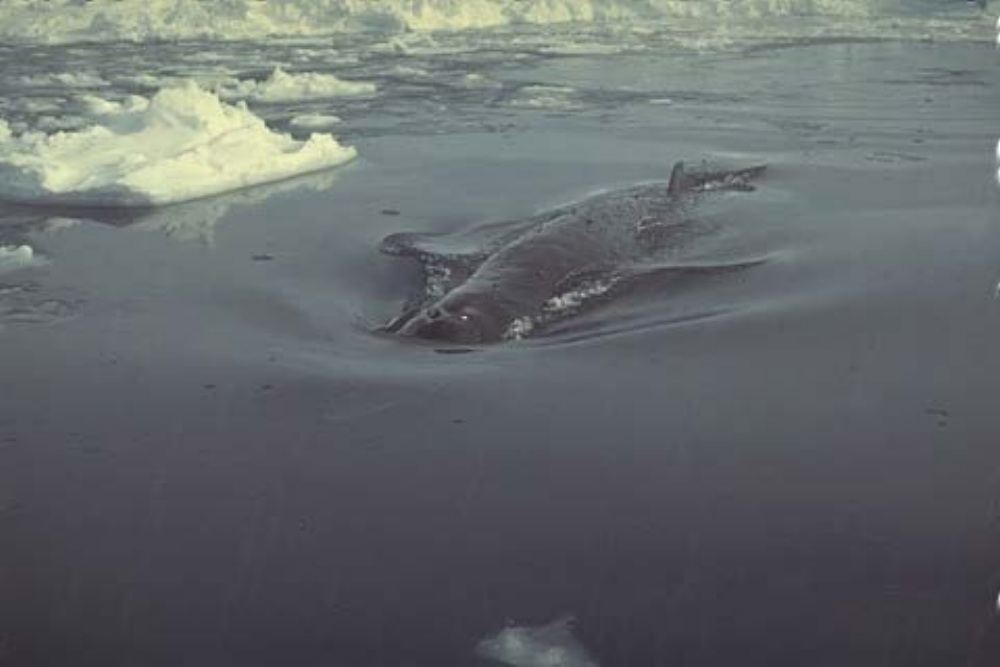}   \\
\vspace{0.4em}
\small~Degraded  & \small~Label & \small~PromptIR~\cite{PromptIR}, 40.81 dB & \small~\textbf{Ours}, \textbf{41.09 dB} \\
\end{tabular}
}
\caption{\small Image Deraining on Rain100L~\cite{Rain200H}.}
\label{fig:suppl_all_in_one_deraining_visual_results_2}
\end{center}
\vspace{-7mm}
\end{figure*}

%% file: suppl_figures/suppl_002_all_in_one_dehazing_visual_results.tex
\begin{figure*}[t]
\begin{center}
\setlength{\tabcolsep}{1.5pt}
\scalebox{0.88}{
\begin{tabular}[b]{c c c c}
\vspace{0.4em}
\includegraphics[width=.25\textwidth,valign=t]{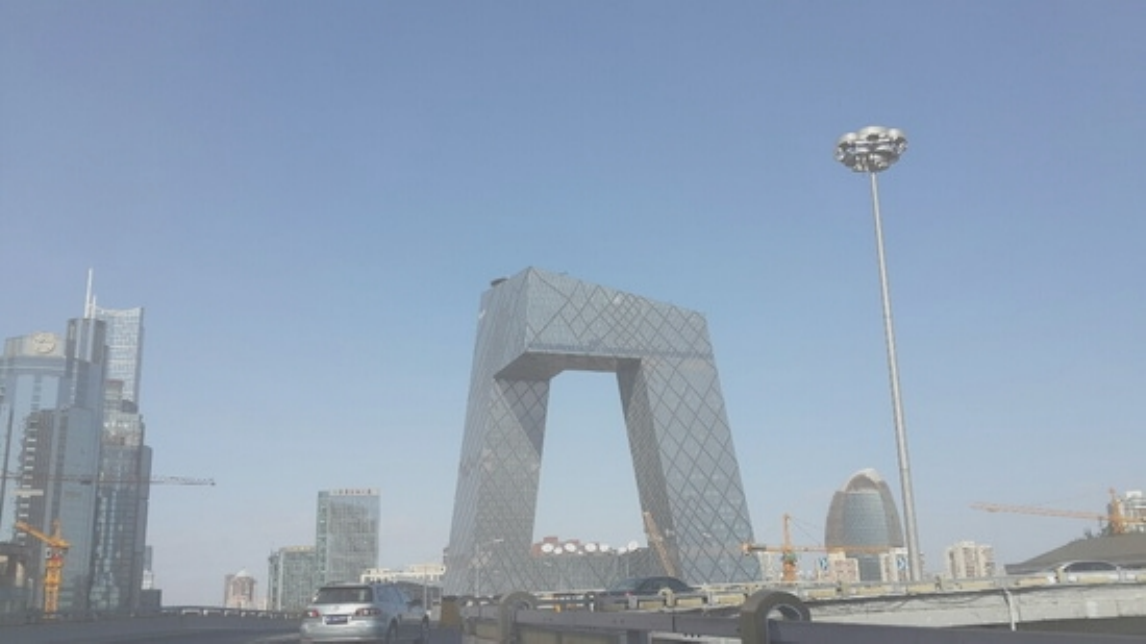} &   
\includegraphics[width=.25\textwidth,valign=t]{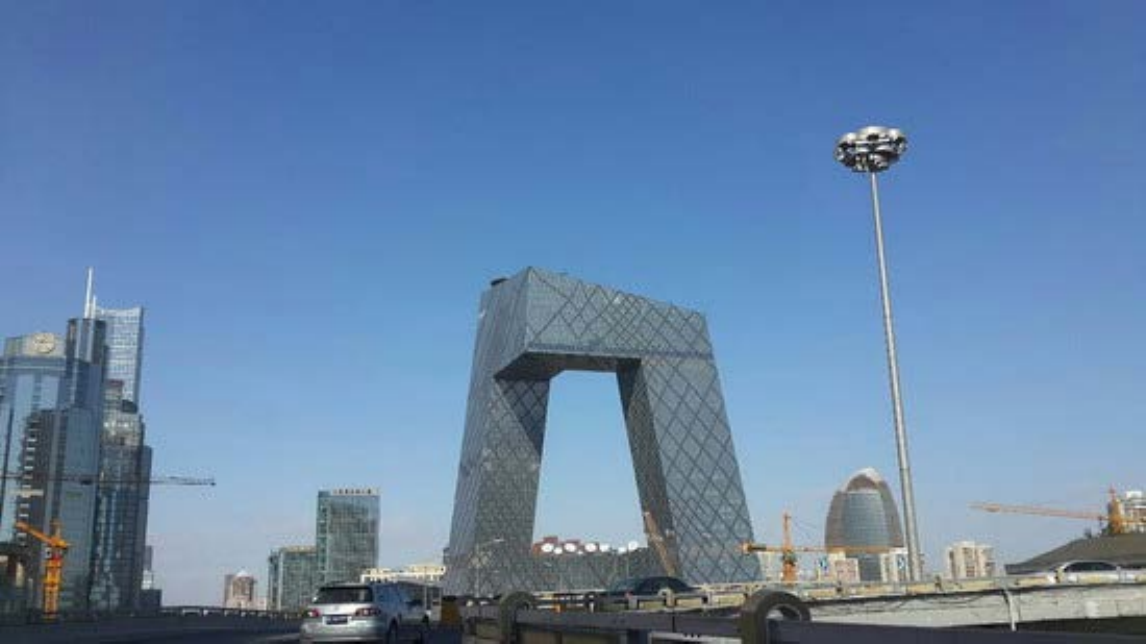} &   
\includegraphics[width=.252\textwidth,valign=t]{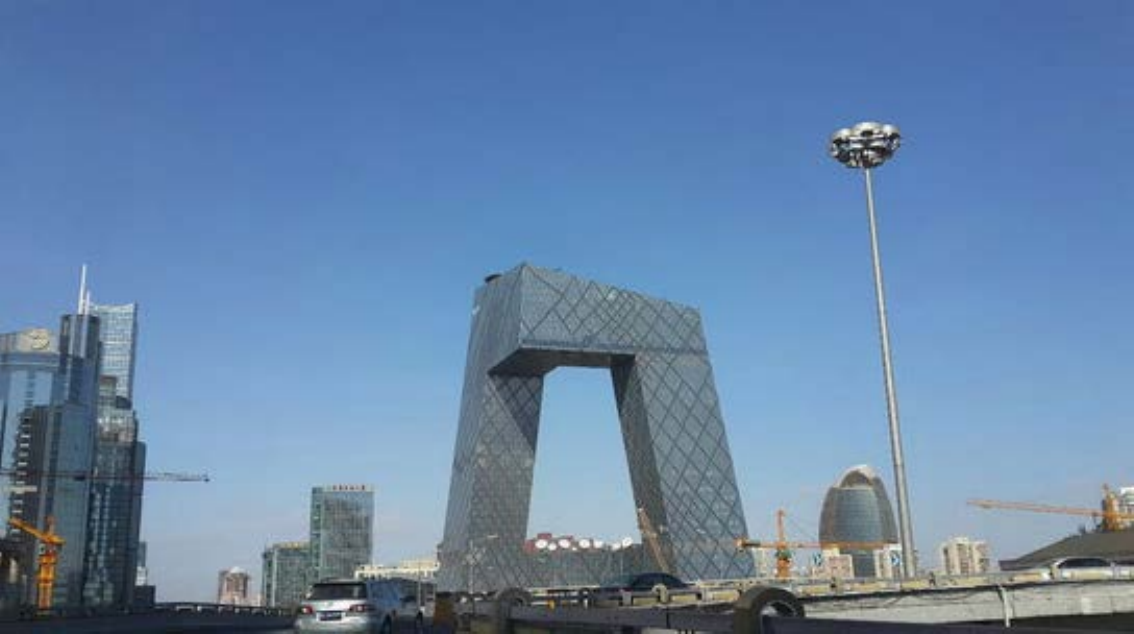} &   
\includegraphics[width=.25\textwidth,valign=t]{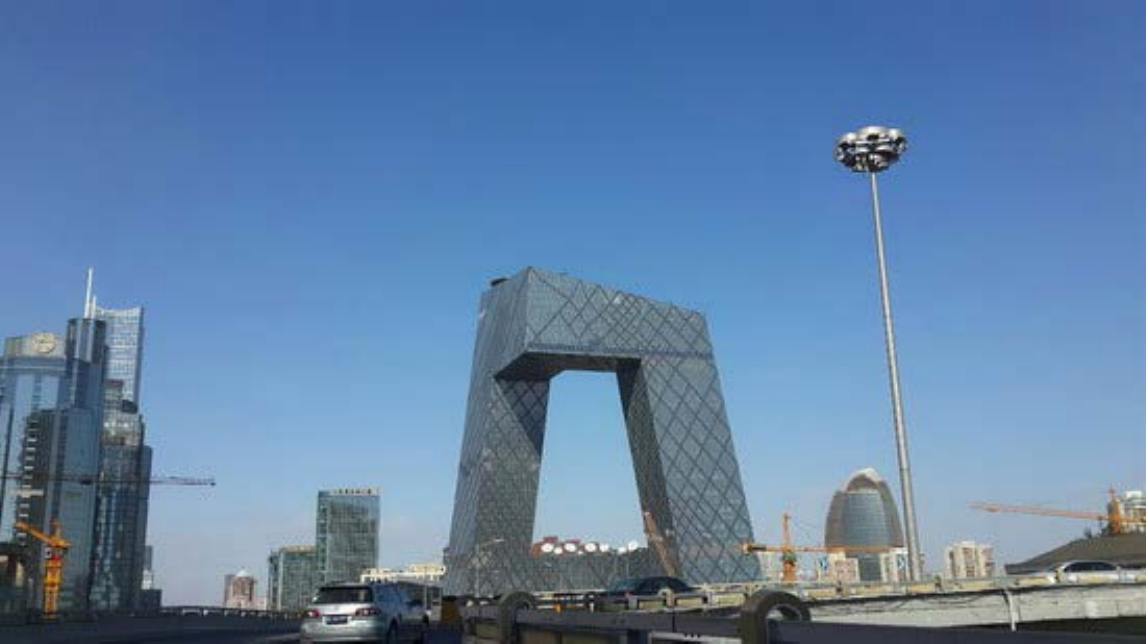}   \\
\vspace{0.4em}
\small~Degraded  & \small~Label & \small~PromptIR~\cite{PromptIR}, 36.58 dB & \small~\textbf{Ours}, \textbf{36.73 dB} \\
\vspace{0.4em}
\includegraphics[width=.25\textwidth,valign=t]{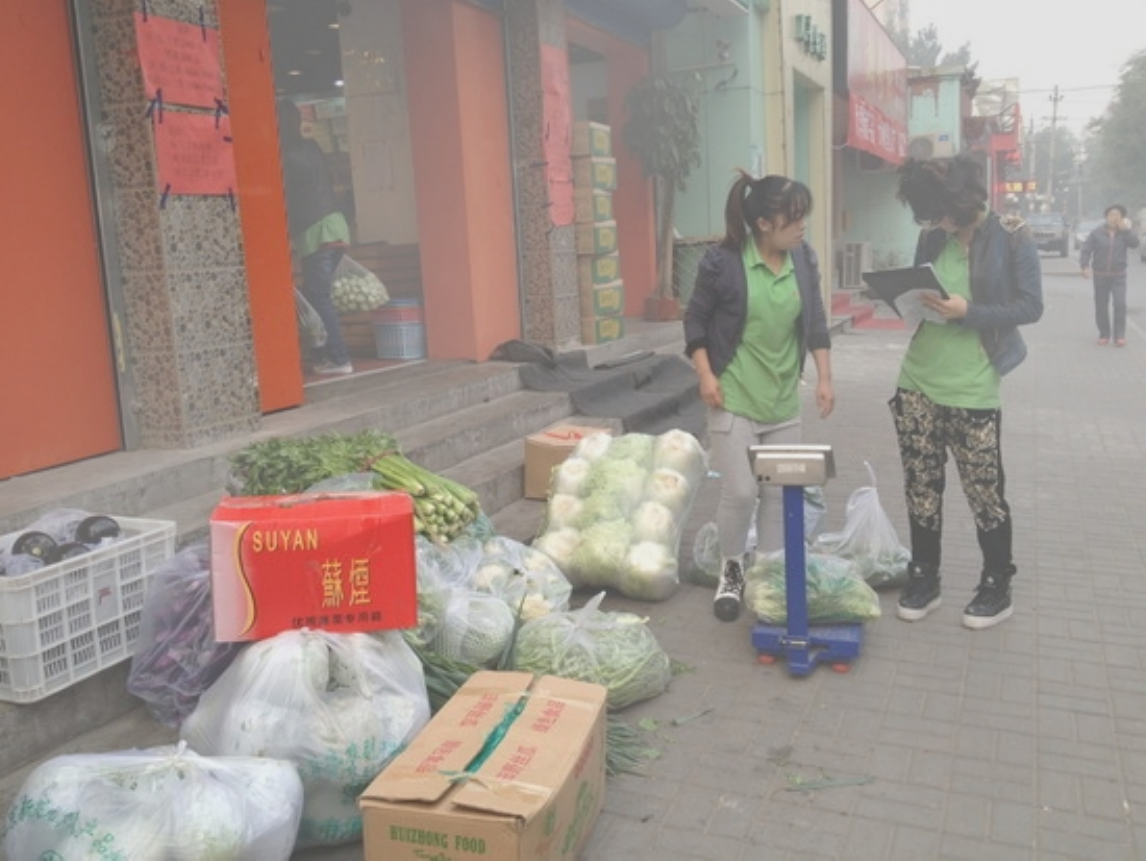} &   
\includegraphics[width=.25\textwidth,valign=t]{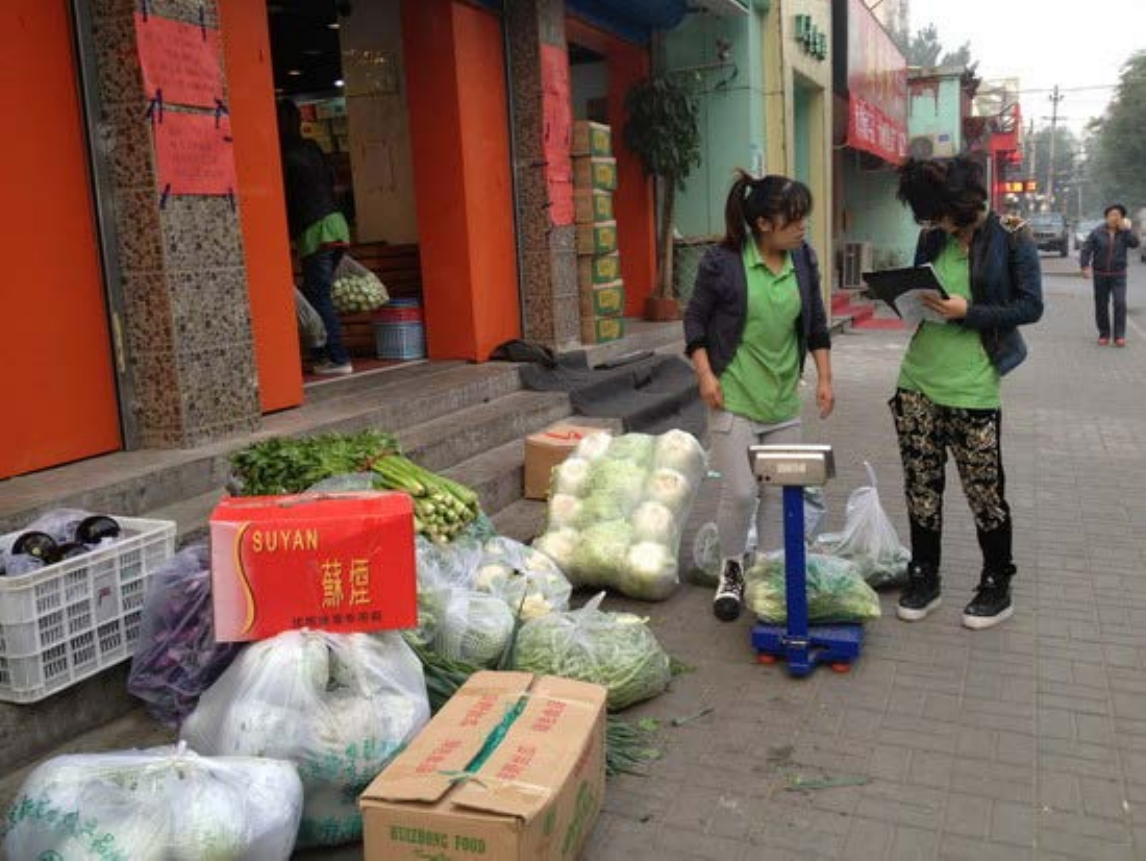} &   
\includegraphics[width=.254\textwidth,valign=t]{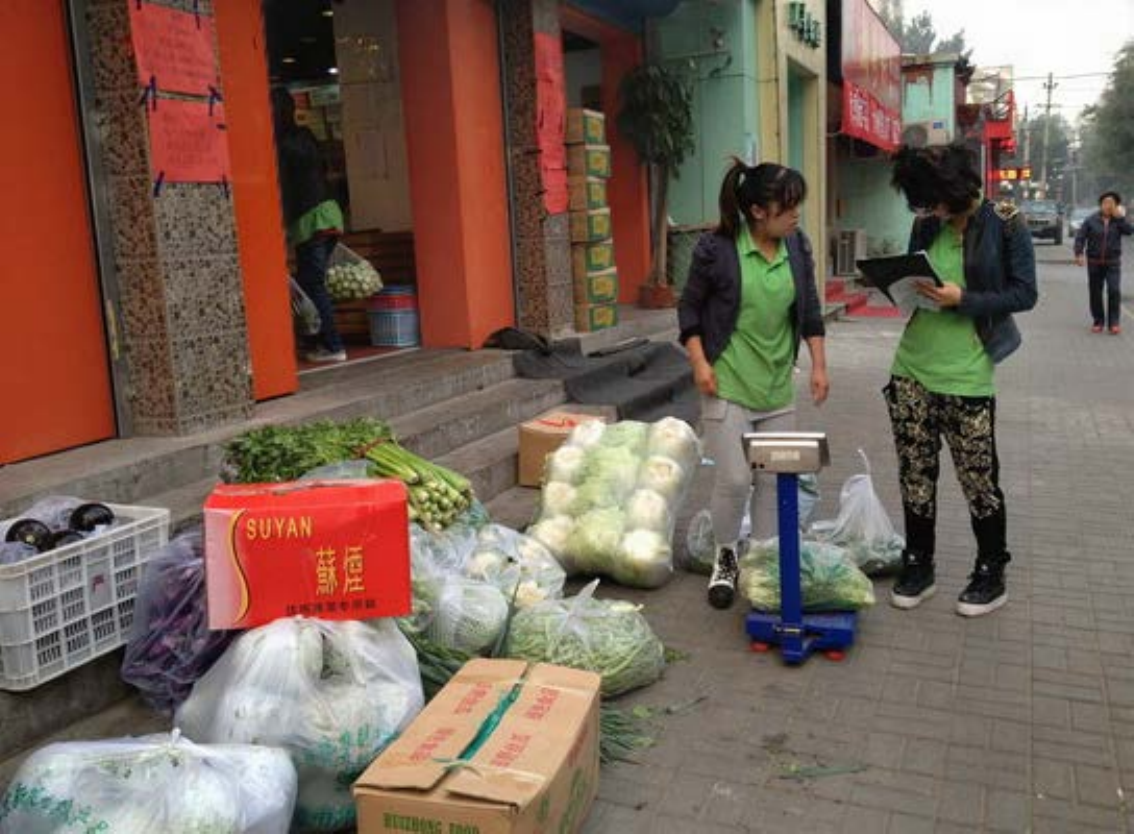} &   
\includegraphics[width=.25\textwidth,valign=t]{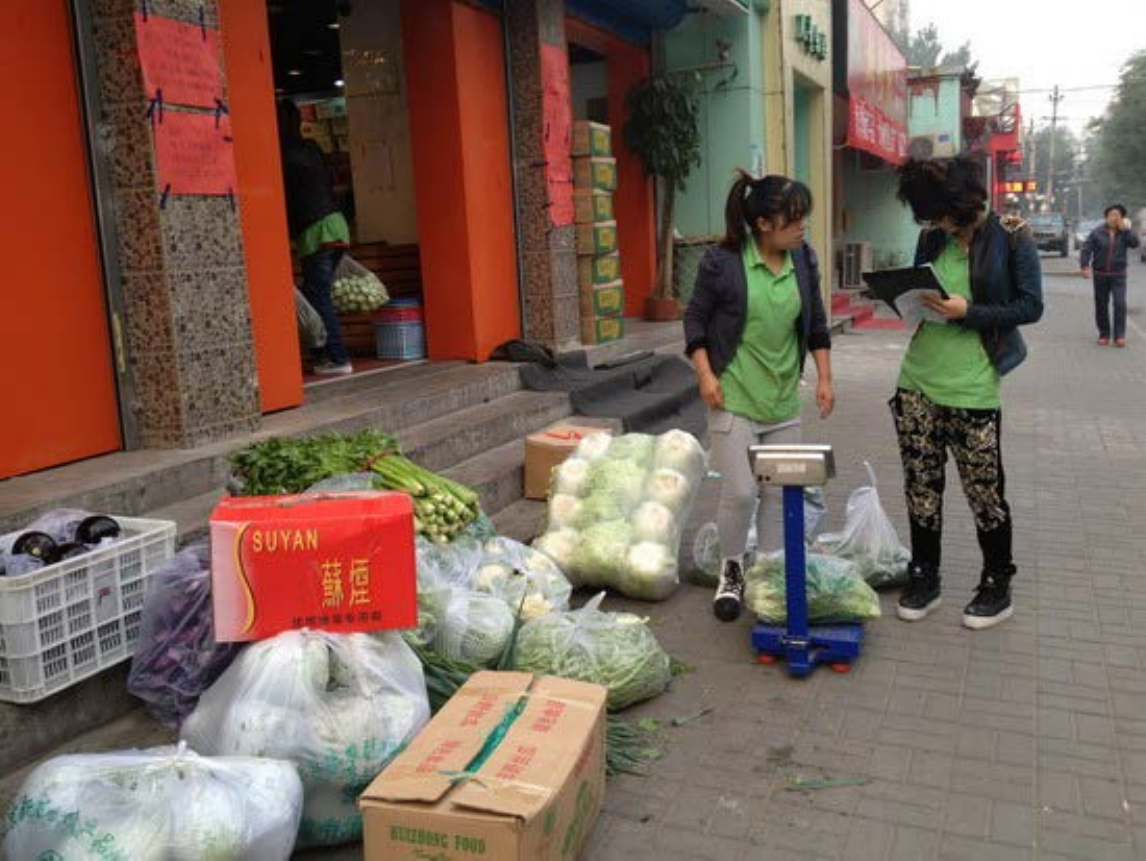}   \\
\vspace{0.4em}
\small~Degraded  & \small~Label & \small~PromptIR~\cite{PromptIR}, 39.69 dB & \small~\textbf{Ours}, \textbf{43.64 dB} \\
\vspace{0.4em}
\includegraphics[width=.25\textwidth,valign=t]{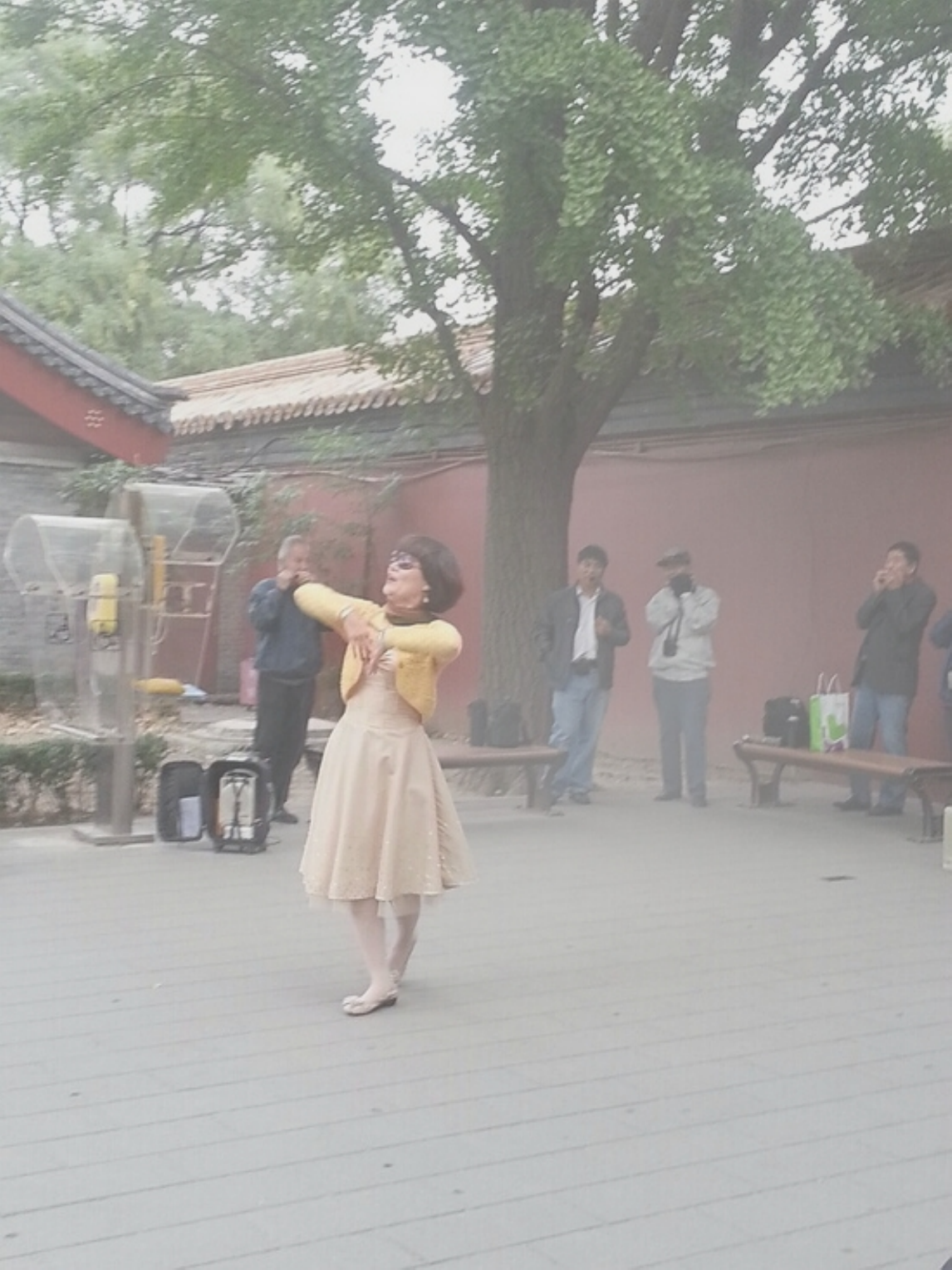} &   
\includegraphics[width=.25\textwidth,valign=t]{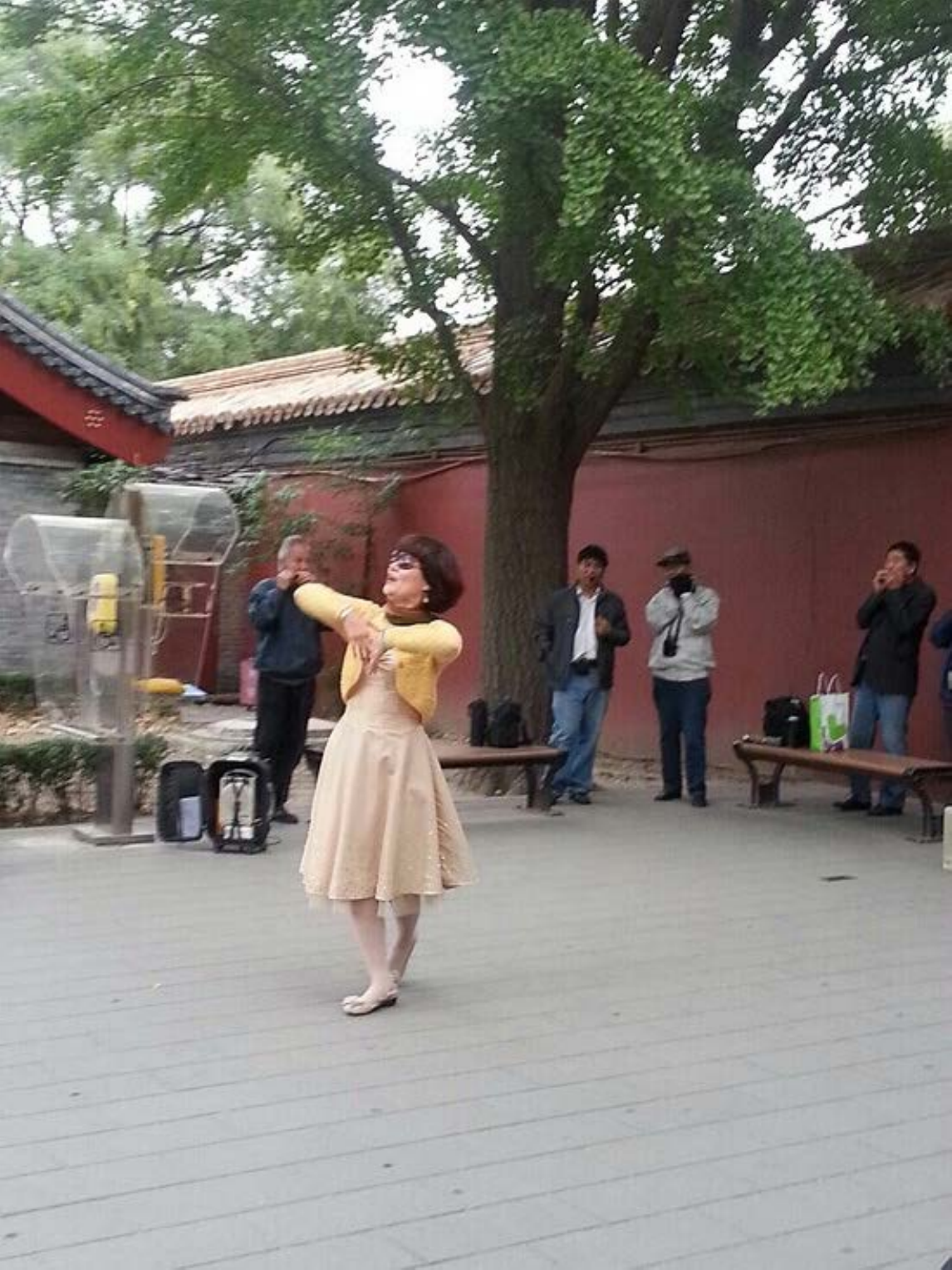} &   
\includegraphics[width=.252\textwidth,valign=t]{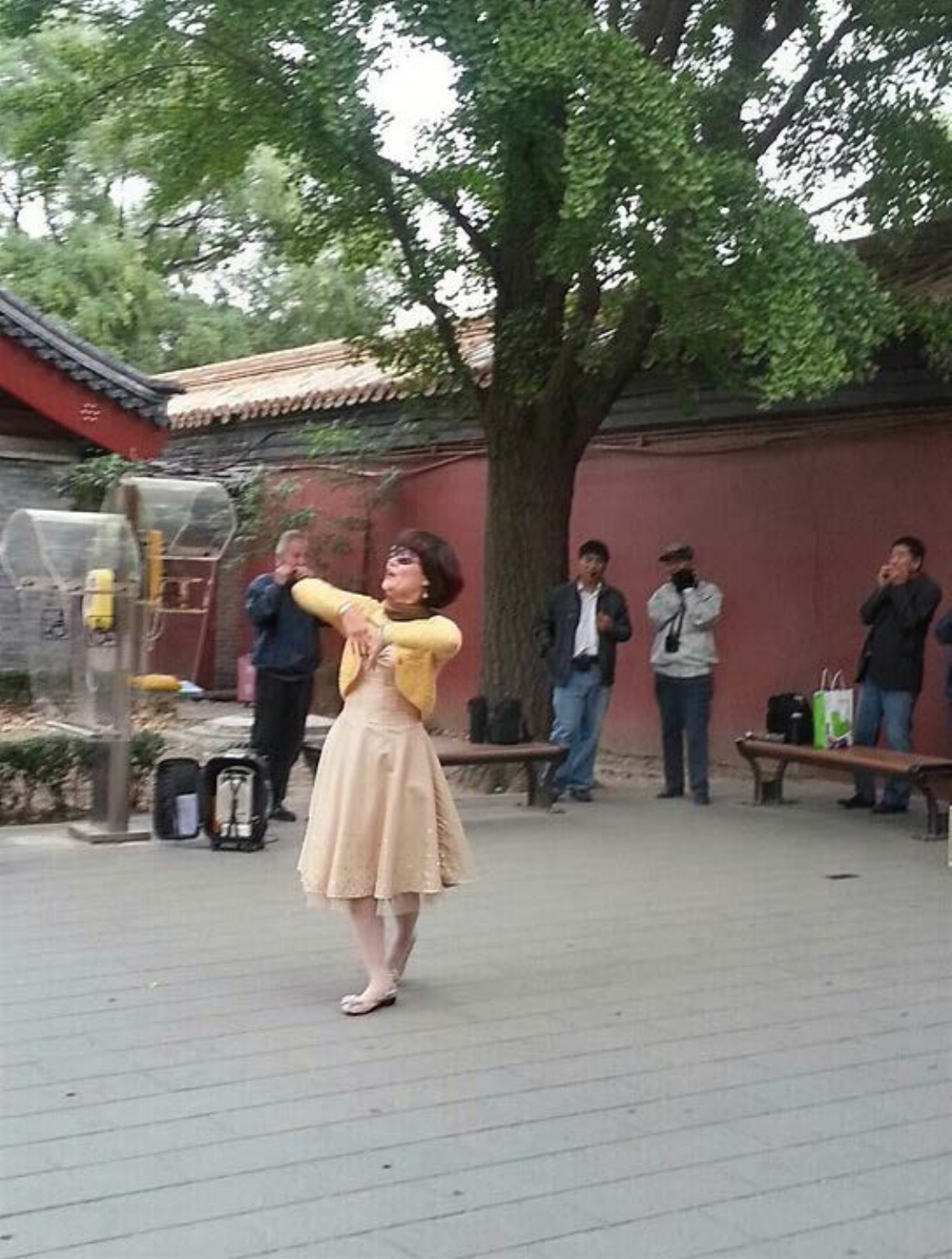} &   
\includegraphics[width=.25\textwidth,valign=t]{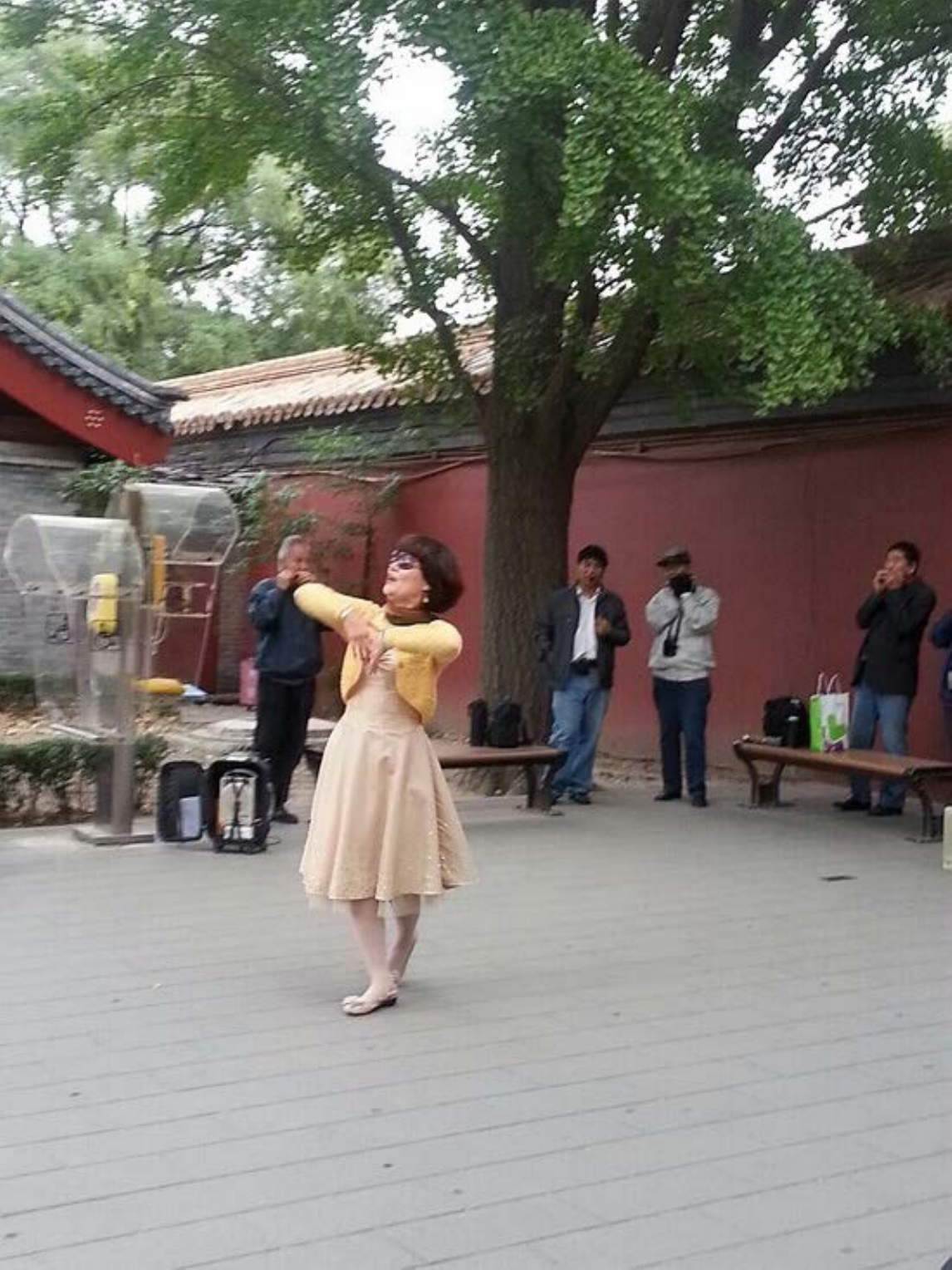}   \\
\vspace{0.4em}
\small~Degraded  & \small~Label & \small~PromptIR~\cite{PromptIR}, 30.05 dB & \small~\textbf{Ours}, \textbf{42.38 dB} \\
\end{tabular}
}
\caption{\small Image Dehazing on SOTS-outdoor~\cite{reside}.}
\label{fig:suppl_all_in_one_dehazing_visual_results}
\end{center}
\vspace{-7mm}
\end{figure*}

%% file: suppl_figures/suppl_003_all_in_one_denoising_visual_results.tex
\begin{figure*}[t]
\begin{center}
\setlength{\tabcolsep}{1.5pt}
\scalebox{0.88}{
\begin{tabular}[b]{c c c c}
\vspace{0.4em}
\includegraphics[width=.25\textwidth,valign=t]{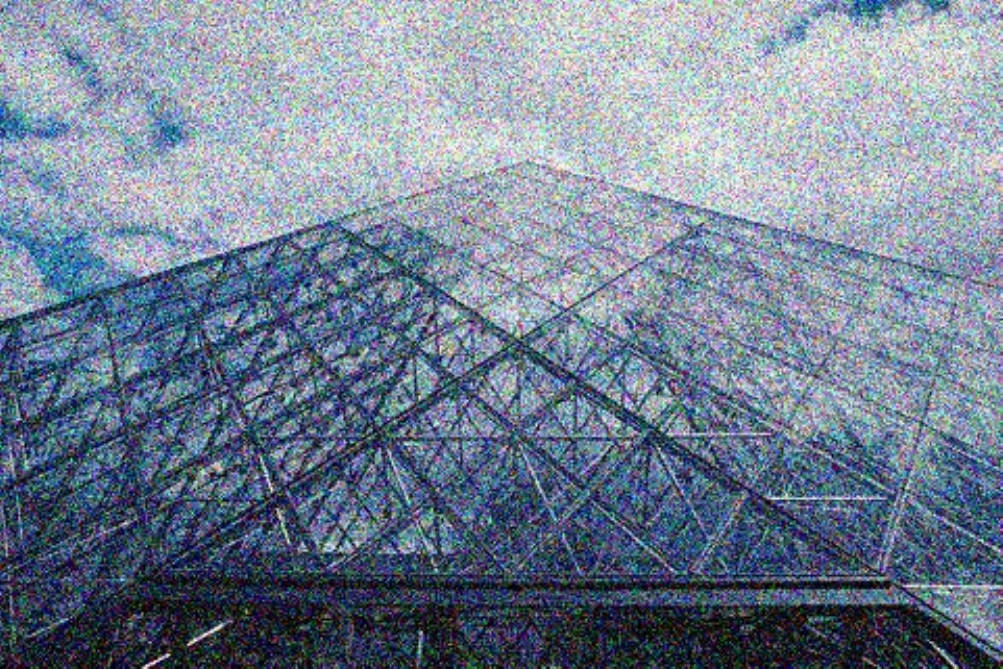} &   
\includegraphics[width=.25\textwidth,valign=t]{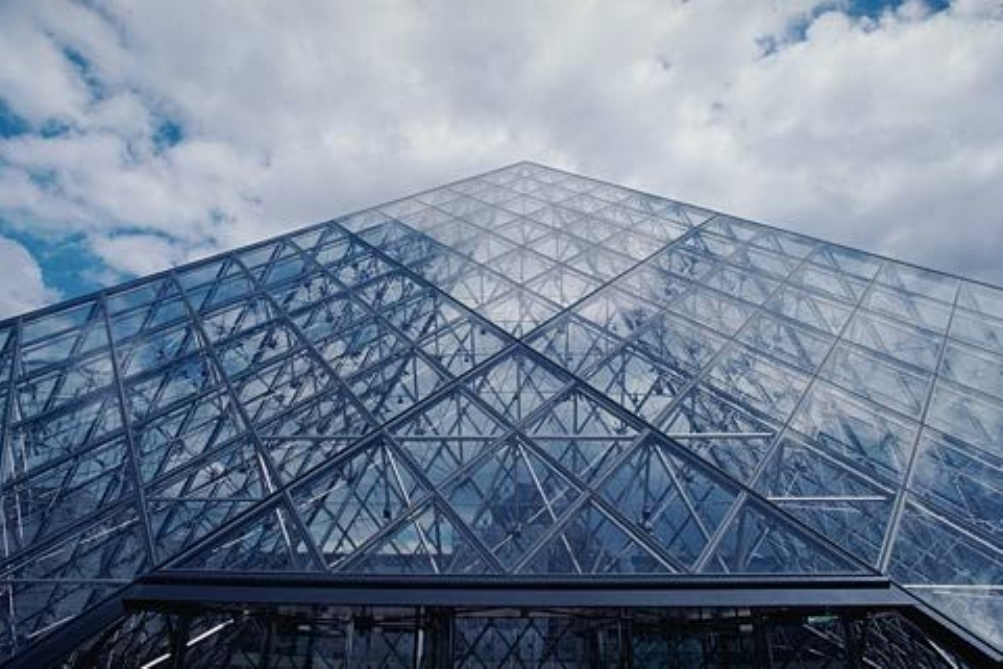} &   
\includegraphics[width=.25\textwidth,valign=t]{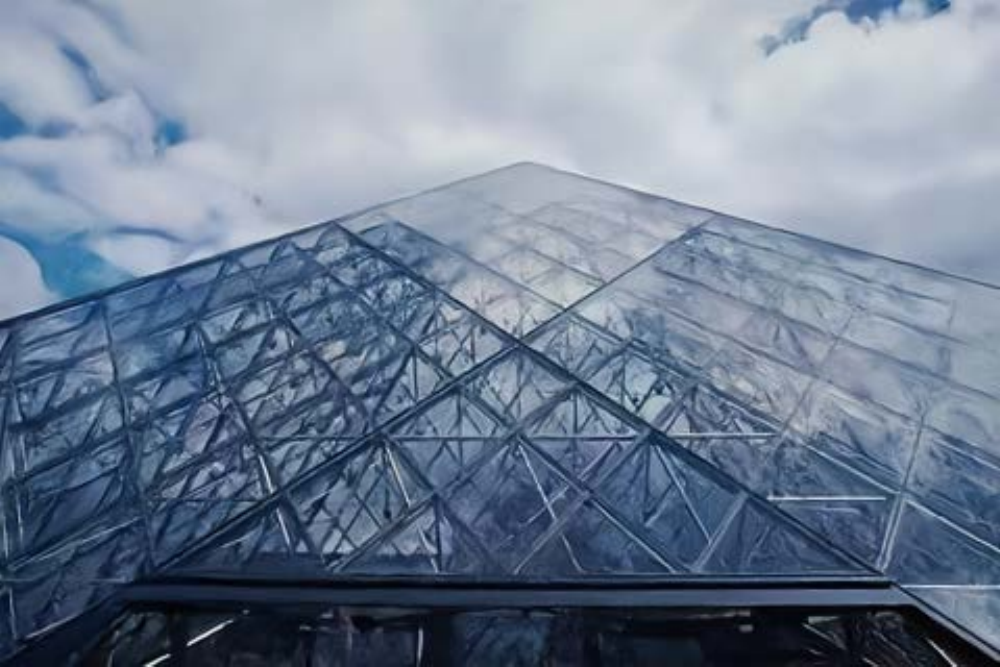} &   
\includegraphics[width=.25\textwidth,valign=t]{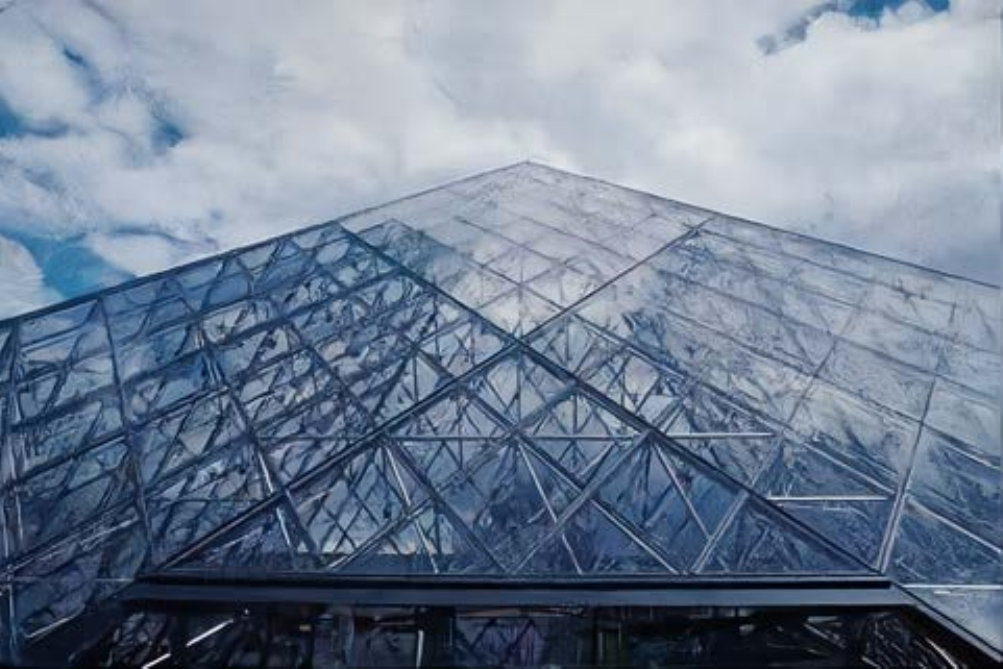}   \\
\vspace{0.4em}
\small~Degraded  & \small~Label & \small~PromptIR~\cite{PromptIR}, 26.46 dB & \small~\textbf{Ours}, \textbf{26.58 dB} \\
\vspace{0.4em}
\includegraphics[width=.25\textwidth,valign=t]{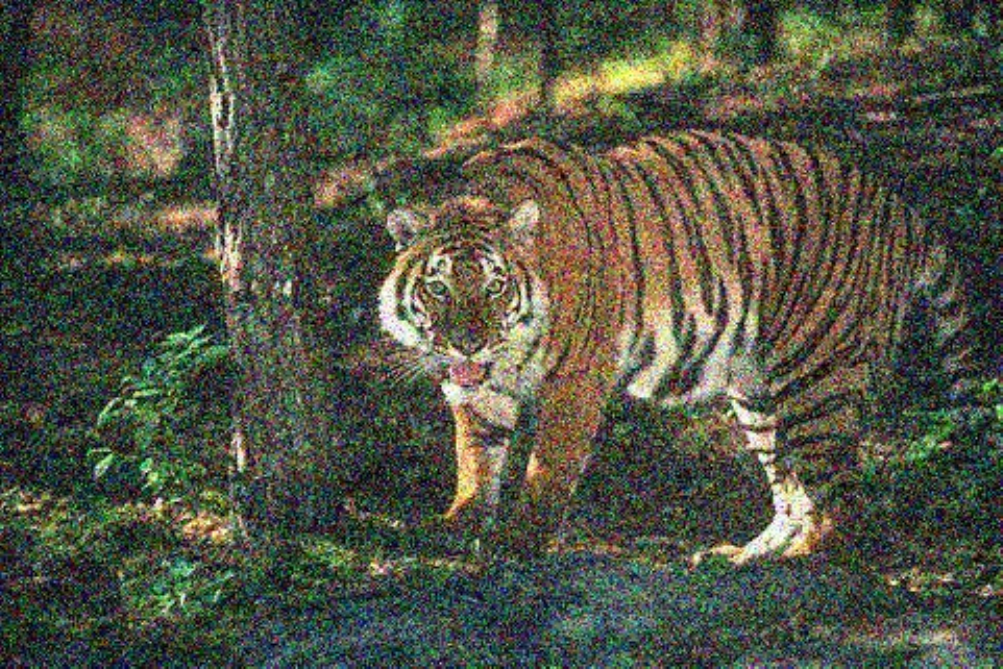} &   
\includegraphics[width=.25\textwidth,valign=t]{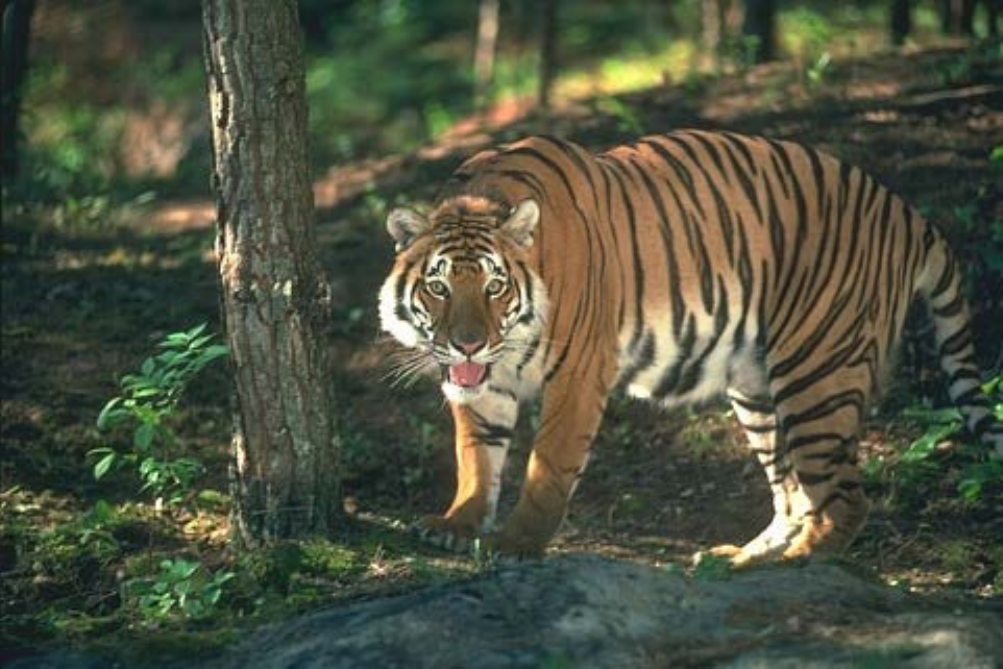} &   
\includegraphics[width=.25\textwidth,valign=t]{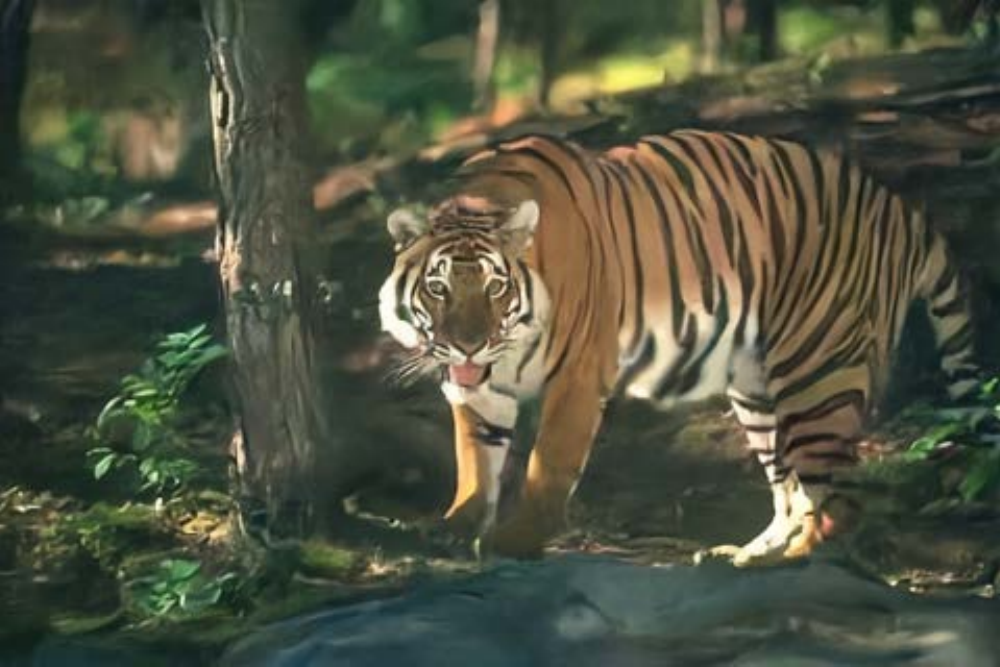} &   
\includegraphics[width=.25\textwidth,valign=t]{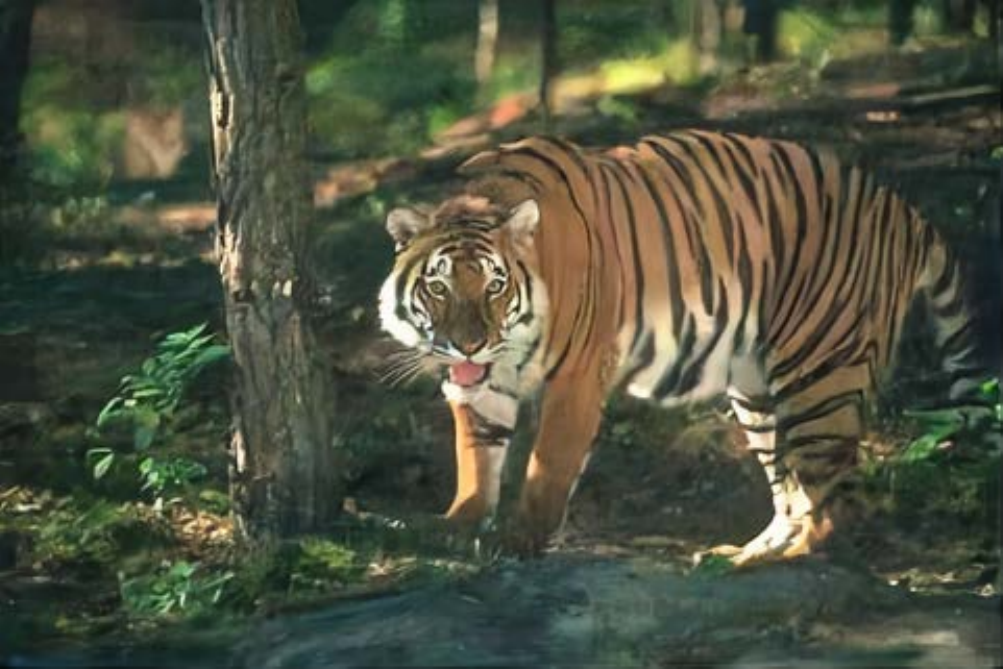}   \\
\vspace{0.4em}
\small~Degraded  & \small~Label & \small~PromptIR~\cite{PromptIR}, 27.41 dB & \small~\textbf{Ours}, \textbf{27.53 dB} \\
\vspace{0.4em}
\includegraphics[width=.25\textwidth,valign=t]{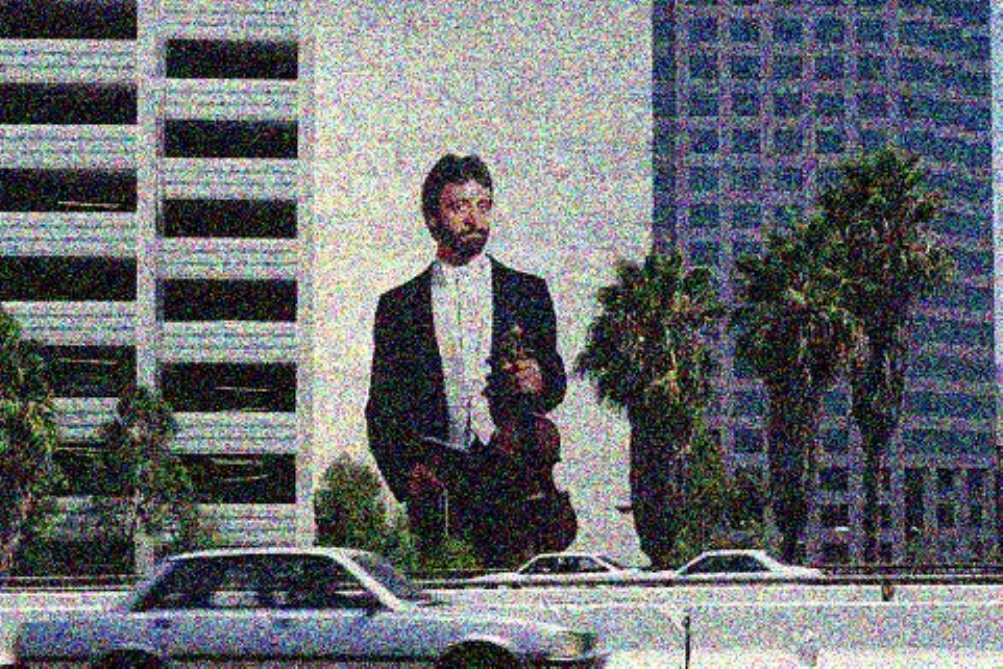} &   
\includegraphics[width=.25\textwidth,valign=t]{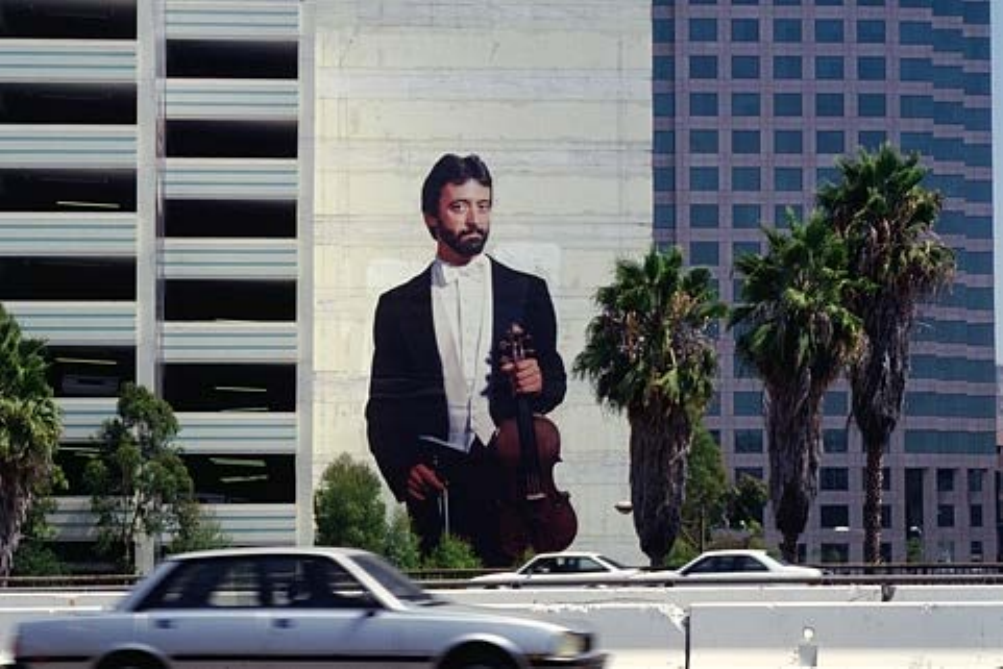} &   
\includegraphics[width=.25\textwidth,valign=t]{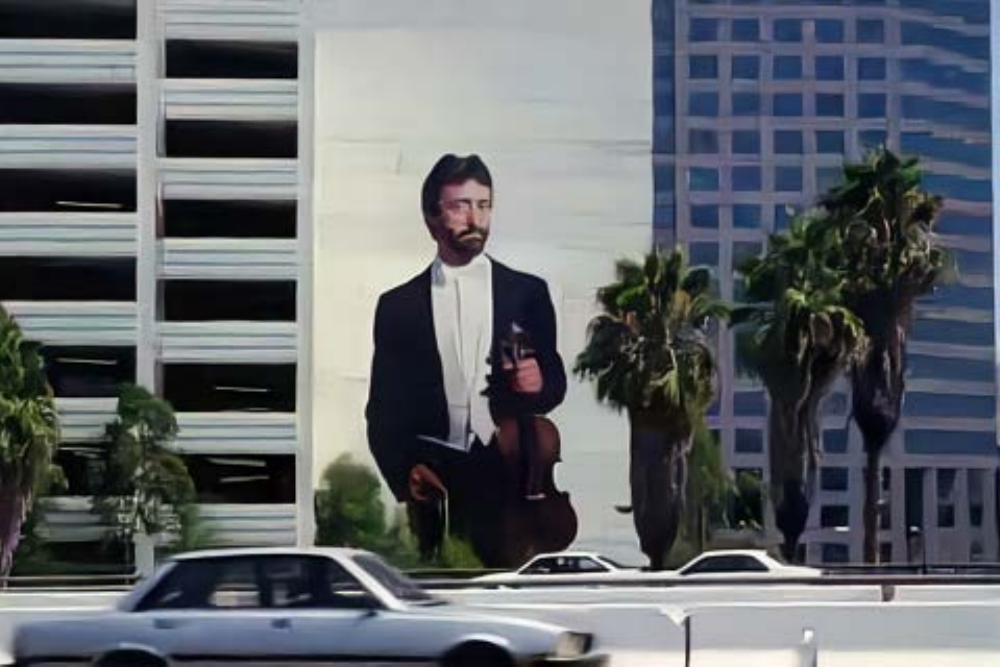} &   
\includegraphics[width=.25\textwidth,valign=t]{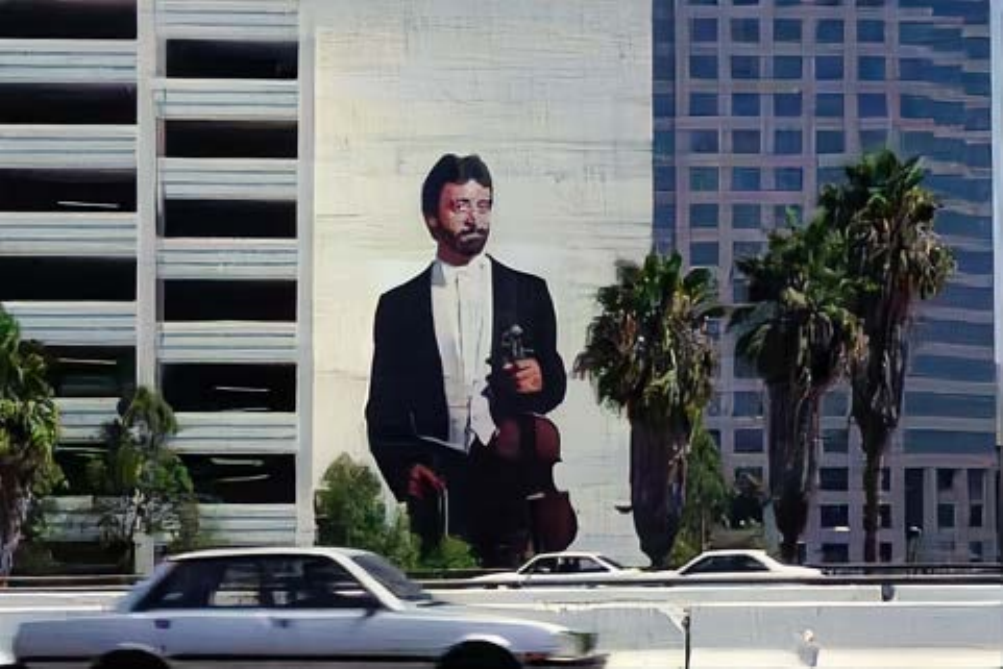}   \\
\vspace{0.4em}
\small~Degraded  & \small~Label & \small~PromptIR~\cite{PromptIR}, 27.69 dB & \small~\textbf{Ours}, \textbf{27.86 dB} \\
\end{tabular}
}
\caption{\small  Image Denoising on CBSD68~\cite{CBSD68}.}
\label{fig:suppl_all_in_one_denoising_visual_results}
\end{center}
\vspace{-7mm}
\end{figure*}

%% file: suppl_figures/suppl_004_01_motion_deblurring_visual_results.tex
\begin{figure*}[t]
\begin{center}
\setlength{\tabcolsep}{1.5pt}
\scalebox{0.88}{
\begin{tabular}[b]{c c c c}
\vspace{0.4em}
\includegraphics[width=.25\textwidth,valign=t]{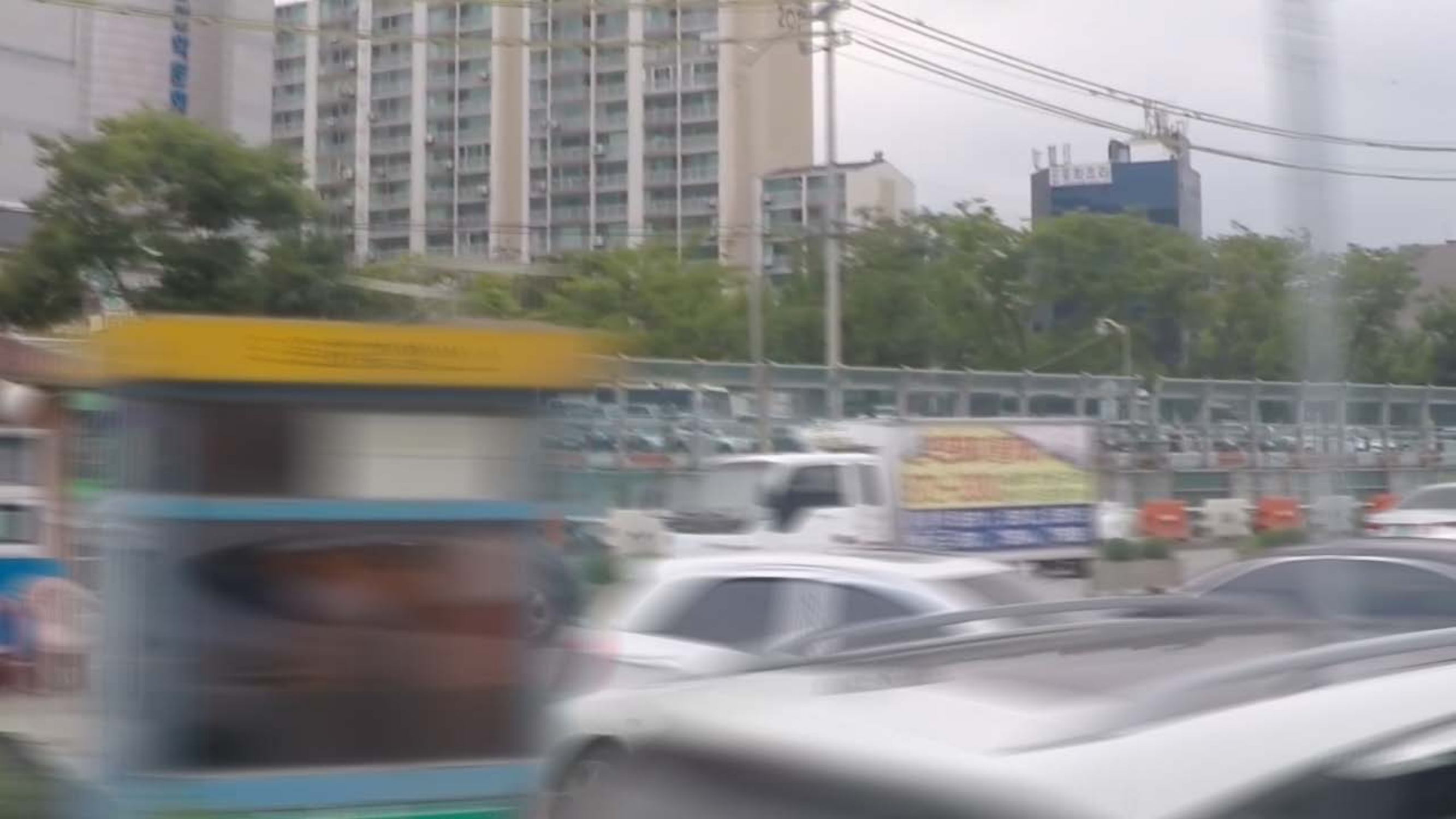}  &   
\includegraphics[width=.25\textwidth,valign=t]{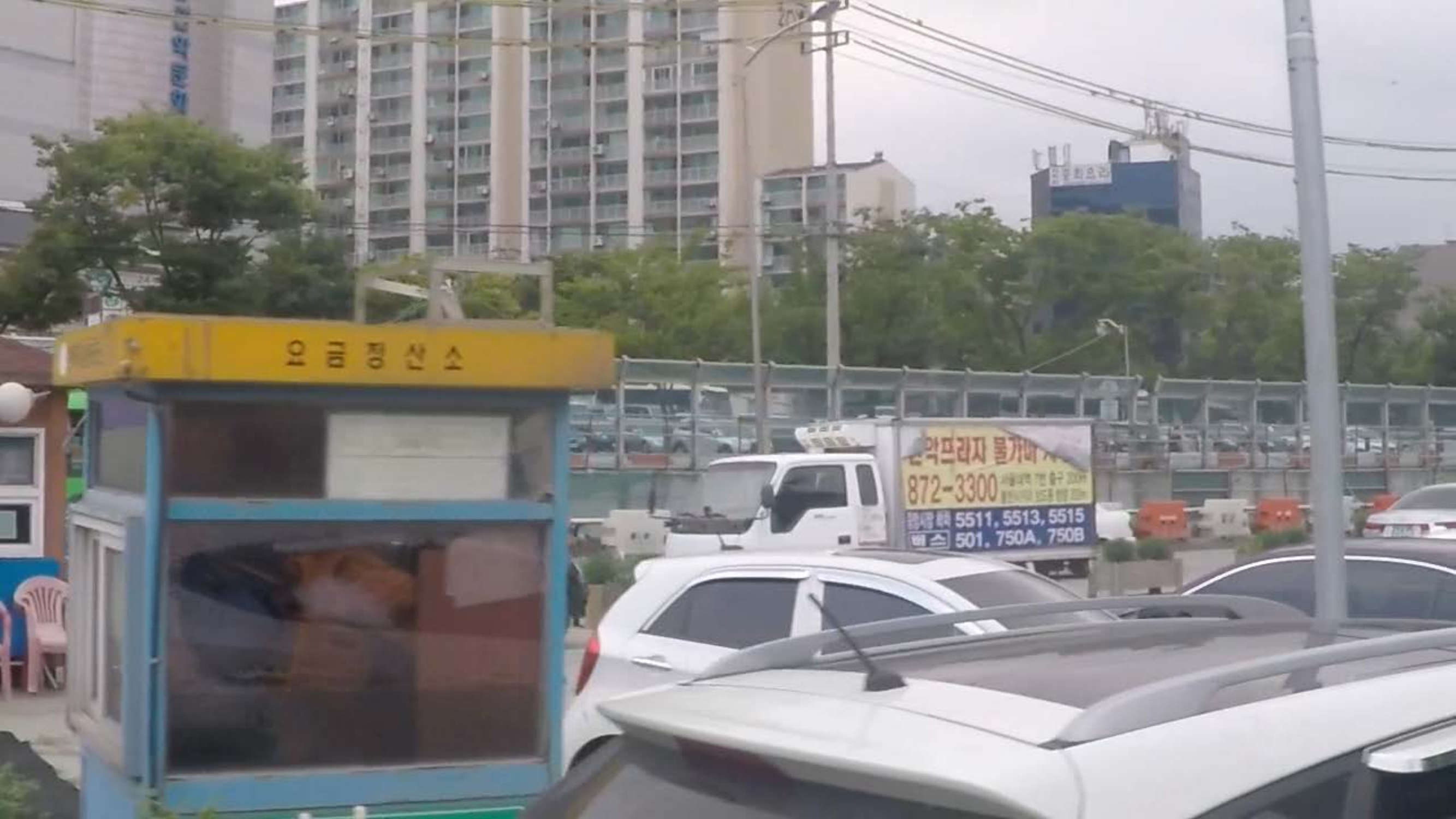} &   
\includegraphics[width=.25\textwidth,valign=t]{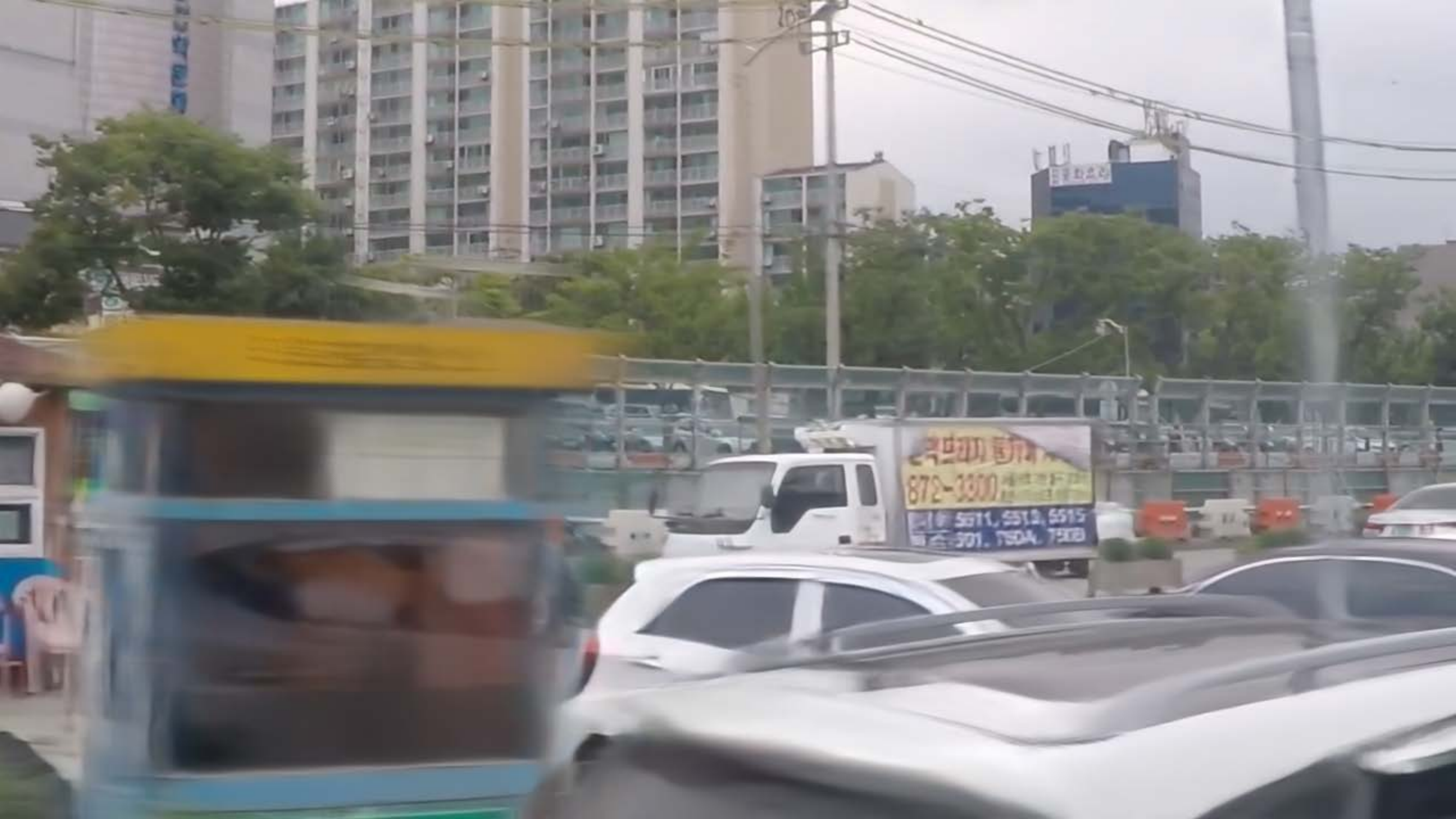}  &
\includegraphics[width=.25\textwidth,valign=t]{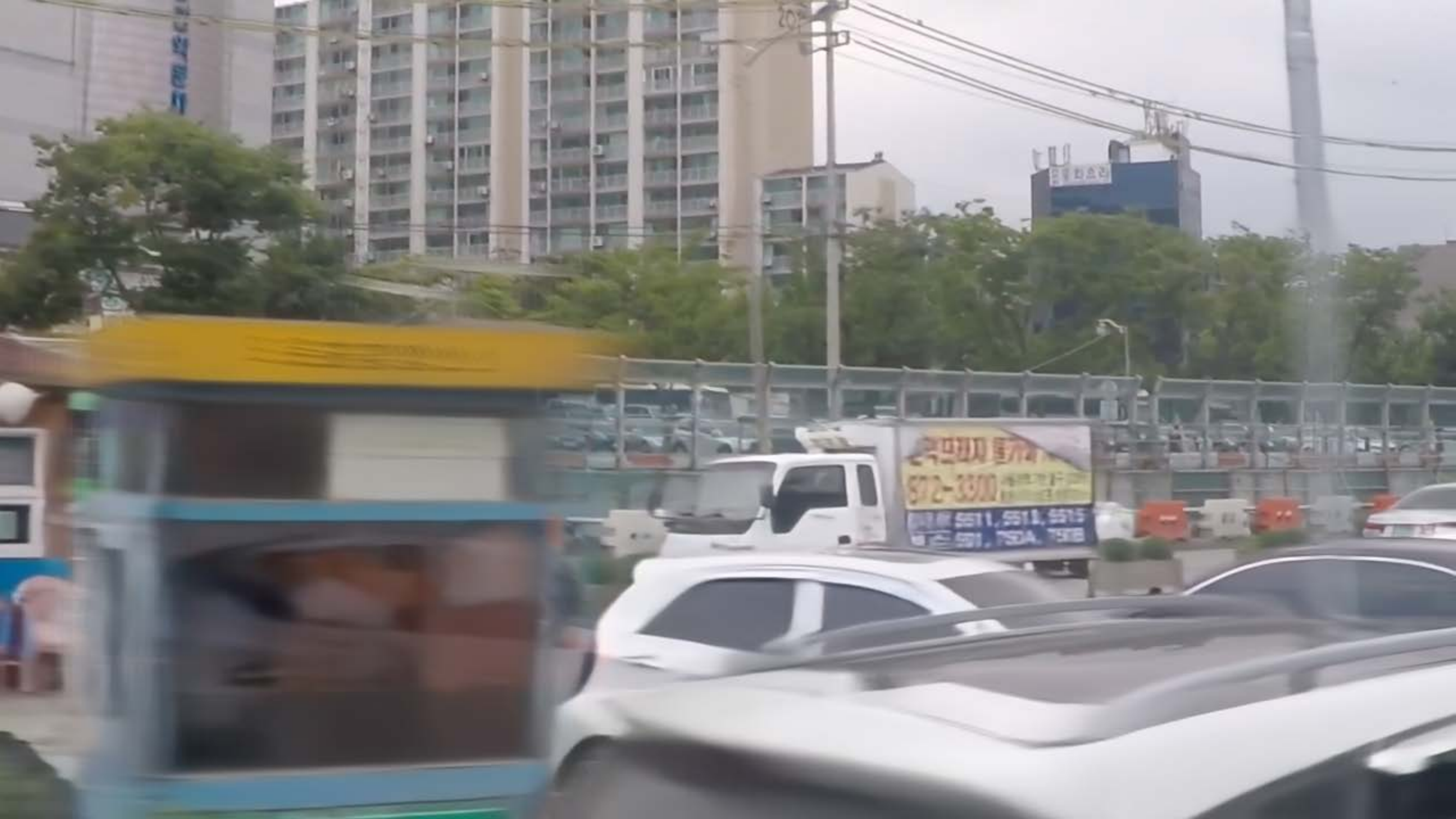} \\
\vspace{0.4em}
\small~Degraded  & \small~Label & \small~DMPHN~\cite{DMPHN}, 28.80 dB & \small~MTRNN~\cite{MT-RNN}, 28.92 dB \\
\vspace{0.4em}
\includegraphics[width=.25\textwidth,valign=t]{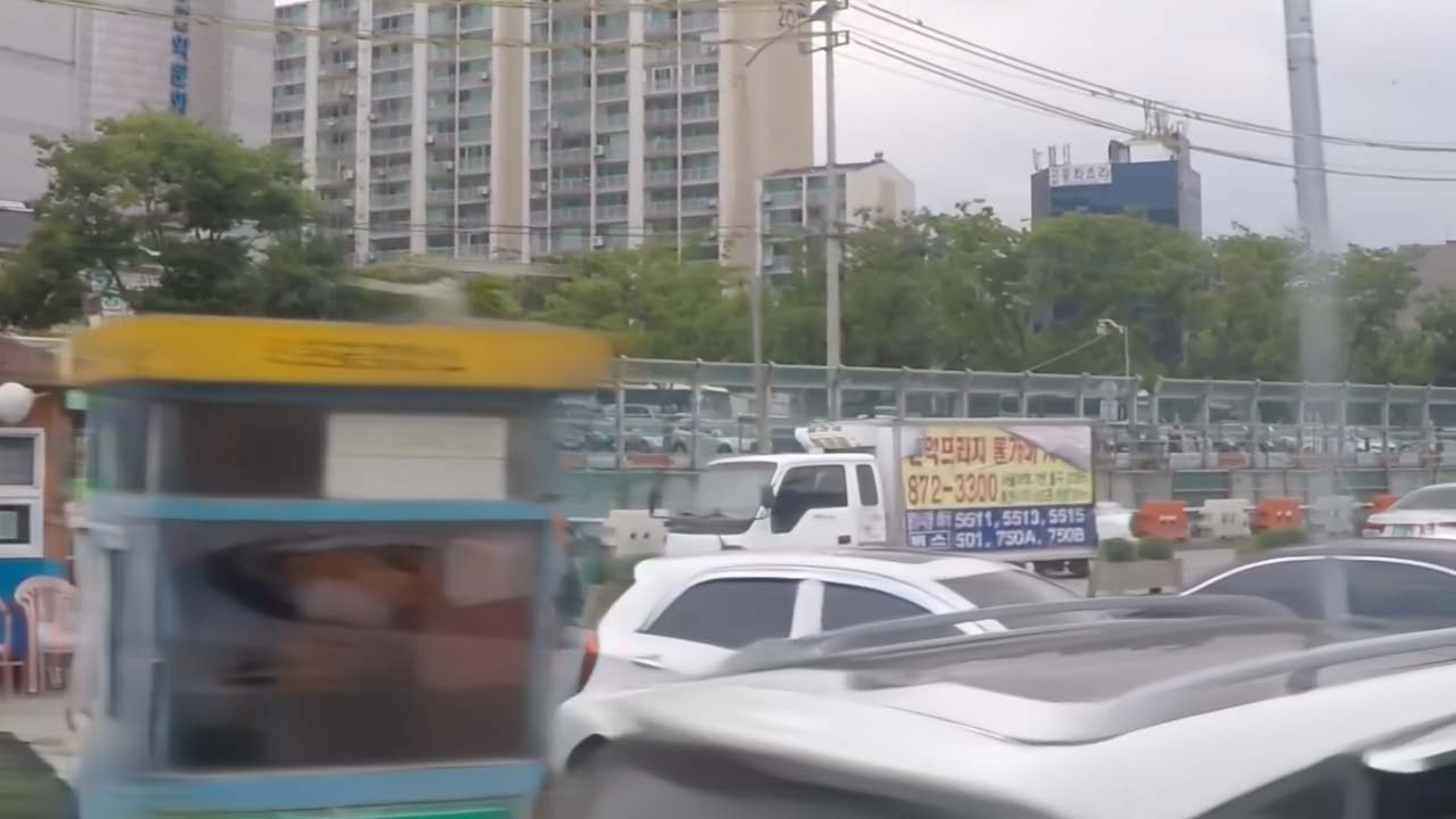}  &   
\includegraphics[width=.25\textwidth,valign=t]{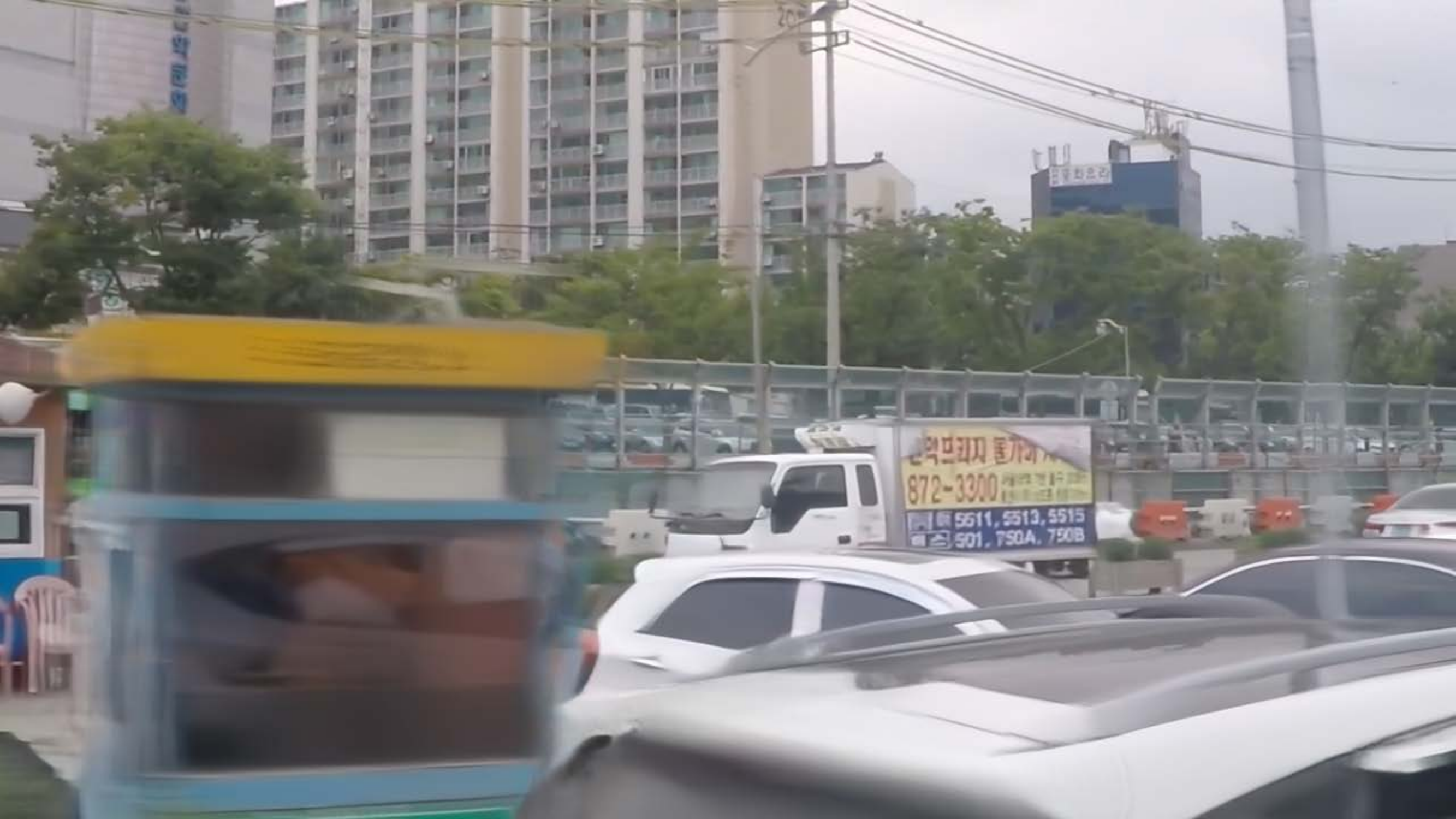} &   
\includegraphics[width=.25\textwidth,valign=t]{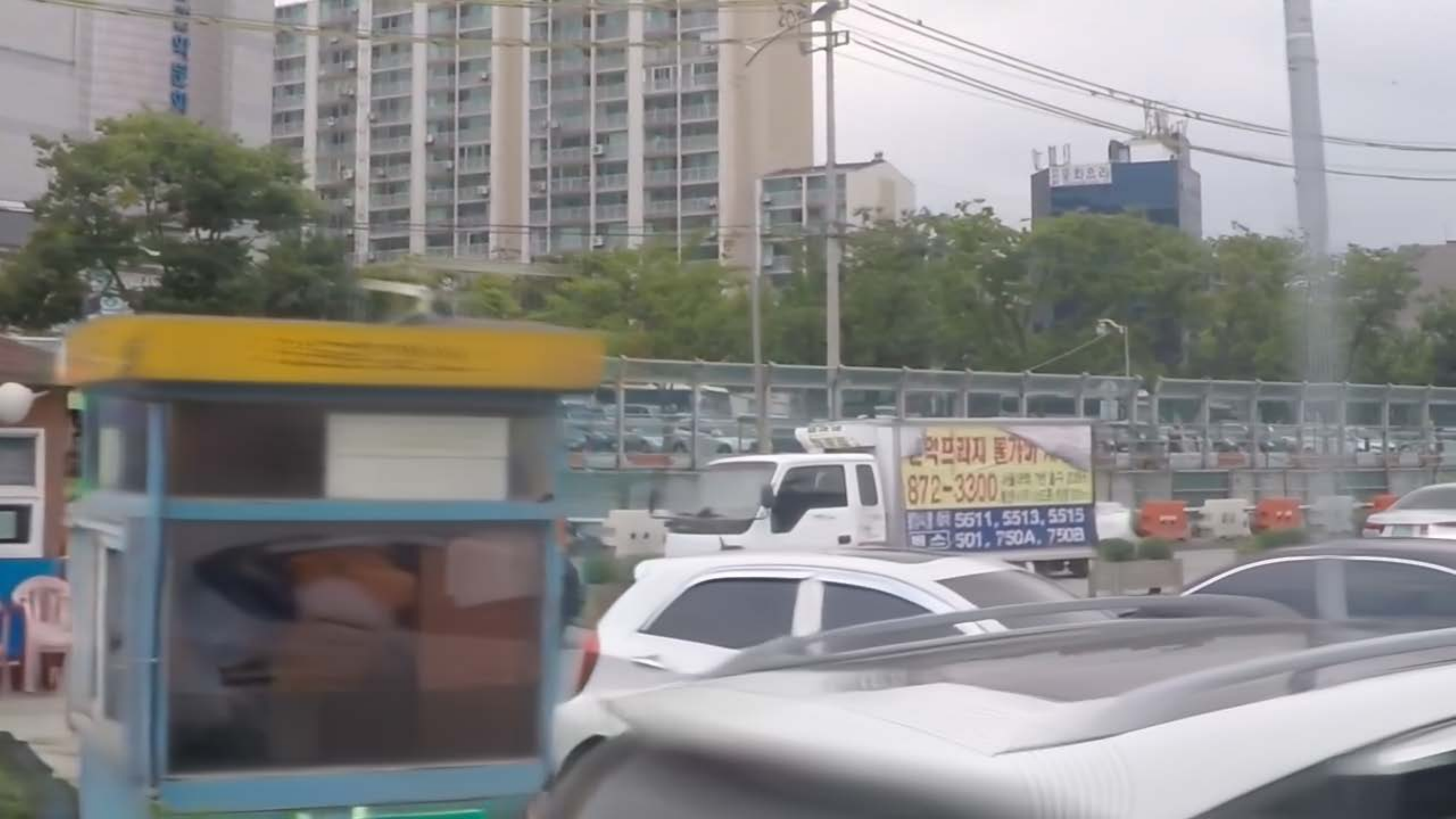}  &
\includegraphics[width=.25\textwidth,valign=t]{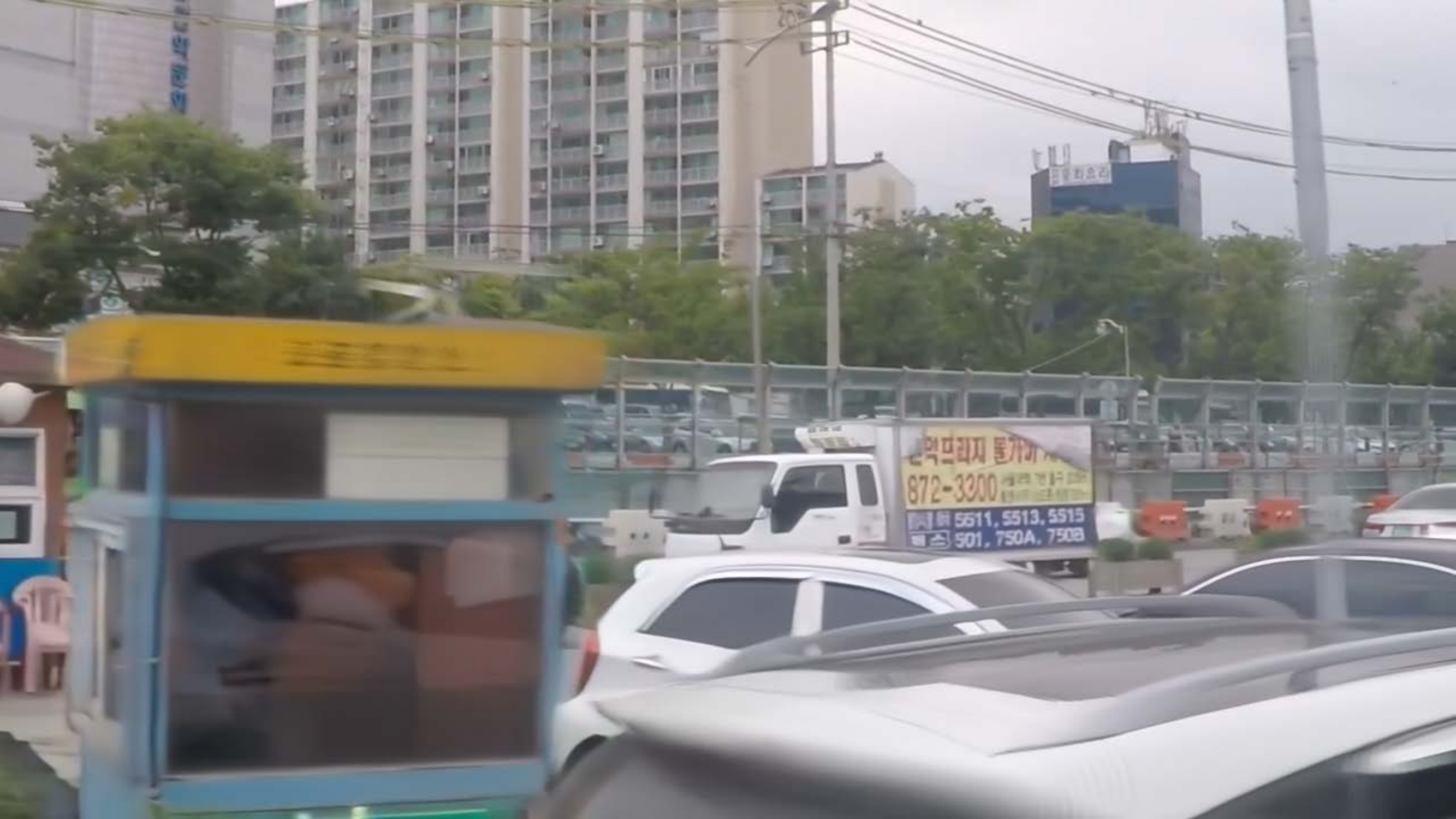} \\
\vspace{0.6em}
\small~MIMO-UNet+~\cite{MIMO-Unet+}, 29.34 dB & \small~HINet~\cite{HINet}, 29.36 dB & \small~NAFNet~\cite{NAFNet}, 29.41 dB & \small~\textbf{Ours}, \textbf{29.80 dB} \\
\vspace{0.4em}
\includegraphics[width=.25\textwidth,valign=t]{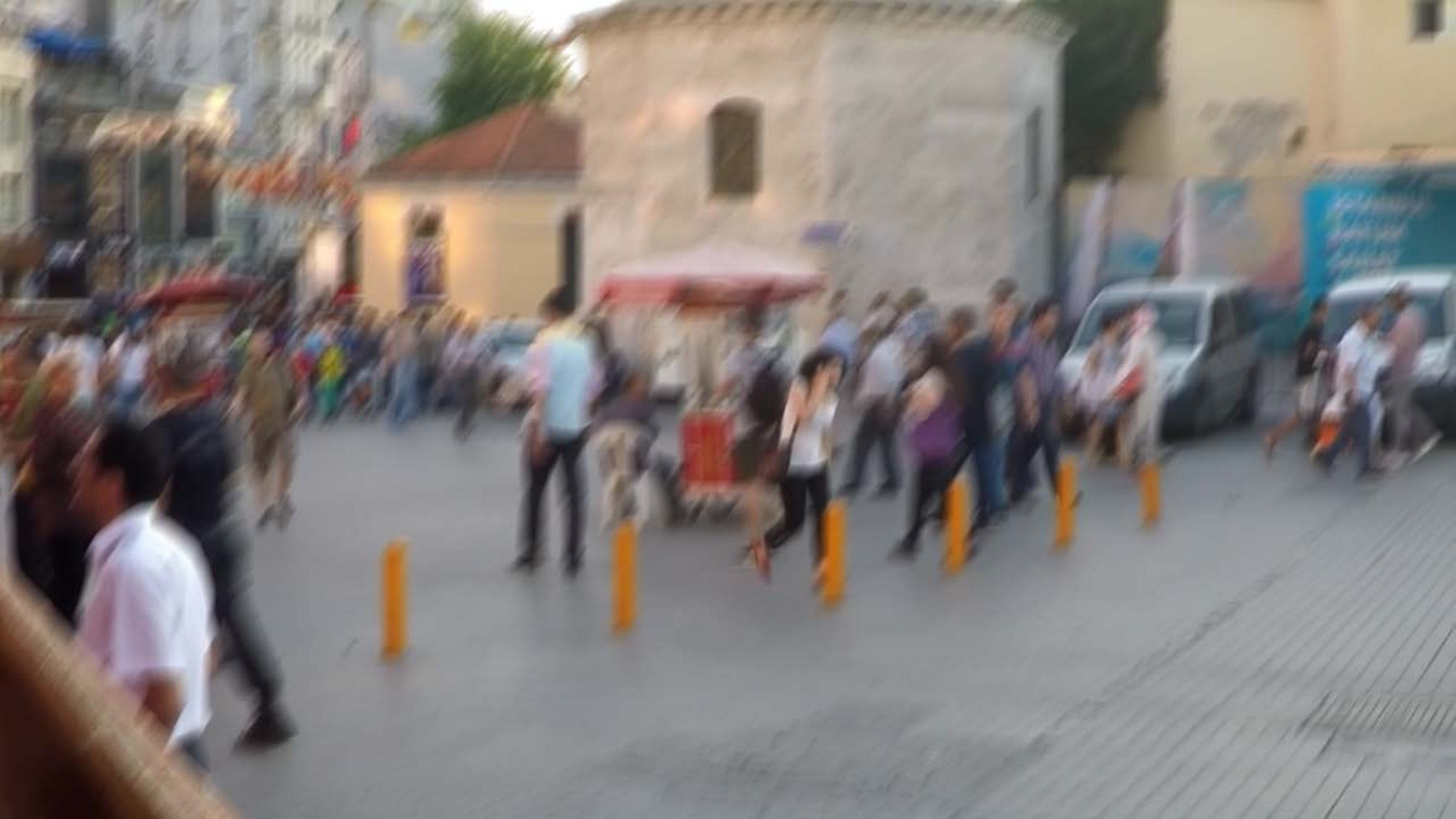}  &   
\includegraphics[width=.25\textwidth,valign=t]{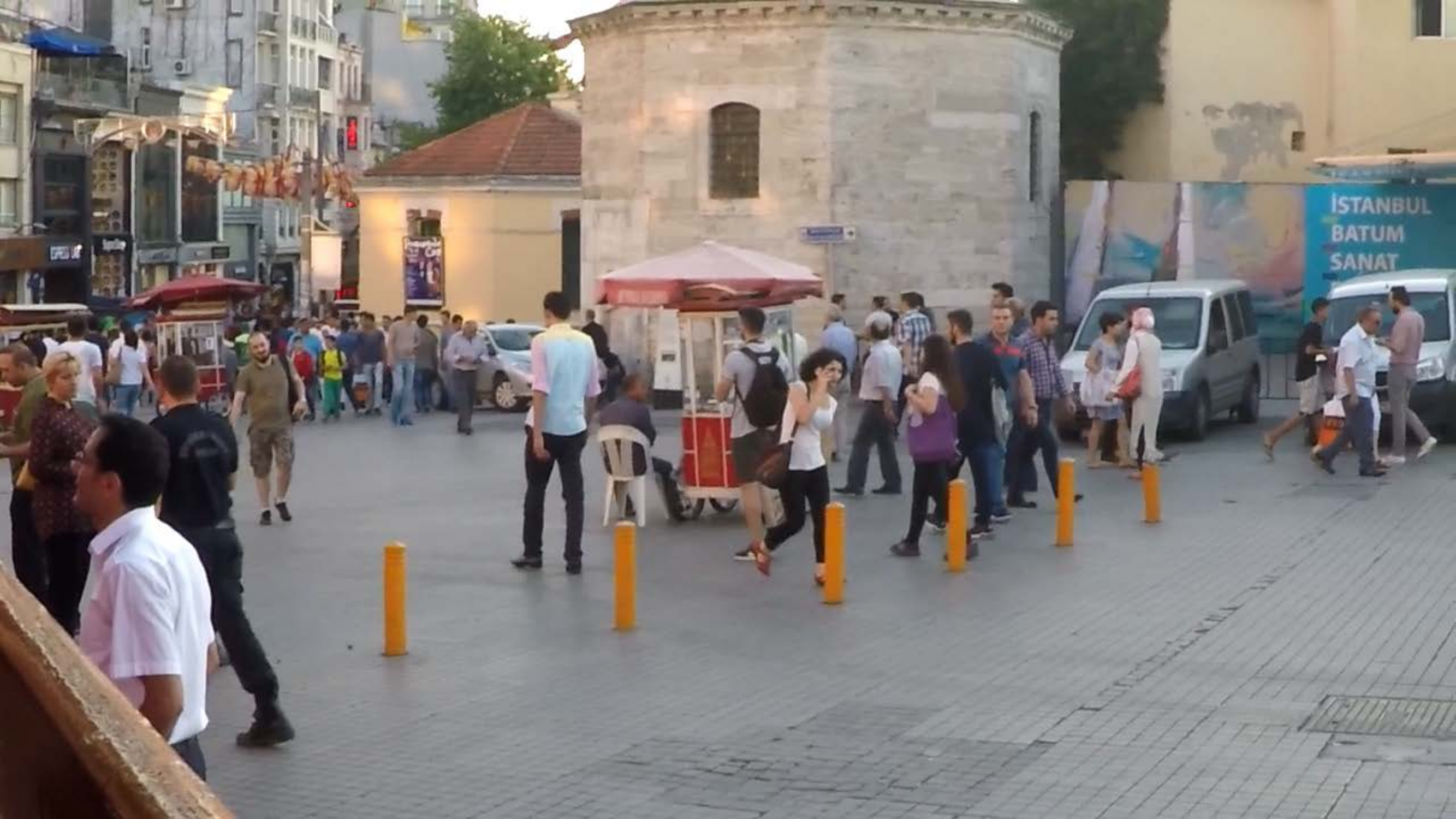} &   
\includegraphics[width=.25\textwidth,valign=t]{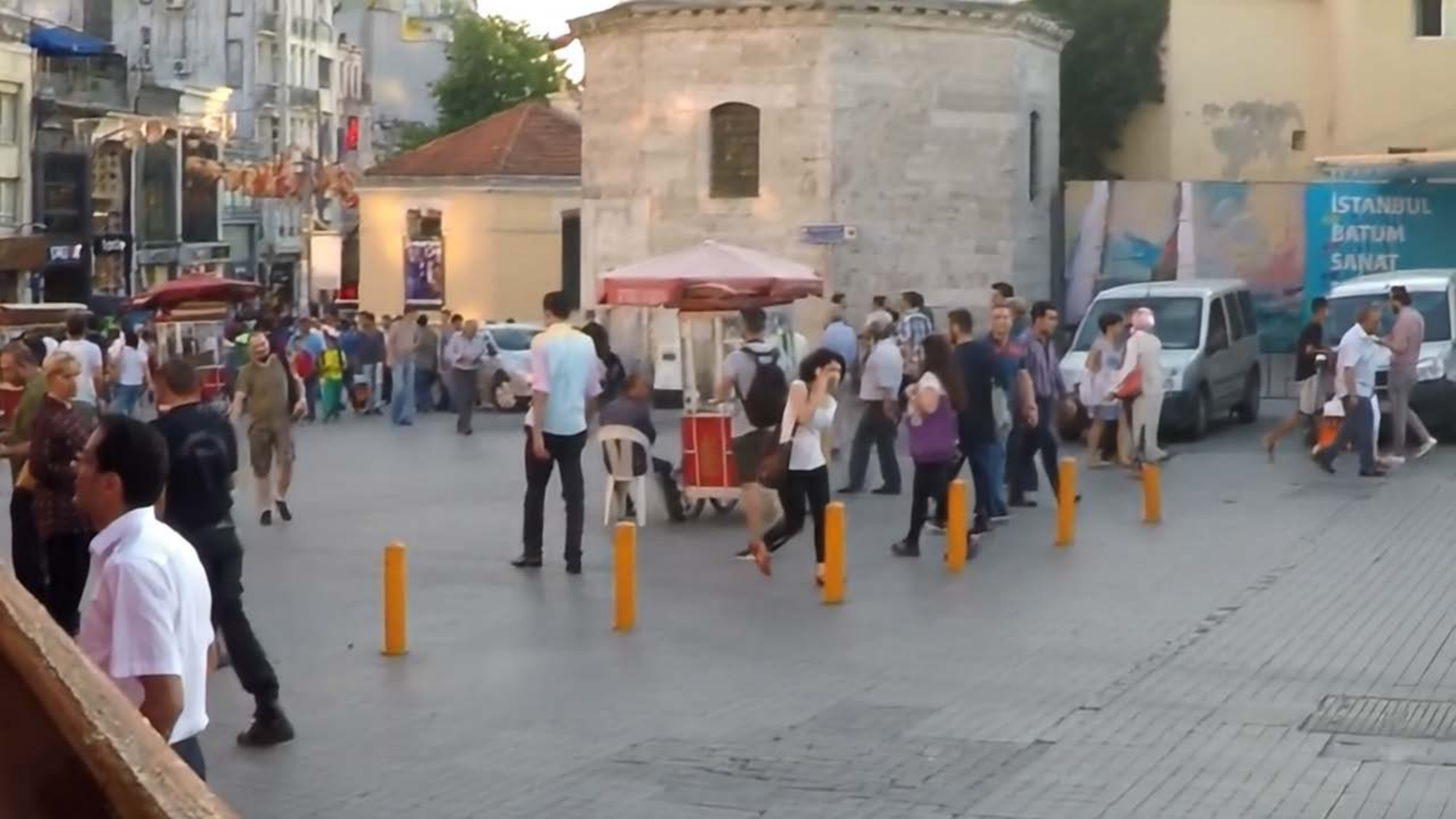}  &
\includegraphics[width=.25\textwidth,valign=t]{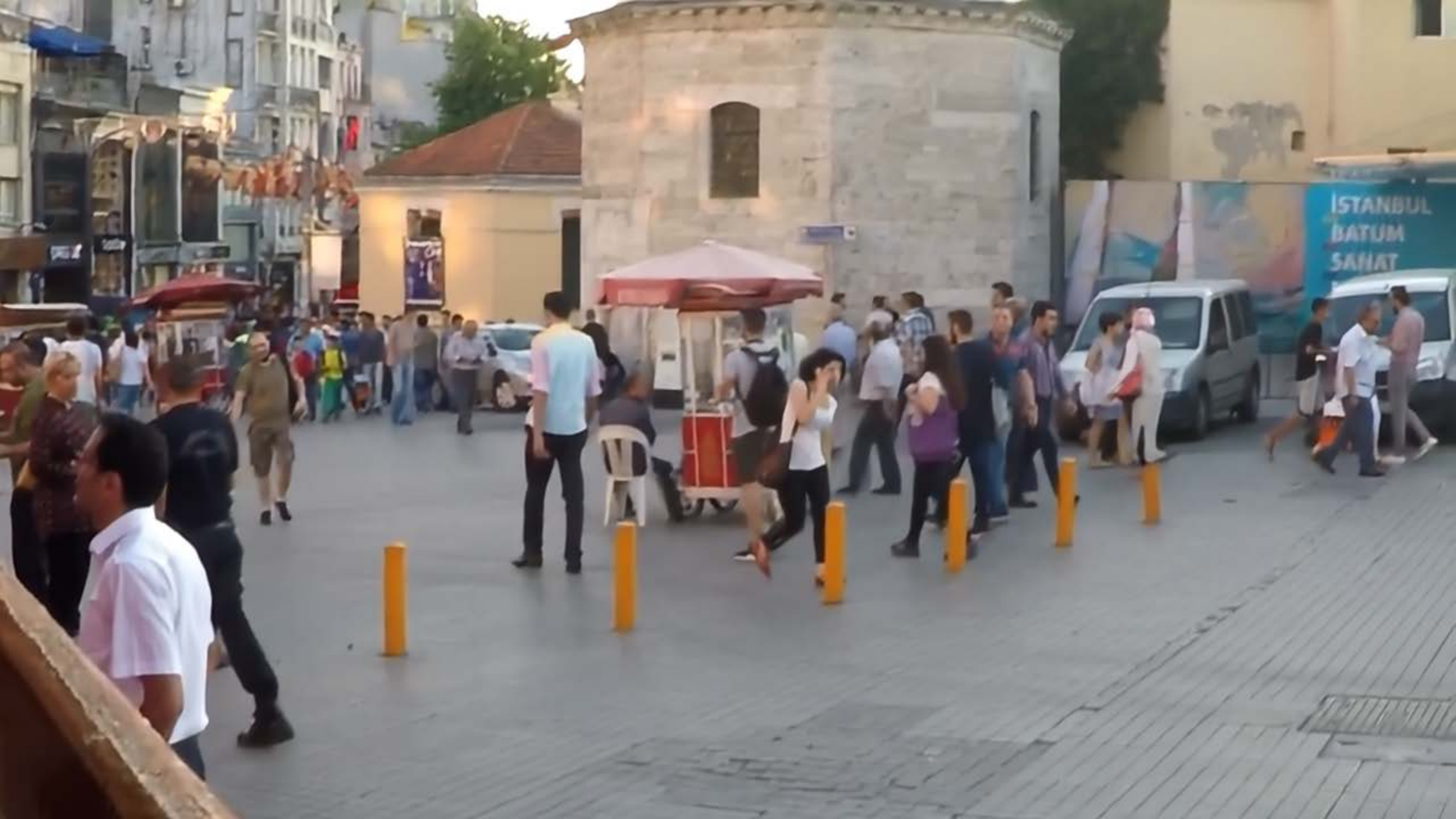} \\
\end{tabular}
}
\caption{\small Single-image motion deblurring on GoPro~\cite{GoPro}.}
\label{fig:suppl_single_image_motion_deblurring_visual_results_1}
\end{center}
\vspace{-7mm}
\end{figure*}

%% file: suppl_figures/suppl_004_02_motion_deblurring_visual_results.tex
\begin{figure*}[t]
\begin{center}
\setlength{\tabcolsep}{1.5pt}
\scalebox{0.88}{
\begin{tabular}[b]{c c c c}
\vspace{0.4em}
\includegraphics[width=.25\textwidth,valign=t]{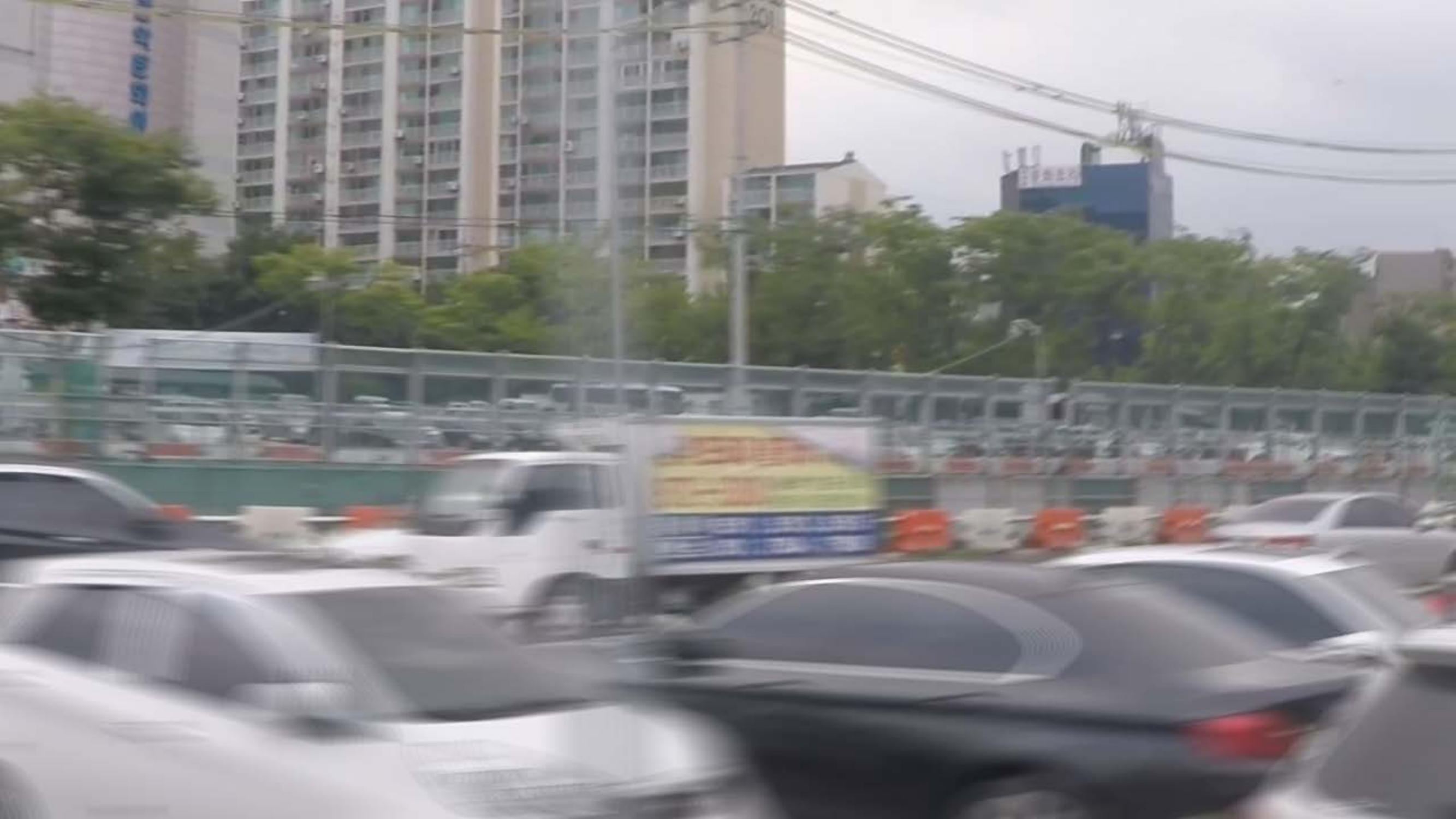}  &   
\includegraphics[width=.25\textwidth,valign=t]{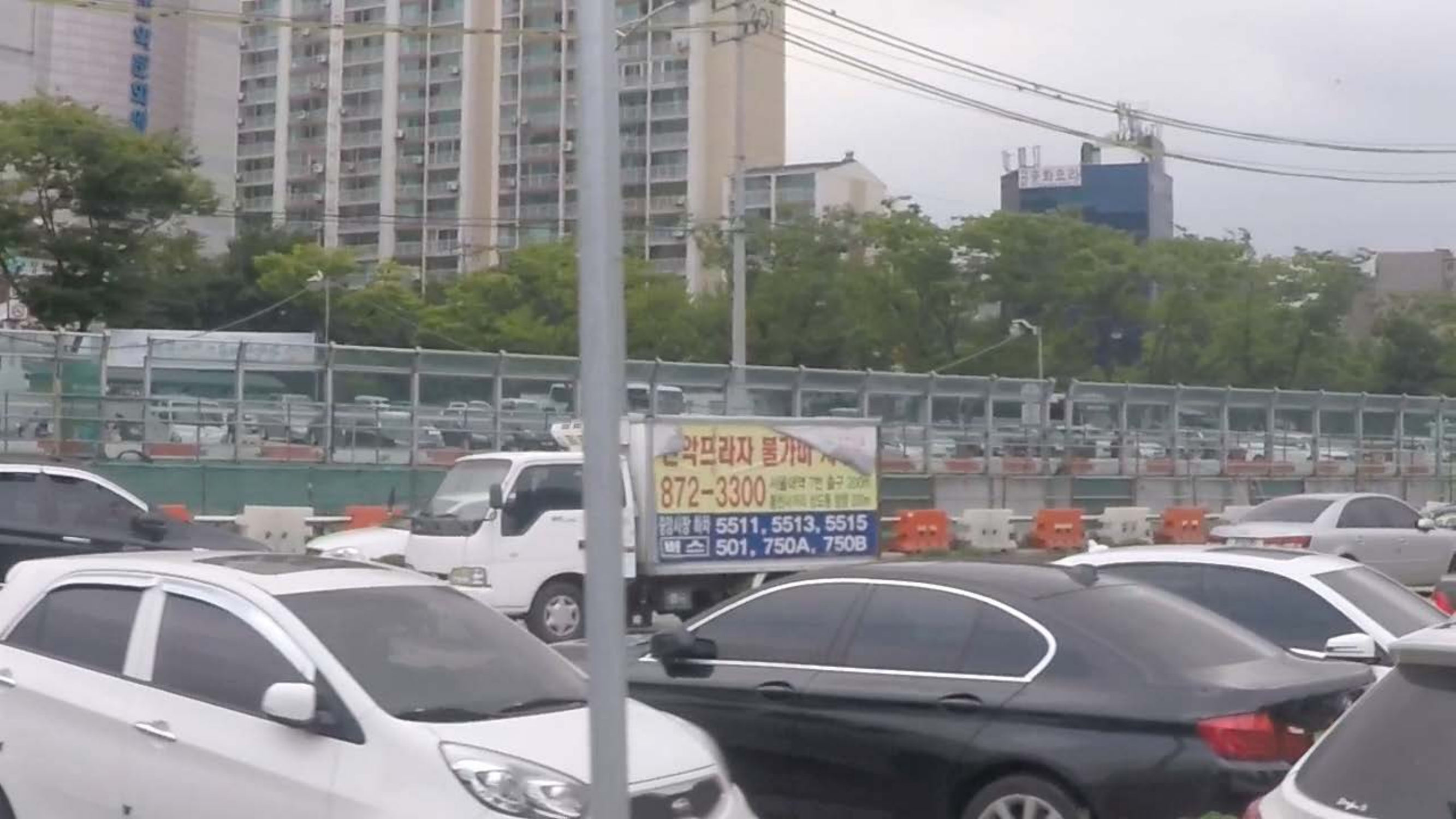} &   
\includegraphics[width=.25\textwidth,valign=t]{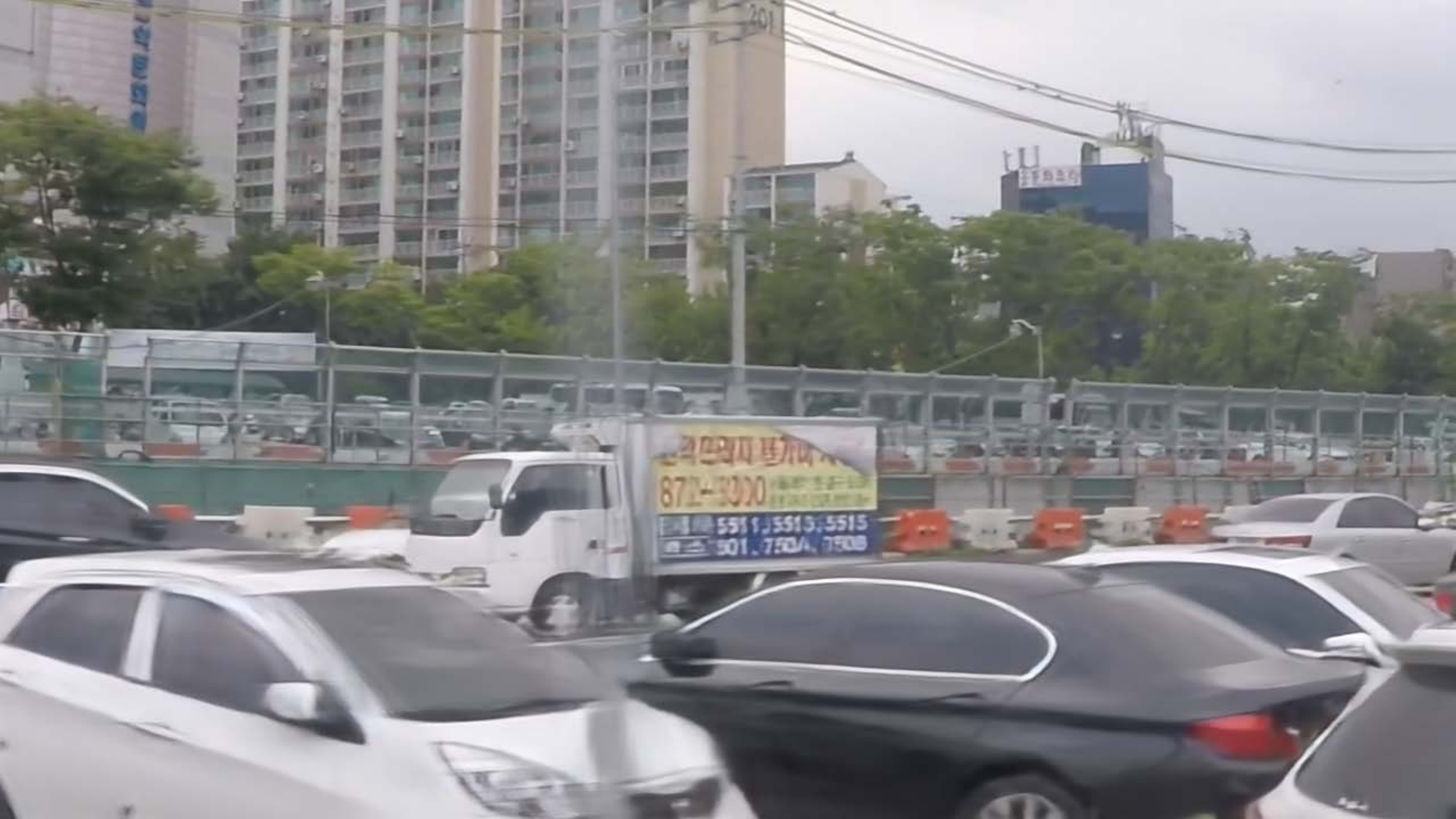}  &
\includegraphics[width=.25\textwidth,valign=t]{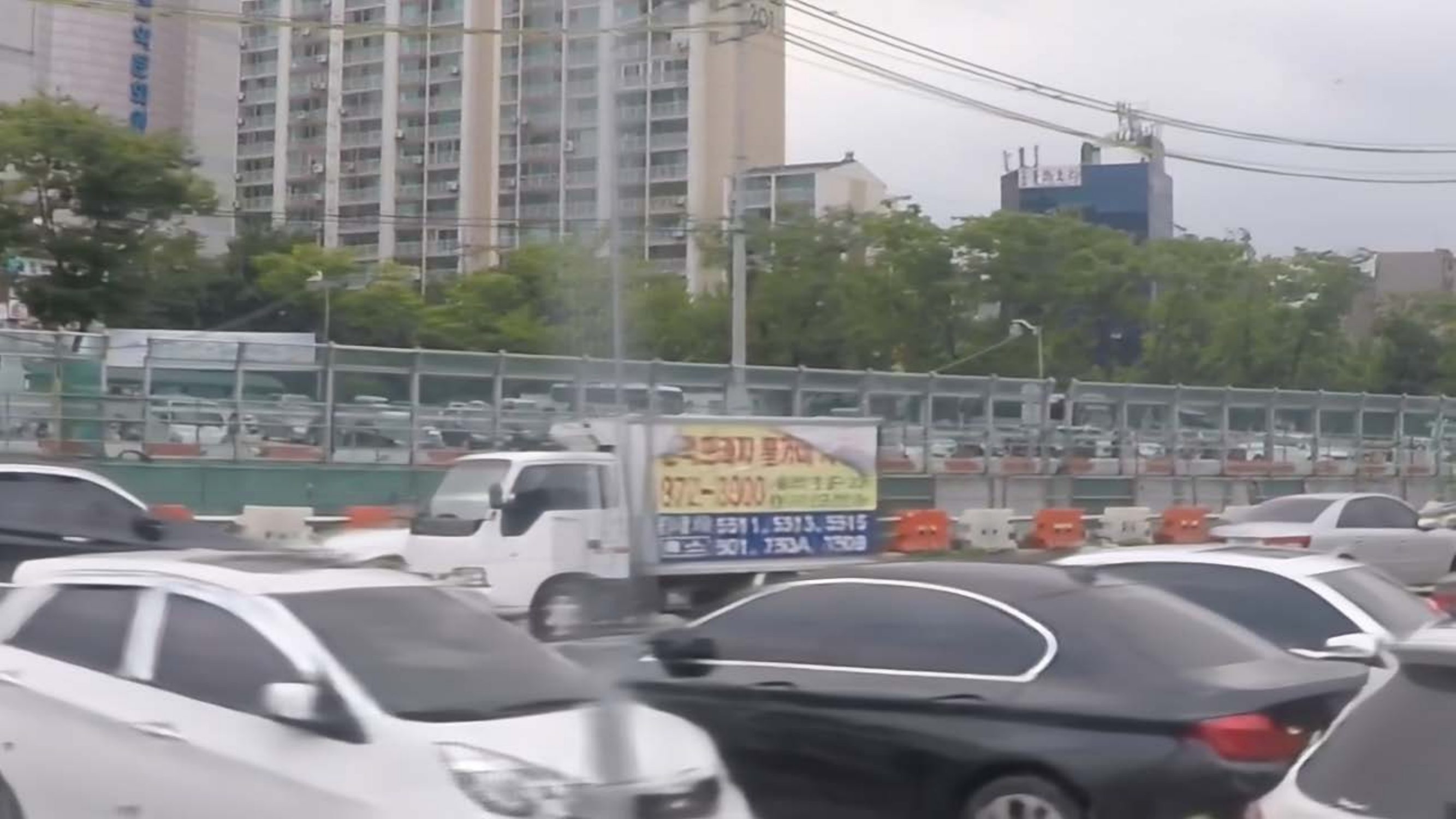} \\
\vspace{0.4em}
\small~Degraded  & \small~Label & \small~DMPHN~\cite{DMPHN}, 27.74 dB & \small~MTRNN~\cite{MT-RNN}, 28.46 dB \\
\vspace{0.4em}
\includegraphics[width=.25\textwidth,valign=t]{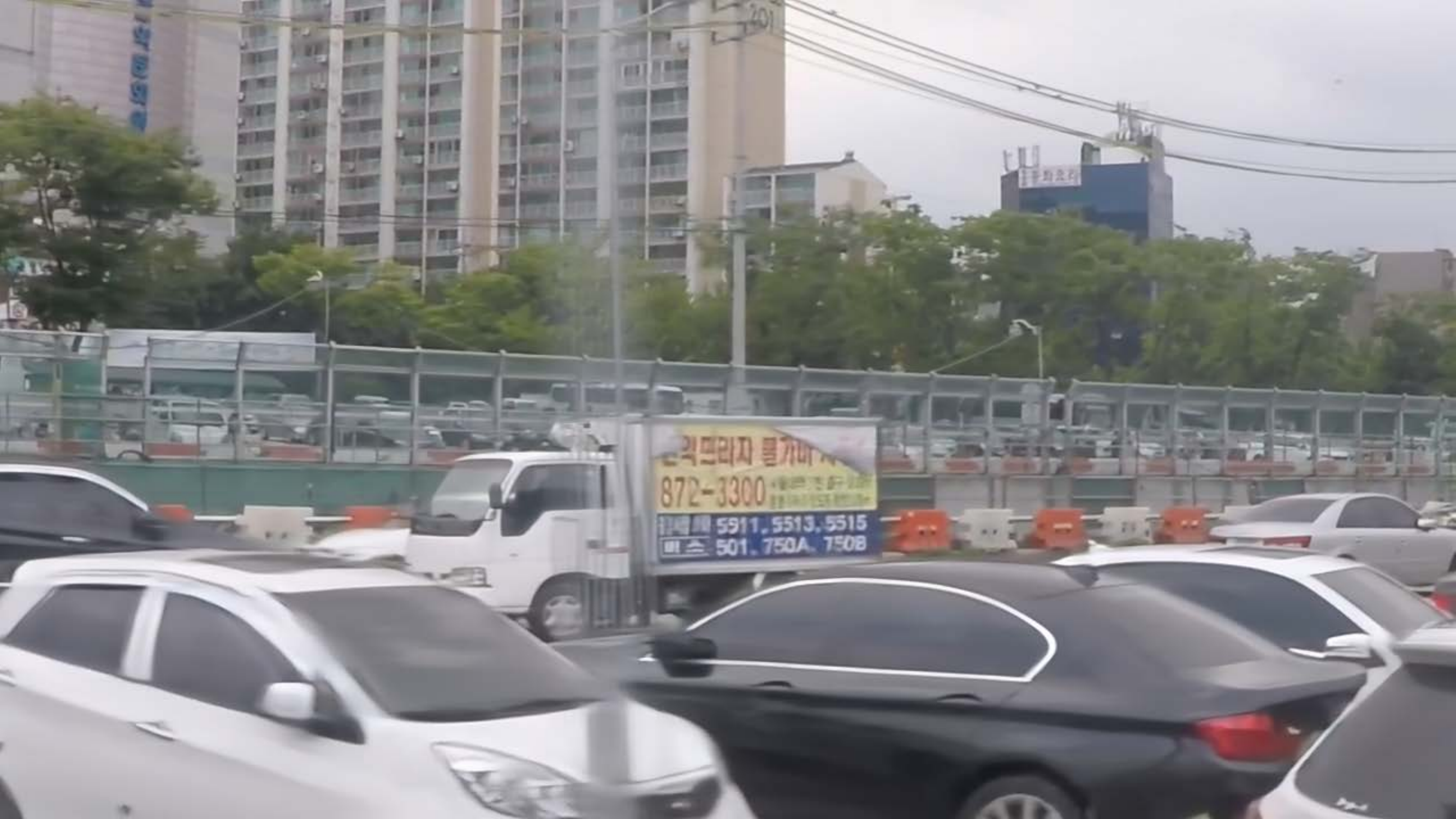}  &   
\includegraphics[width=.25\textwidth,valign=t]{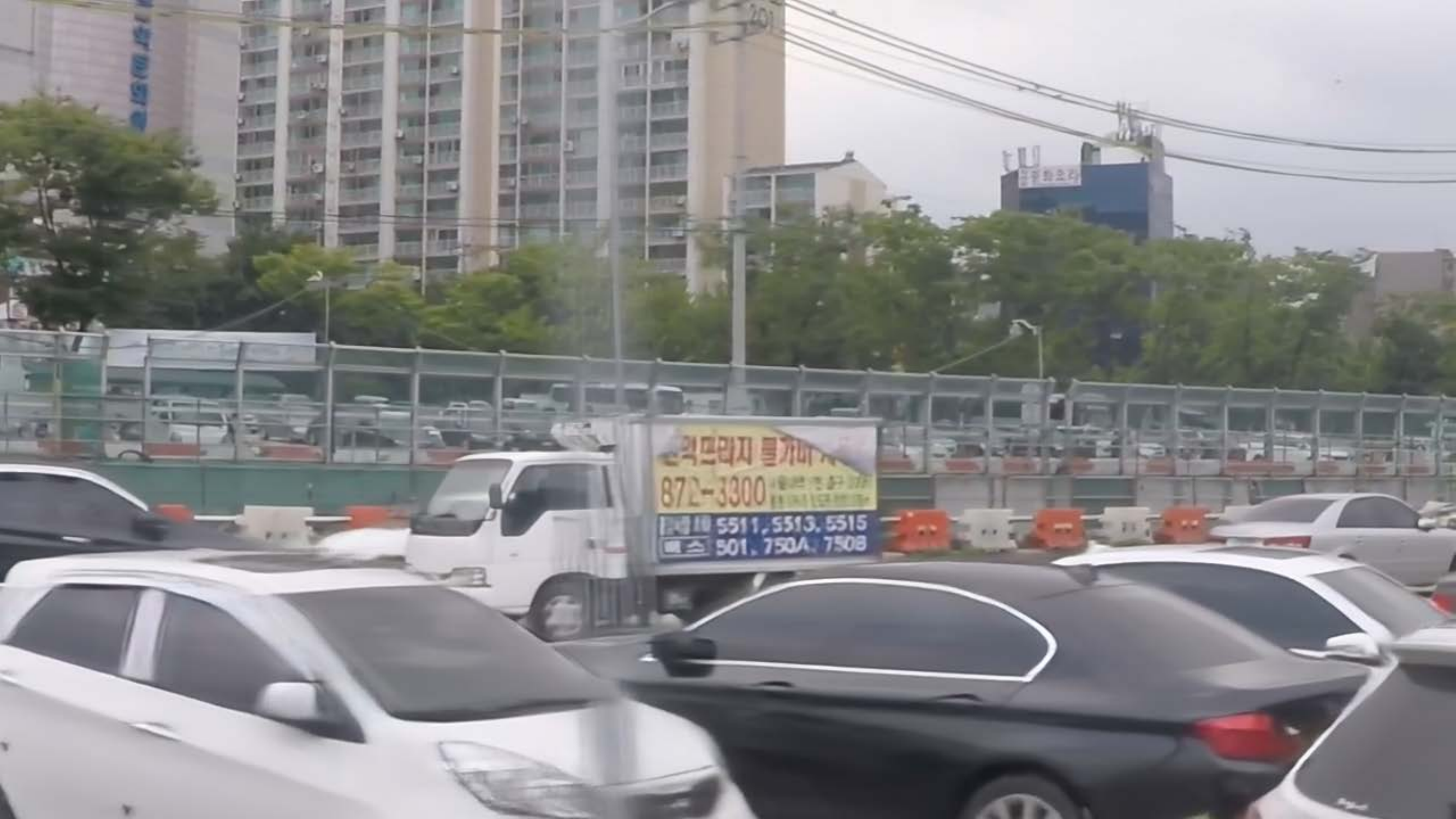} &   
\includegraphics[width=.25\textwidth,valign=t]{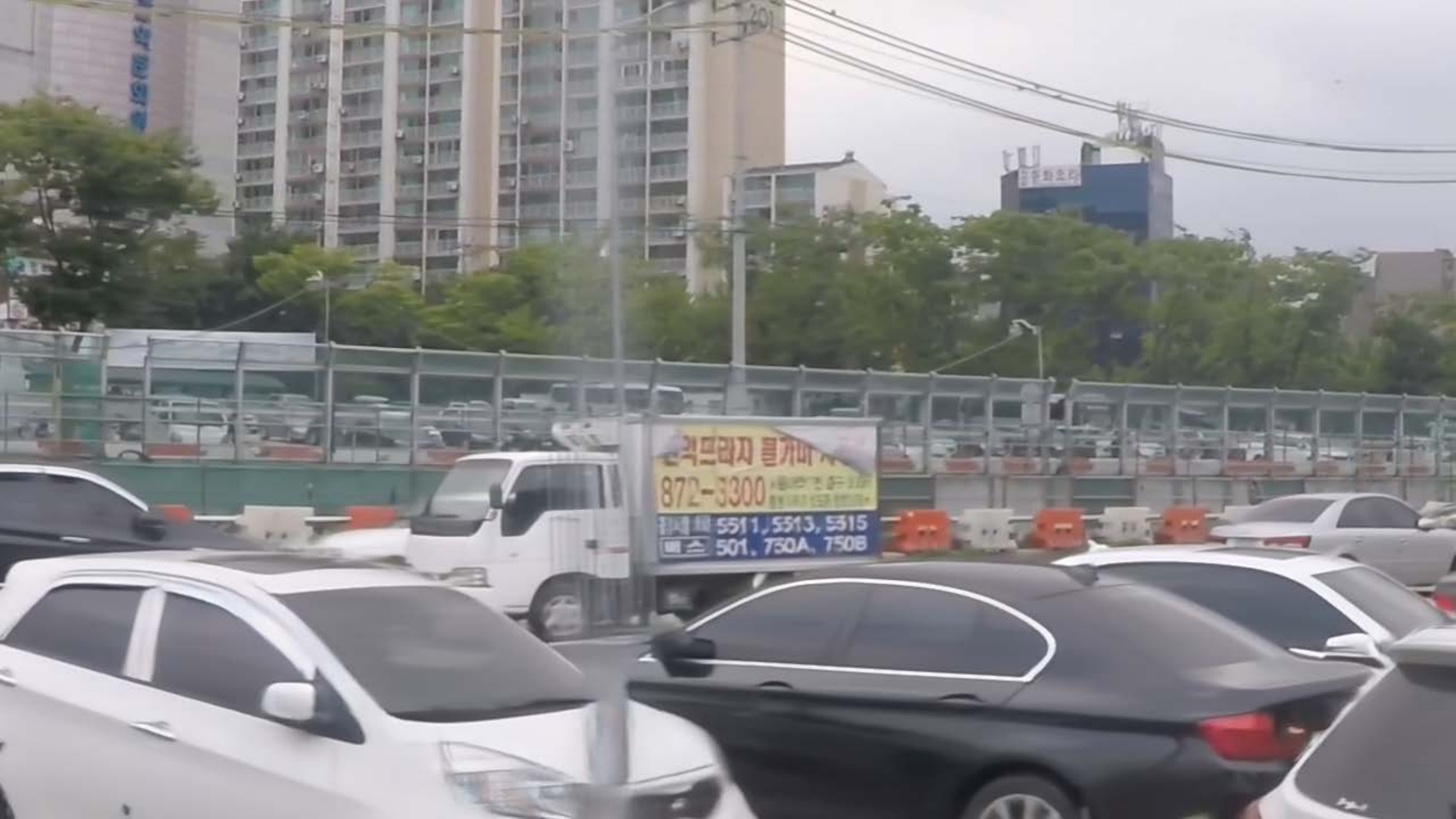}  &
\includegraphics[width=.25\textwidth,valign=t]{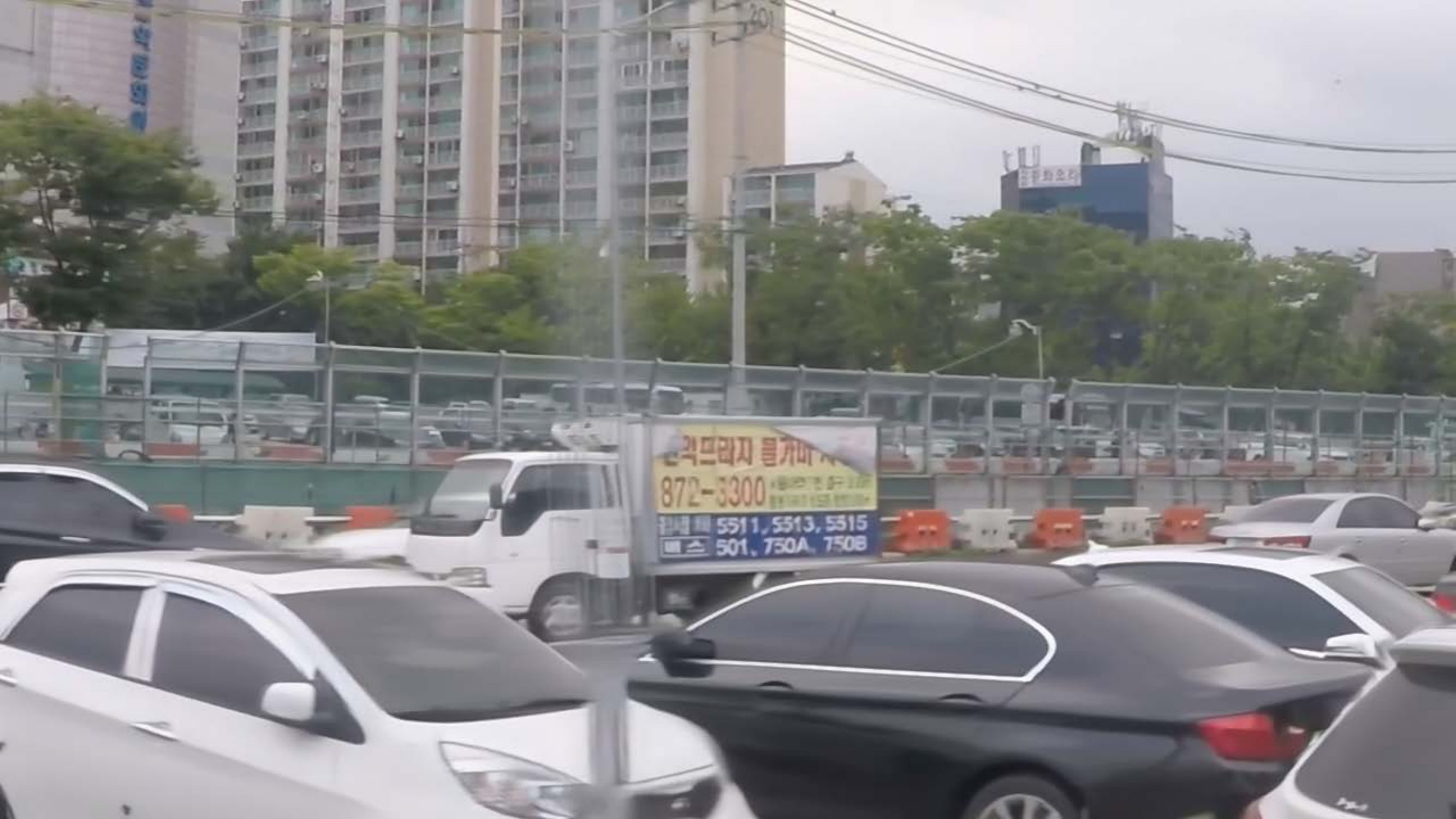} \\
\vspace{0.6em}
\small~MIMO-UNet+~\cite{MIMO-Unet+}, 29.10 dB  & \small~HINet~\cite{HINet}, 28.92 dB & \small~NAFNet~\cite{NAFNet}, 29.81 dB & \small~\textbf{Ours}, \textbf{30.18 dB} \\
\vspace{0.4em}
\includegraphics[width=.25\textwidth,valign=t]{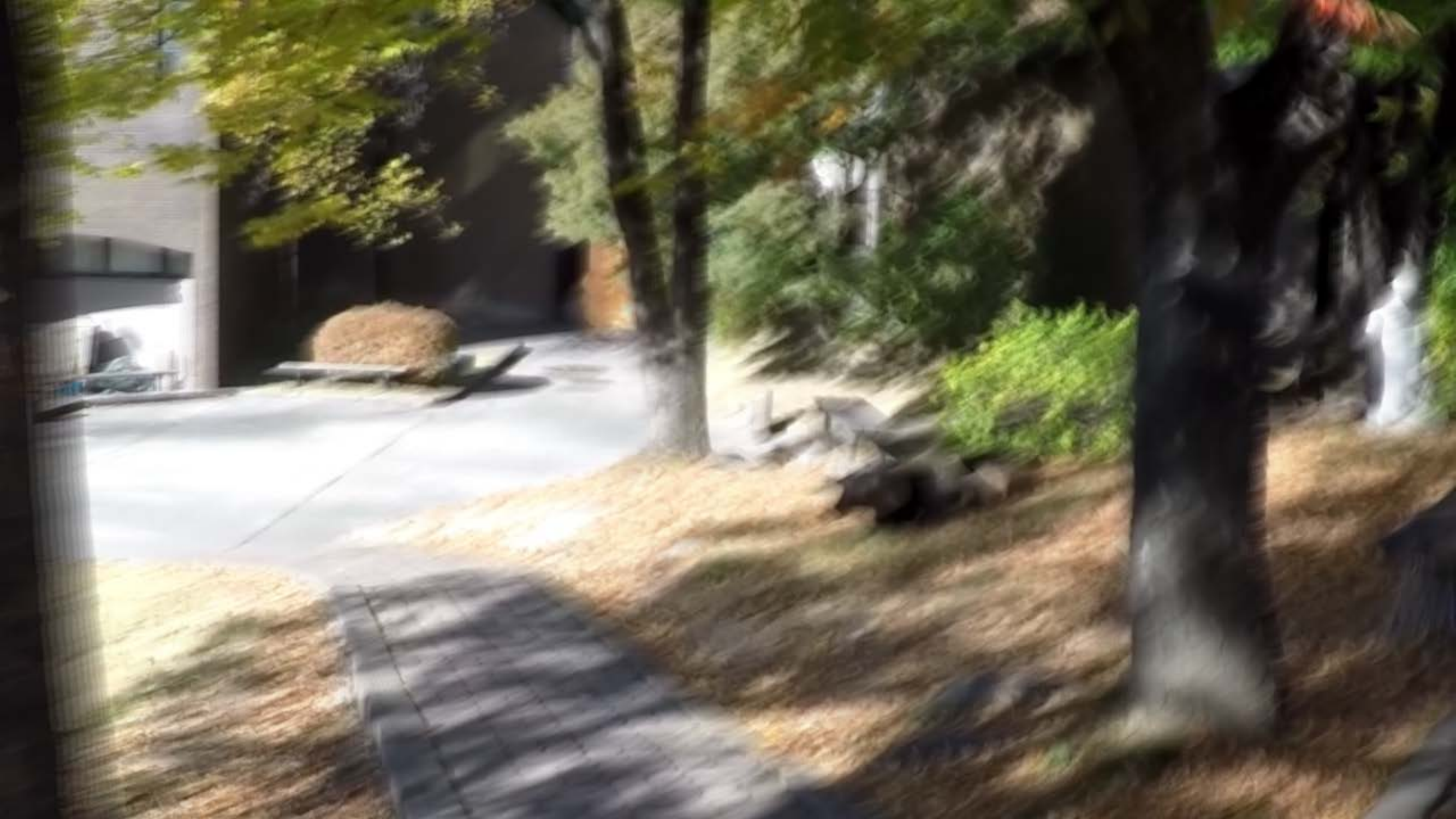}  &   
\includegraphics[width=.25\textwidth,valign=t]{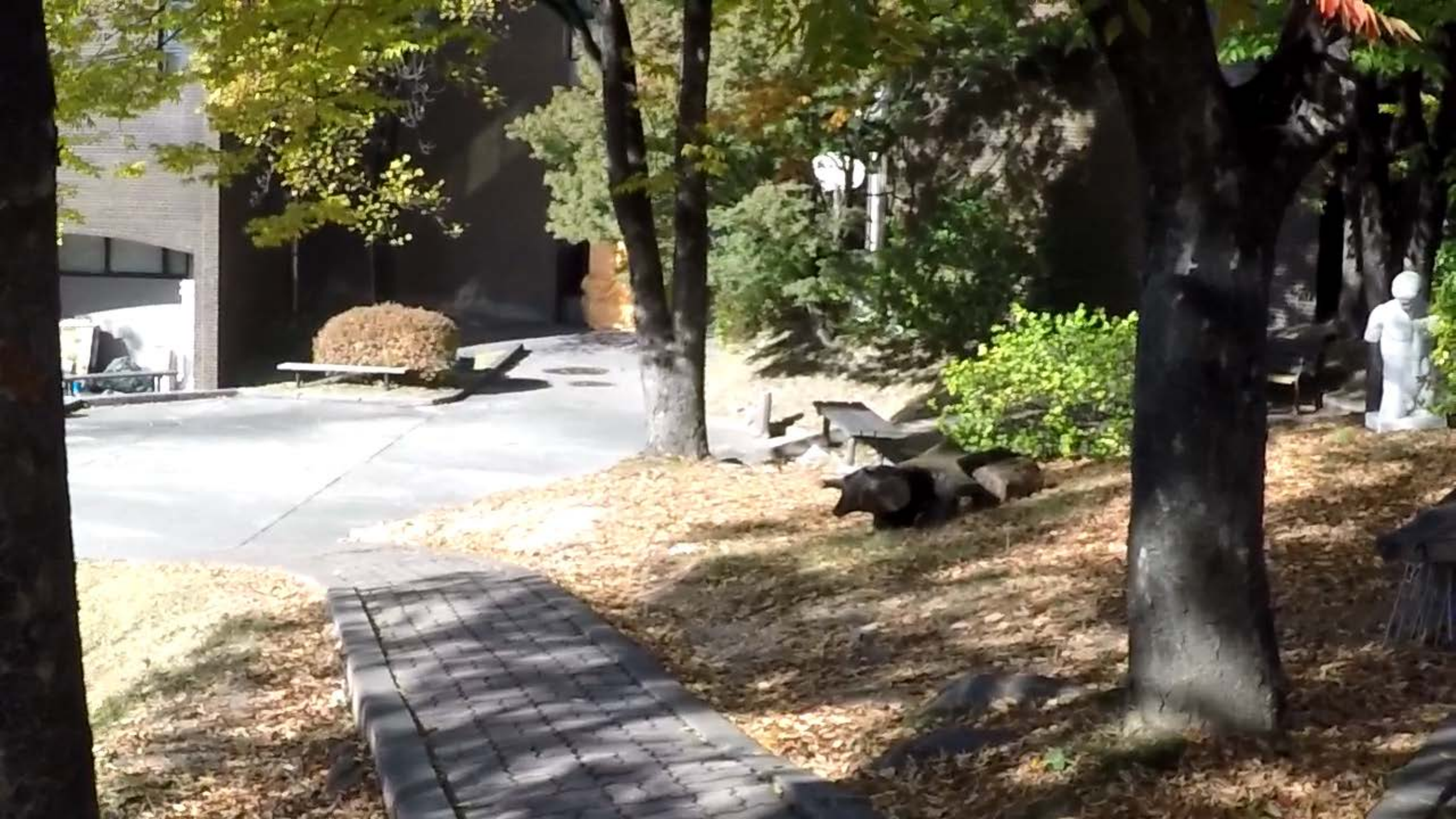} &   
\includegraphics[width=.25\textwidth,valign=t]{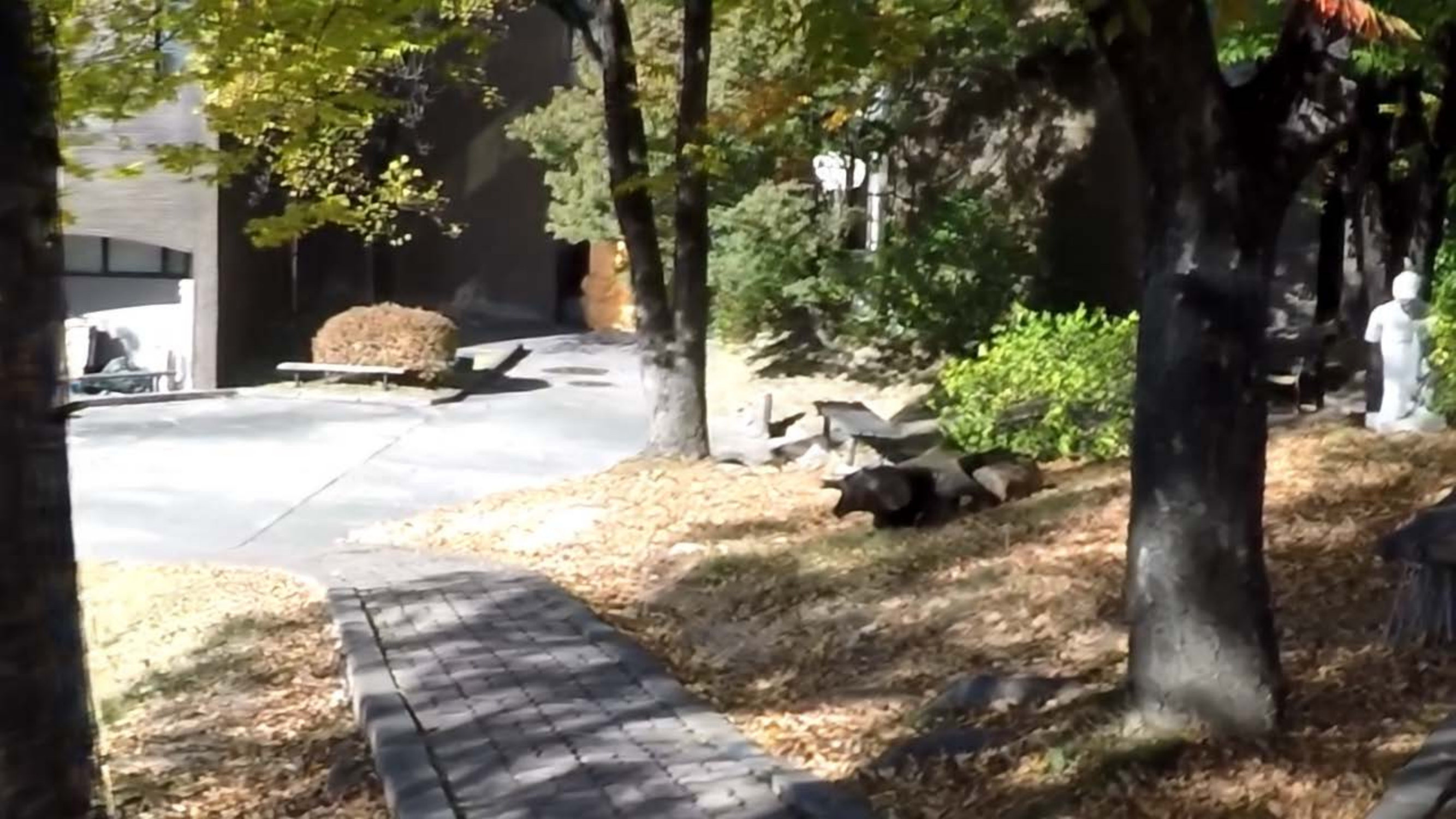}  &
\includegraphics[width=.25\textwidth,valign=t]{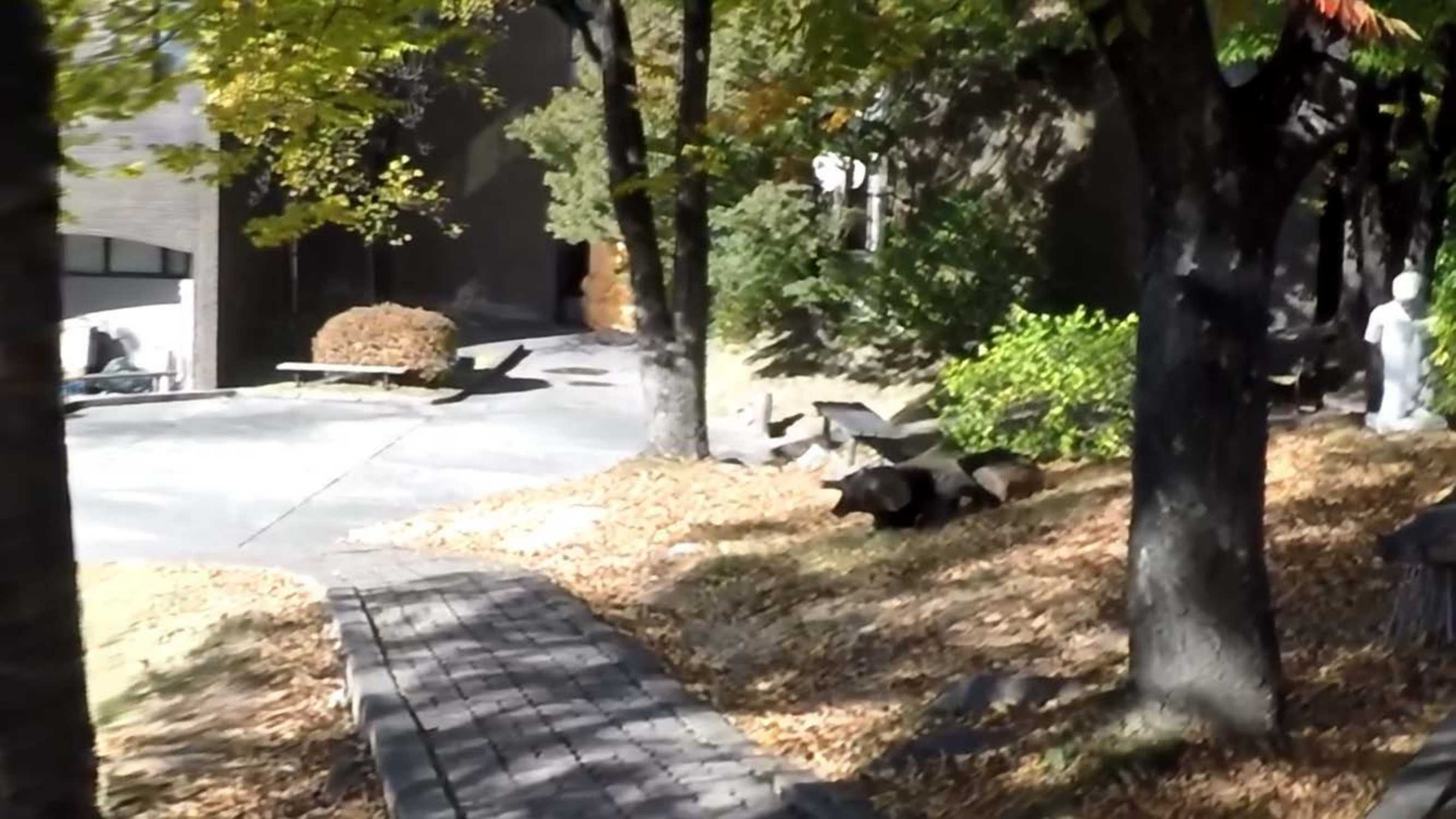} \\
\end{tabular}
}
\caption{\small Single-image motion deblurring on GoPro~\cite{GoPro}.}
\label{fig:suppl_single_image_motion_deblurring_visual_results_2}
\end{center}
\vspace{-7mm}
\end{figure*}

%% file: suppl_figures/suppl_005_01_defocus_deblurring_visual_results.tex
\begin{figure*}[t]
\begin{center}
\setlength{\tabcolsep}{1.5pt}
\scalebox{0.88}{
\begin{tabular}[b]{c c c c}
\vspace{0.4em}
\includegraphics[width=.25\textwidth,valign=t]{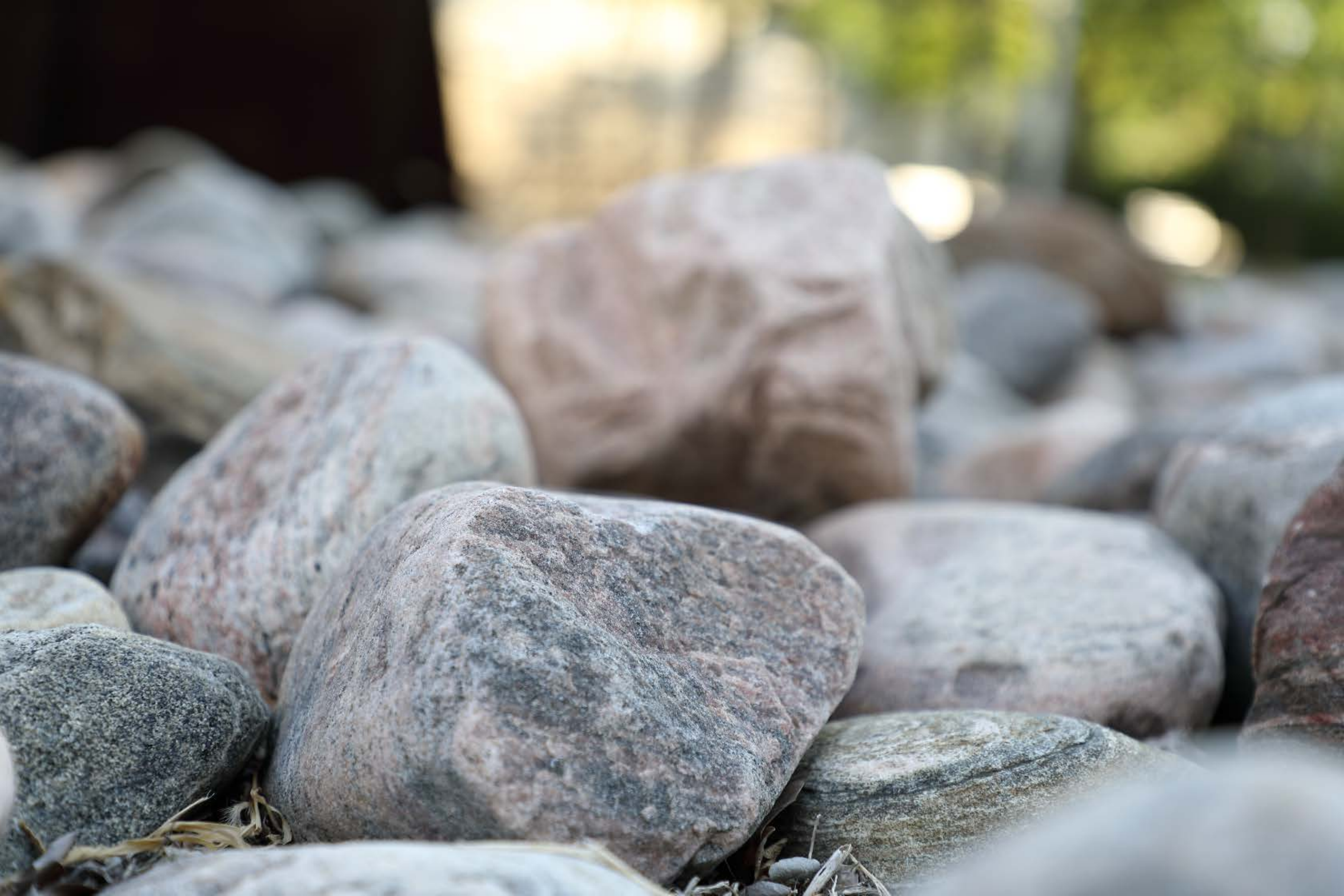}  &   
\includegraphics[width=.25\textwidth,valign=t]{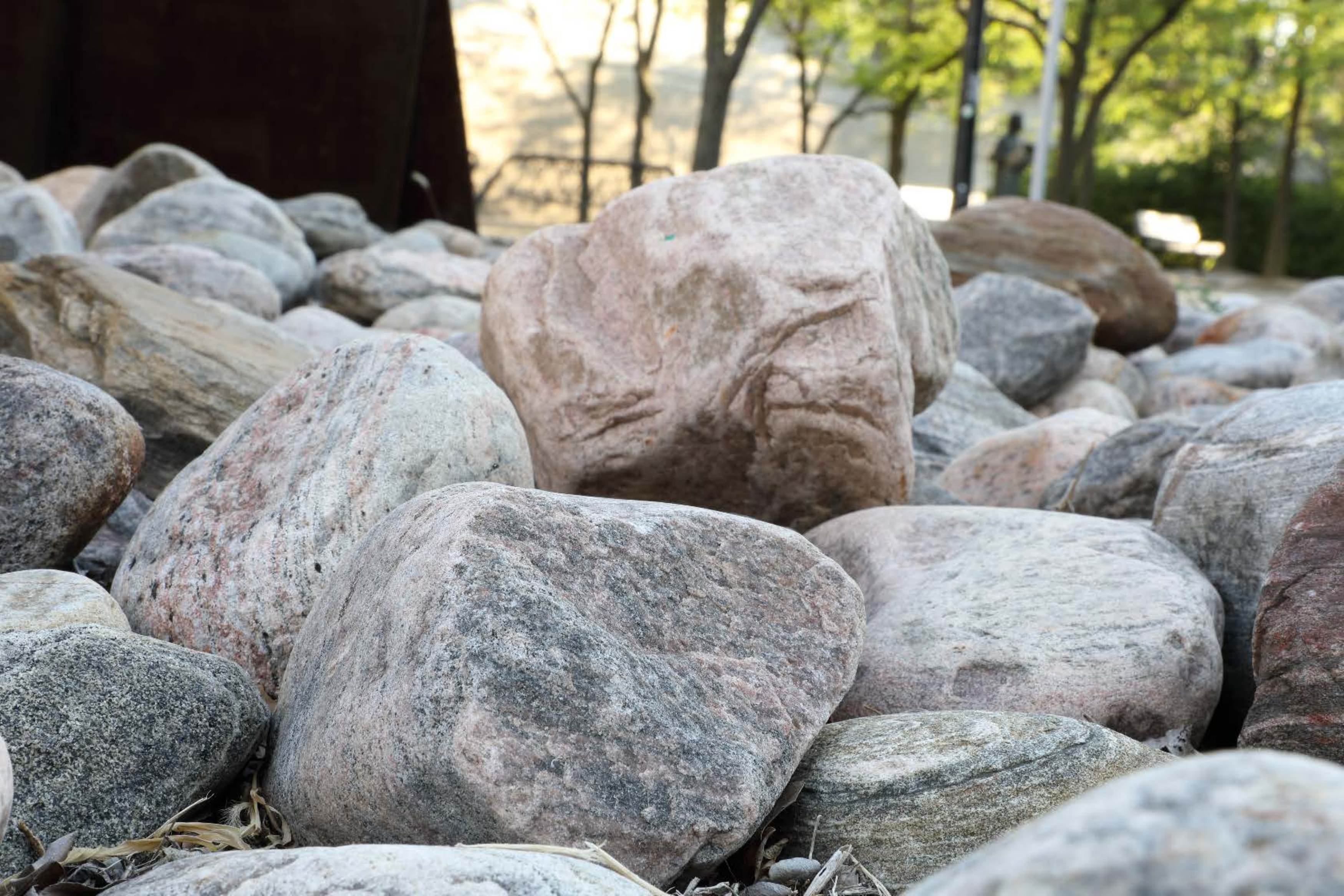} &   
\includegraphics[width=.25\textwidth,valign=t]{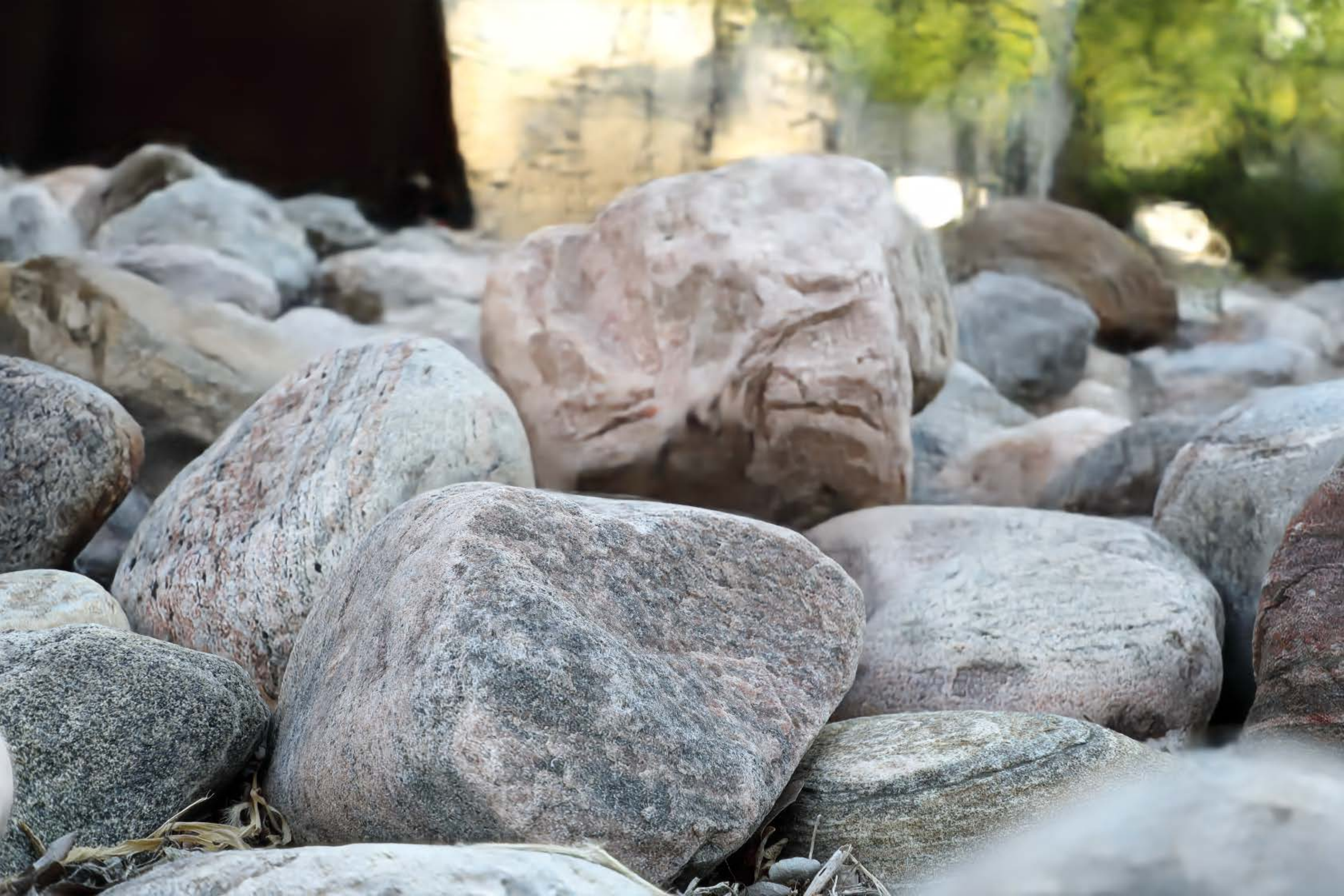}  &
\includegraphics[width=.25\textwidth,valign=t]{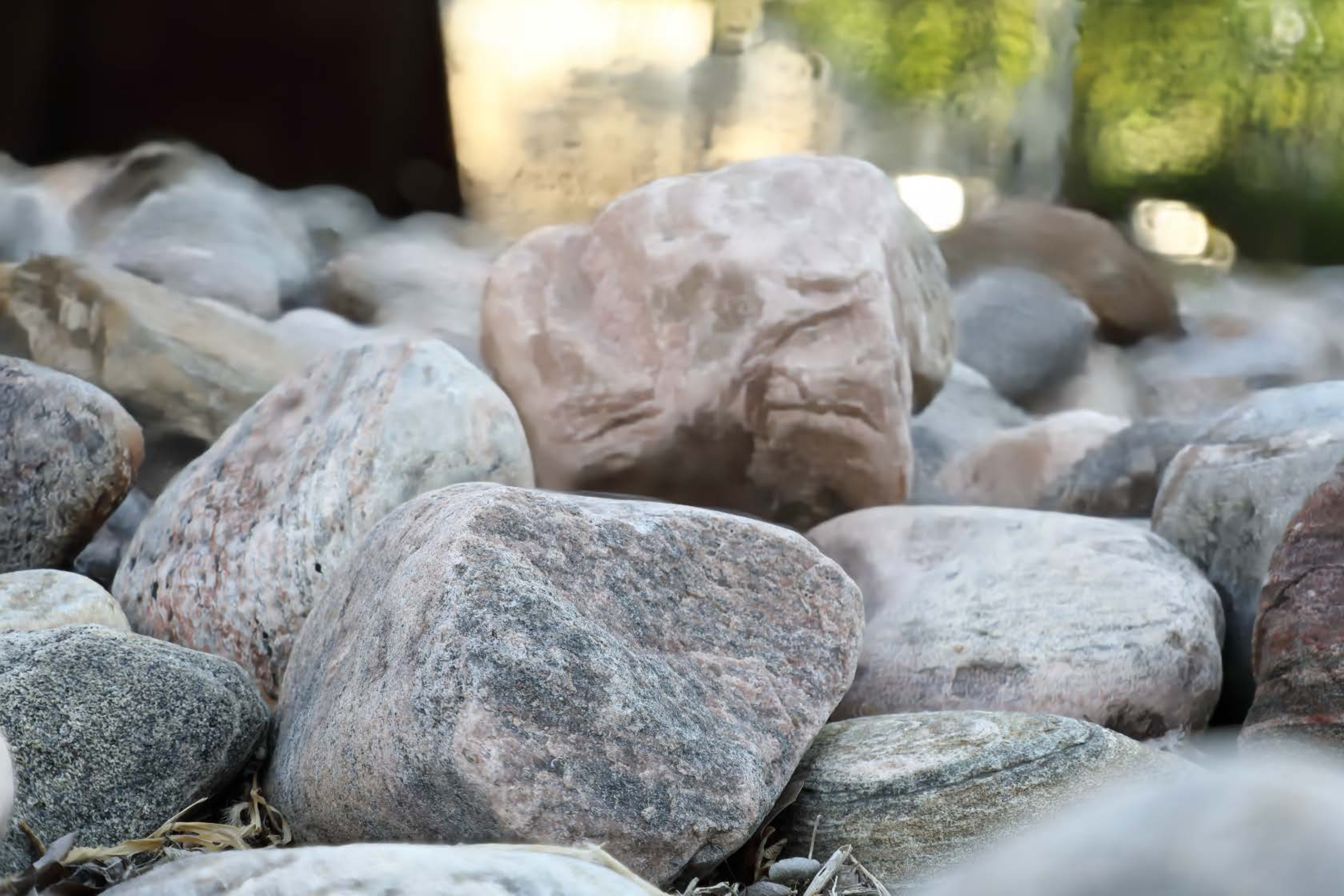} \\
\vspace{0.4em}
\small~Degraded  & \small~Label & \small~DMPHN~\cite{DMPHN}, 21.62 dB & \small~MPRNet~\cite{MPRNet}, 20.91 dB \\
\vspace{0.4em}
\includegraphics[width=.25\textwidth,valign=t]{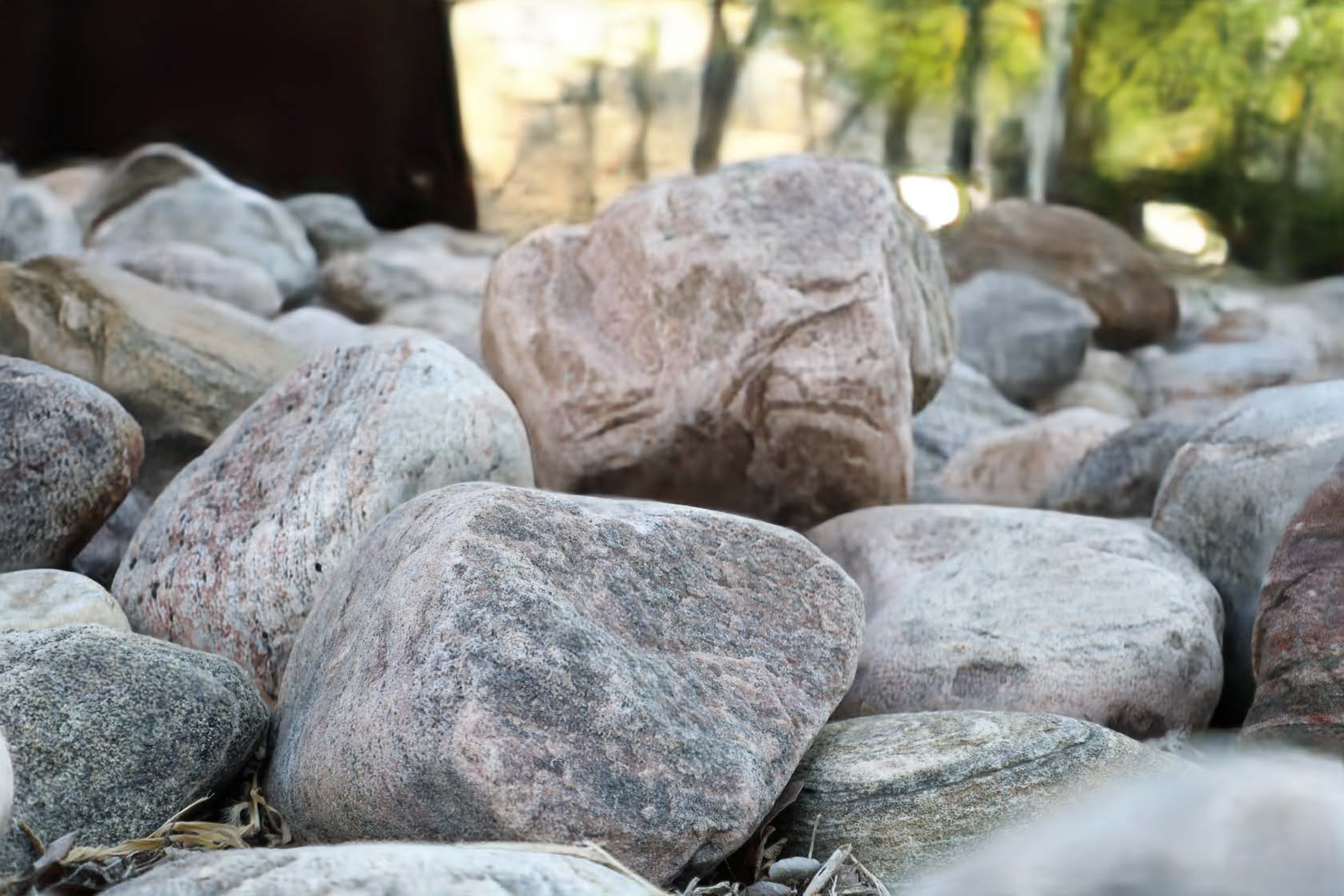}  &   
\includegraphics[width=.25\textwidth,valign=t]{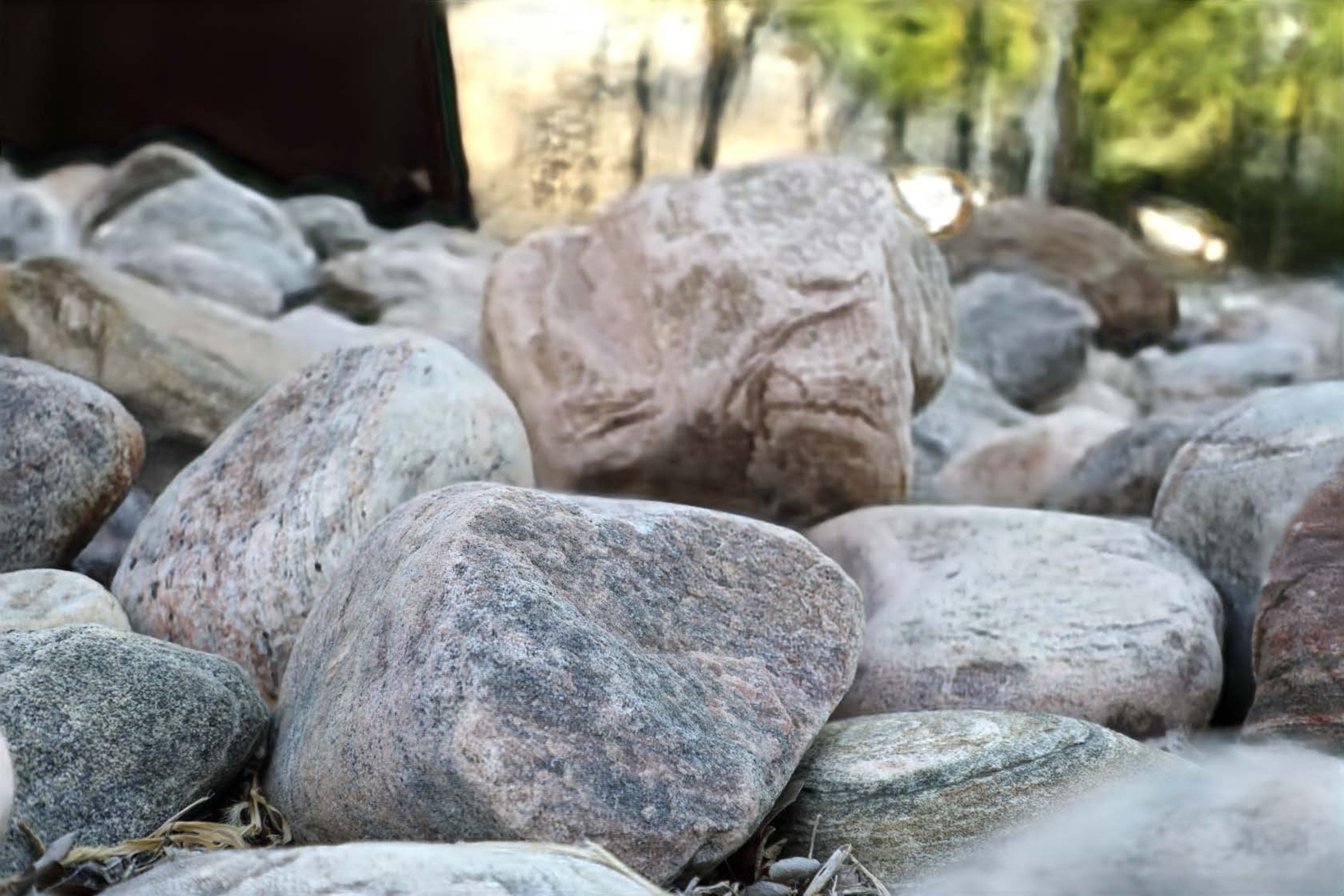} &   
\includegraphics[width=.25\textwidth,valign=t]{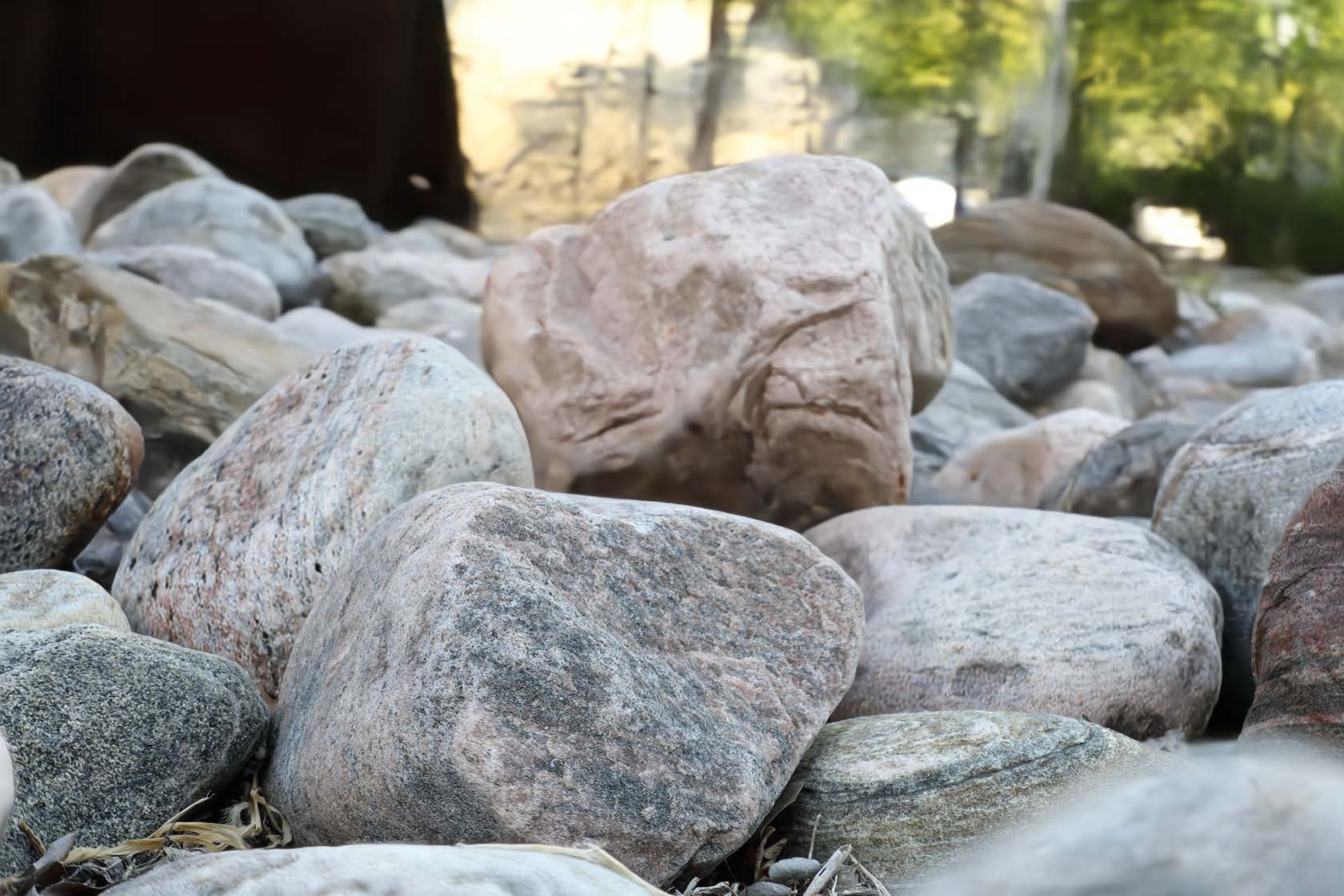}  &
\includegraphics[width=.25\textwidth,valign=t]{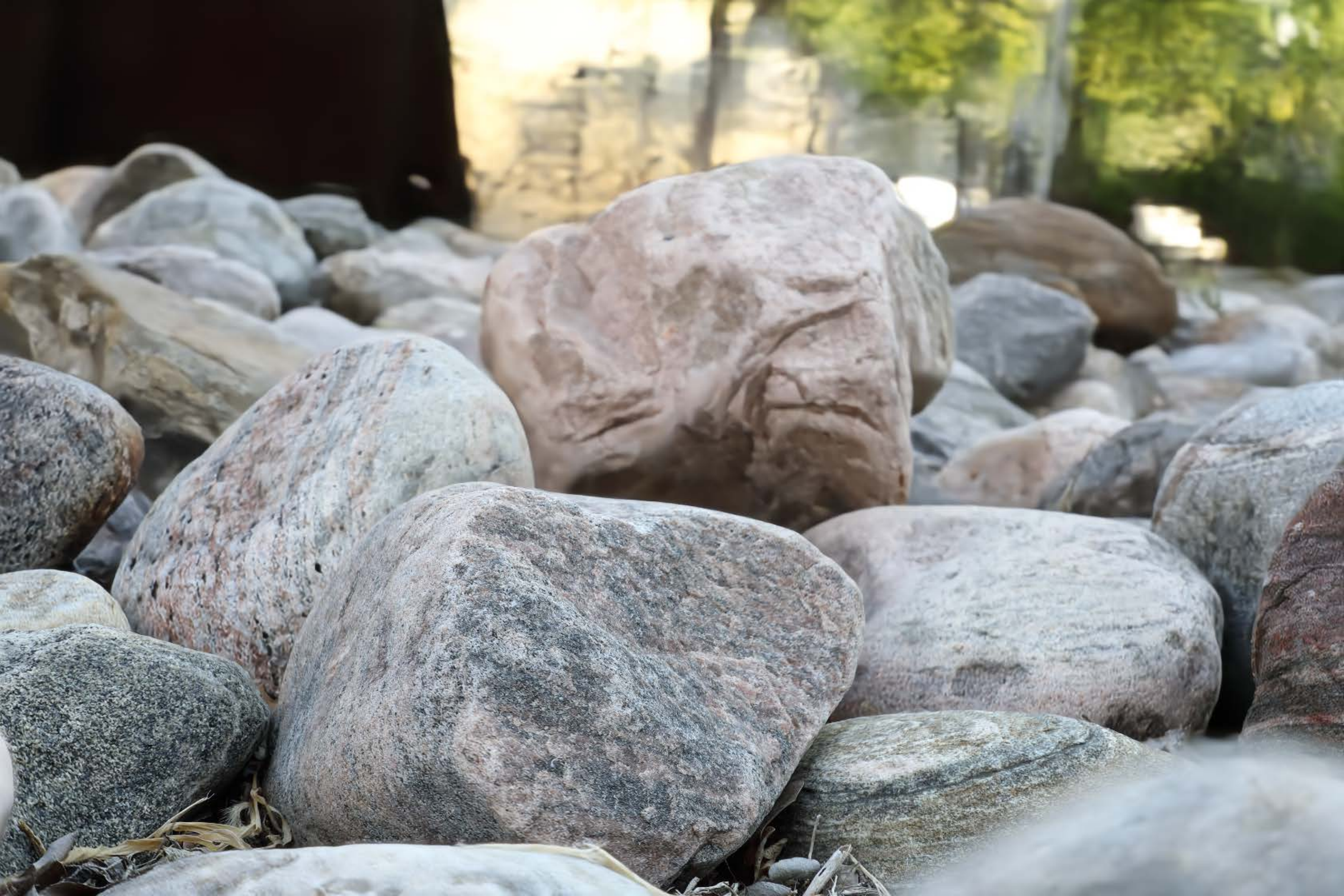} \\
\vspace{0.6em}
\small~DPDNet~\cite{DPDNet}, 22.09 dB  & \small~RDPD~\cite{RDPD}, 22.50 dB & \small~Restormer~\cite{Restormer}, 22.51 dB & \small~\textbf{Ours}, \textbf{22.64 dB} \\
\vspace{0.4em}
\includegraphics[width=.25\textwidth,valign=t]{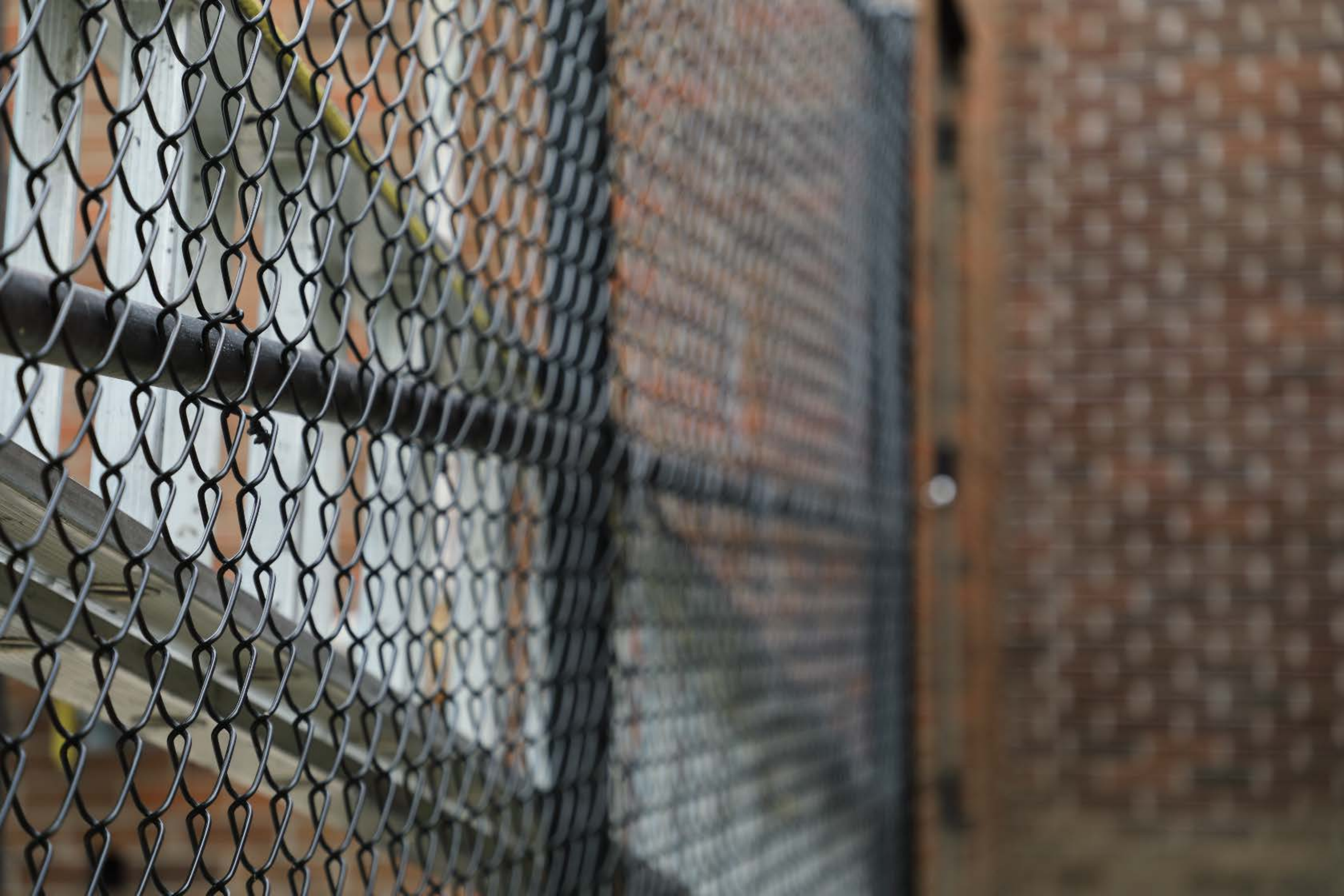}  &   
\includegraphics[width=.25\textwidth,valign=t]{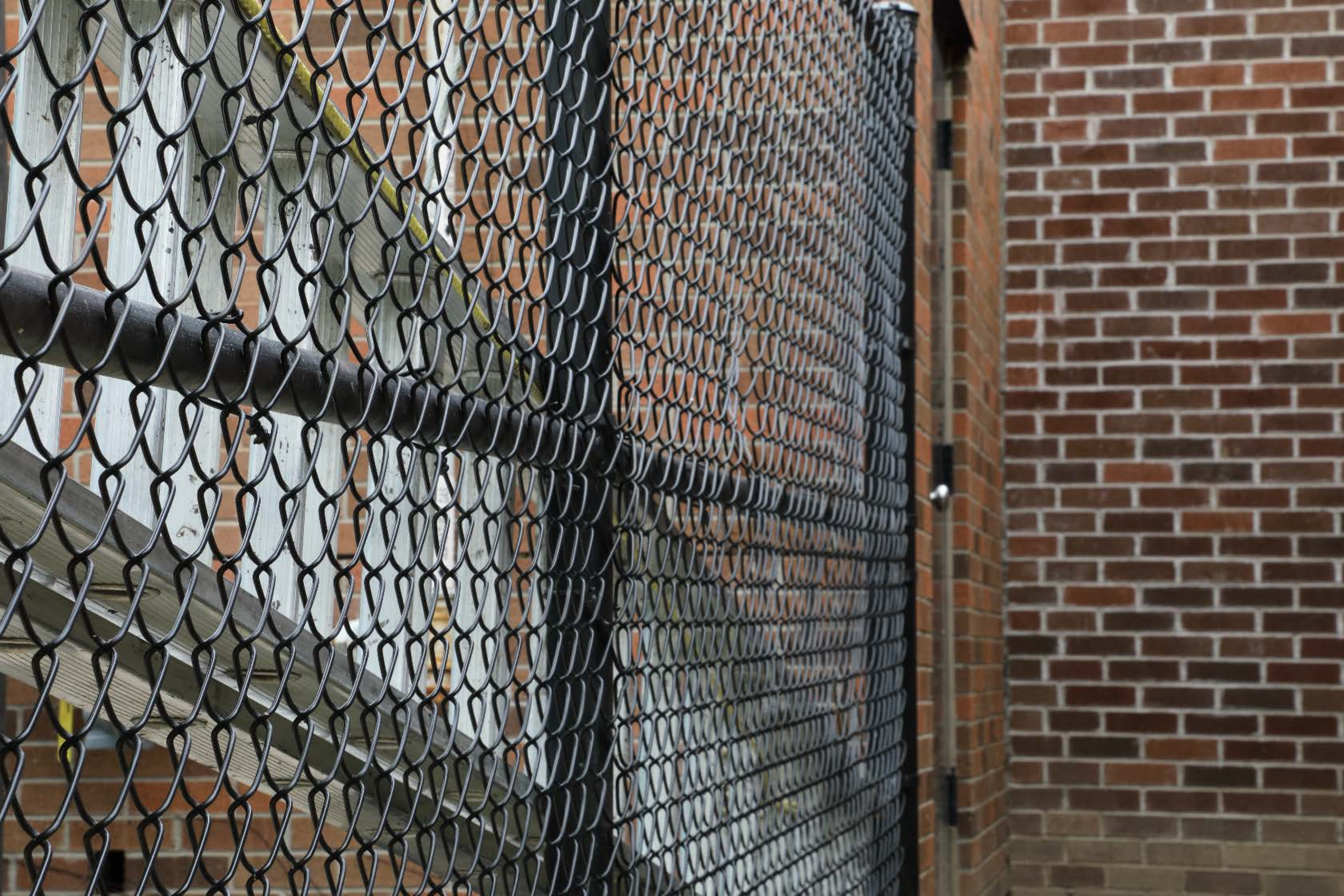} &   
\includegraphics[width=.25\textwidth,valign=t]{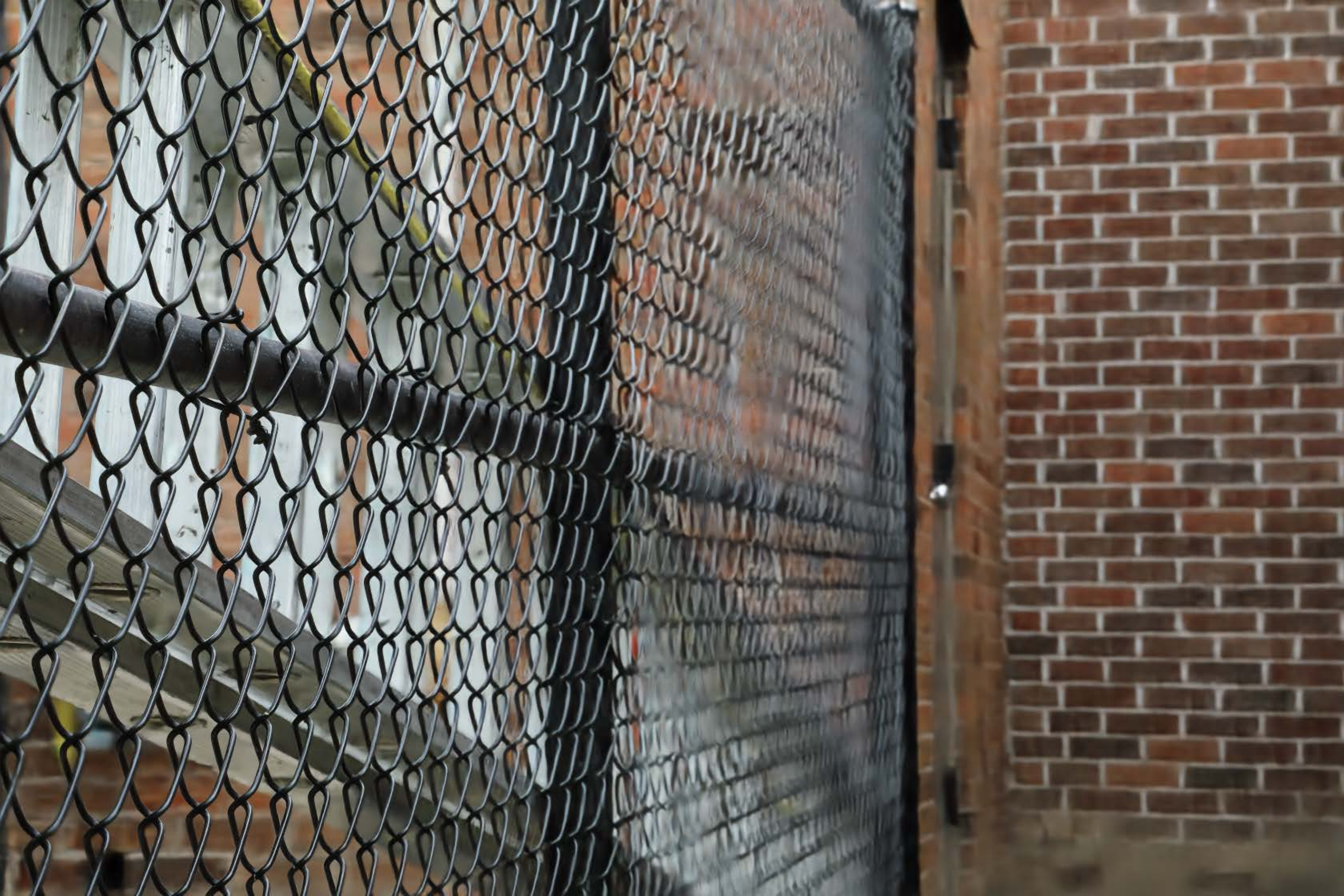}  &
\includegraphics[width=.25\textwidth,valign=t]{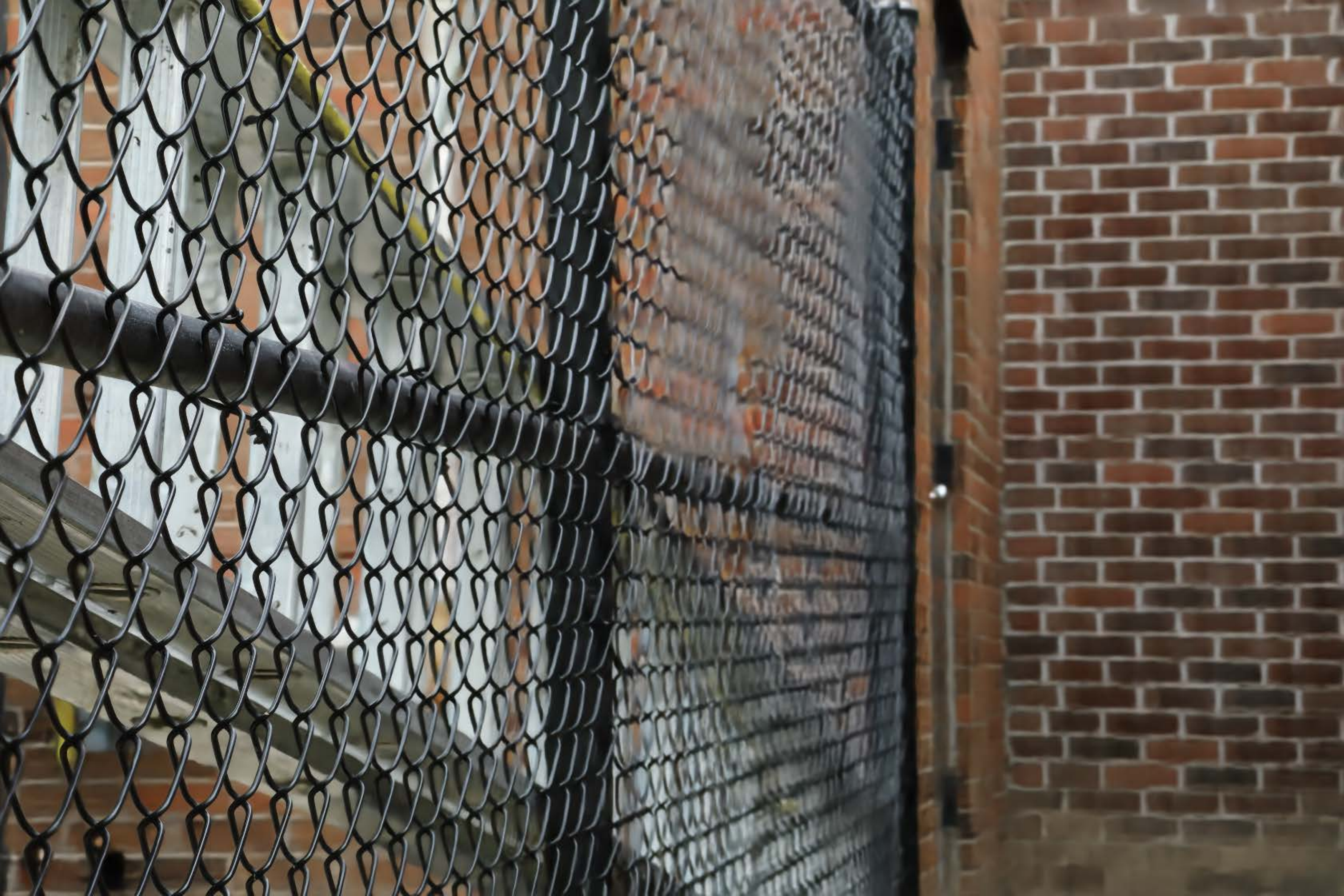} \\
\vspace{0.4em}
\small~Degraded  & \small~Label & \small~DMPHN~\cite{DMPHN}, 21.62 dB & \small~MPRNet~\cite{MPRNet}, 21.82 dB \\
\end{tabular}
}
\caption{\small Defocus deblurring on DPDD~\cite{DPDNet}.}
\label{fig:suppl_defocus_deblurring_visual_results_1}
\end{center}
\vspace{-7mm}
\end{figure*}

%% file: suppl_figures/suppl_005_02_defocus_deblurring_visual_results.tex
\begin{figure*}[t]
\begin{center}
\setlength{\tabcolsep}{1.5pt}
\scalebox{0.88}{
\begin{tabular}[b]{c c c c}
\vspace{0.4em}
\includegraphics[width=.25\textwidth,valign=t]{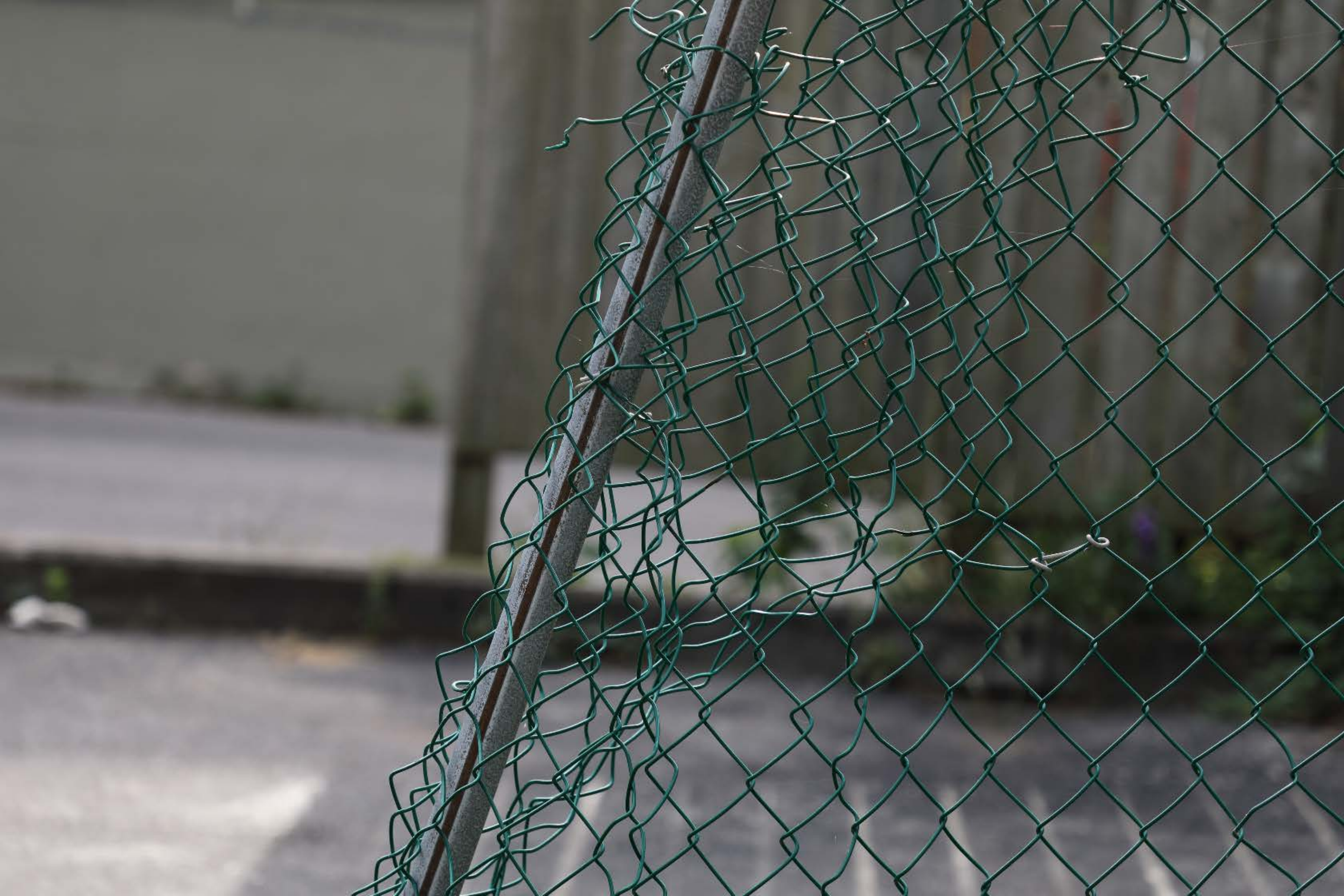}  &   
\includegraphics[width=.25\textwidth,valign=t]{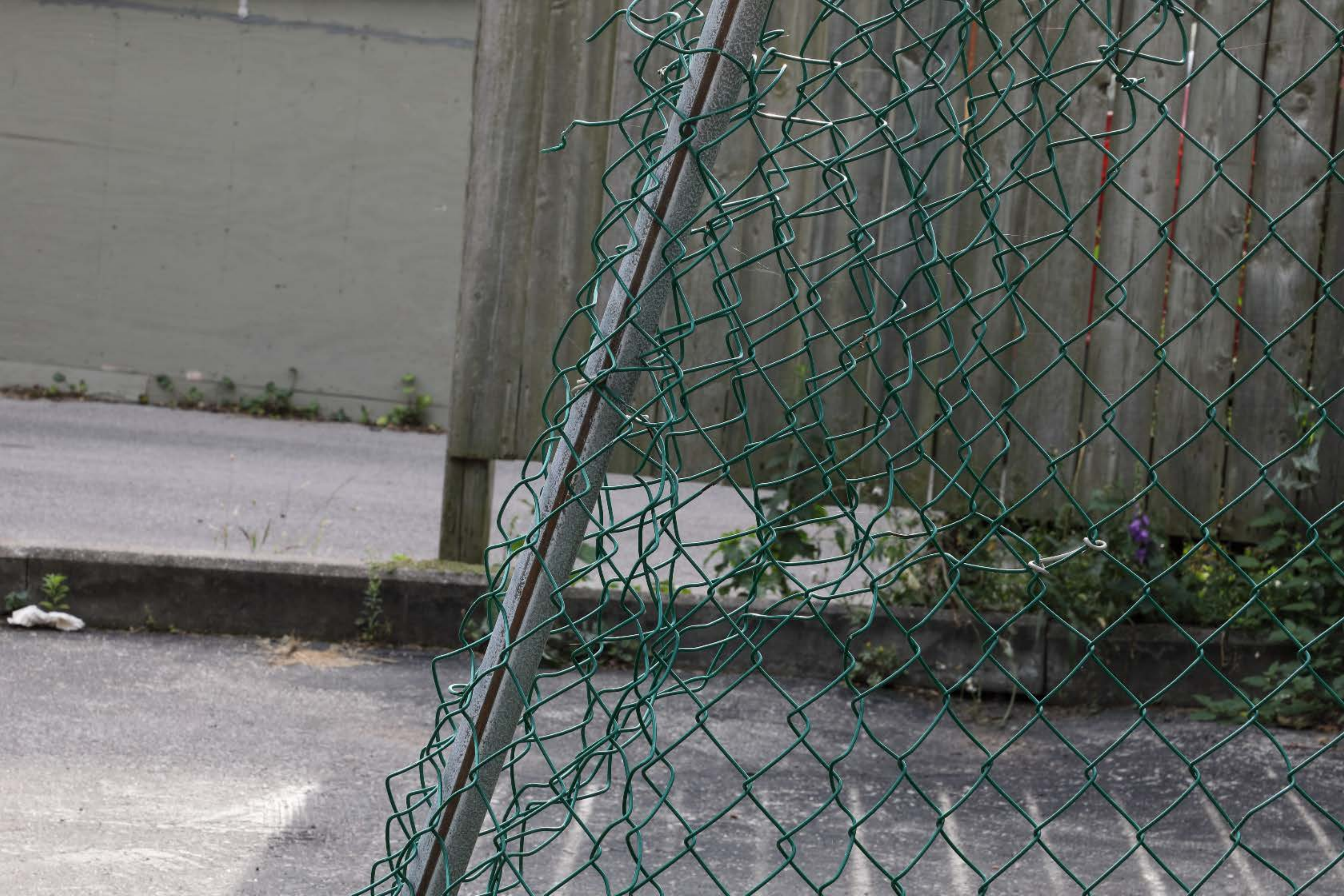} &   
\includegraphics[width=.25\textwidth,valign=t]{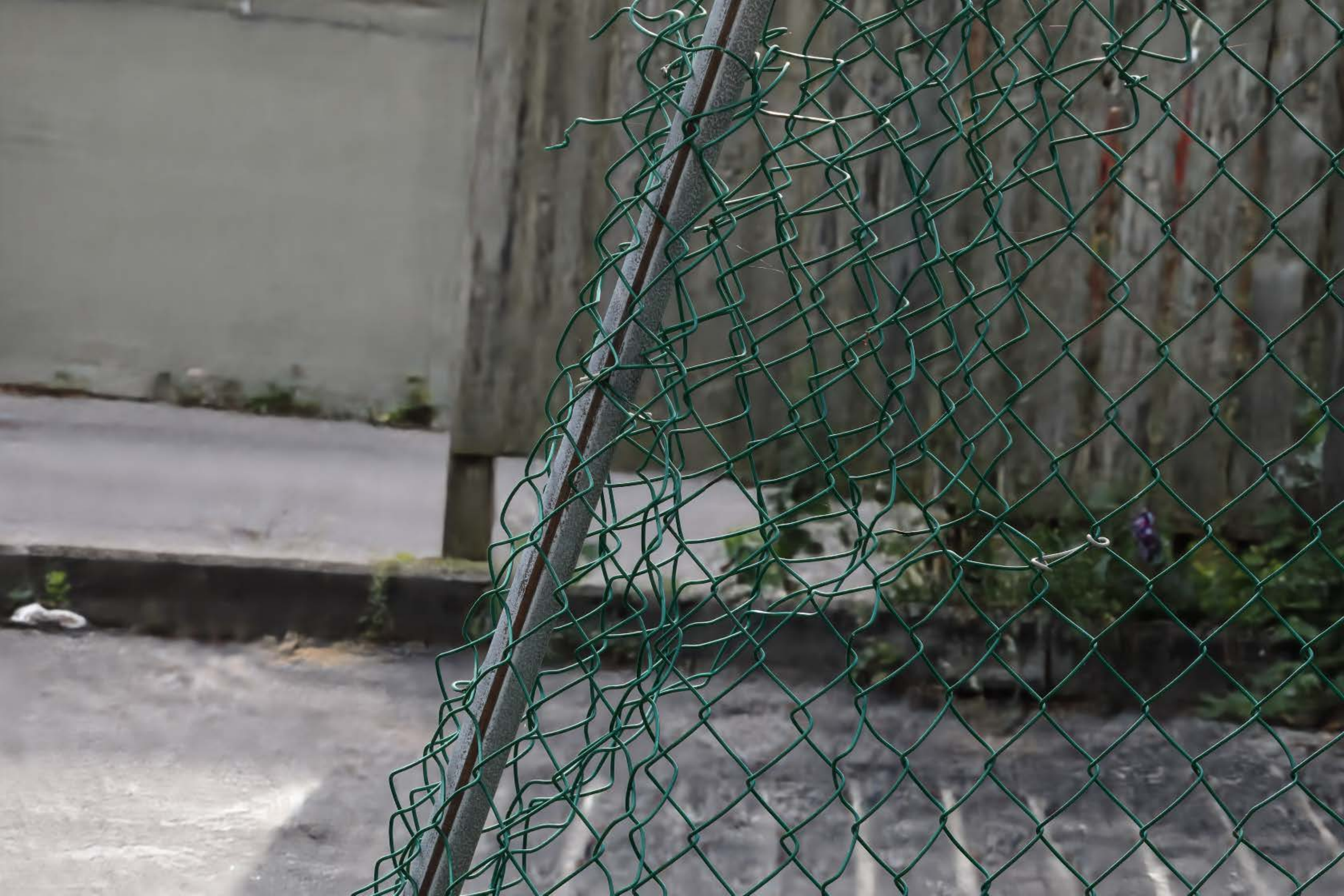}  &
\includegraphics[width=.25\textwidth,valign=t]{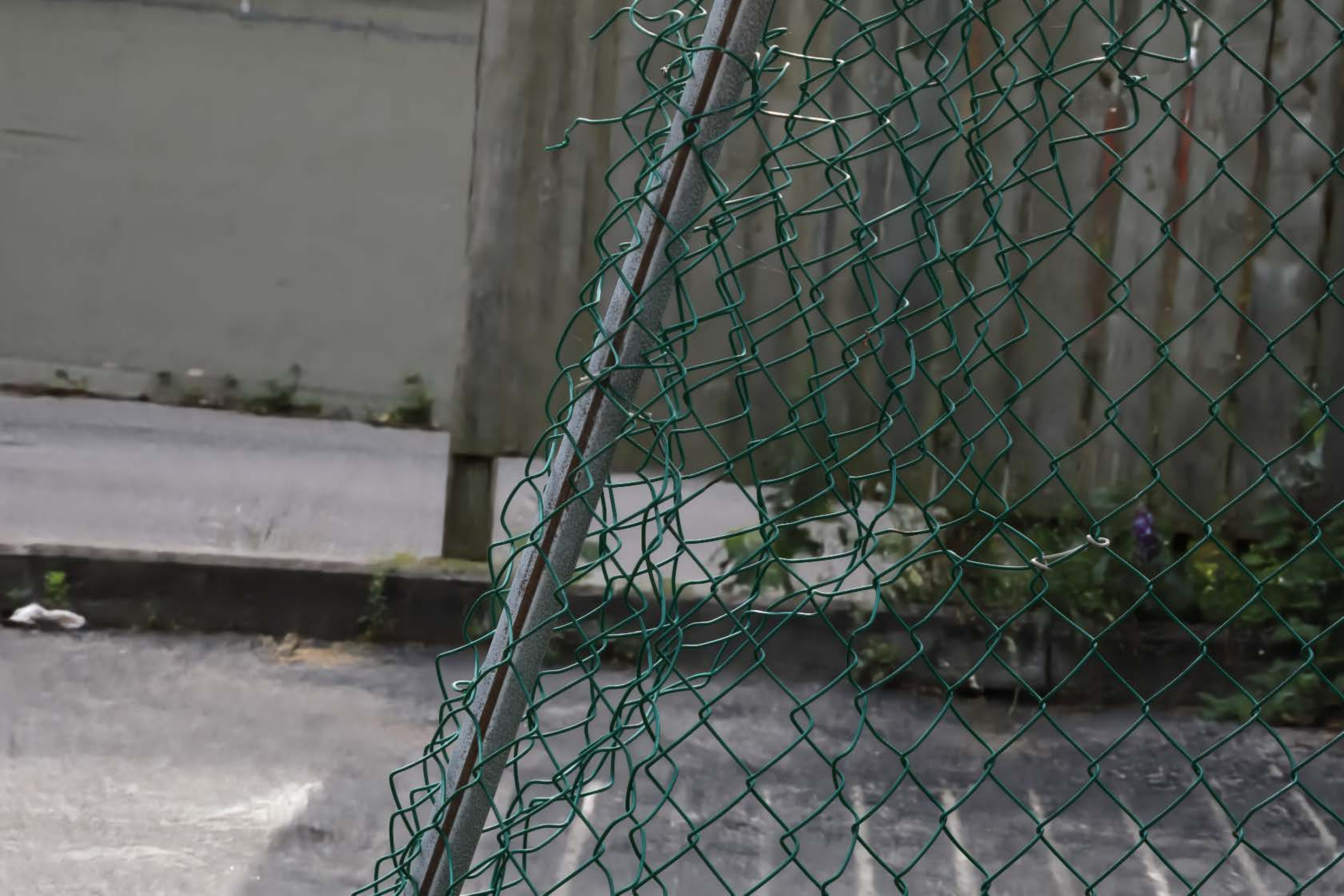} \\
\vspace{0.4em}
\small~Degraded  & \small~Label & \small~DMPHN~\cite{DMPHN}, 18.34 dB & \small~MPRNet~\cite{MPRNet}, 18.46 dB \\
\vspace{0.4em}
\includegraphics[width=.25\textwidth,valign=t]{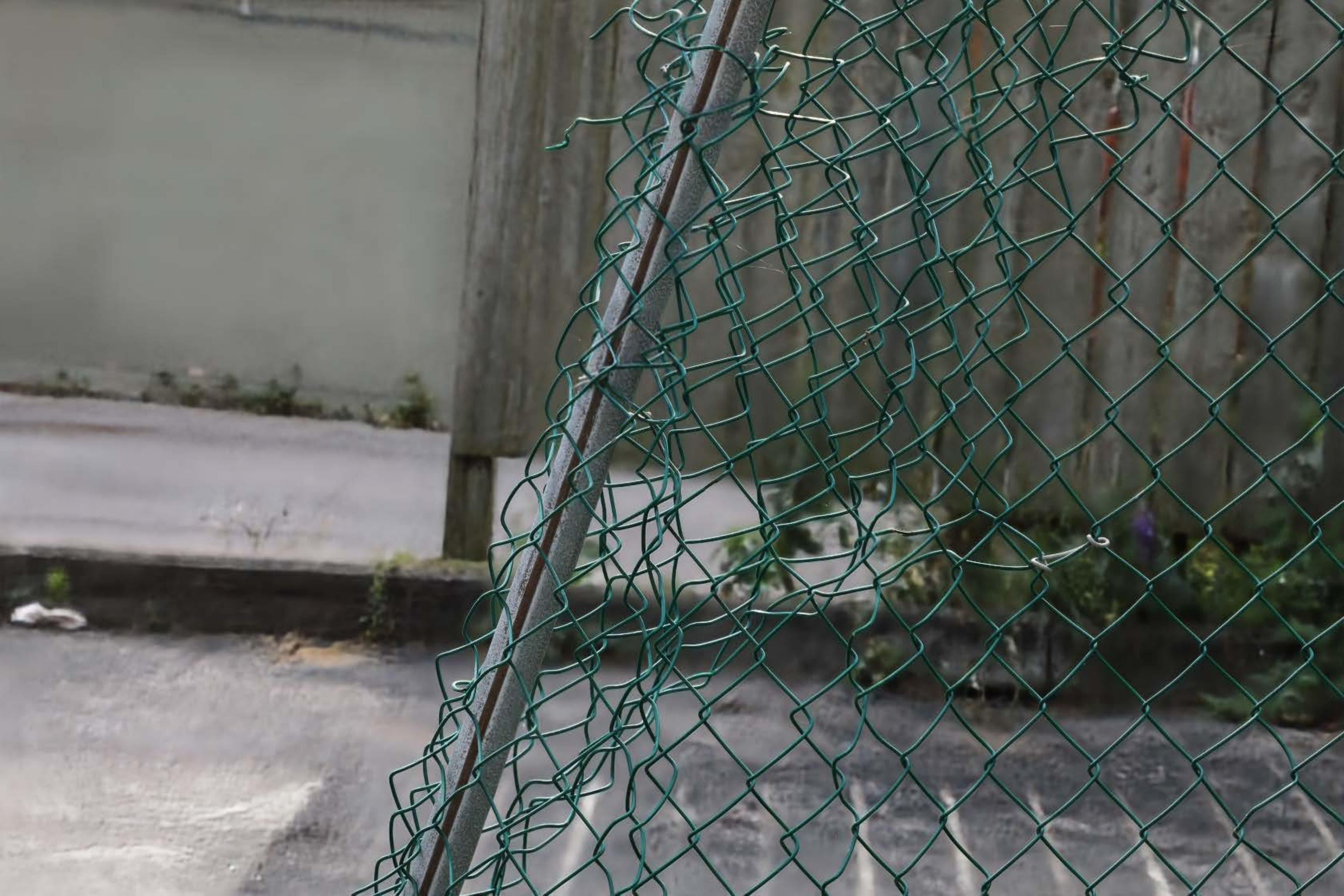}  &   
\includegraphics[width=.25\textwidth,valign=t]{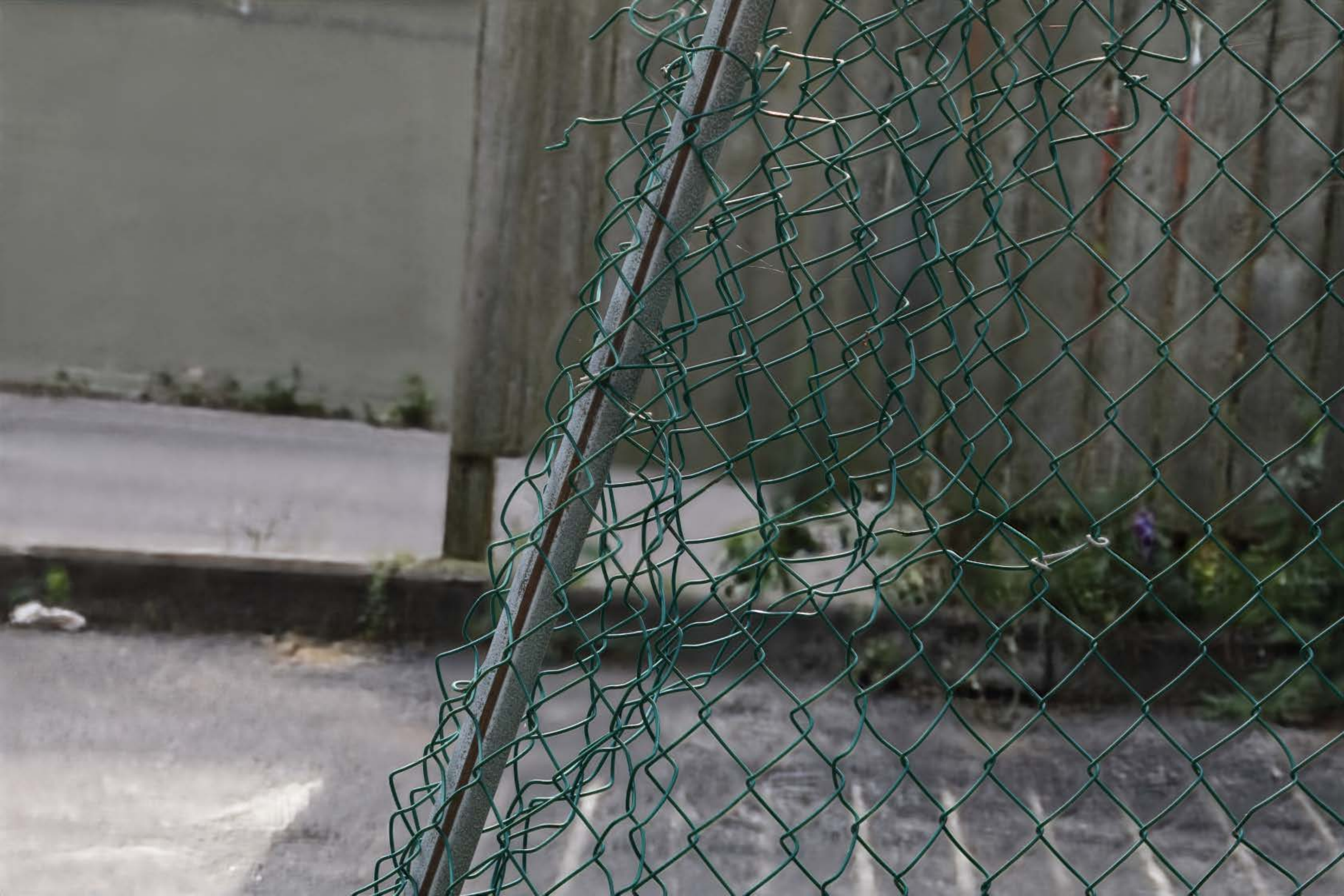} &   
\includegraphics[width=.25\textwidth,valign=t]{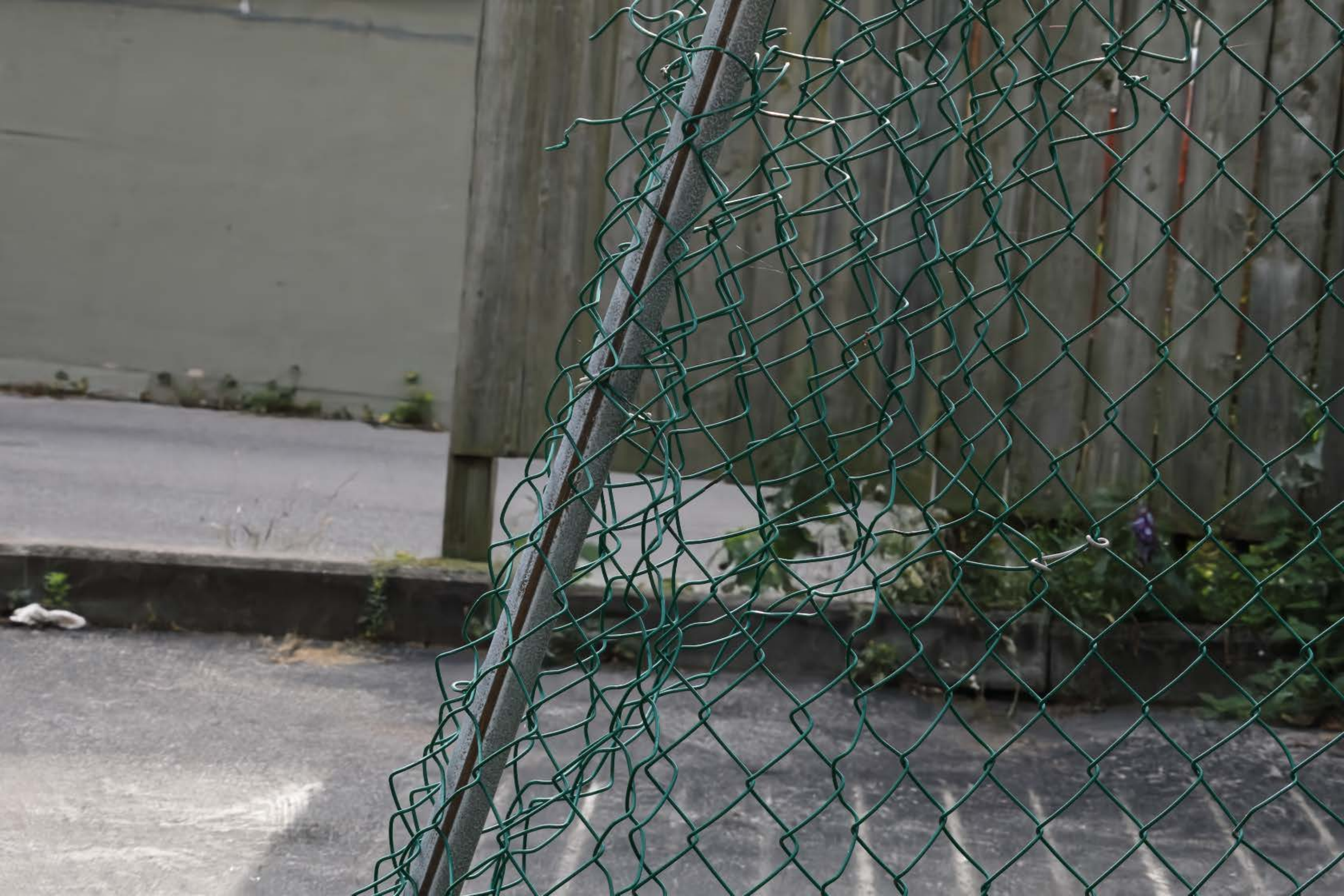}  &
\includegraphics[width=.25\textwidth,valign=t]{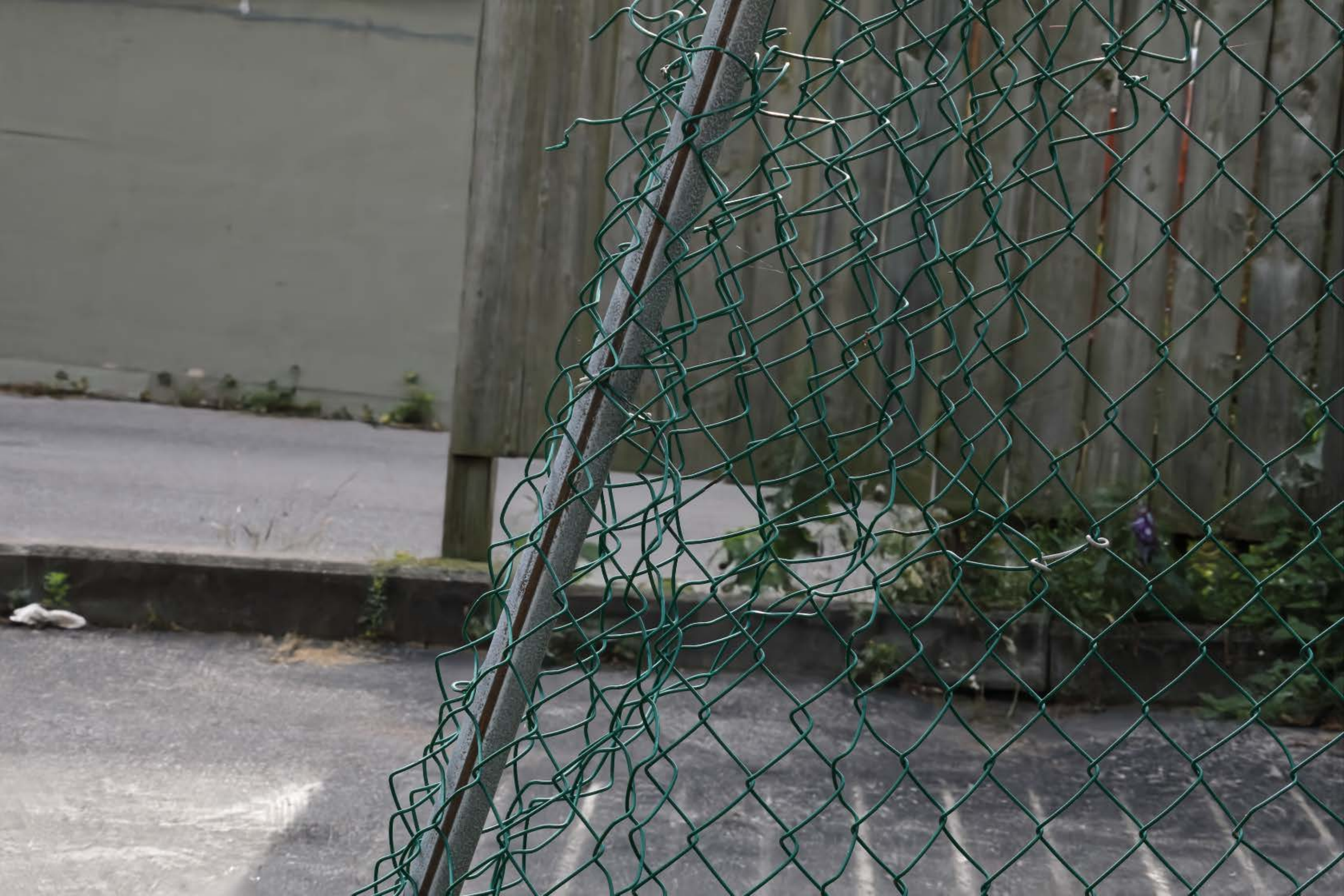} \\
\vspace{0.6em}
\small~DPDNet~\cite{DPDNet}, 18.54 dB  & \small~RDPD~\cite{RDPD}, 18.61 dB & \small~Restormer~\cite{Restormer}, 18.63 dB & \small~\textbf{Ours}, \textbf{18.75 dB} \\
\vspace{0.4em}
\includegraphics[width=.25\textwidth,valign=t]{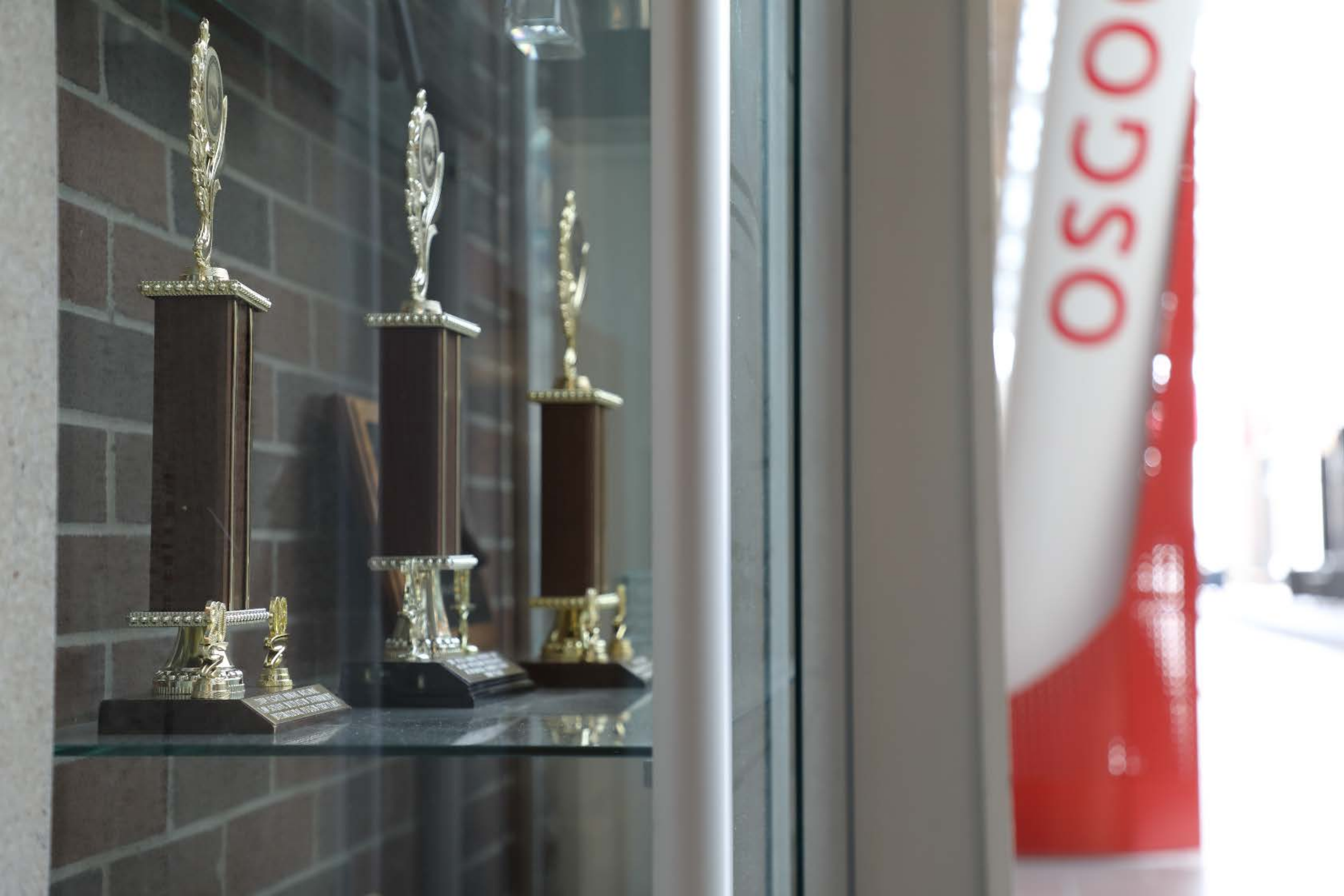}  &   
\includegraphics[width=.25\textwidth,valign=t]{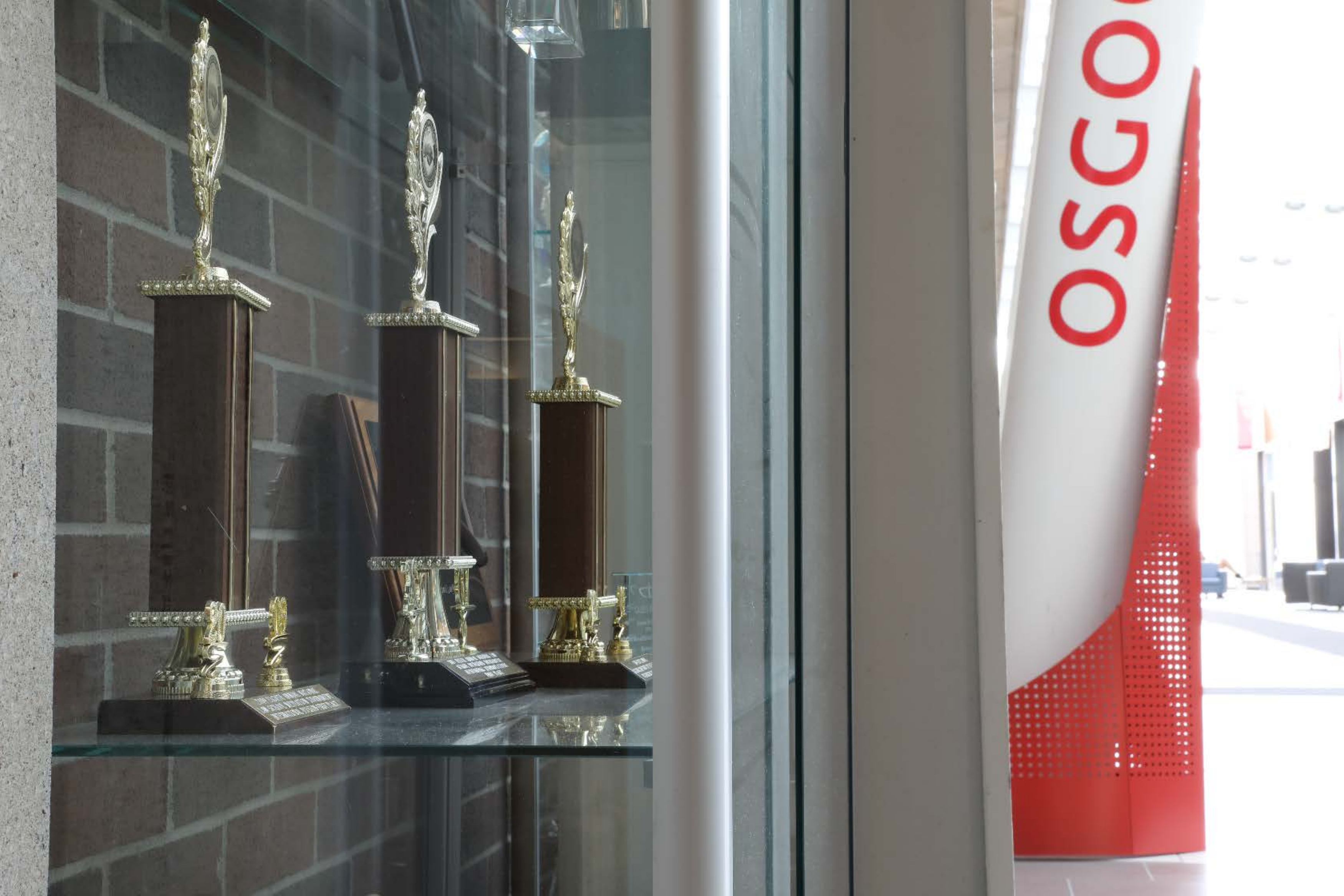} &   
\includegraphics[width=.25\textwidth,valign=t]{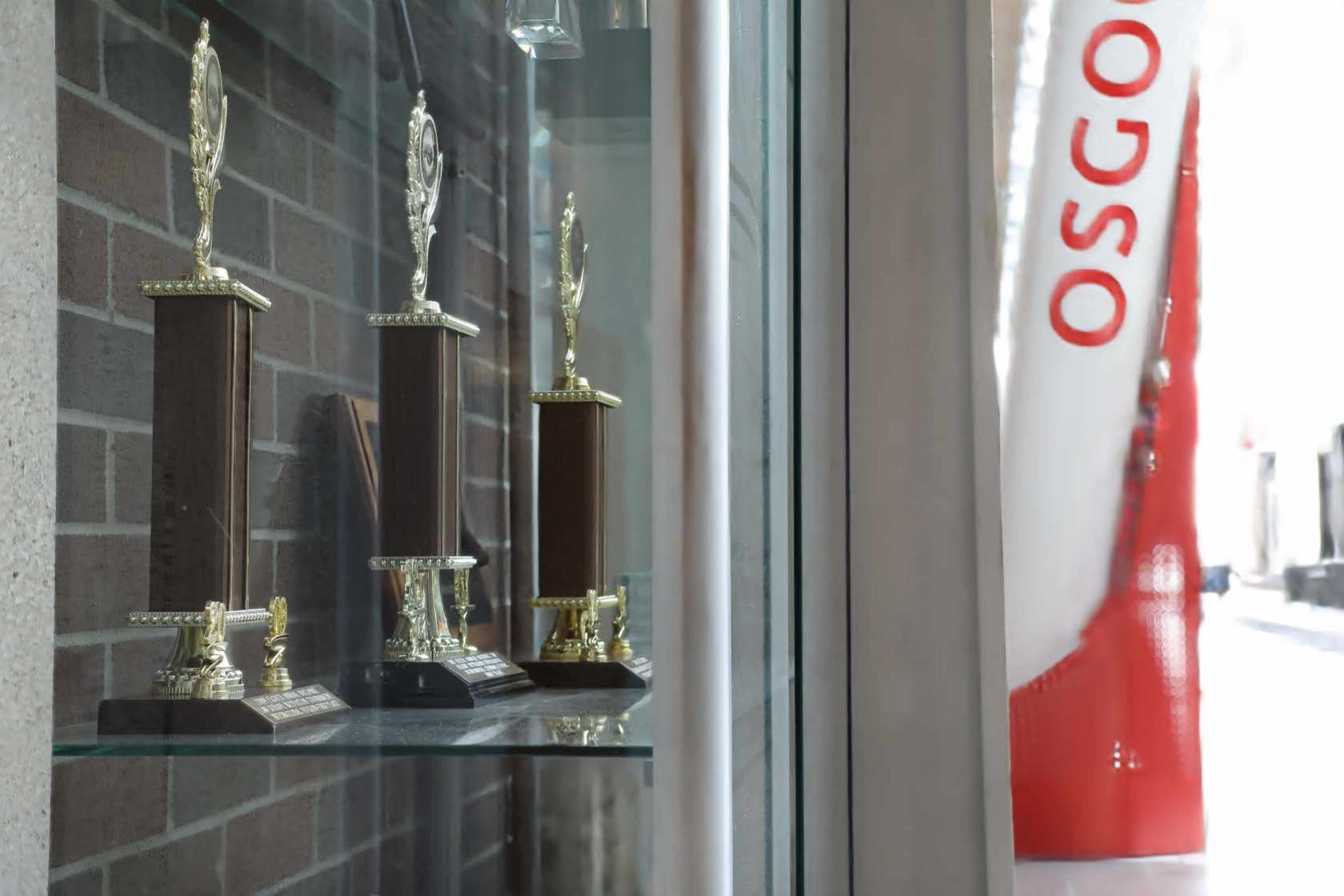}  &
\includegraphics[width=.25\textwidth,valign=t]{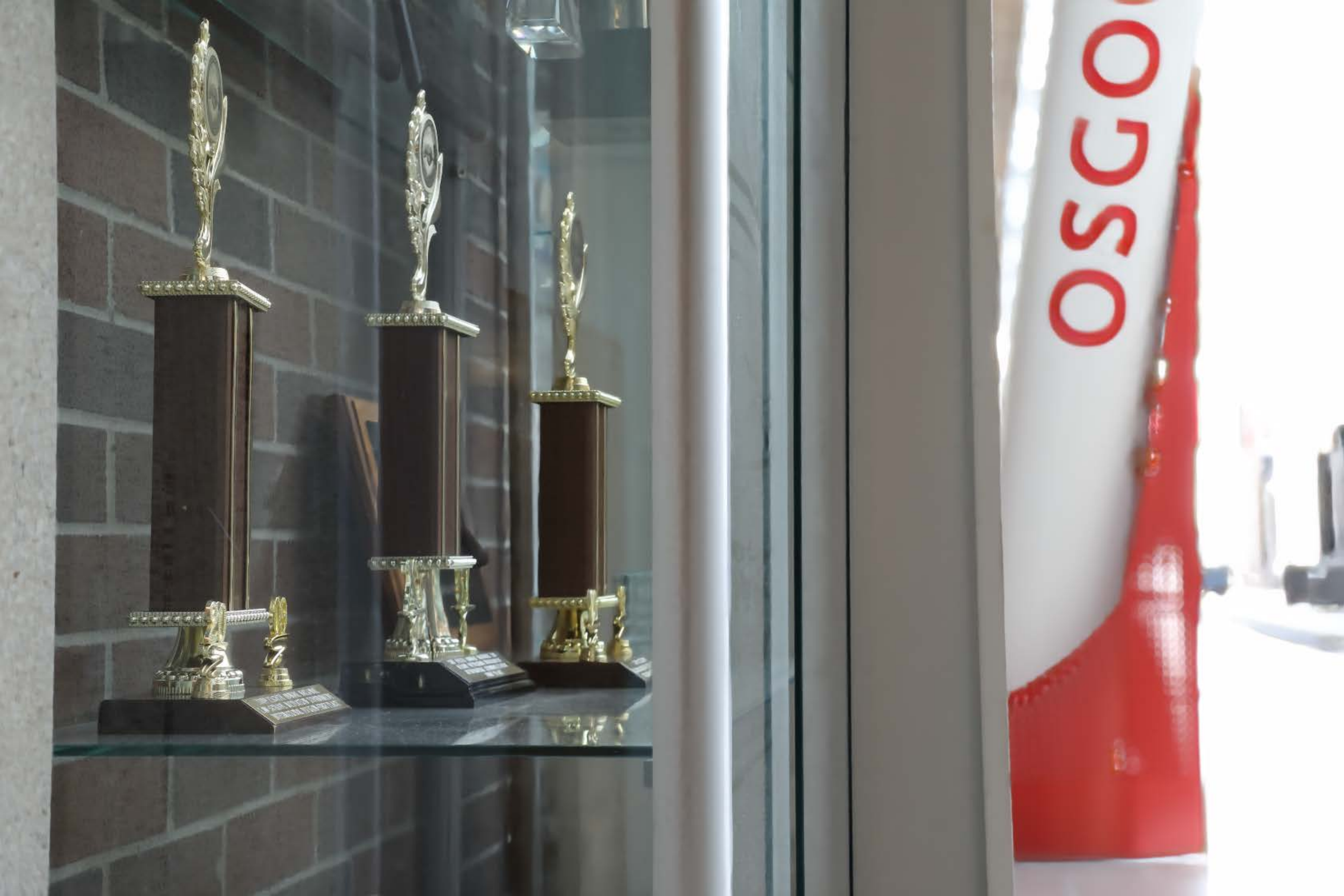} \\
\vspace{0.4em}
\small~Degraded  & \small~Label & \small~DMPHN~\cite{DMPHN}, 27.06 dB & \small~MPRNet~\cite{MPRNet}, 27.81 dB \\
\end{tabular}
}
\caption{\small Defocus deblurring on DPDD~\cite{DPDNet}.}
\label{fig:suppl_defocus_deblurring_visual_results_2}
\end{center}
\vspace{-7mm}
\end{figure*}

%% file: suppl_figures/suppl_007_dehazing_visual_results.tex
\begin{figure*}[t]
\begin{center}
\setlength{\tabcolsep}{1.5pt}
\scalebox{0.88}{
\begin{tabular}[b]{c c c c c c}
\hspace{-4mm}
\vspace{0.4em}
\includegraphics[width=.16\textwidth,valign=t]{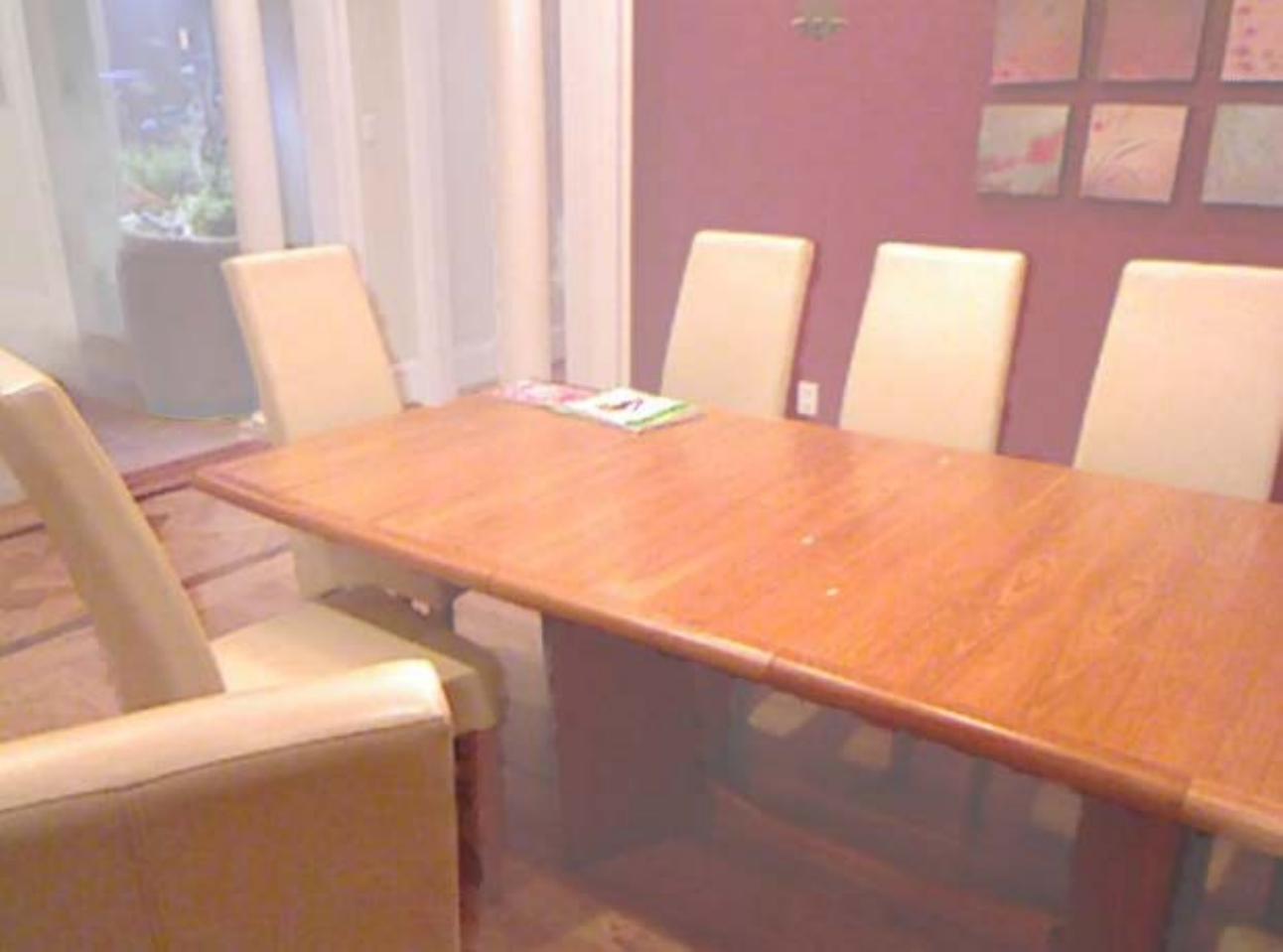}  &   
\includegraphics[width=.16\textwidth,valign=t]{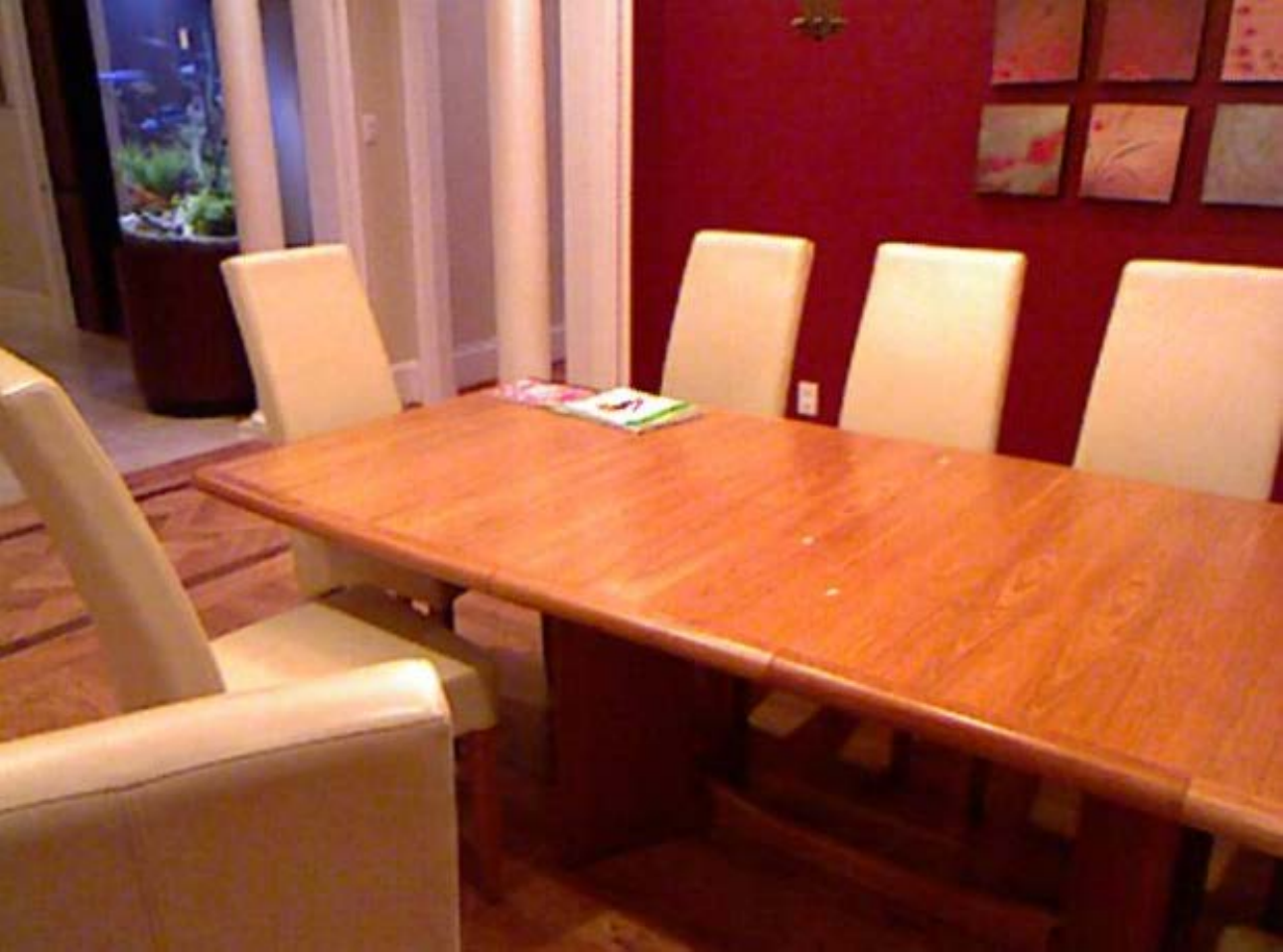} &   
\includegraphics[width=.16\textwidth,valign=t]{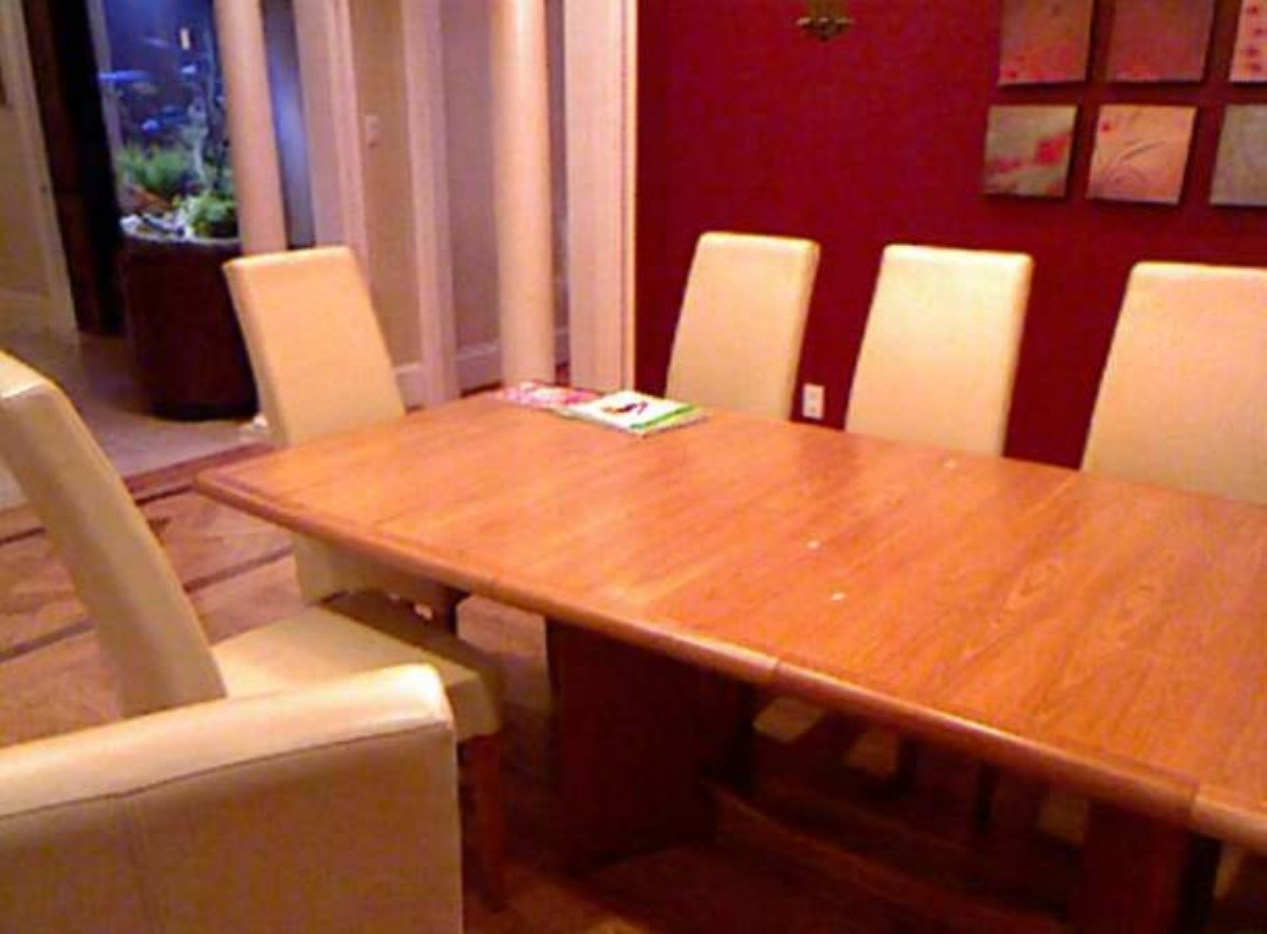}  &
\includegraphics[width=.16\textwidth,valign=t]{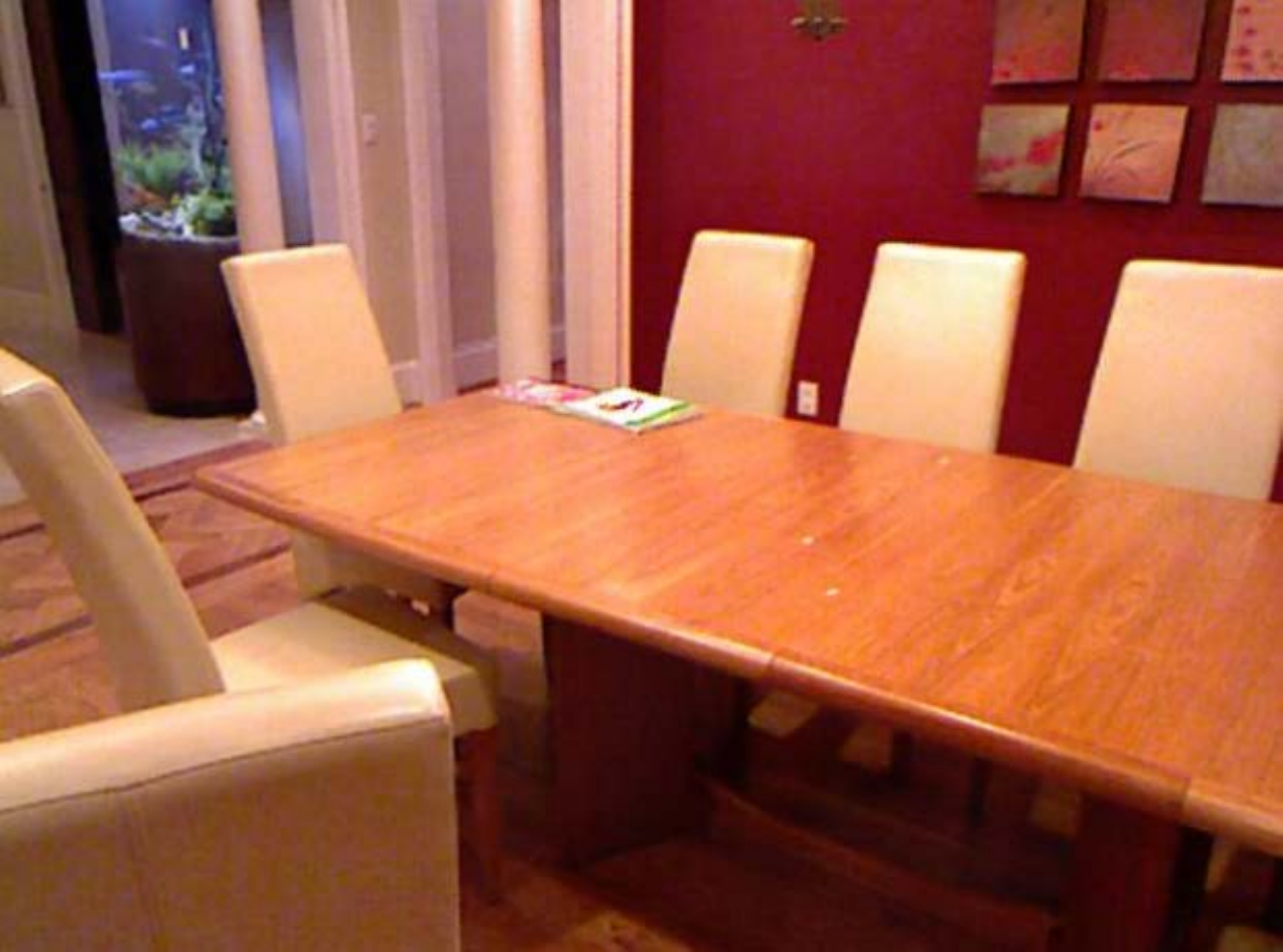}  &
\includegraphics[width=.16\textwidth,valign=t]{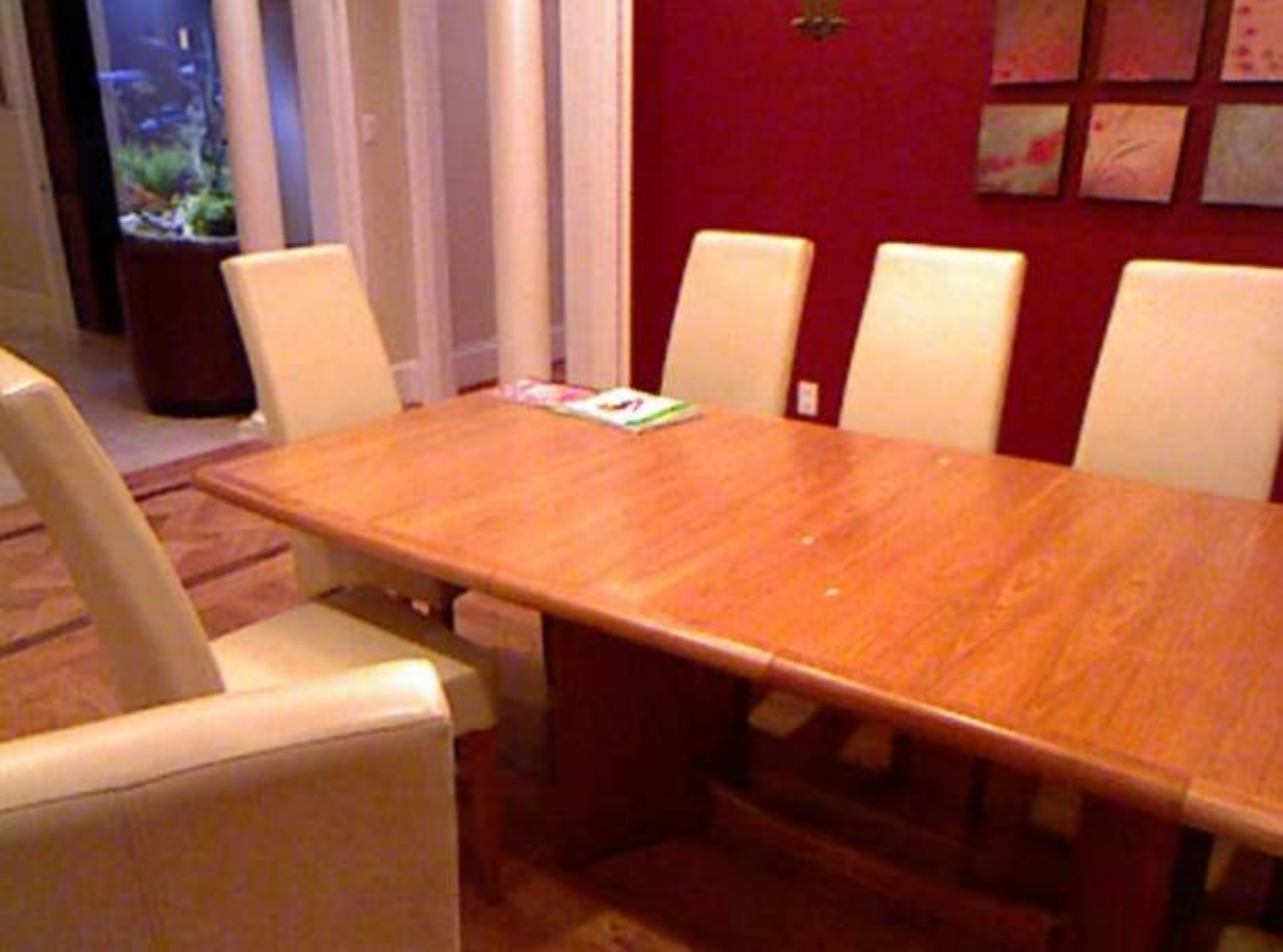}  &
\includegraphics[width=.16\textwidth,valign=t]{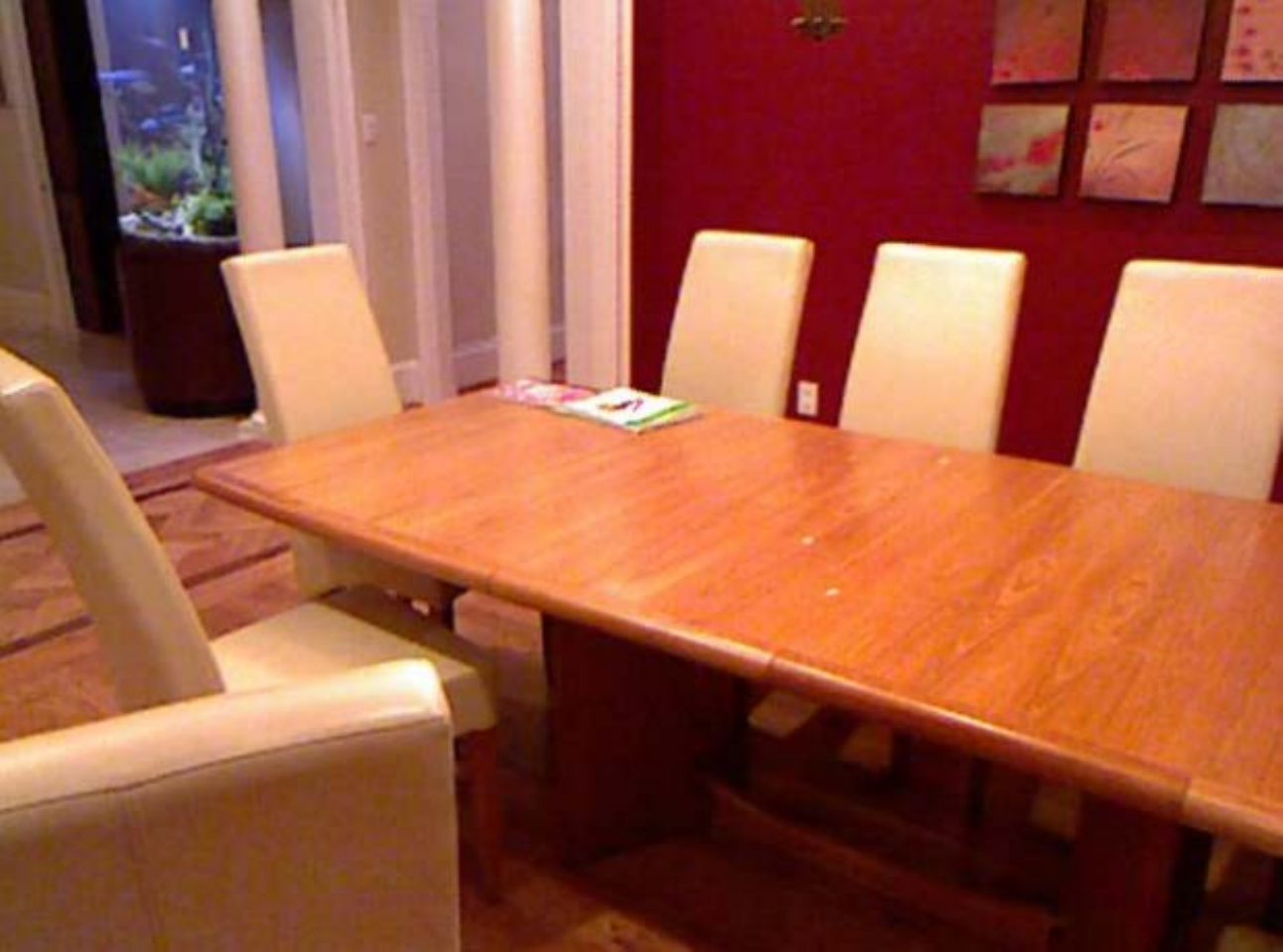} \\
\vspace{0.6em}
\small~Degraded  & \small~Label & \small~Dehamer~\cite{Dehamer} & \small~Dehazeformer~\cite{Dehazeformer} & \small~SFNet~\cite{SFNet} 40.19 & \small~\textbf{Ours} \textbf{41.44}\\

\vspace{0.4em}
\includegraphics[width=.16\textwidth,valign=t]{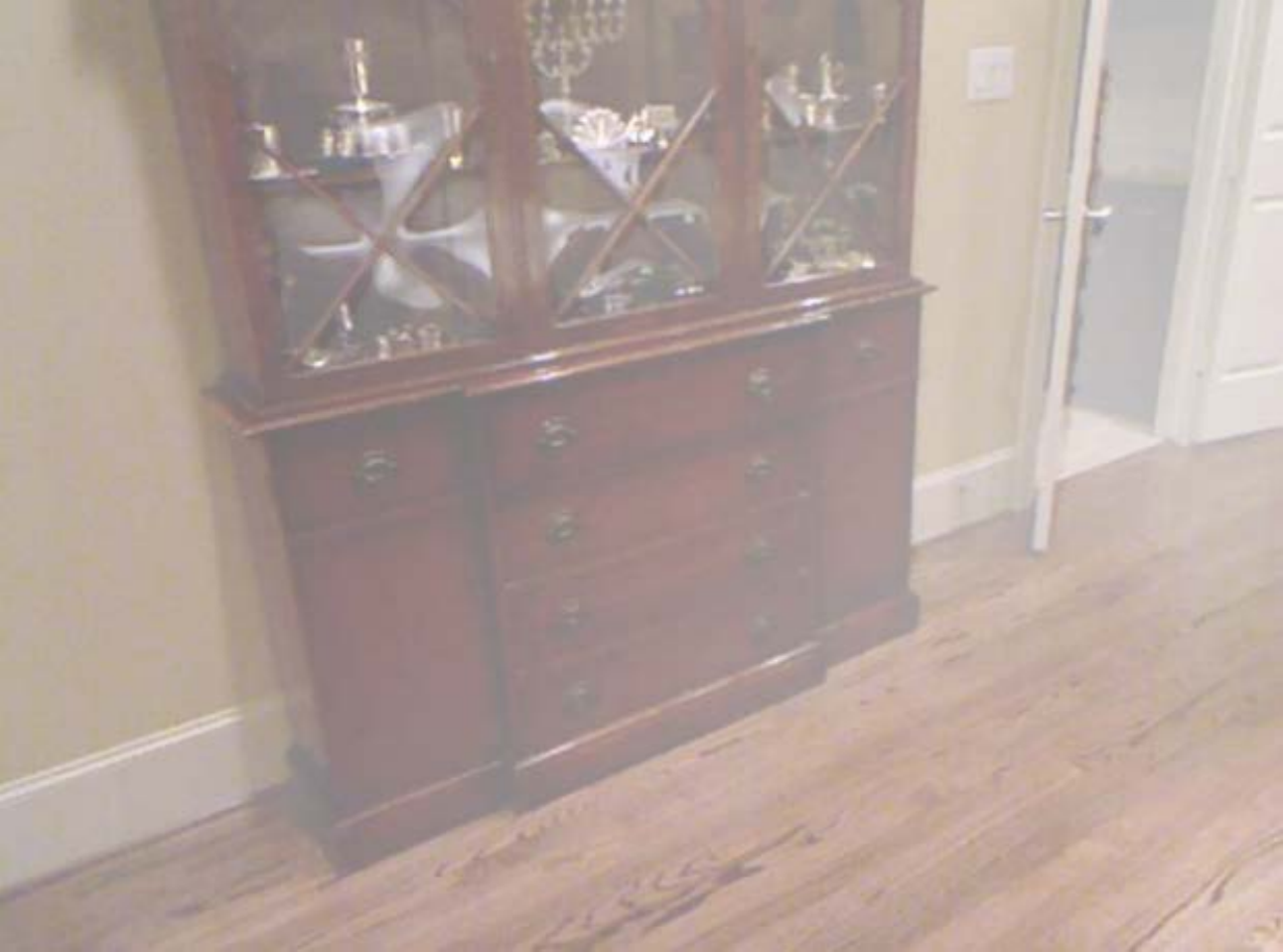}  &   
\includegraphics[width=.16\textwidth,valign=t]{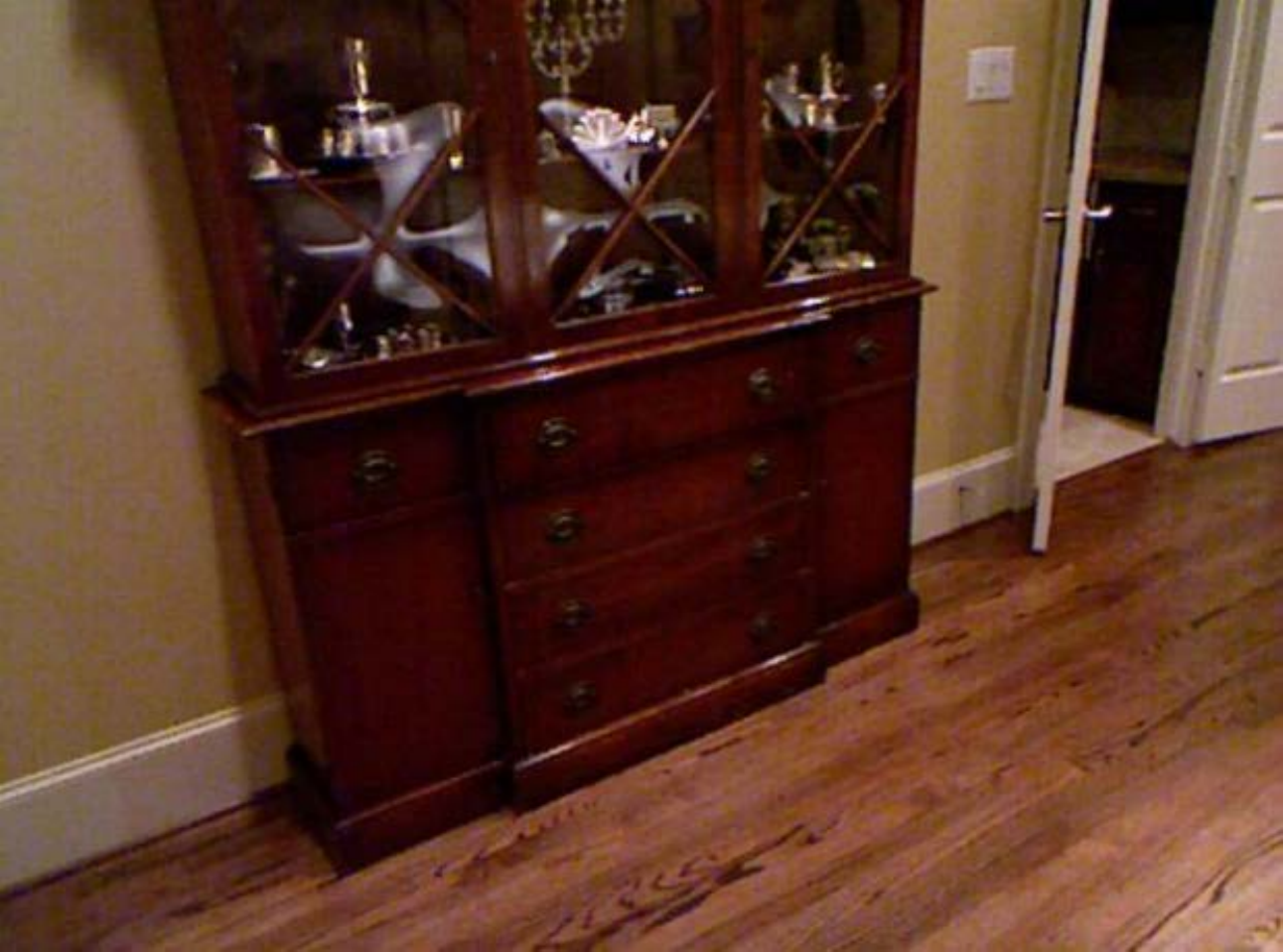} &   
\includegraphics[width=.16\textwidth,valign=t]{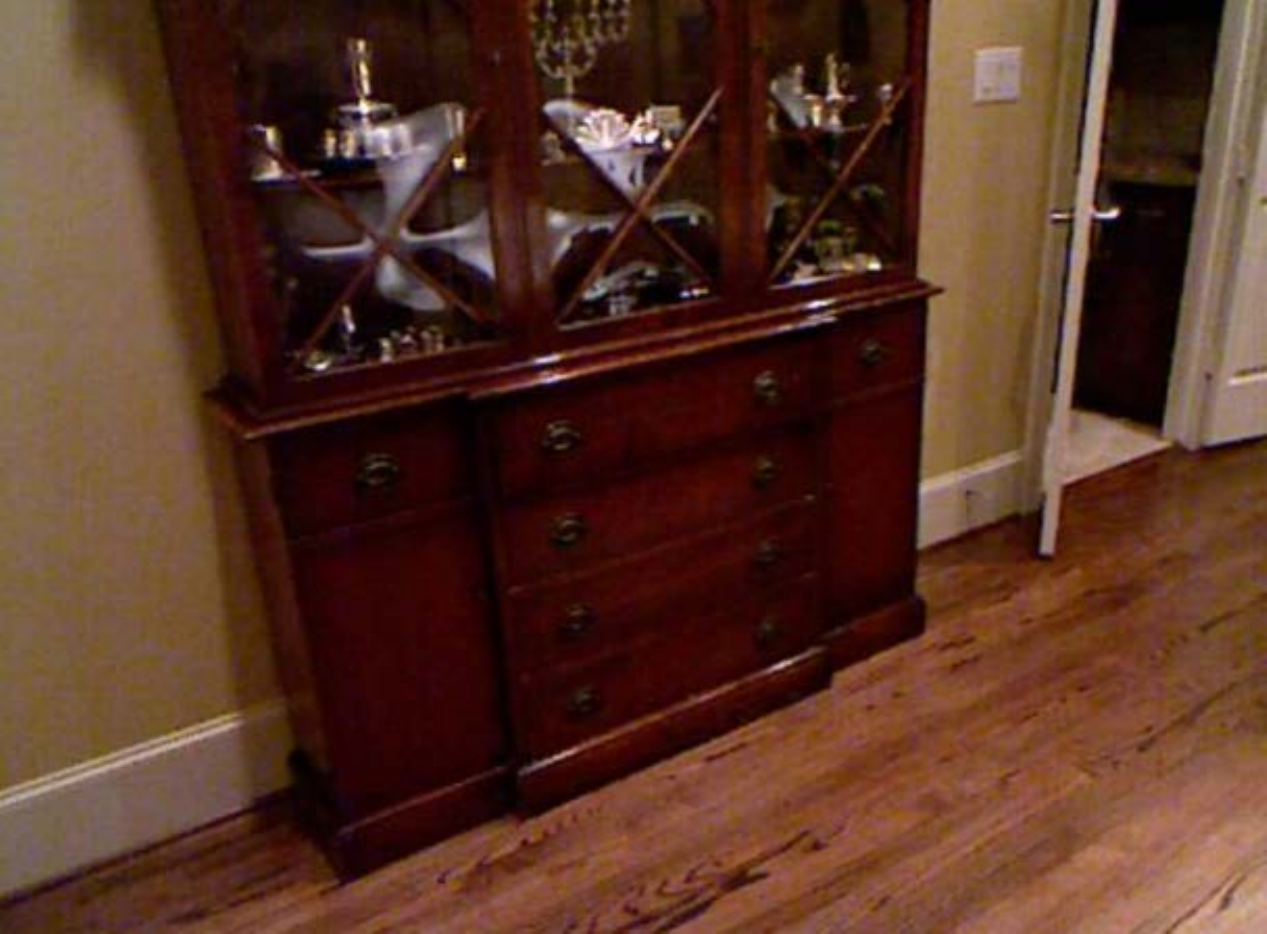}  &
\includegraphics[width=.16\textwidth,valign=t]{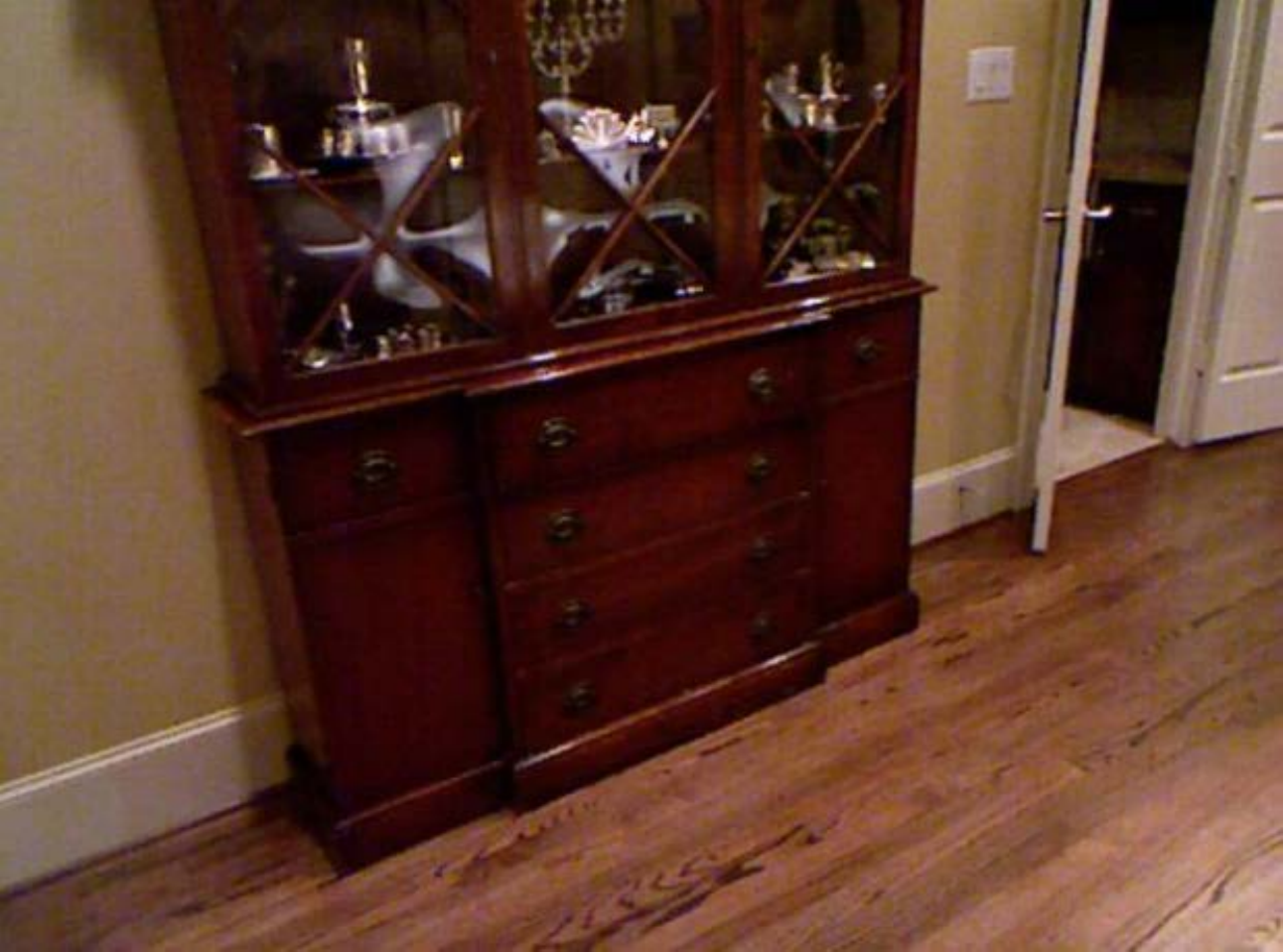}  &
\includegraphics[width=.16\textwidth,valign=t]{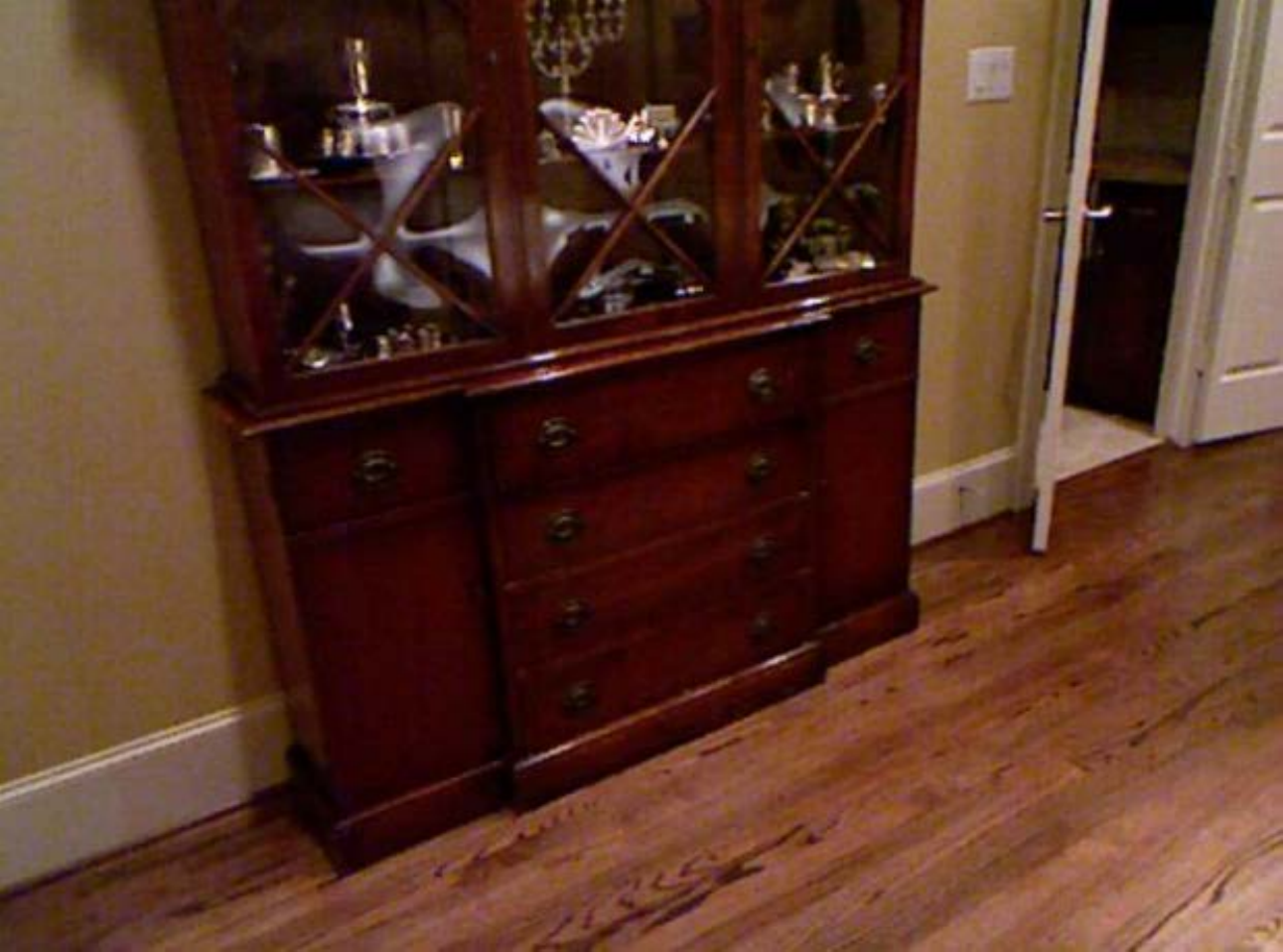}  &
\includegraphics[width=.16\textwidth,valign=t]{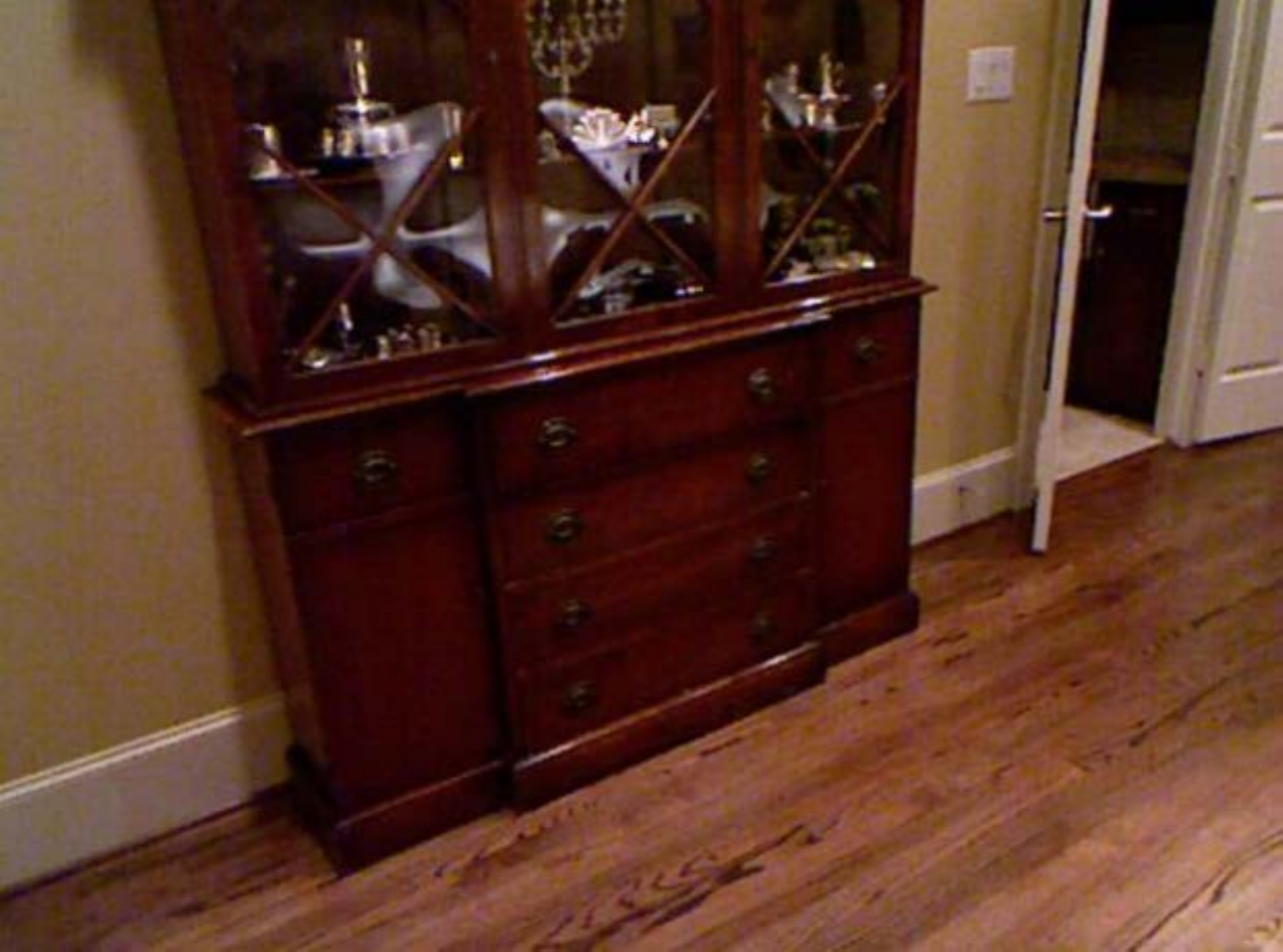} \\
\vspace{0.6em}
\small~Degraded  & \small~Label & \small~Dehamer~\cite{Dehamer} & \small~Dehazeformer~\cite{Dehazeformer} & \small~SFNet~\cite{SFNet} 41.65 & \small~\textbf{Ours} \textbf{42.78}\\
\end{tabular}
}
\caption{\small Image dehazing results on SOTS~\cite{reside}.}
\label{fig:suppl_dehazing_visual_results}
\end{center}
\vspace{-7mm}
\end{figure*}

%% file: suppl_figures/suppl_006_01_deraining_visual_results.tex
\begin{figure*}[t]
\begin{center}
\setlength{\tabcolsep}{1.5pt}
\scalebox{0.88}{
\begin{tabular}[b]{c c c c}
\vspace{0.2em}
\includegraphics[width=.25\textwidth,valign=t]{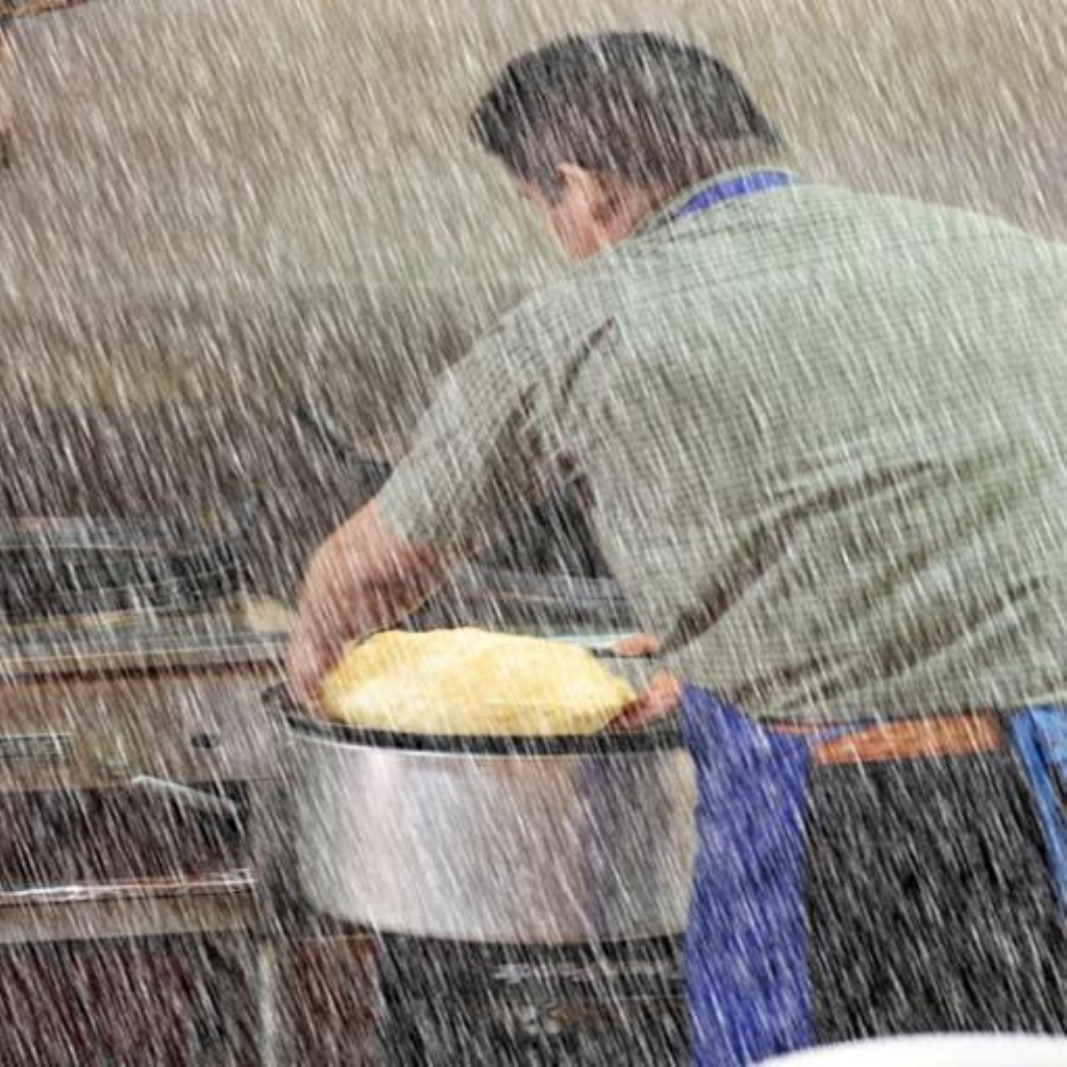}  &   
\includegraphics[width=.25\textwidth,valign=t]{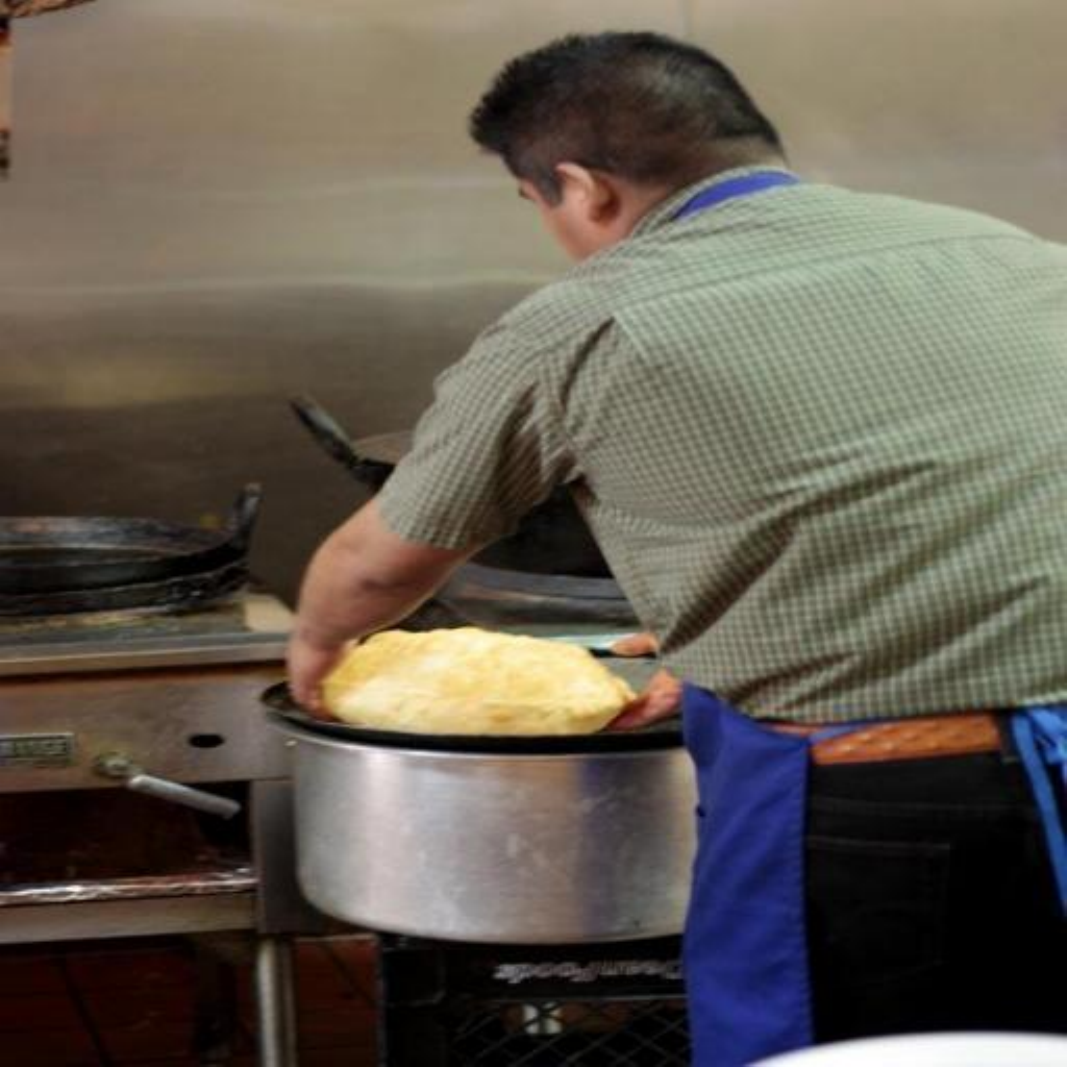} &   
\includegraphics[width=.25\textwidth,valign=t]{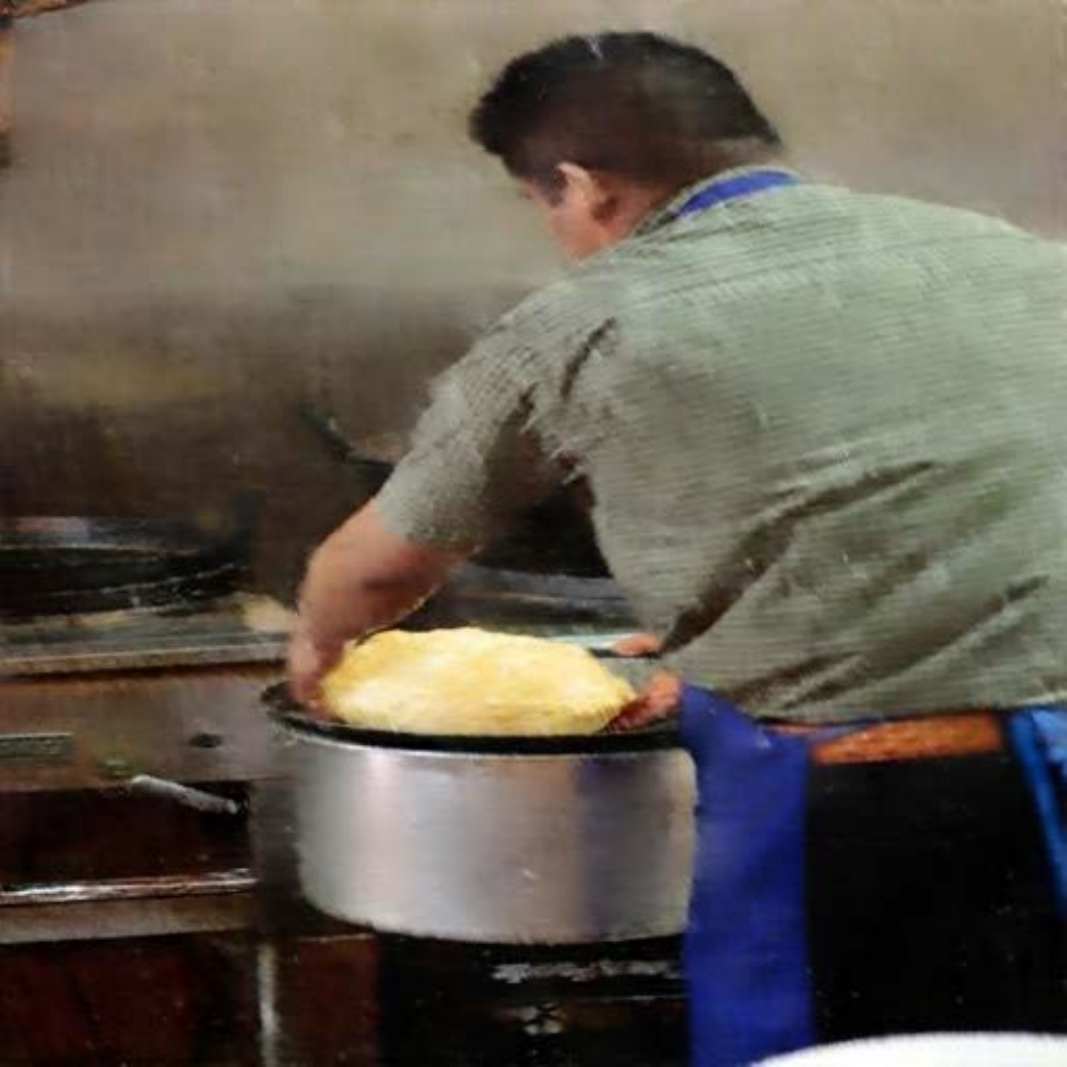}  &
\includegraphics[width=.25\textwidth,valign=t]{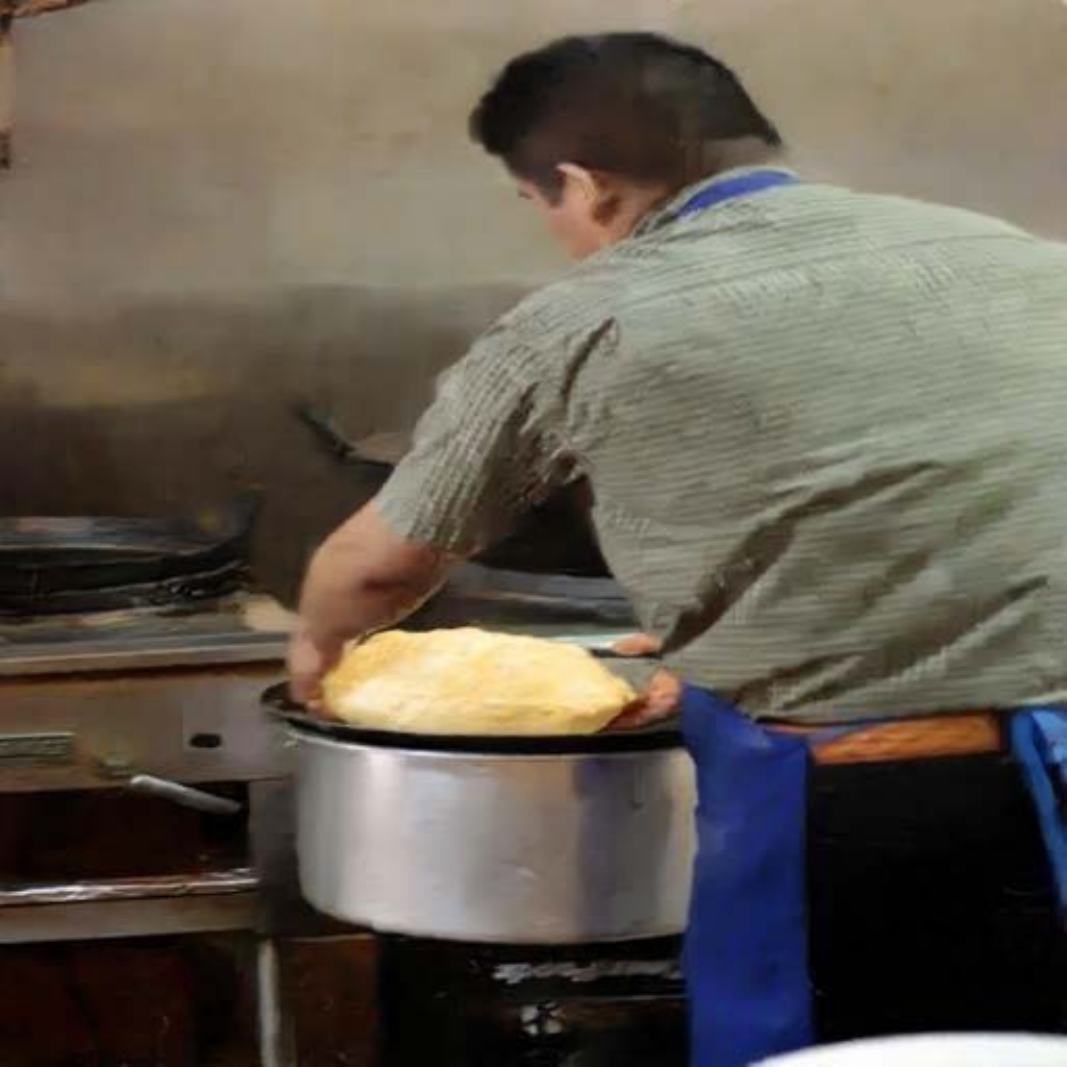} \\
\vspace{0.2em}
\small~Degraded  & \small~Label & \small~RESCAN~\cite{RESCAN}, 30.36 dB & \small~PReNet~\cite{PReNet}, 31.50 dB \\
\vspace{0.4em}
\includegraphics[width=.25\textwidth,valign=t]{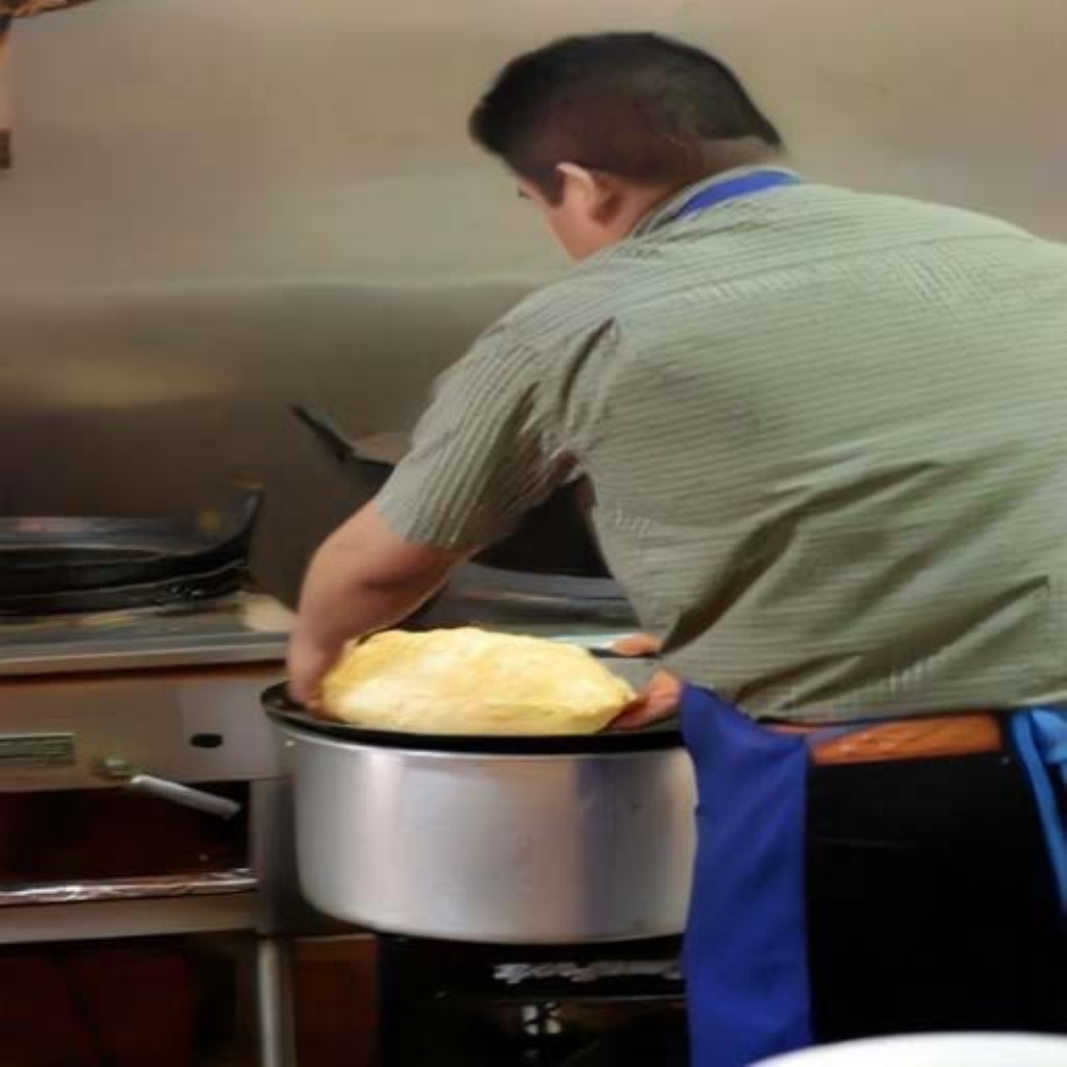}  &   
\includegraphics[width=.25\textwidth,valign=t]{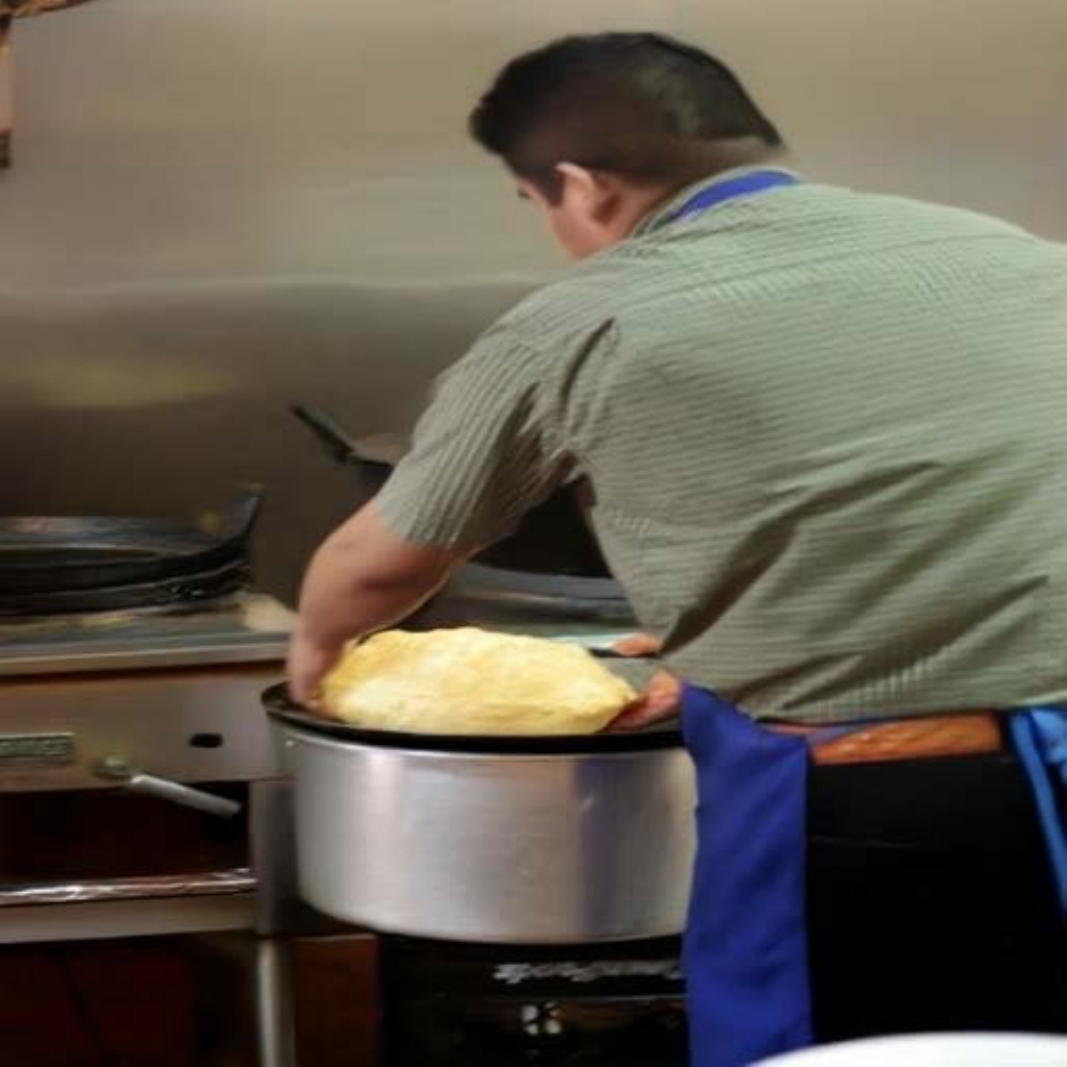} &   
\includegraphics[width=.25\textwidth,valign=t]{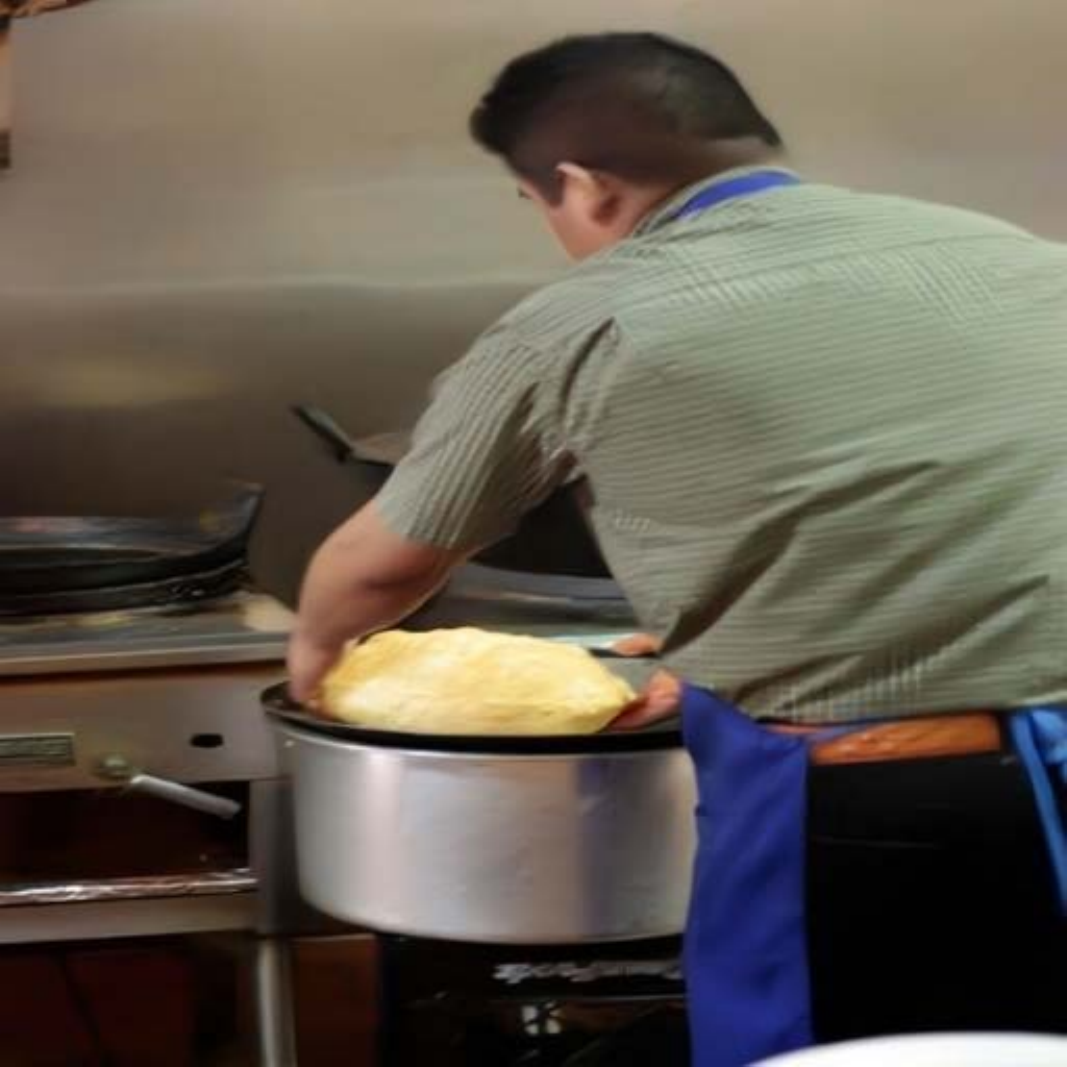}  &
\includegraphics[width=.25\textwidth,valign=t]{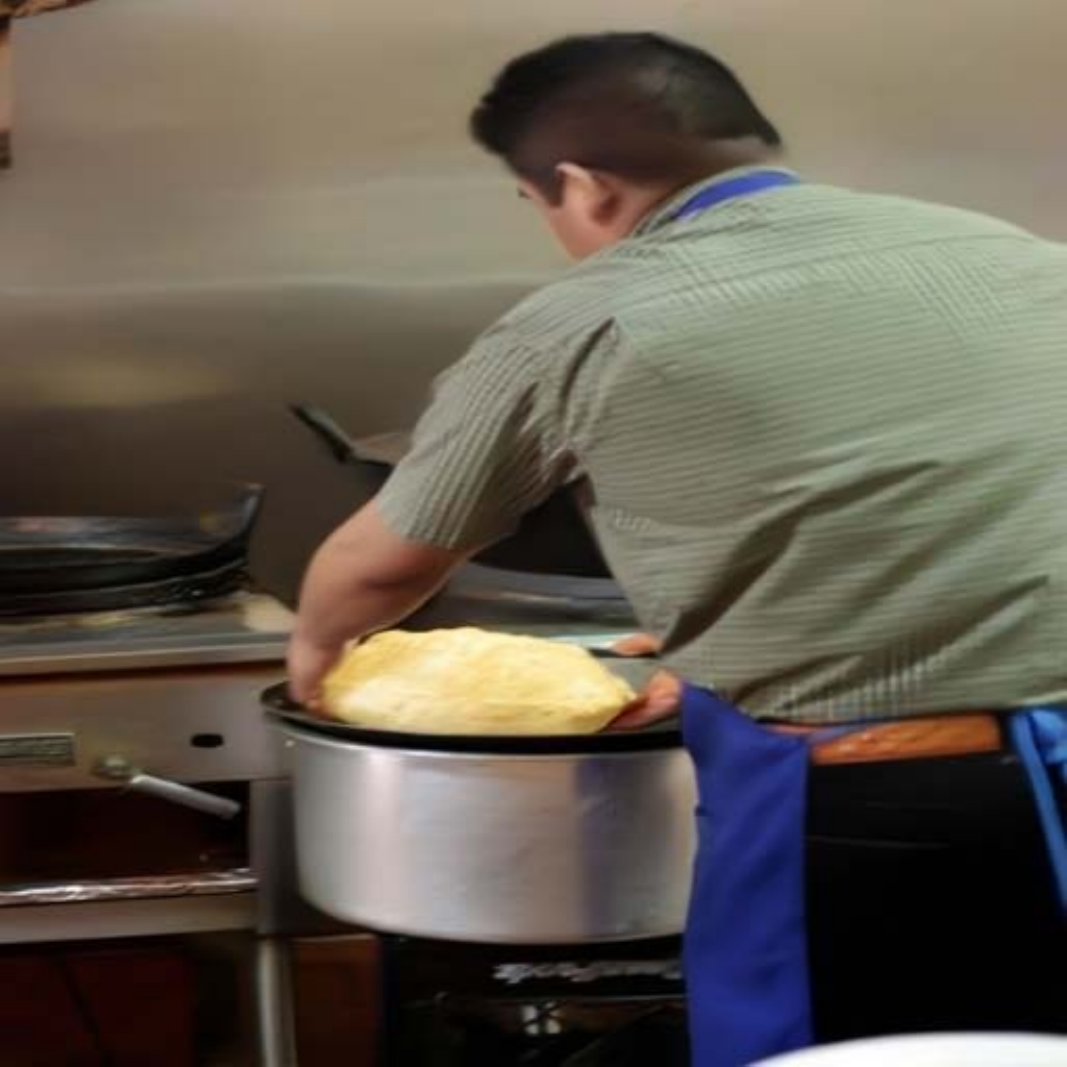} \\
\vspace{0.2em}
\small~MPRNet~\cite{MPRNet}, 33.02 dB  & \small~Uformer~\cite{Uformer}, 32.99 dB & \small~DRSformer~\cite{DRSformer}, 33.30 dB & \small~\textbf{Ours}, 33.50 dB \\
\vspace{0.2em}
\includegraphics[width=.25\textwidth,valign=t]{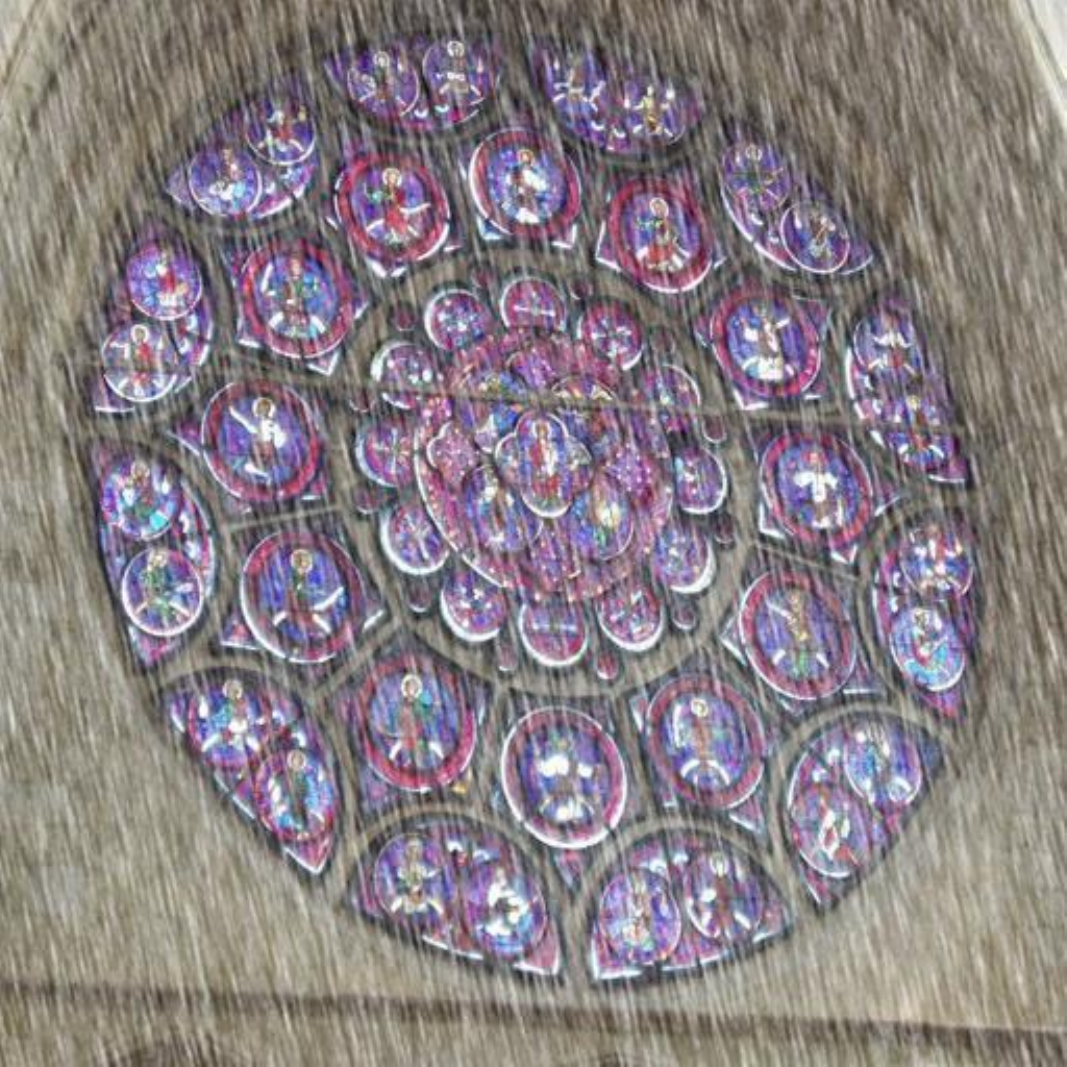}  &   
\includegraphics[width=.25\textwidth,valign=t]{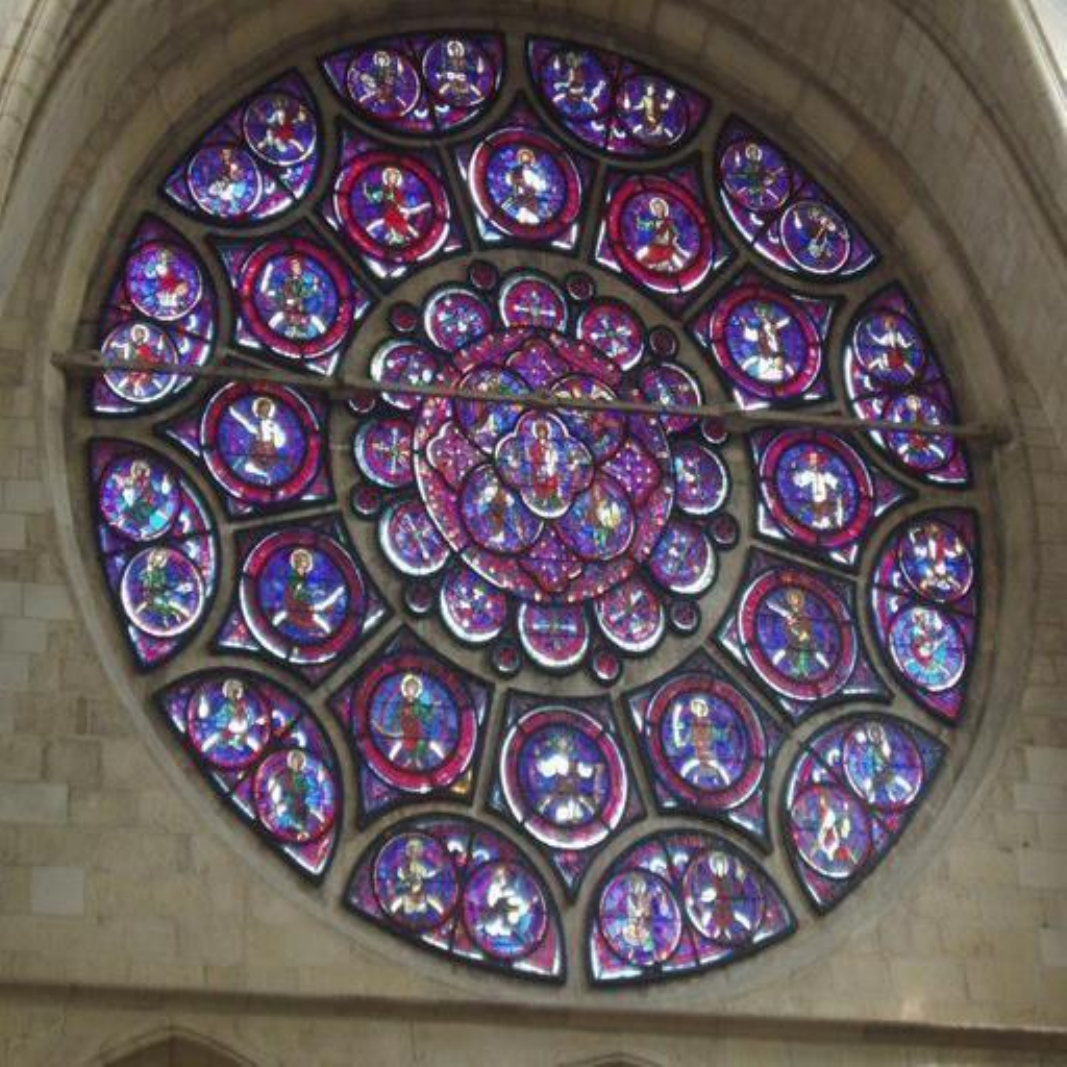} &   
\includegraphics[width=.25\textwidth,valign=t]{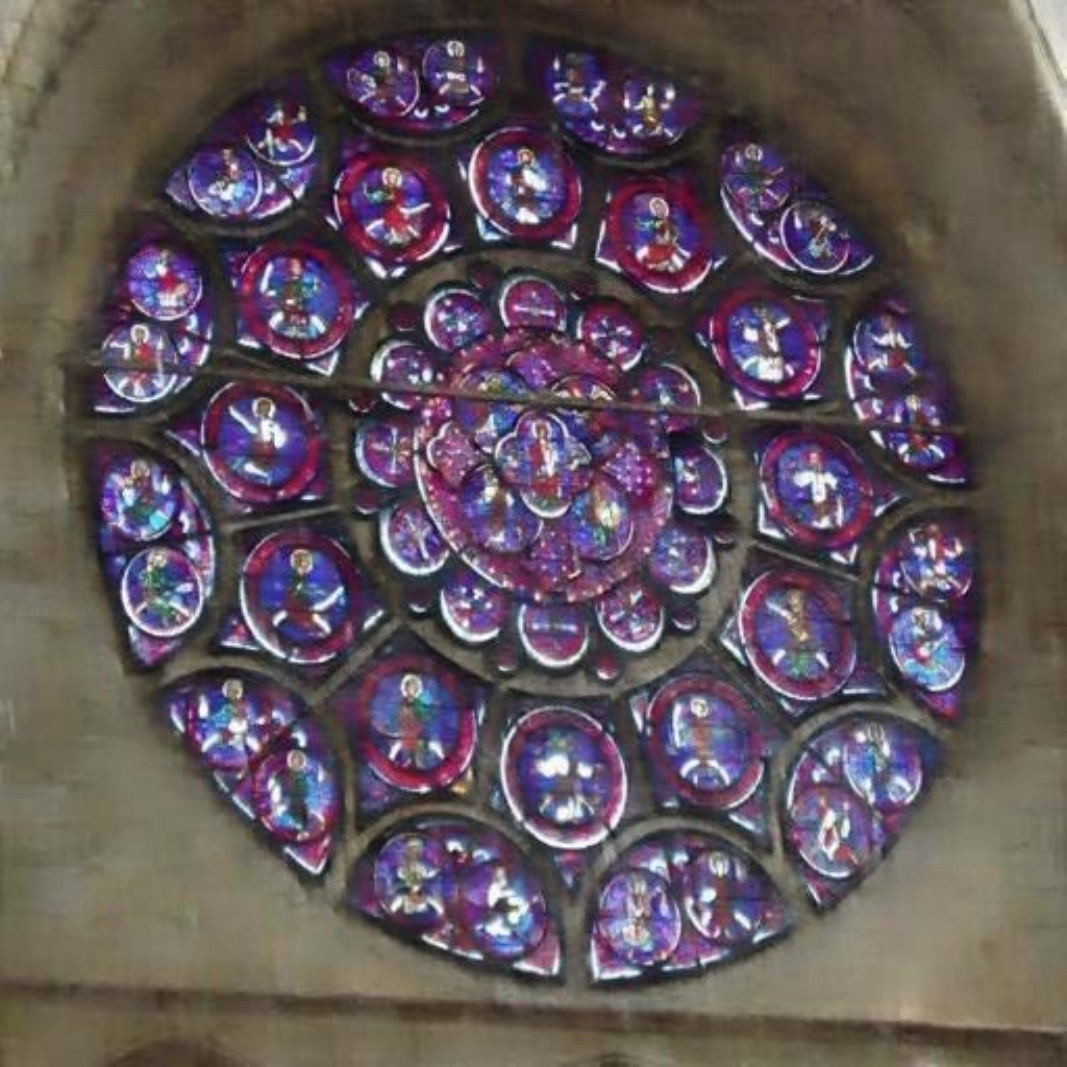}  &
\includegraphics[width=.25\textwidth,valign=t]{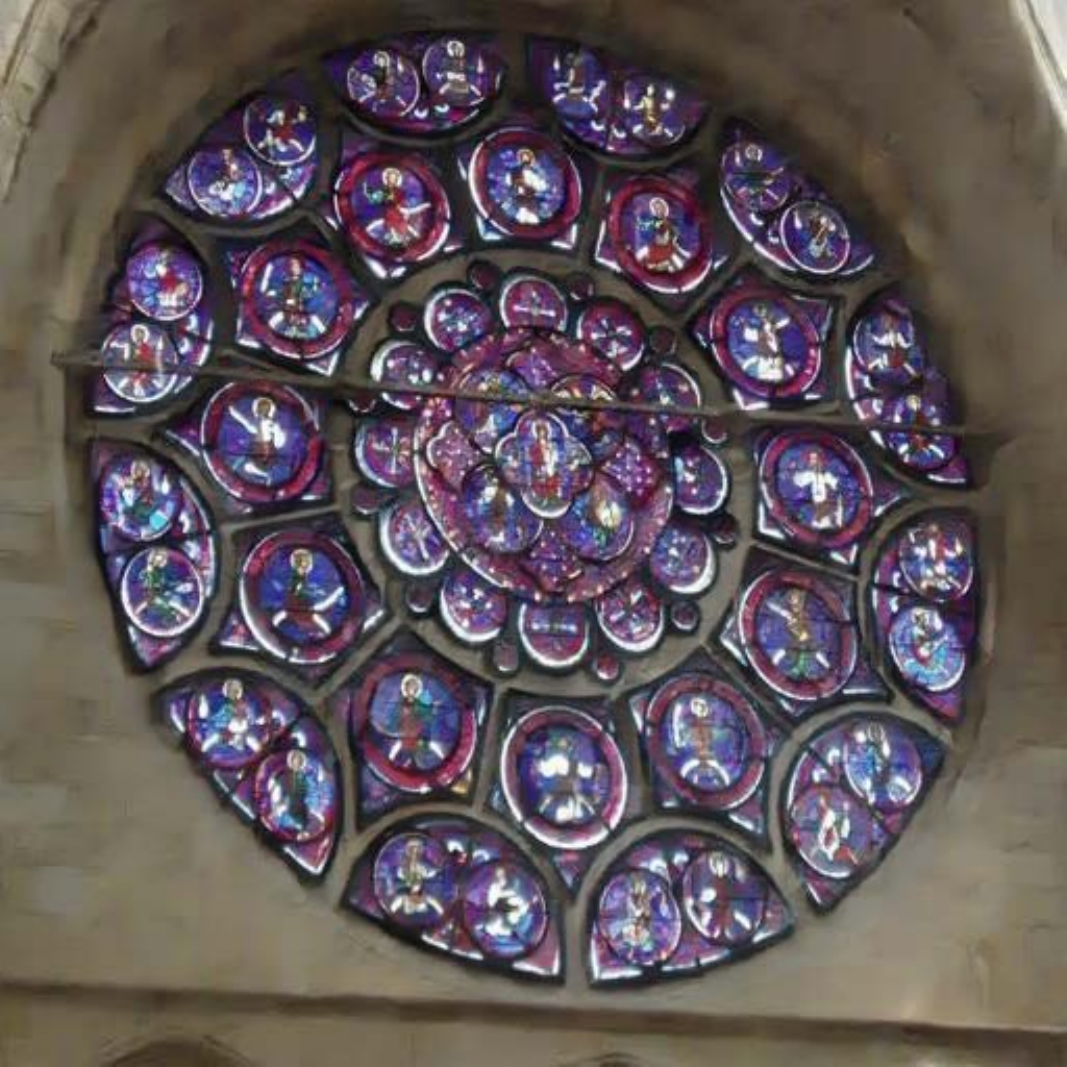} \\
\vspace{0.2em}
\small~Degraded  & \small~Label & \small~RESCAN~\cite{RESCAN}, 28.65 dB & \small~PReNet~\cite{PReNet}, 27.64 dB \\
\end{tabular}
}
\caption{\small Image deraining results on DID-Data~\cite{DID-Data}}
\label{fig:suppl_deraining_visual_results_1}
\end{center}
\vspace{-7mm}
\end{figure*}

%% file: suppl_figures/suppl_006_02_deraining_visual_results.tex
\begin{figure*}[t]
\begin{center}
\setlength{\tabcolsep}{1.5pt}
\scalebox{0.88}{
\begin{tabular}[b]{c c c c}
\vspace{0.2em}

\vspace{0.2em}
\includegraphics[width=.25\textwidth,valign=t]{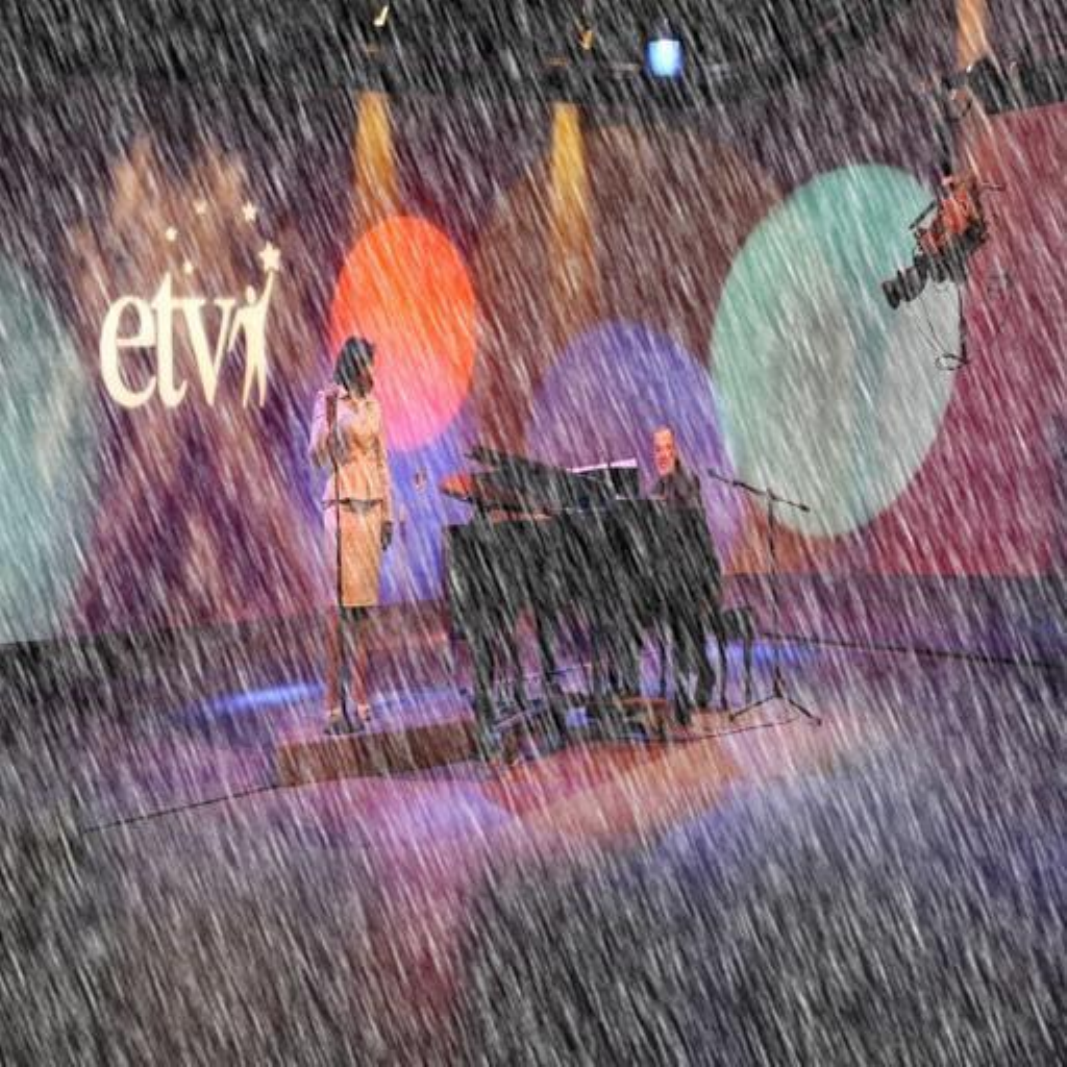}  &   
\includegraphics[width=.25\textwidth,valign=t]{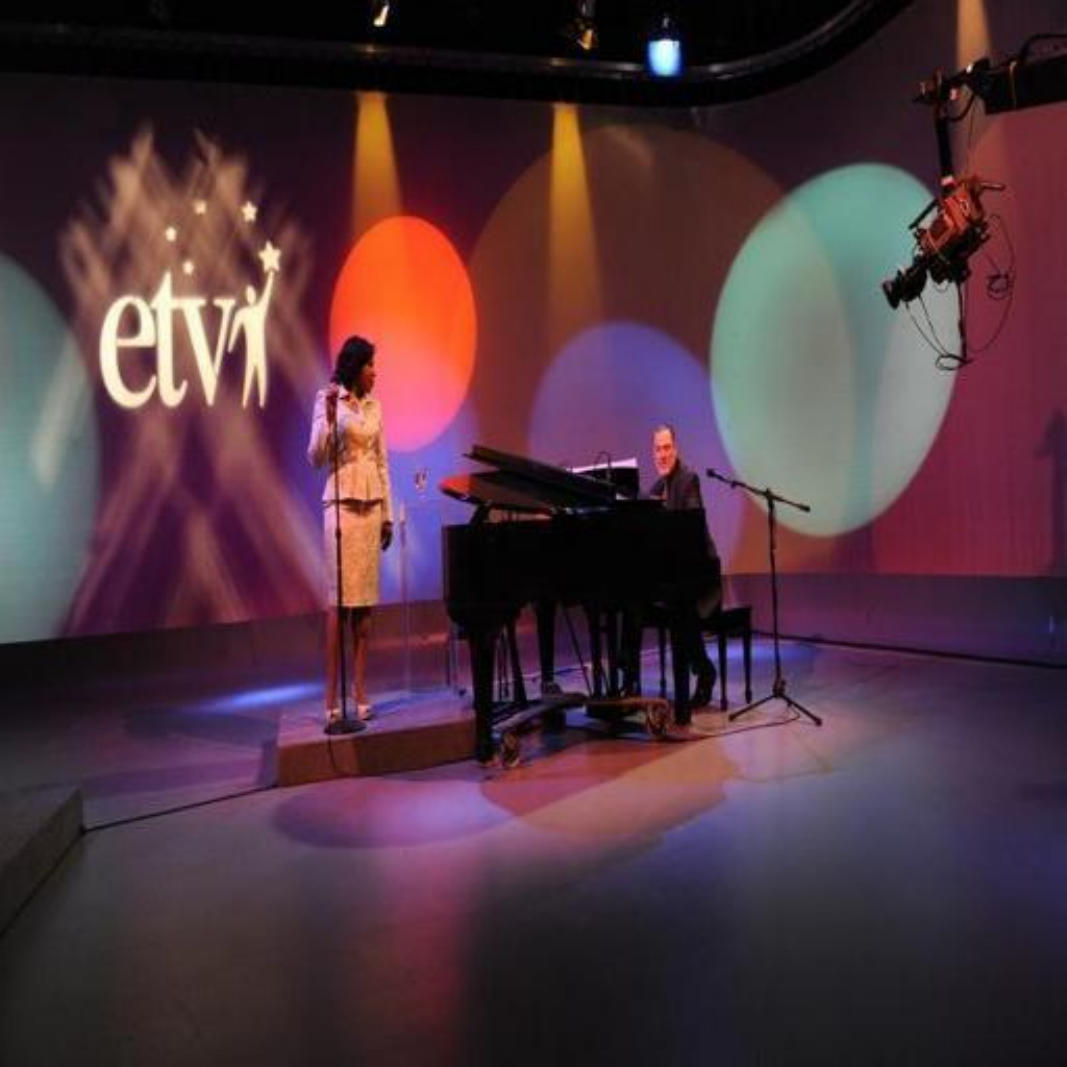} &   
\includegraphics[width=.25\textwidth,valign=t]{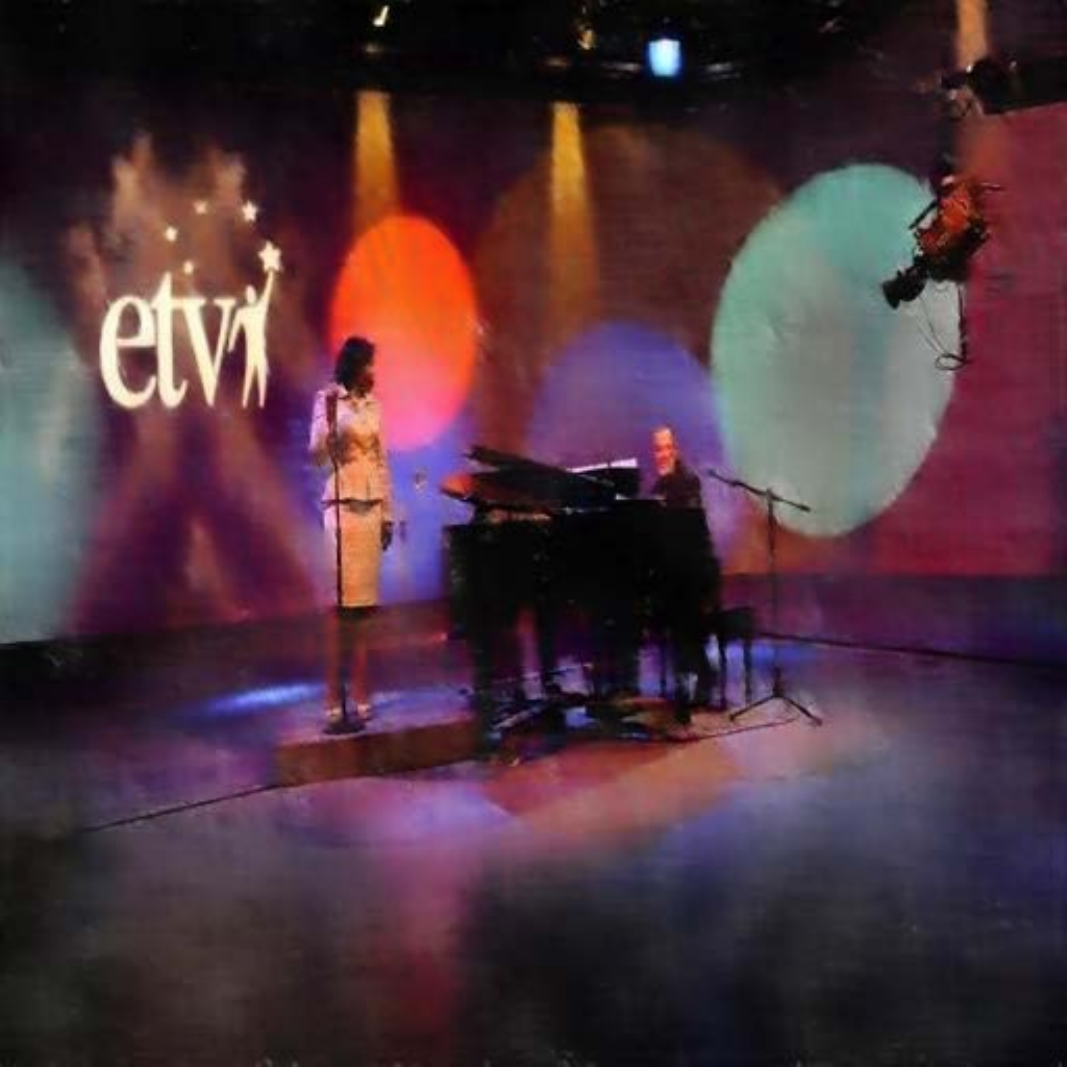}  &
\includegraphics[width=.25\textwidth,valign=t]{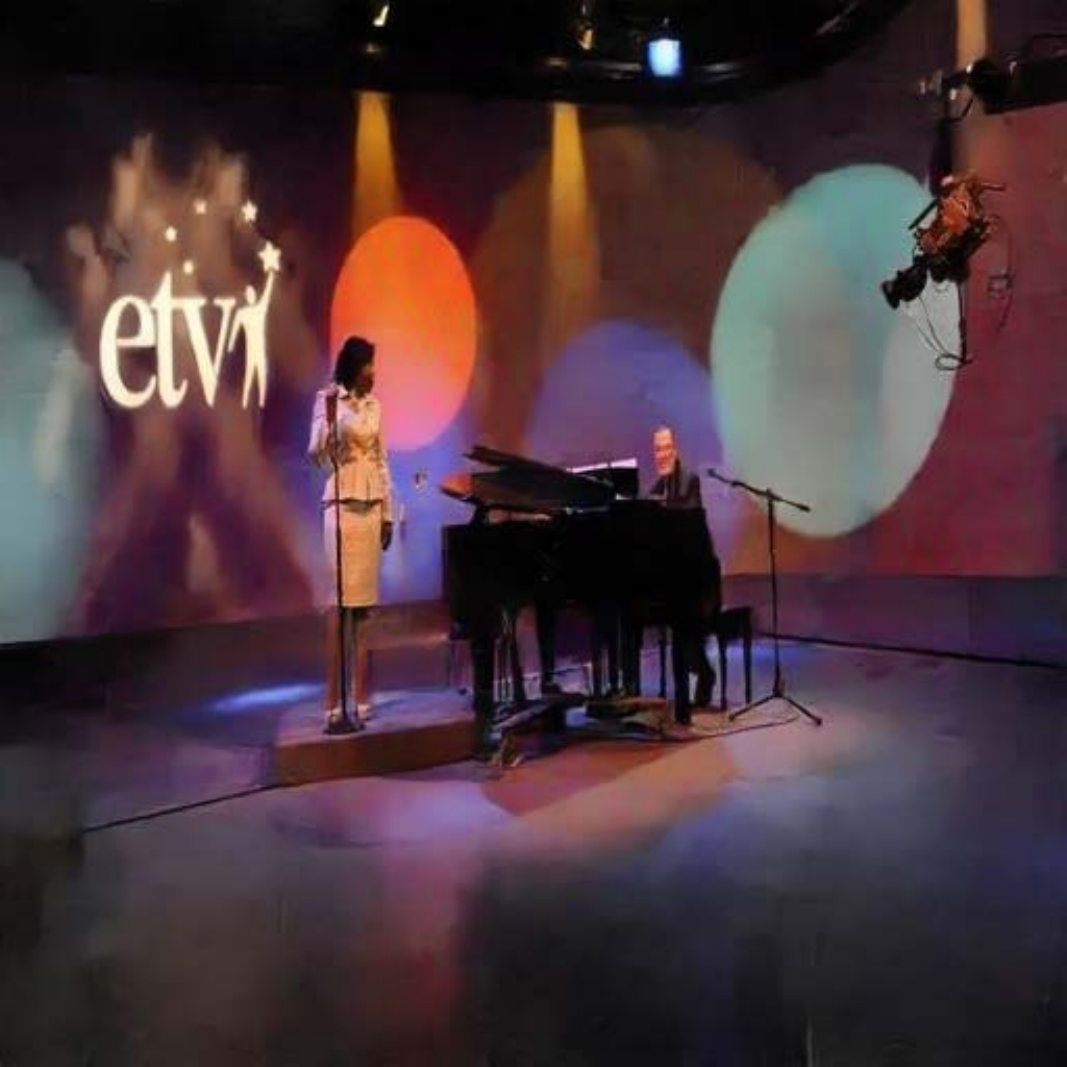} \\
\vspace{0.2em}
\small~Degraded  & \small~Label & \small~RESCAN~\cite{RESCAN}, 30.72 dB & \small~PReNet~\cite{PReNet}, 32.68 dB \\
\vspace{0.2em}
\includegraphics[width=.25\textwidth,valign=t]{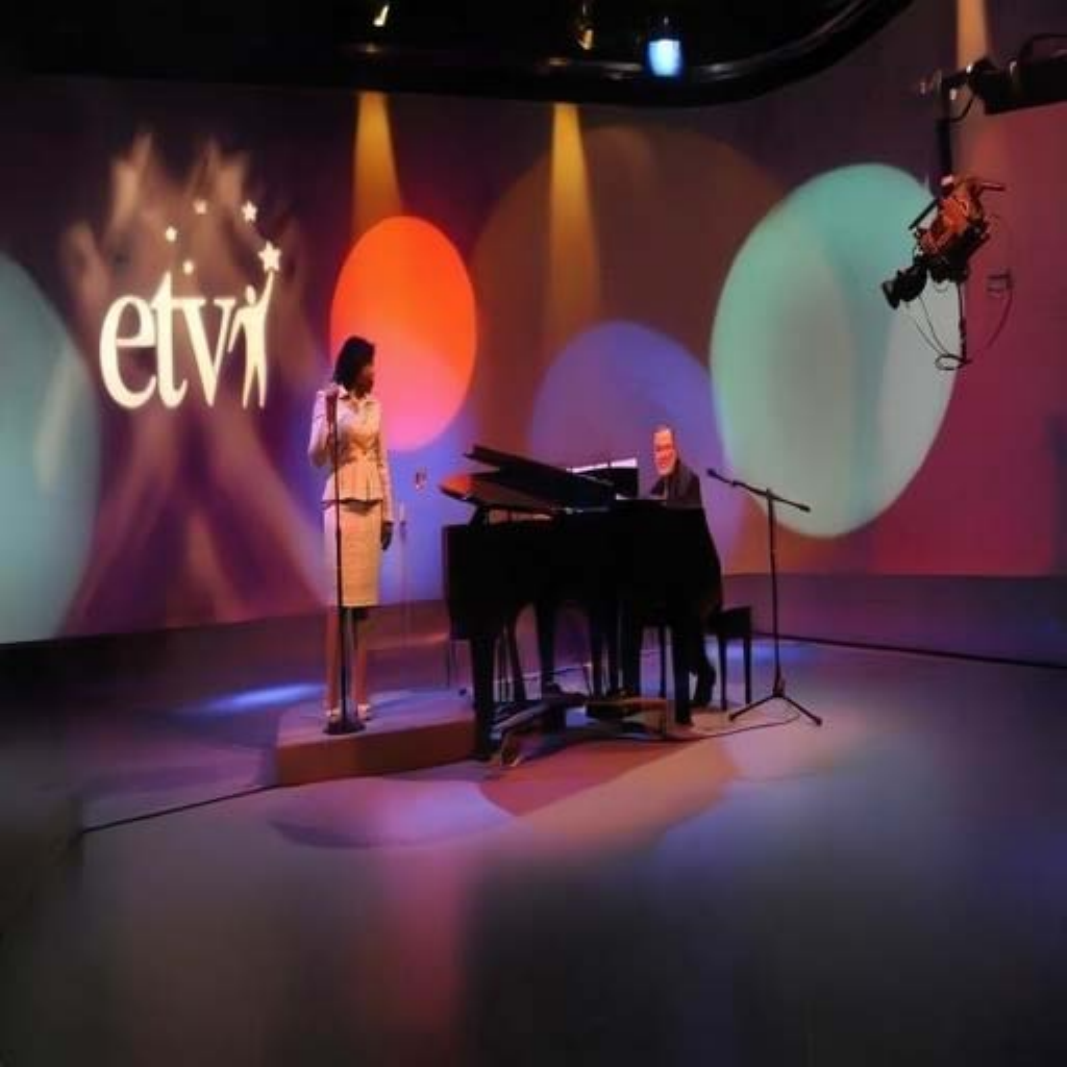}  &   
\includegraphics[width=.25\textwidth,valign=t]{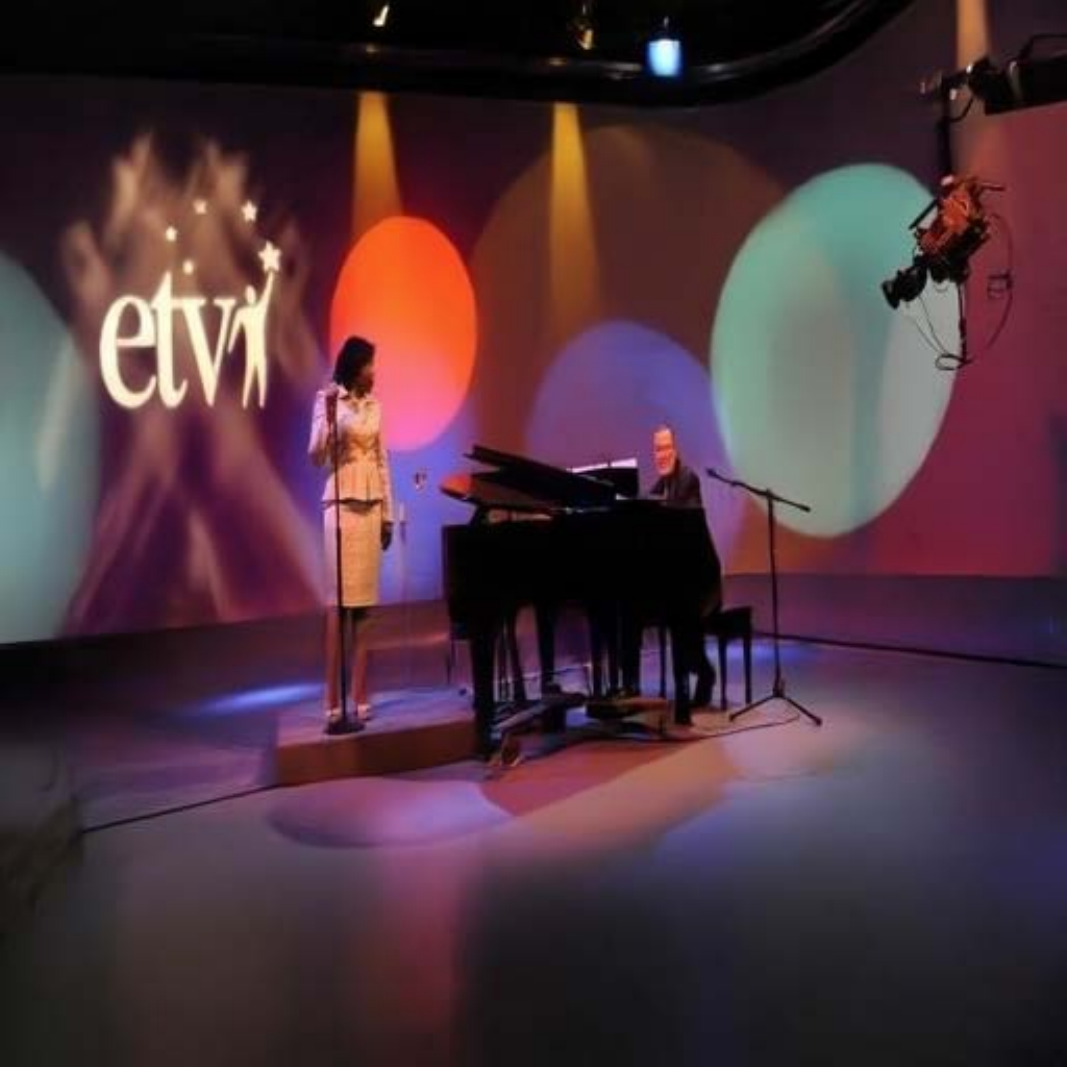} &   
\includegraphics[width=.25\textwidth,valign=t]{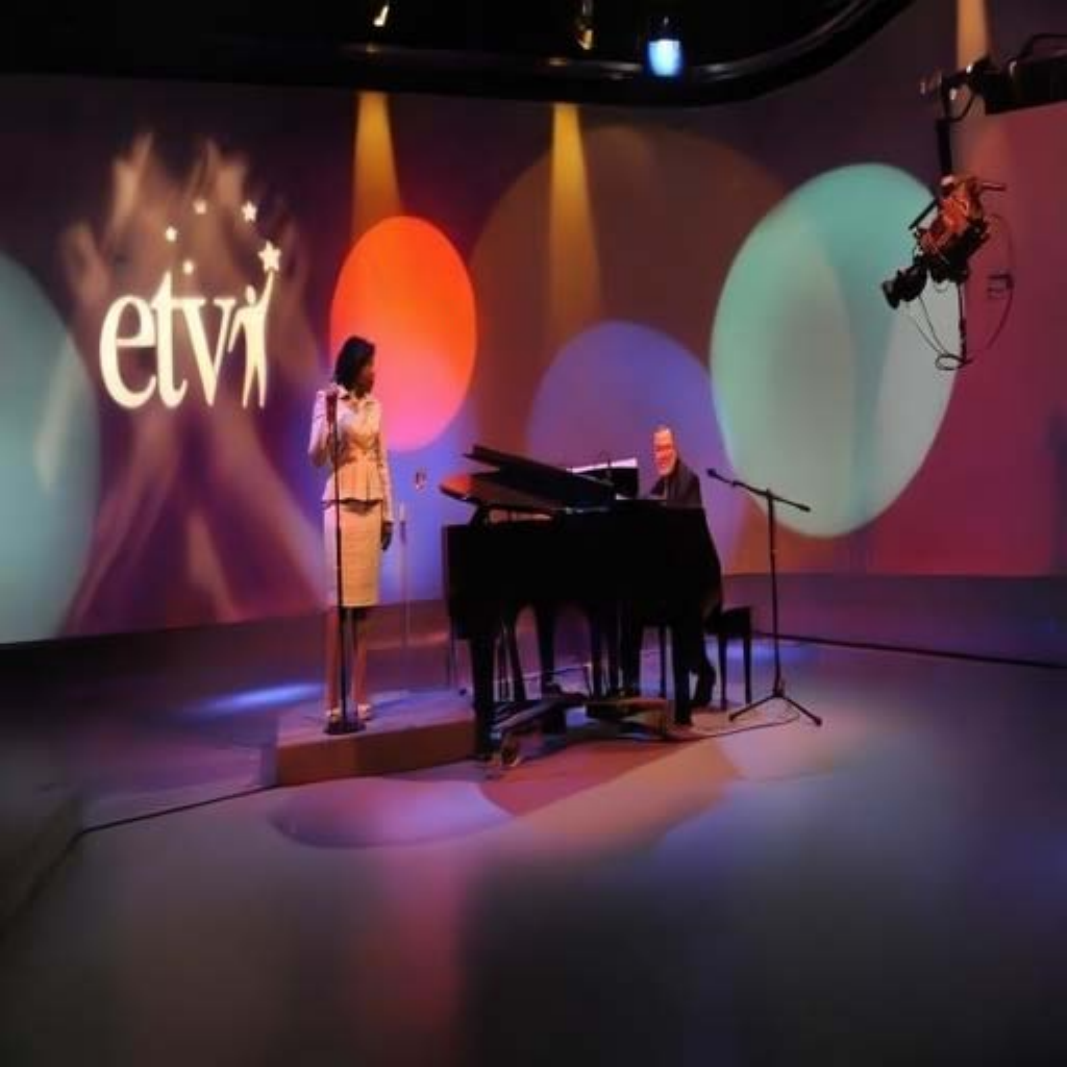}  &
\includegraphics[width=.25\textwidth,valign=t]{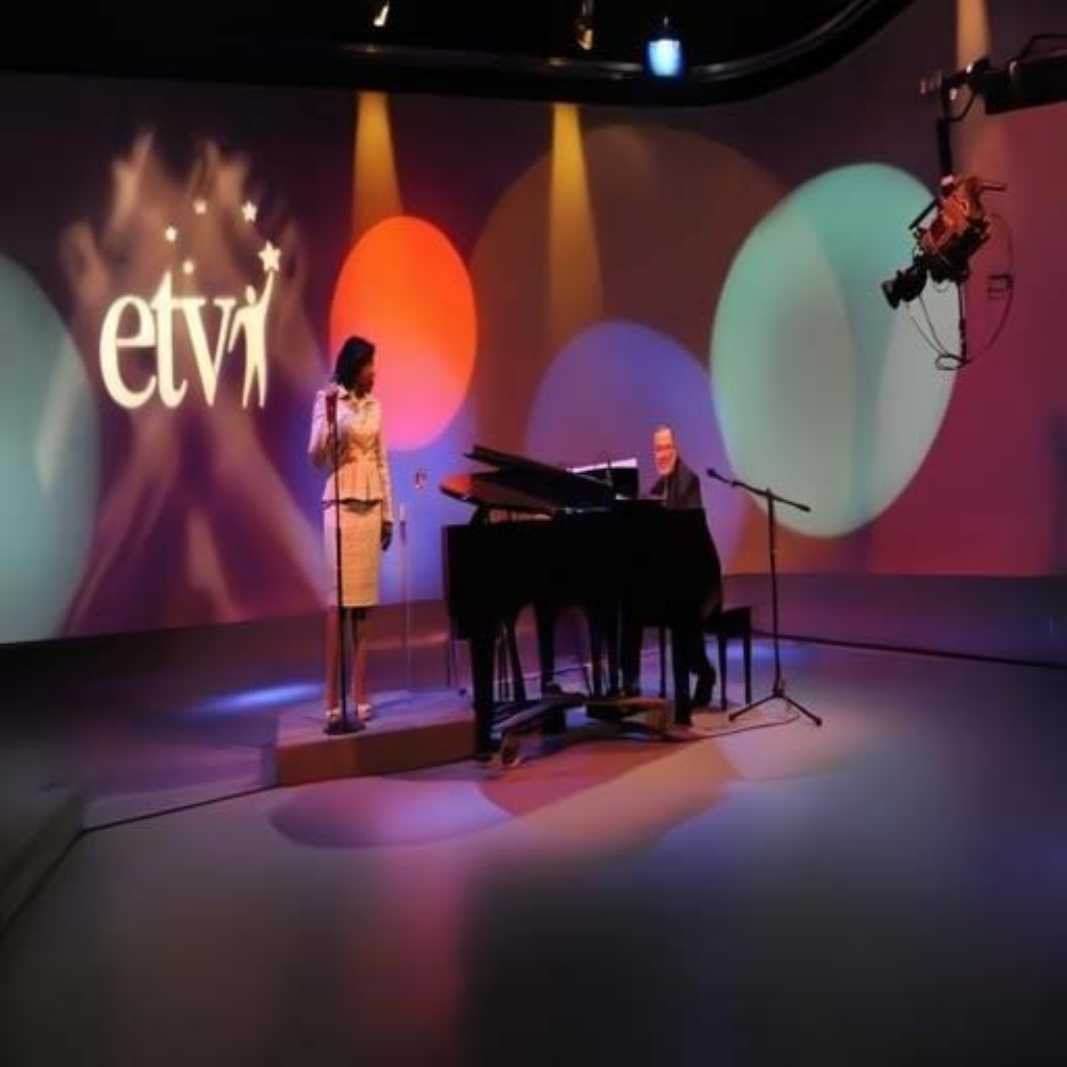} \\
\vspace{0.2em}
\small~MPRNet~\cite{MPRNet}, 35.11 dB  & \small~Uformer~\cite{Uformer}, 35.14 dB & \small~DRSformer~\cite{DRSformer}, 35.72 dB & \small~\textbf{Ours}, \textbf{36.20 dB} \\
\vspace{0.2em}
\includegraphics[width=.25\textwidth,valign=t]{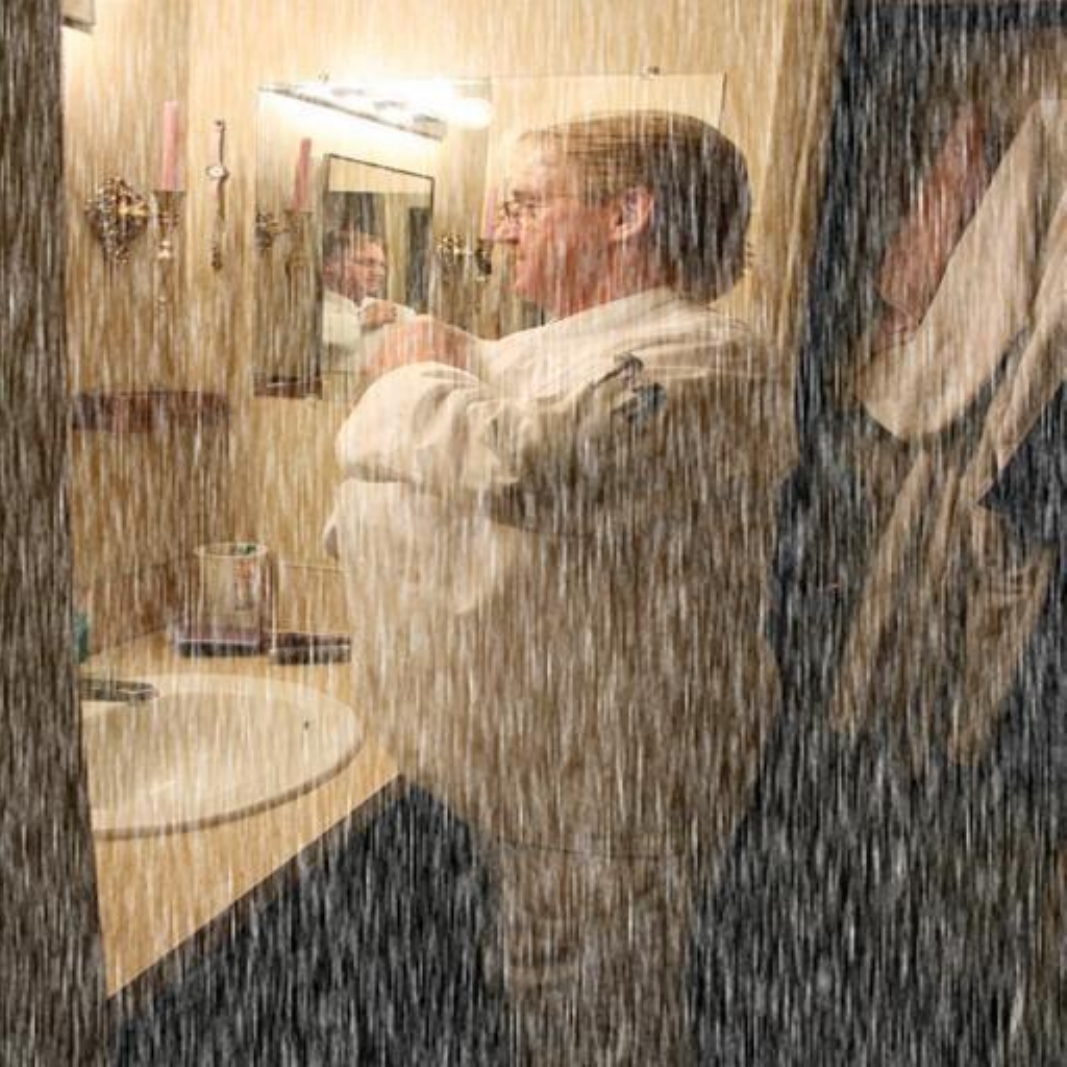}  &   
\includegraphics[width=.25\textwidth,valign=t]{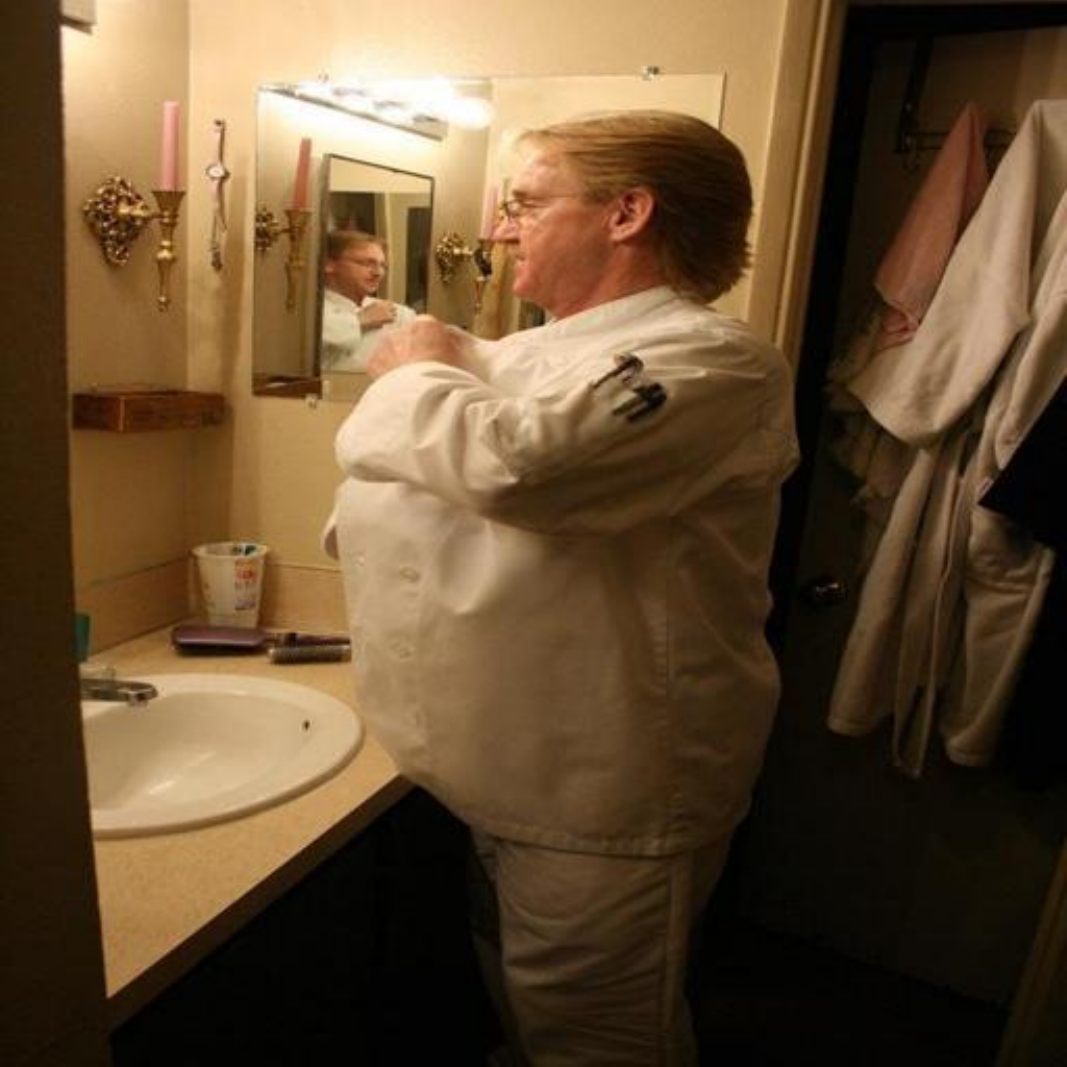} &   
\includegraphics[width=.25\textwidth,valign=t]{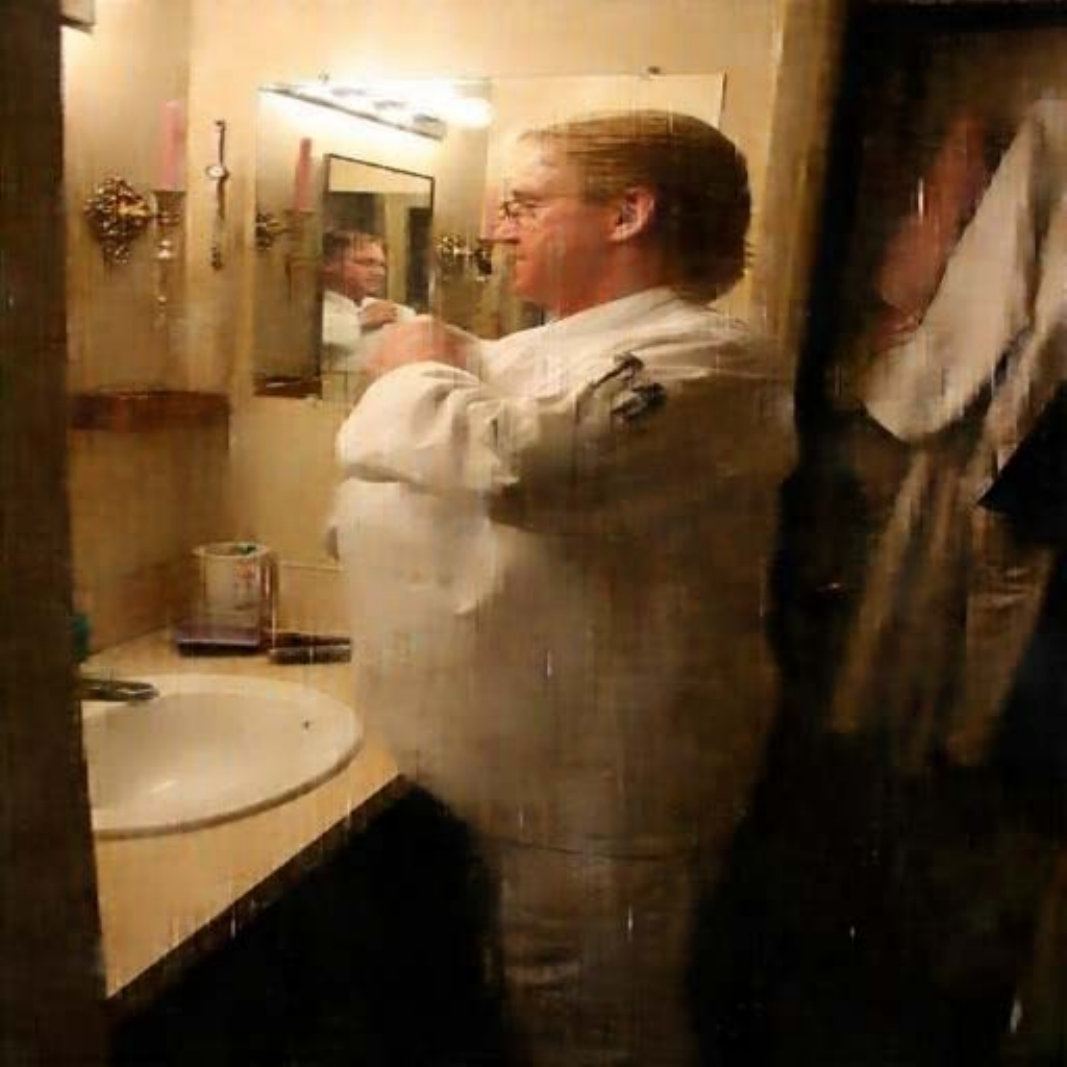}  &
\includegraphics[width=.25\textwidth,valign=t]{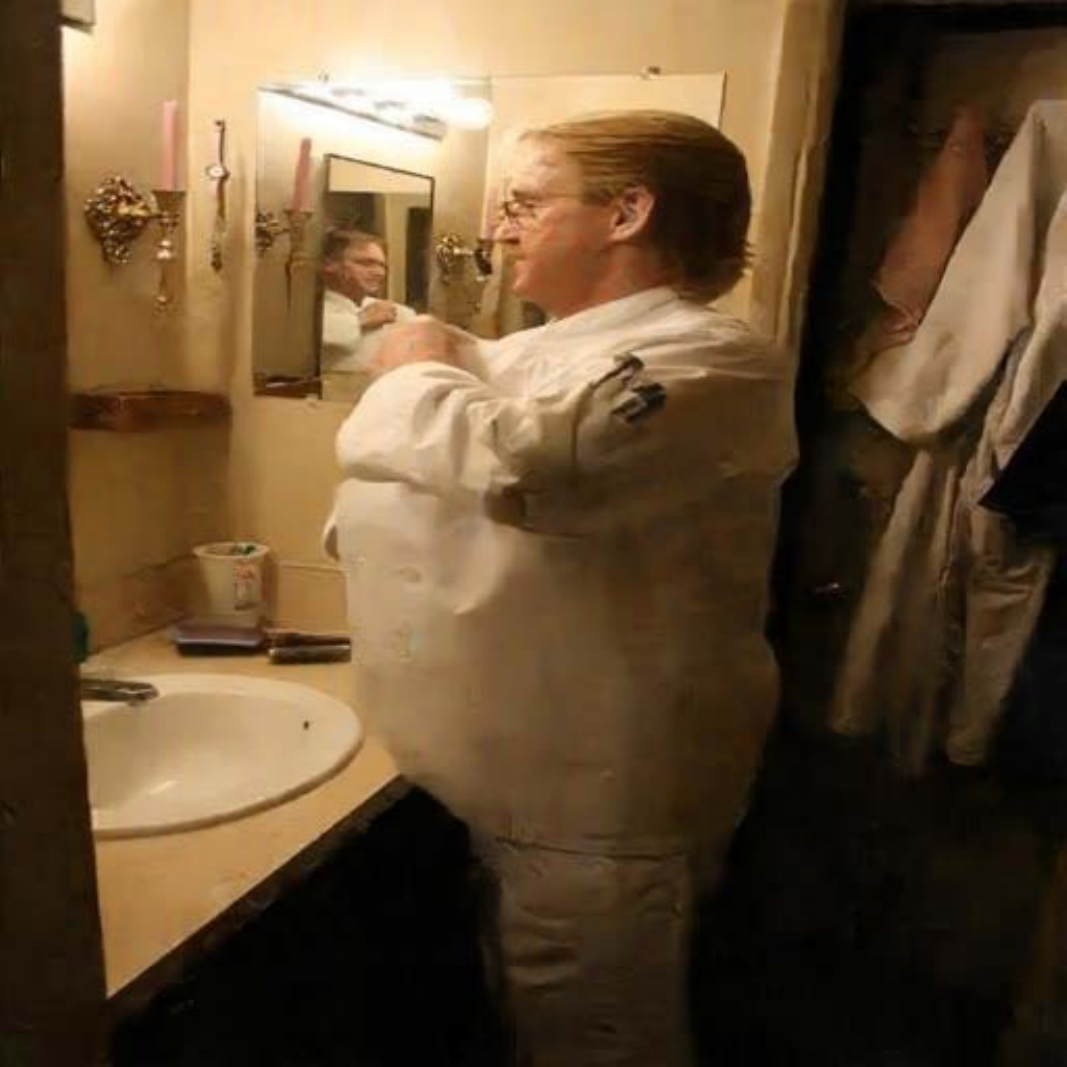} \\
\vspace{0.2em}
\small~Degraded  & \small~Label & \small~RESCAN~\cite{RESCAN}, 30.87 dB & \small~PReNet~\cite{PReNet}, 33.09 dB \\
\end{tabular}
}
\caption{\small Image deraining results on DID-Data~\cite{DID-Data}}
\label{fig:suppl_deraining_visual_results_2}
\end{center}
\vspace{-7mm}
\end{figure*}

%% file: suppl_figures/suppl_006_03_deraining_visual_results.tex
\begin{figure*}[t]
\begin{center}
\setlength{\tabcolsep}{1.5pt}
\scalebox{0.88}{
\begin{tabular}[b]{c c c c}
\vspace{0.2em}

\vspace{0.2em}
\includegraphics[width=.25\textwidth,valign=t]{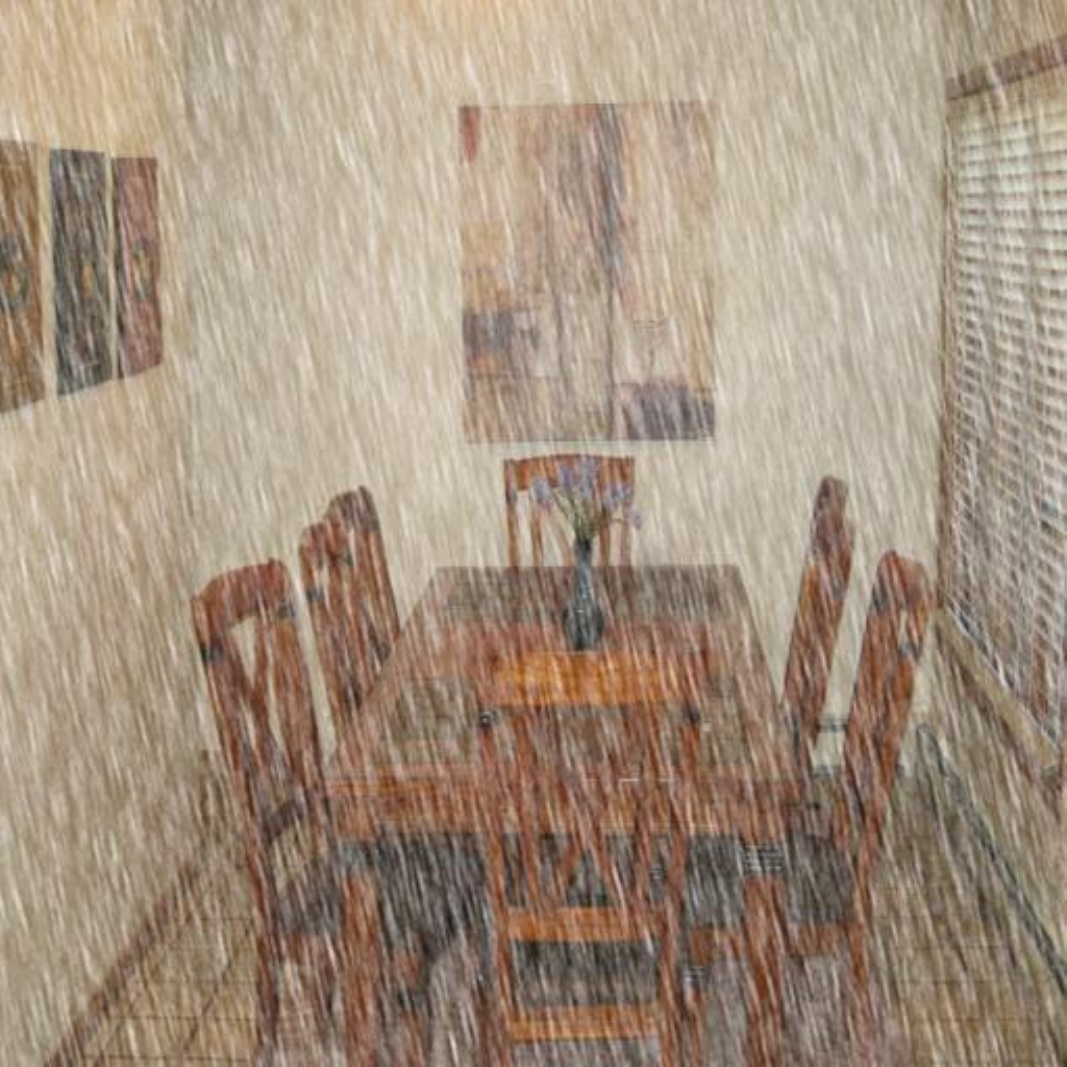}  &   
\includegraphics[width=.25\textwidth,valign=t]{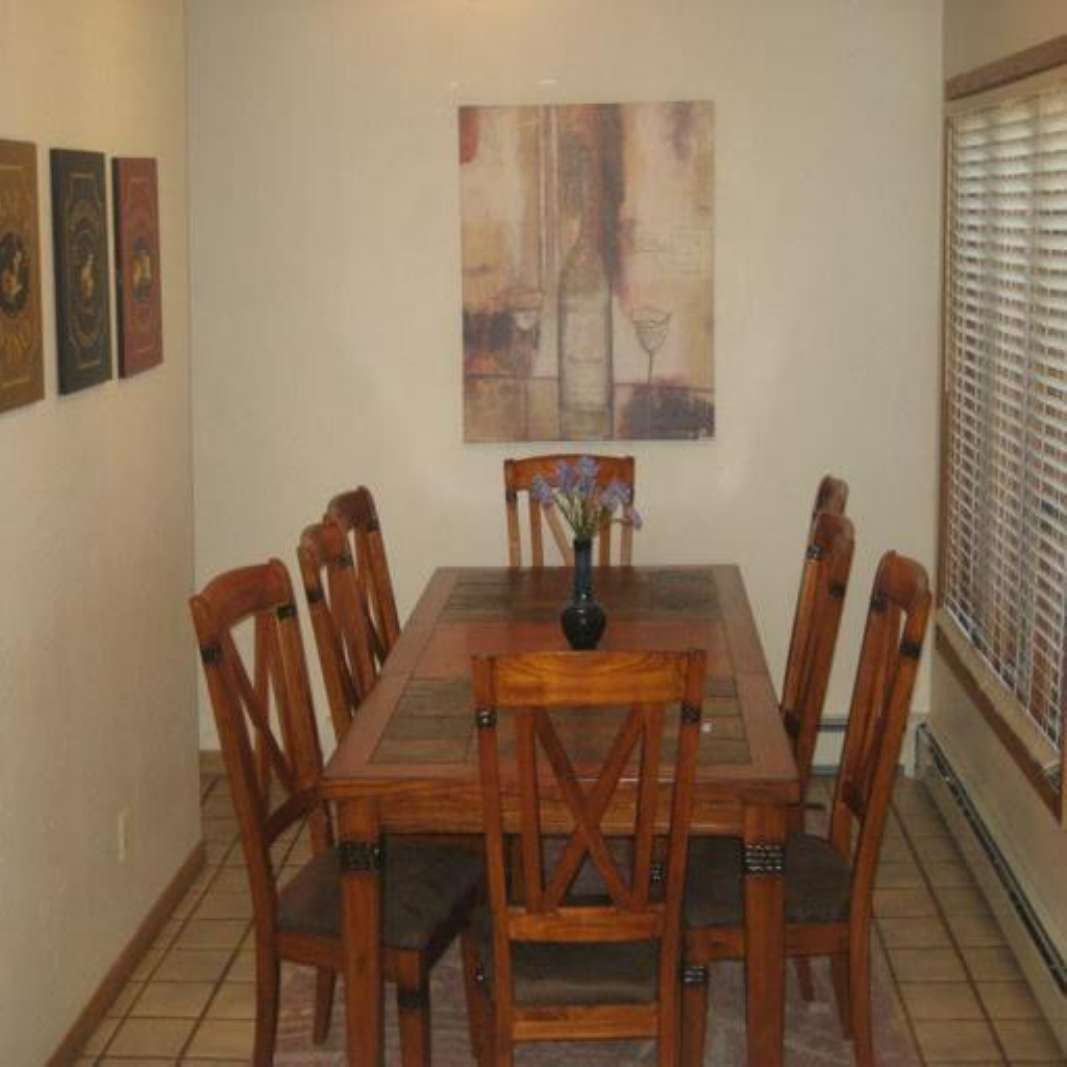} &   
\includegraphics[width=.25\textwidth,valign=t]{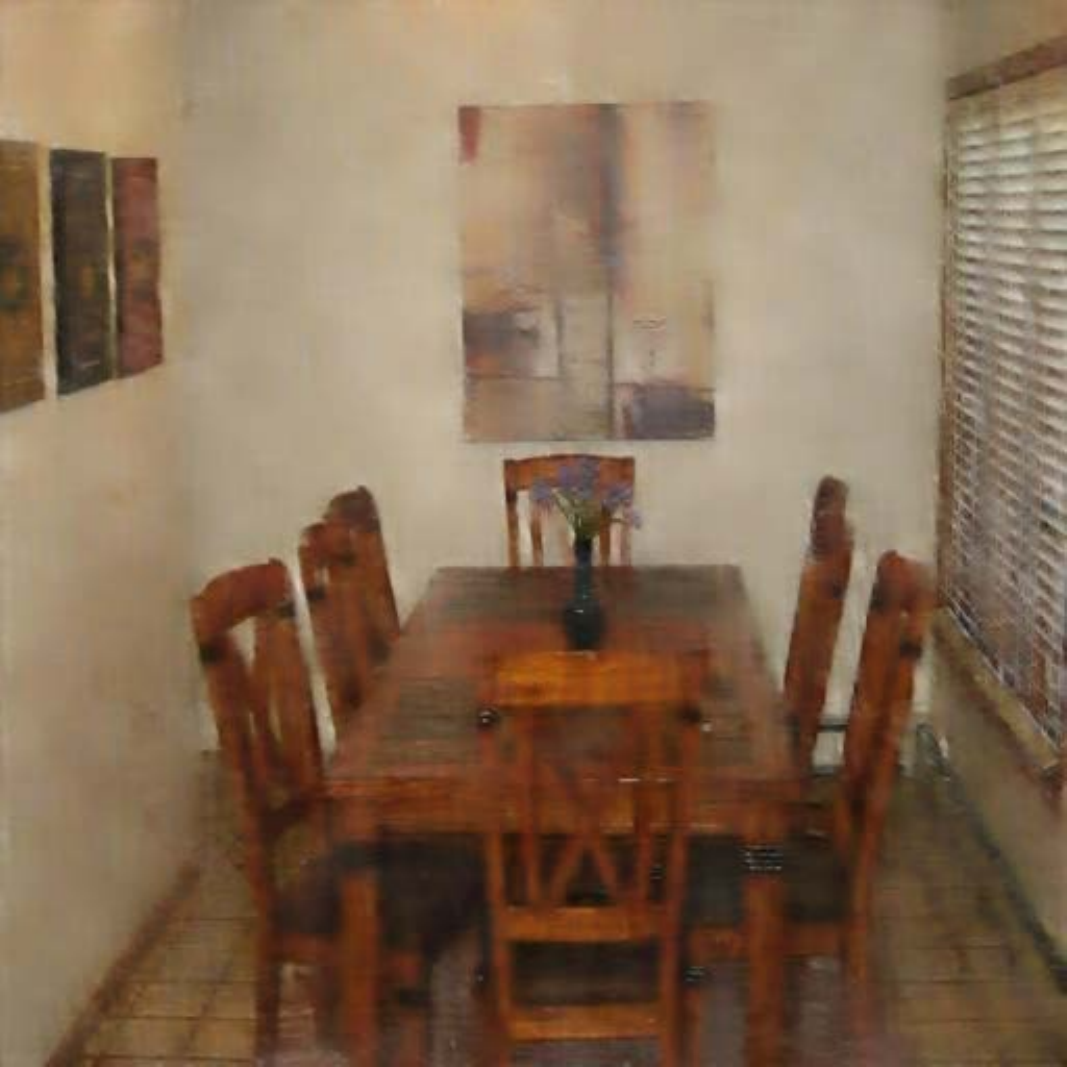}  &
\includegraphics[width=.25\textwidth,valign=t]{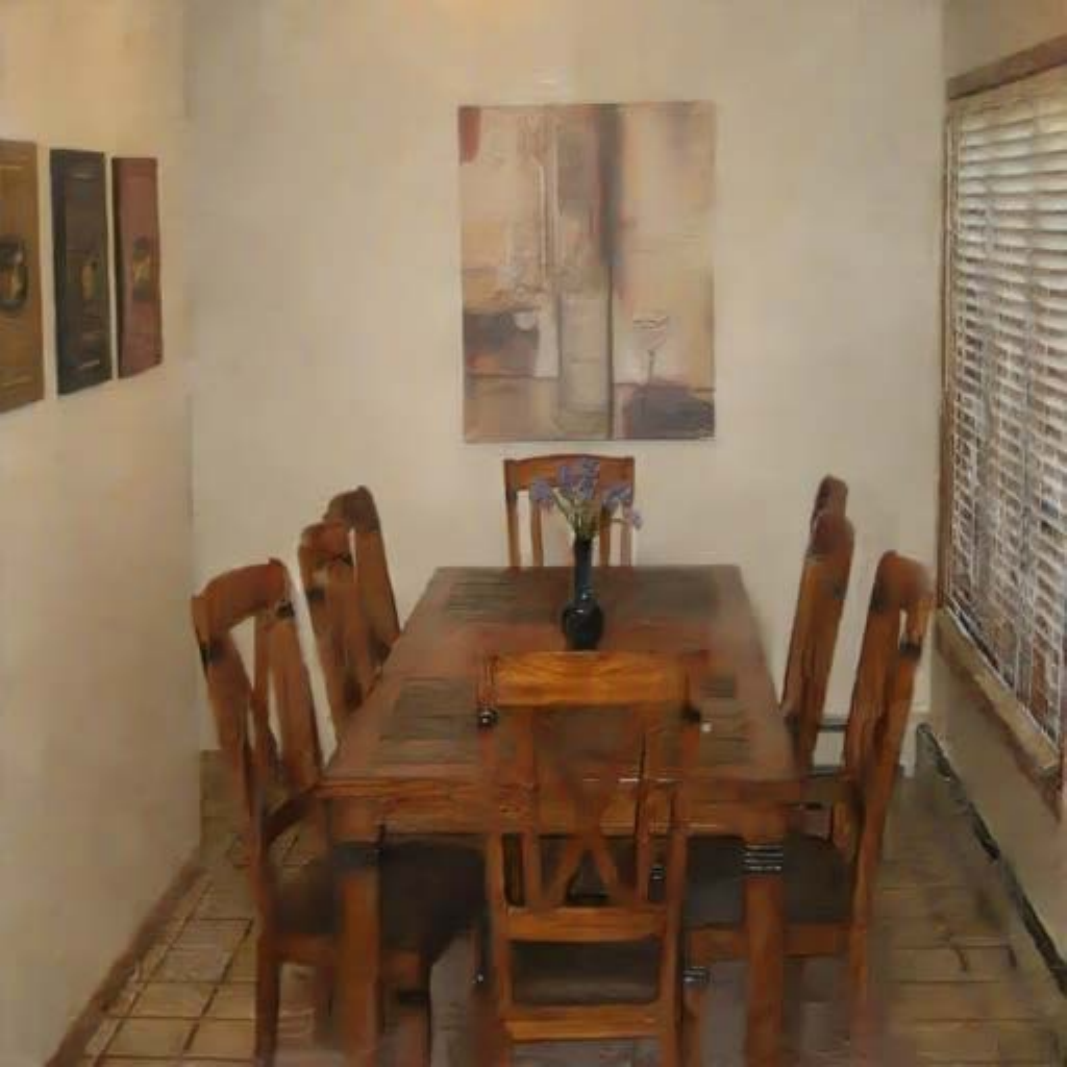} \\
\vspace{0.2em}
\small~Degraded  & \small~Label & \small~RESCAN~\cite{RESCAN}, 30.42 dB & \small~PReNet~\cite{PReNet}, 31.59 dB \\
\vspace{0.2em}
\includegraphics[width=.25\textwidth,valign=t]{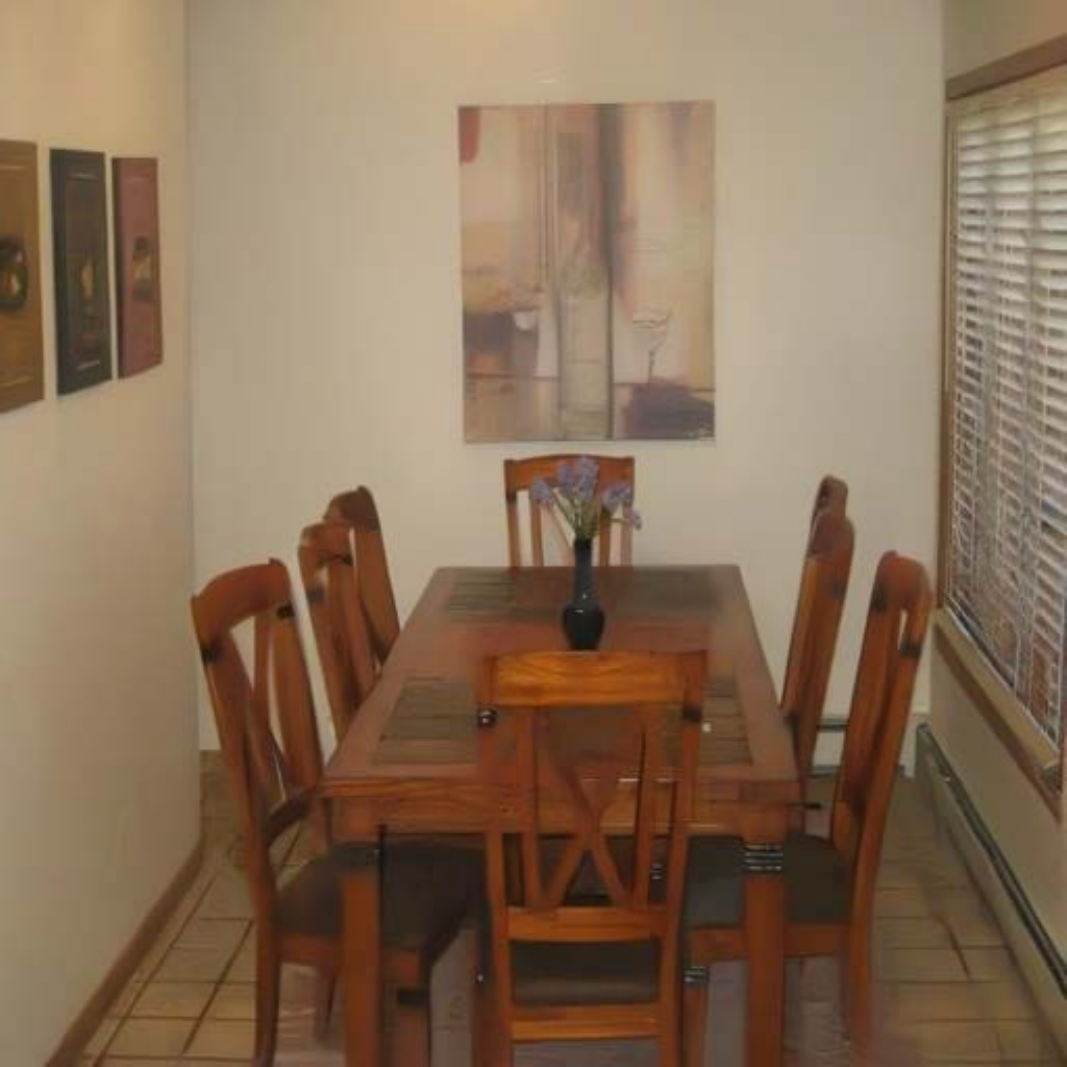}  &   
\includegraphics[width=.25\textwidth,valign=t]{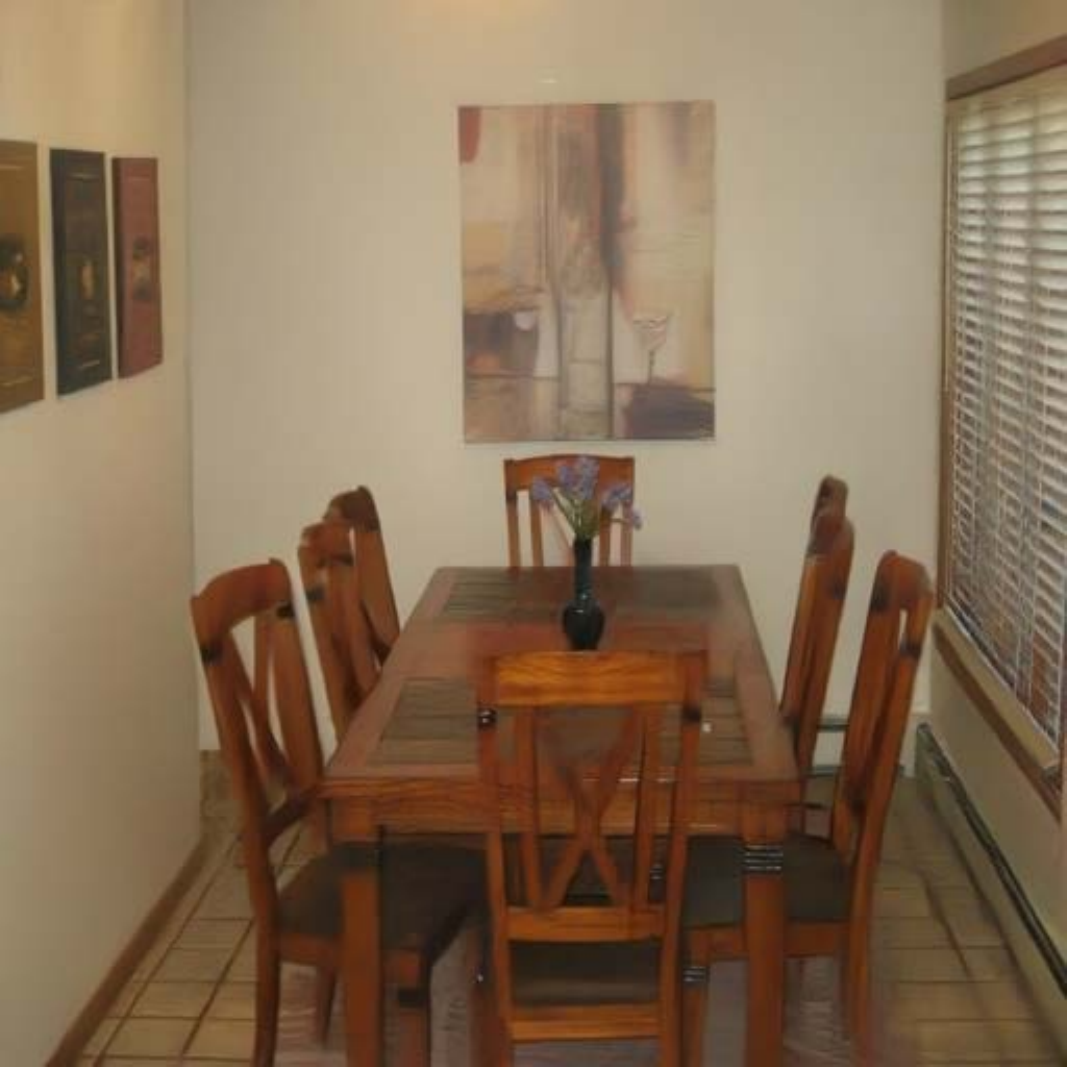} &   
\includegraphics[width=.25\textwidth,valign=t]{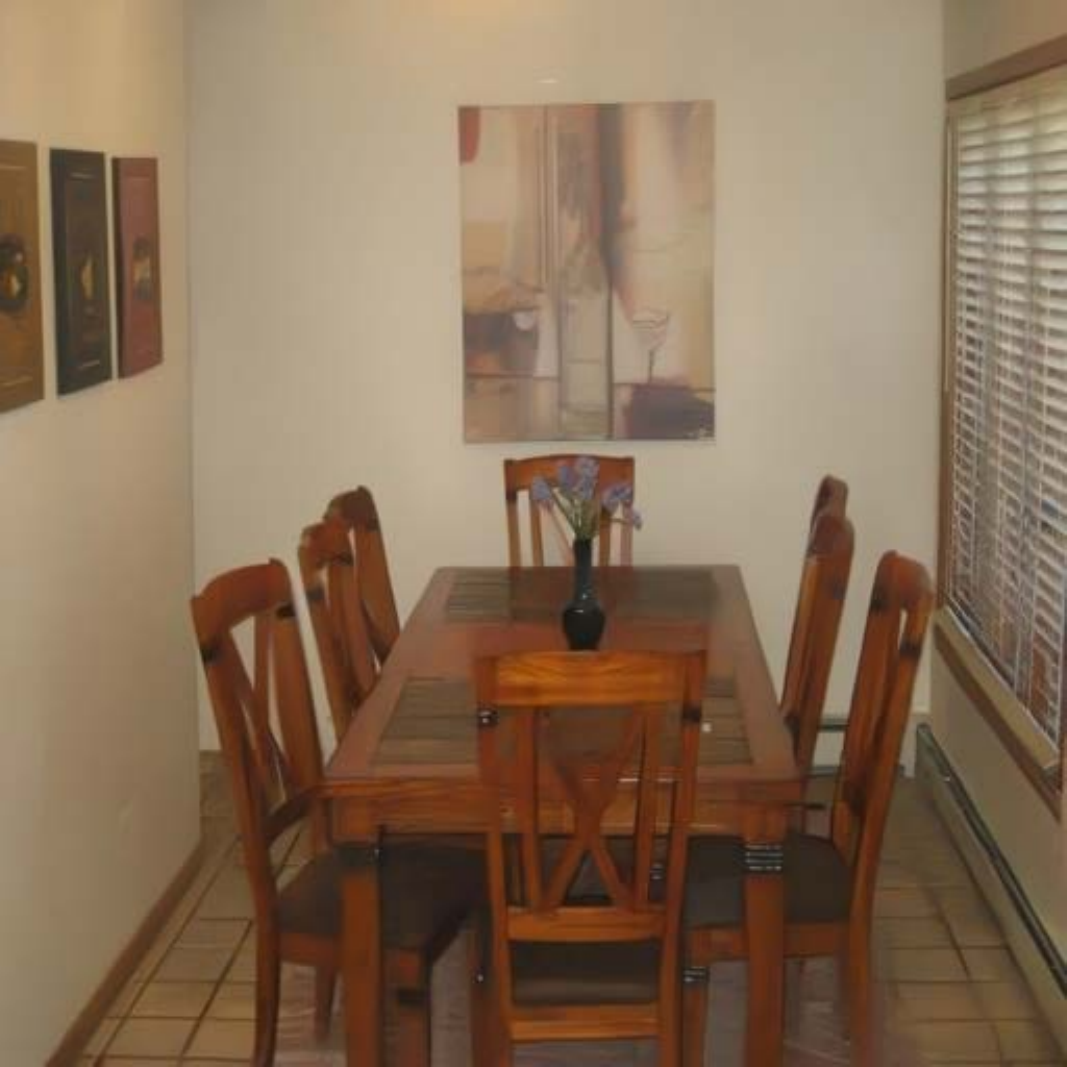}  &
\includegraphics[width=.25\textwidth,valign=t]{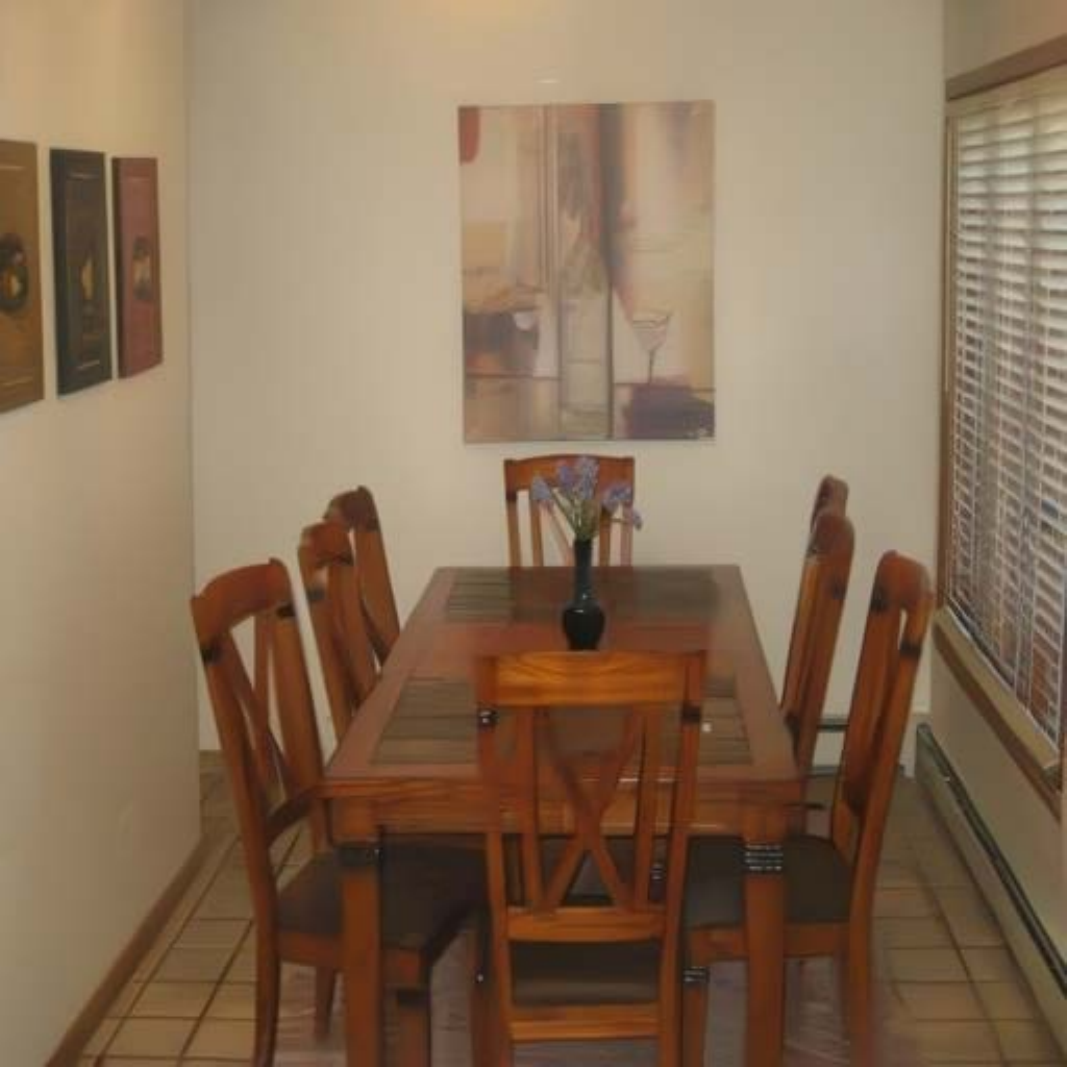} \\
\vspace{0.2em}
\small~MPRNet~\cite{MPRNet}, 33.71 dB  & \small~Uformer~\cite{Uformer}, 34.05 dB & \small~DRSformer~\cite{DRSformer}, 34.44 dB & \small~\textbf{Ours}, \textbf{34.65 dB} \\
\vspace{0.2em}
\includegraphics[width=.25\textwidth,valign=t]{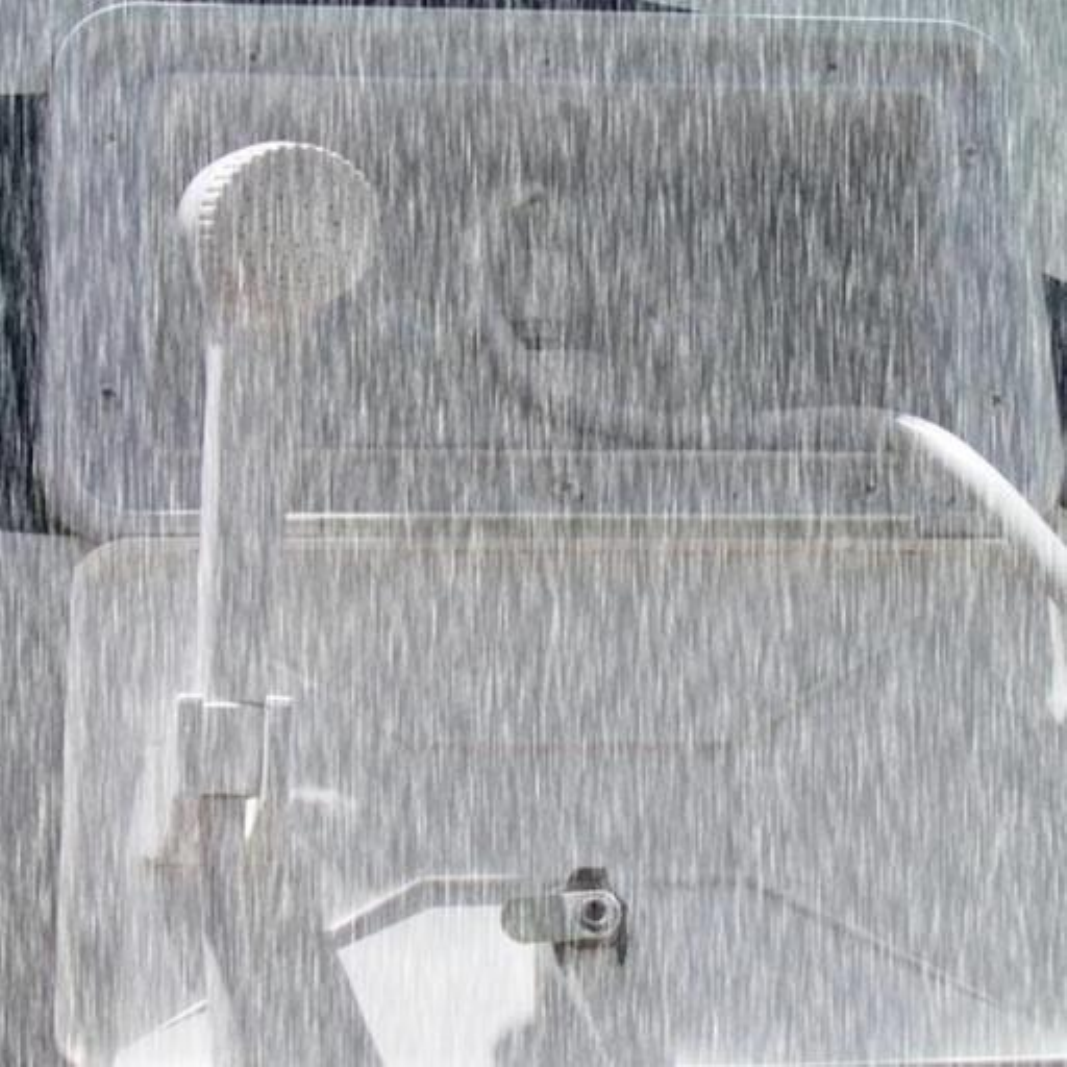}  &   
\includegraphics[width=.25\textwidth,valign=t]{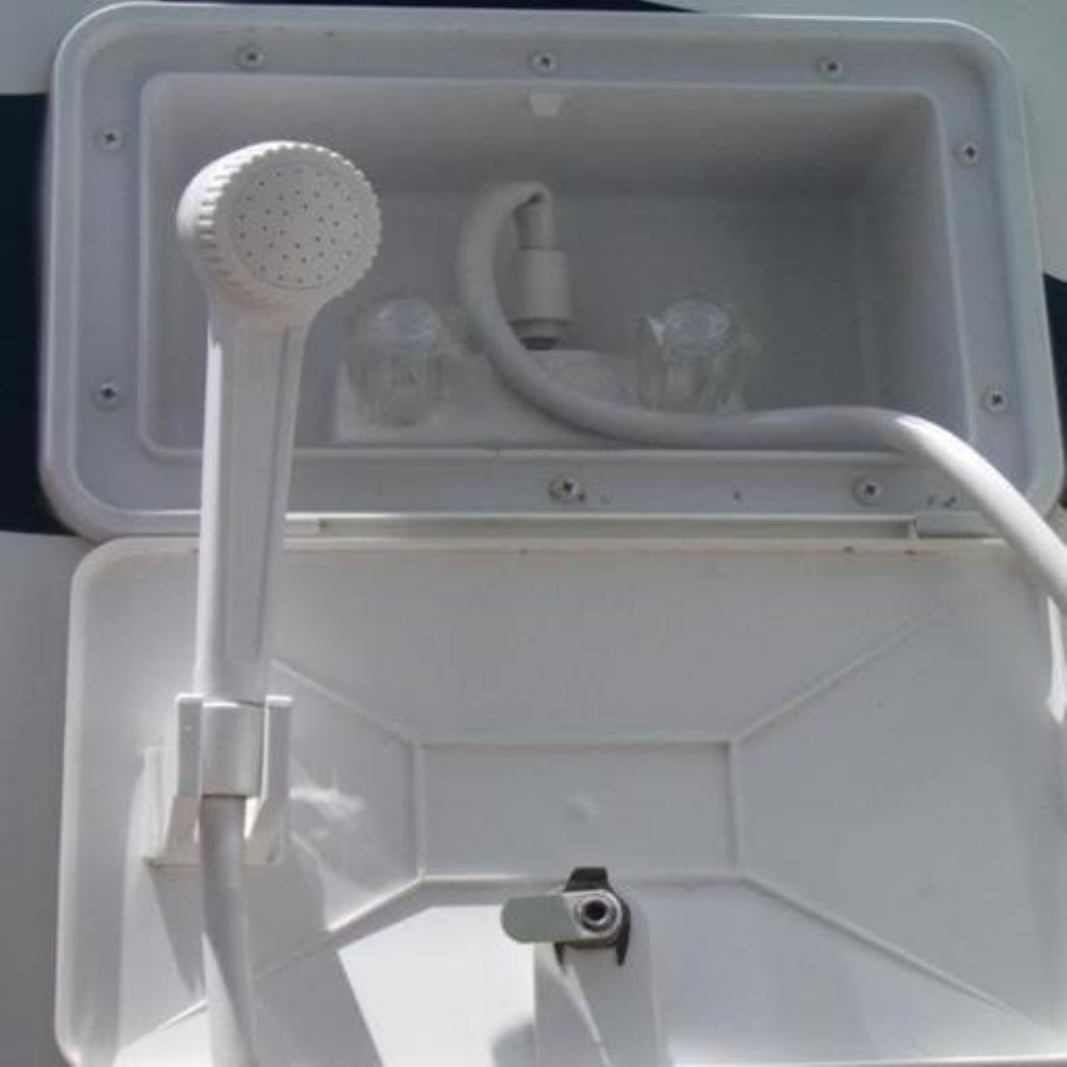} &   
\includegraphics[width=.25\textwidth,valign=t]{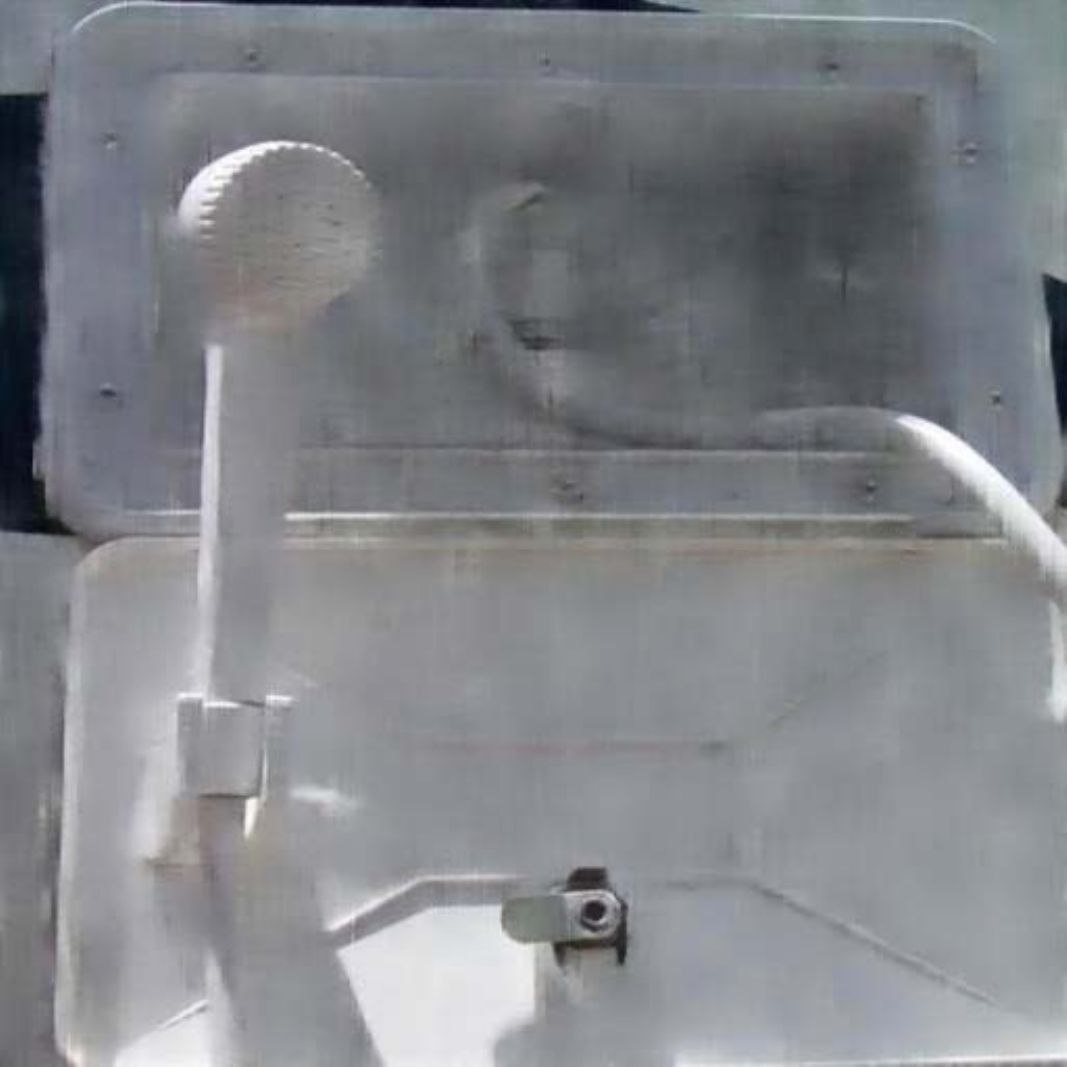}  &
\includegraphics[width=.25\textwidth,valign=t]{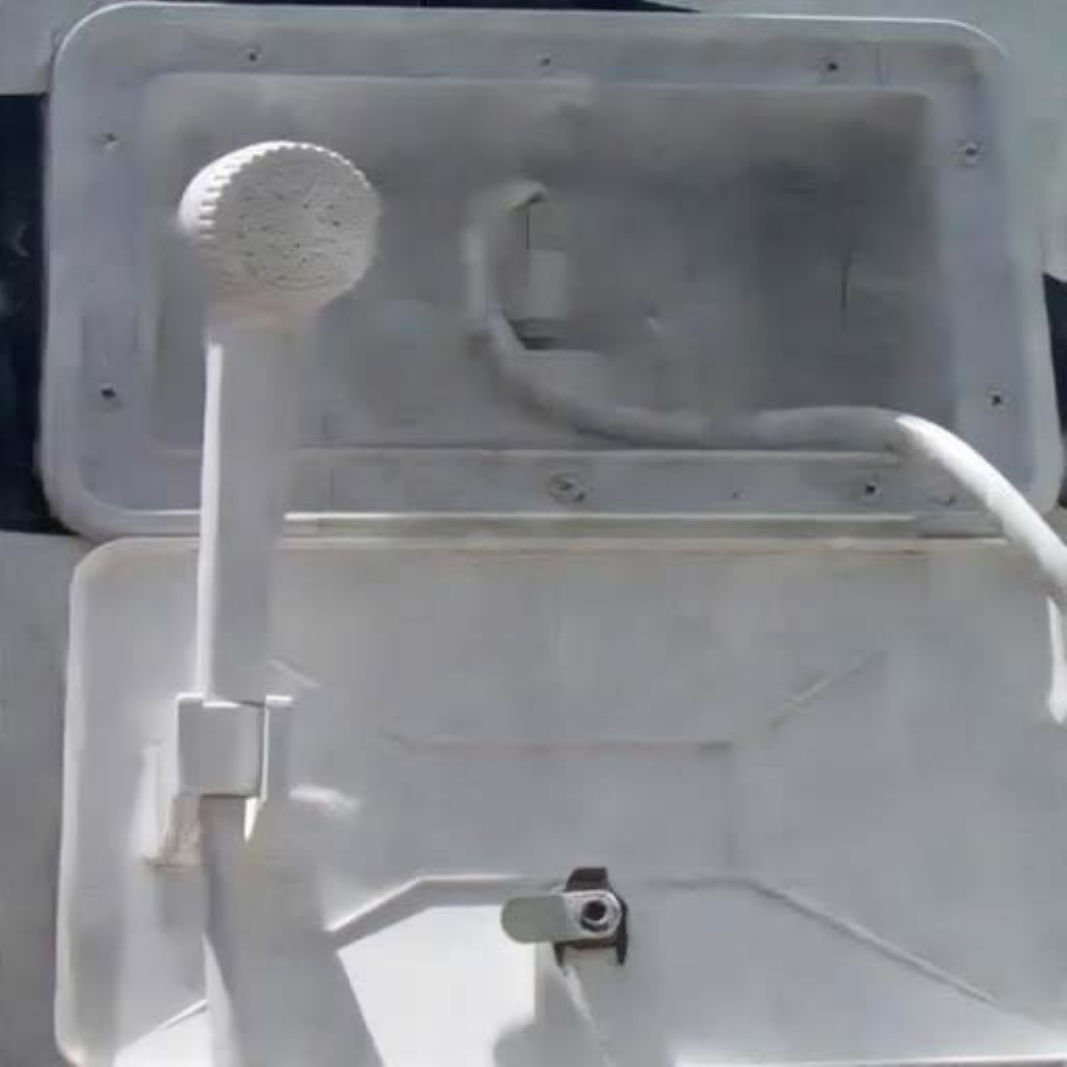} \\
\vspace{0.2em}
\small~Degraded  & \small~Label & \small~RESCAN~\cite{RESCAN}, 33.07 dB & \small~PReNet~\cite{PReNet}, 34.97 dB \\
\end{tabular}
}
\caption{\small Image deraining results on DID-Data~\cite{DID-Data}}
\label{fig:suppl_deraining_visual_results_3}
\end{center}
\vspace{-7mm}
\end{figure*}

%% file: suppl_figures/suppl_008_denoising_visual_results.tex
\begin{figure*}[t]
\begin{center}
\setlength{\tabcolsep}{1.5pt}
\scalebox{0.88}{
\begin{tabular}[b]{c c c}
\hspace{-4mm}
\vspace{0.4em}
\includegraphics[width=.33\textwidth,valign=t]{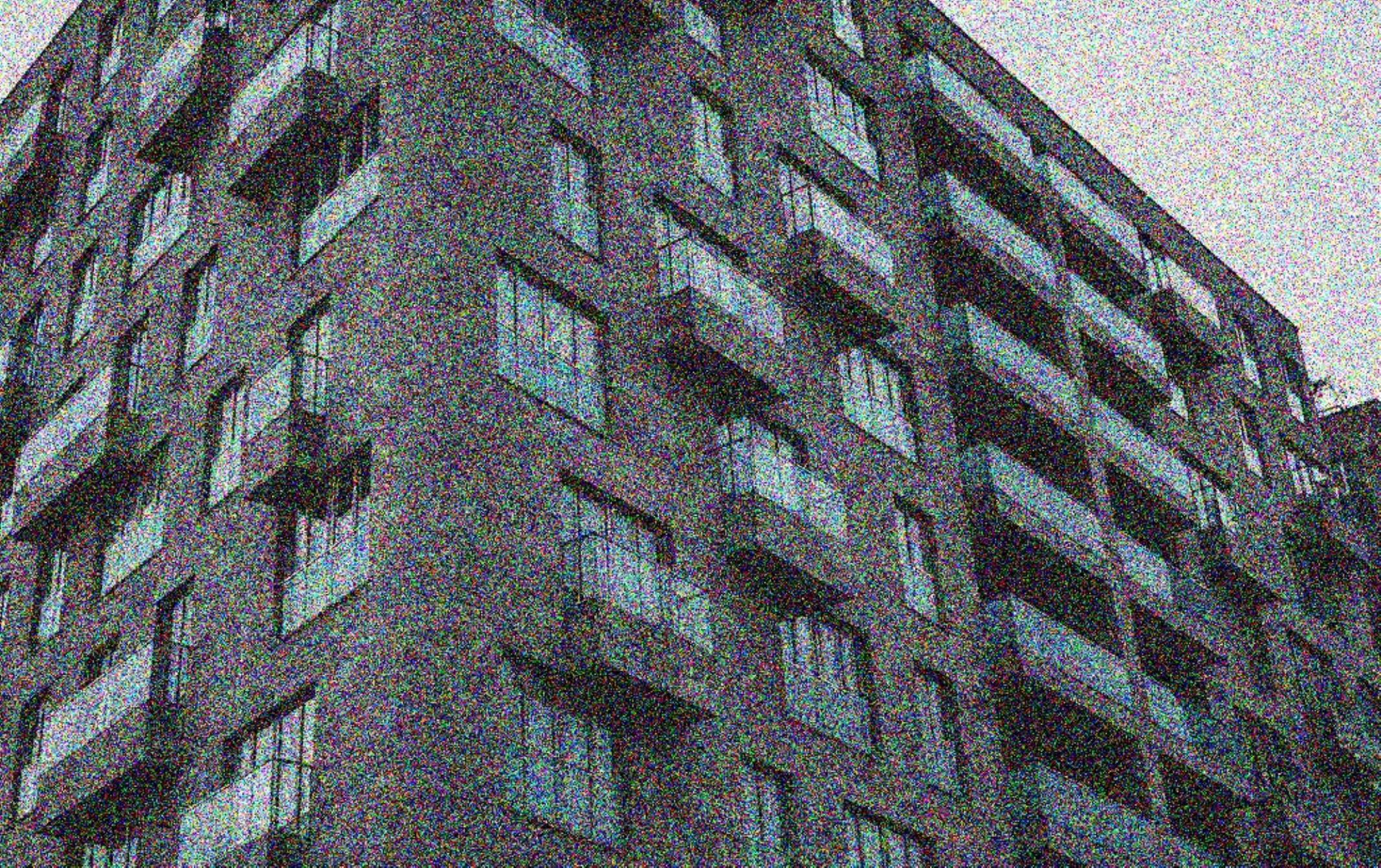}  &   
\includegraphics[width=.33\textwidth,valign=t]{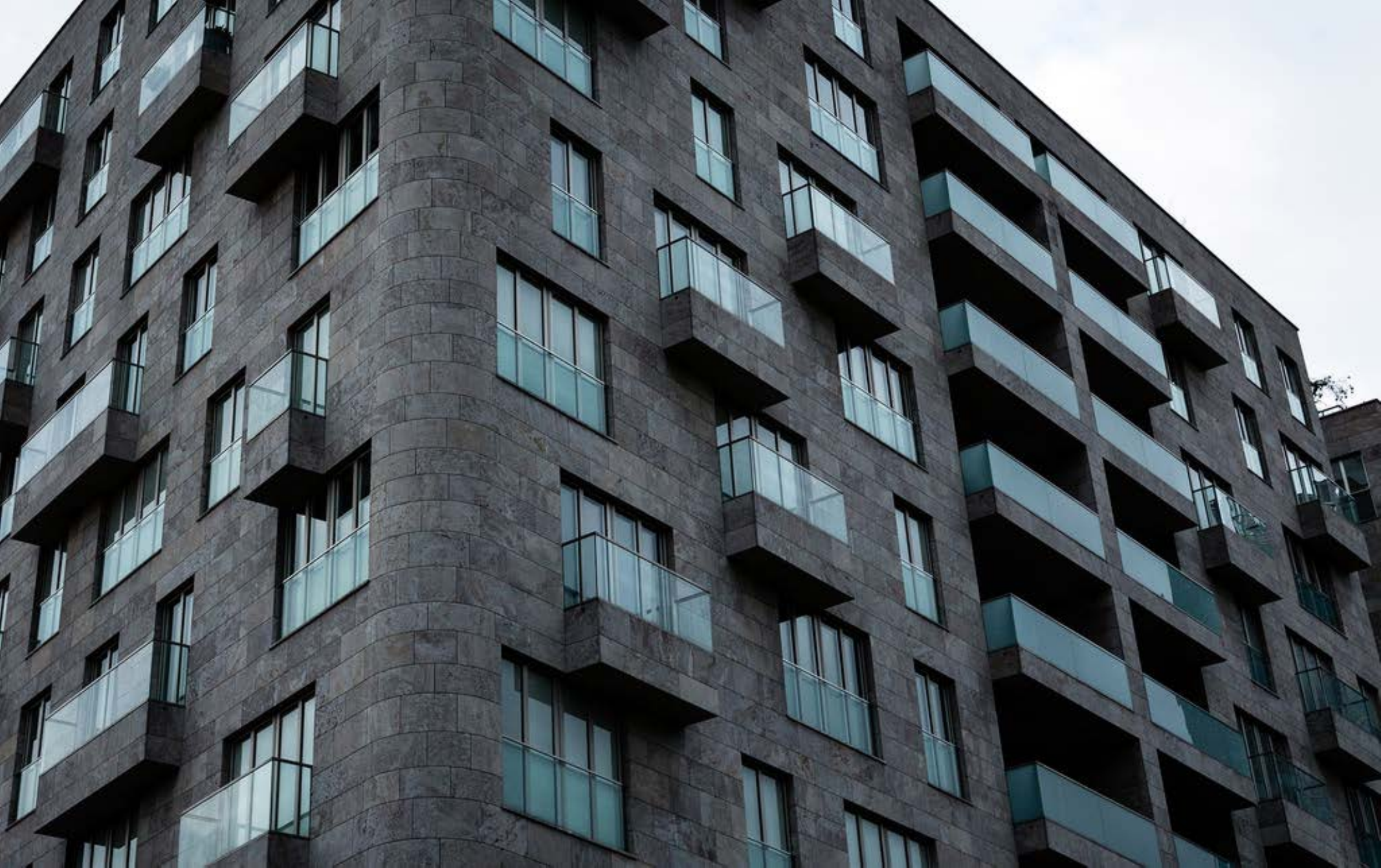} &   
\includegraphics[width=.33\textwidth,valign=t]{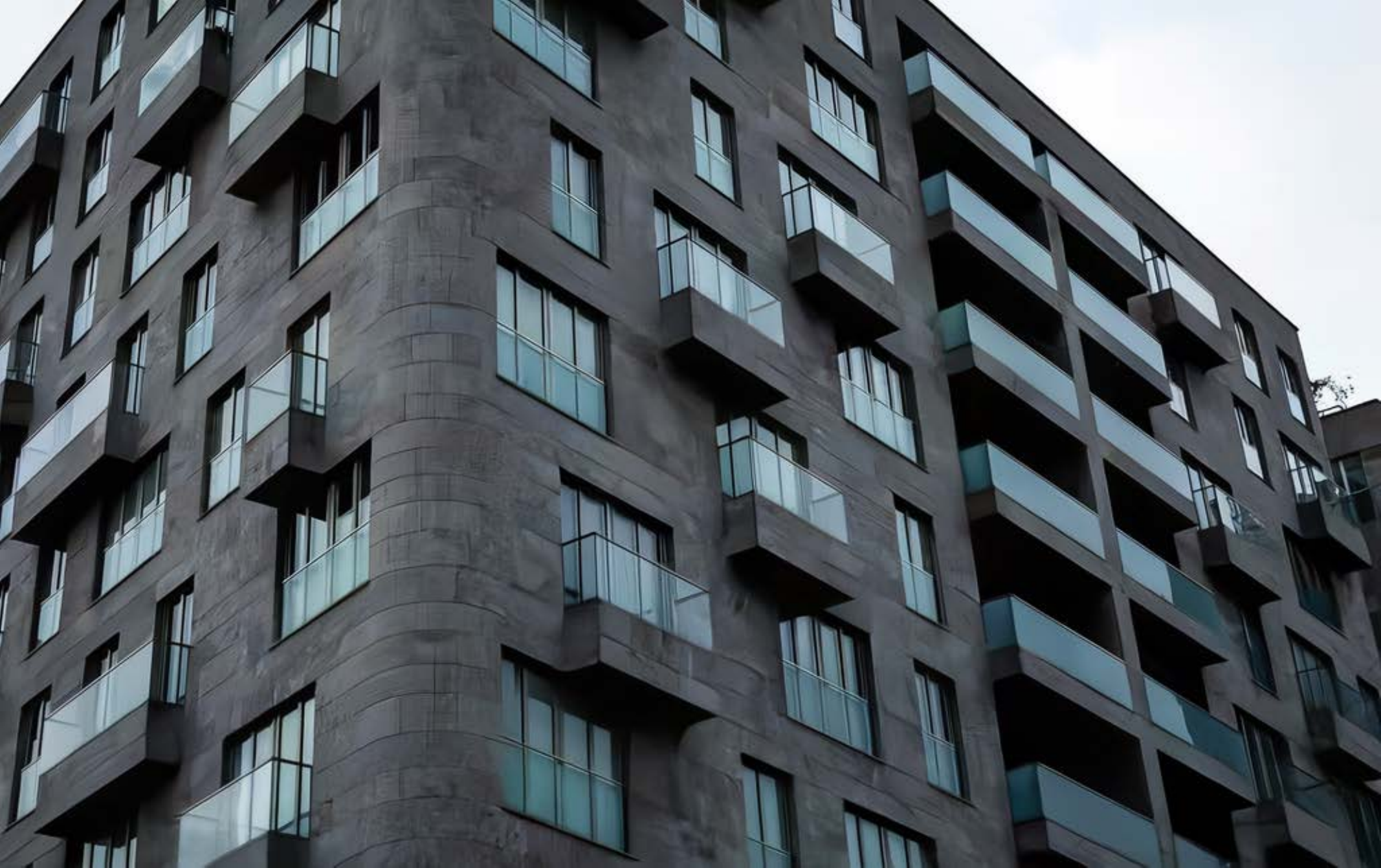} \\
\vspace{0.6em}
\small~Degraded  & \small~Label & \small~RNAN~\cite{RNAN}, 29.62 dB \\

\vspace{0.4em}
\includegraphics[width=.33\textwidth,valign=t]{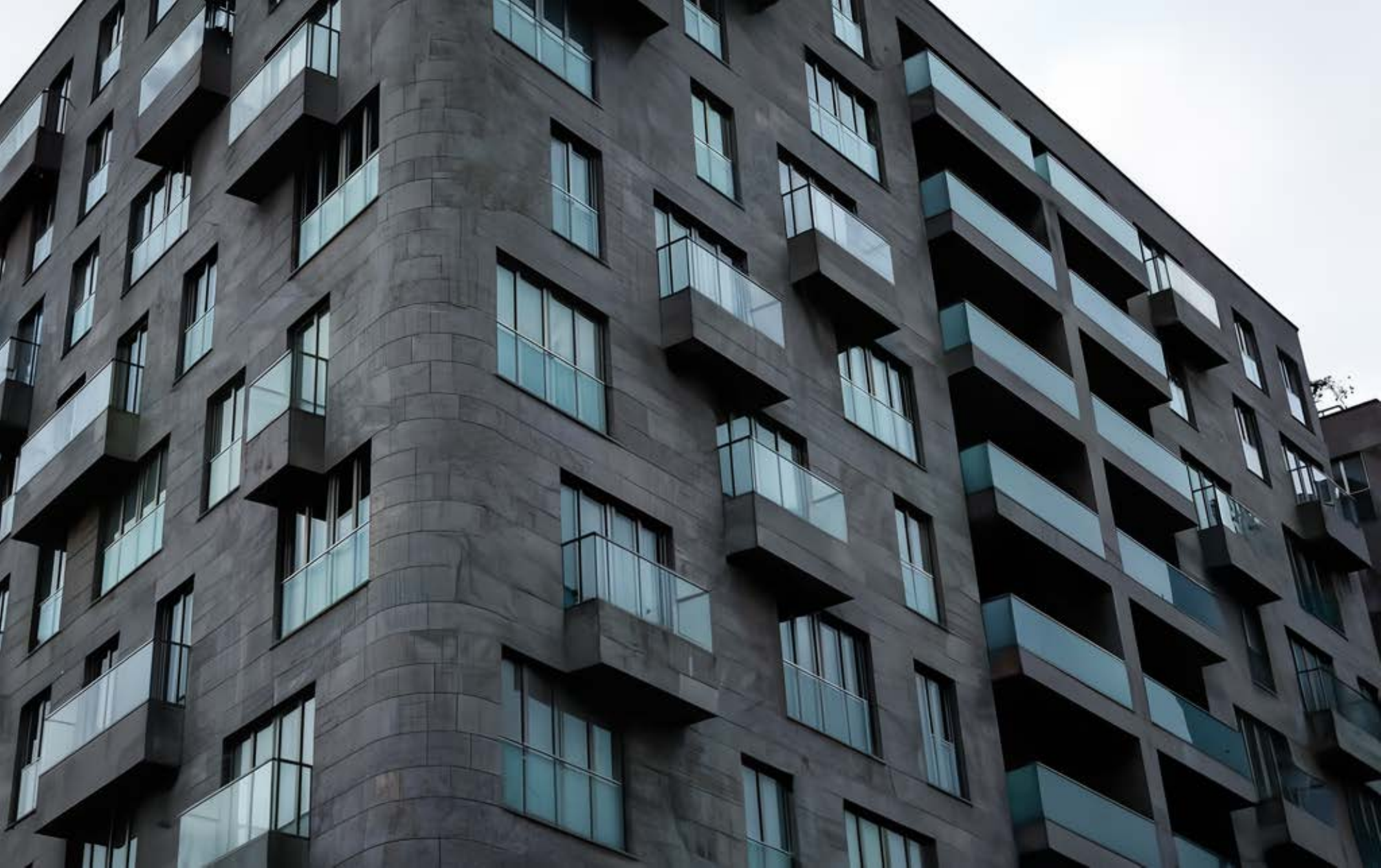}  &   
\includegraphics[width=.33\textwidth,valign=t]{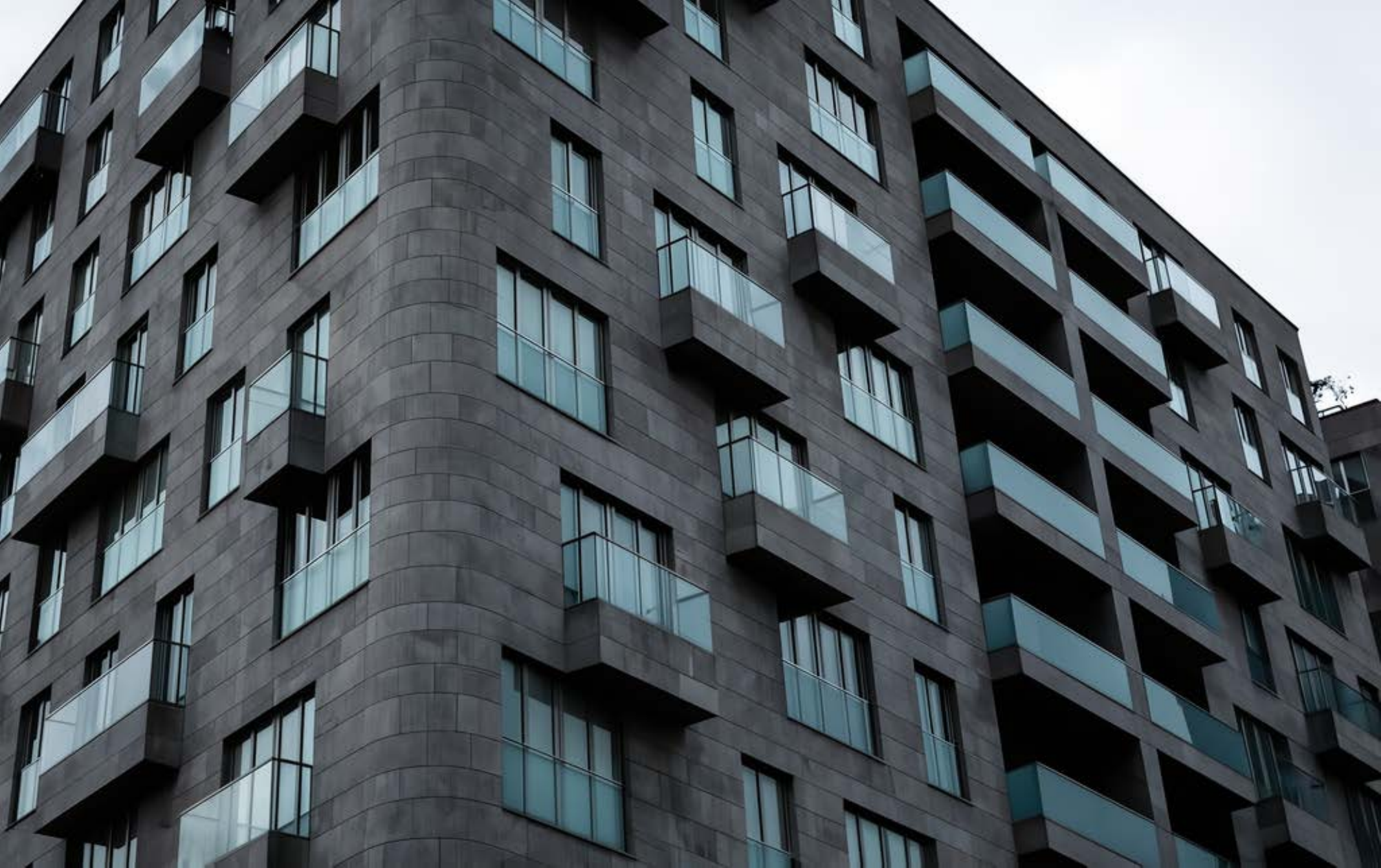} &   
\includegraphics[width=.33\textwidth,valign=t]{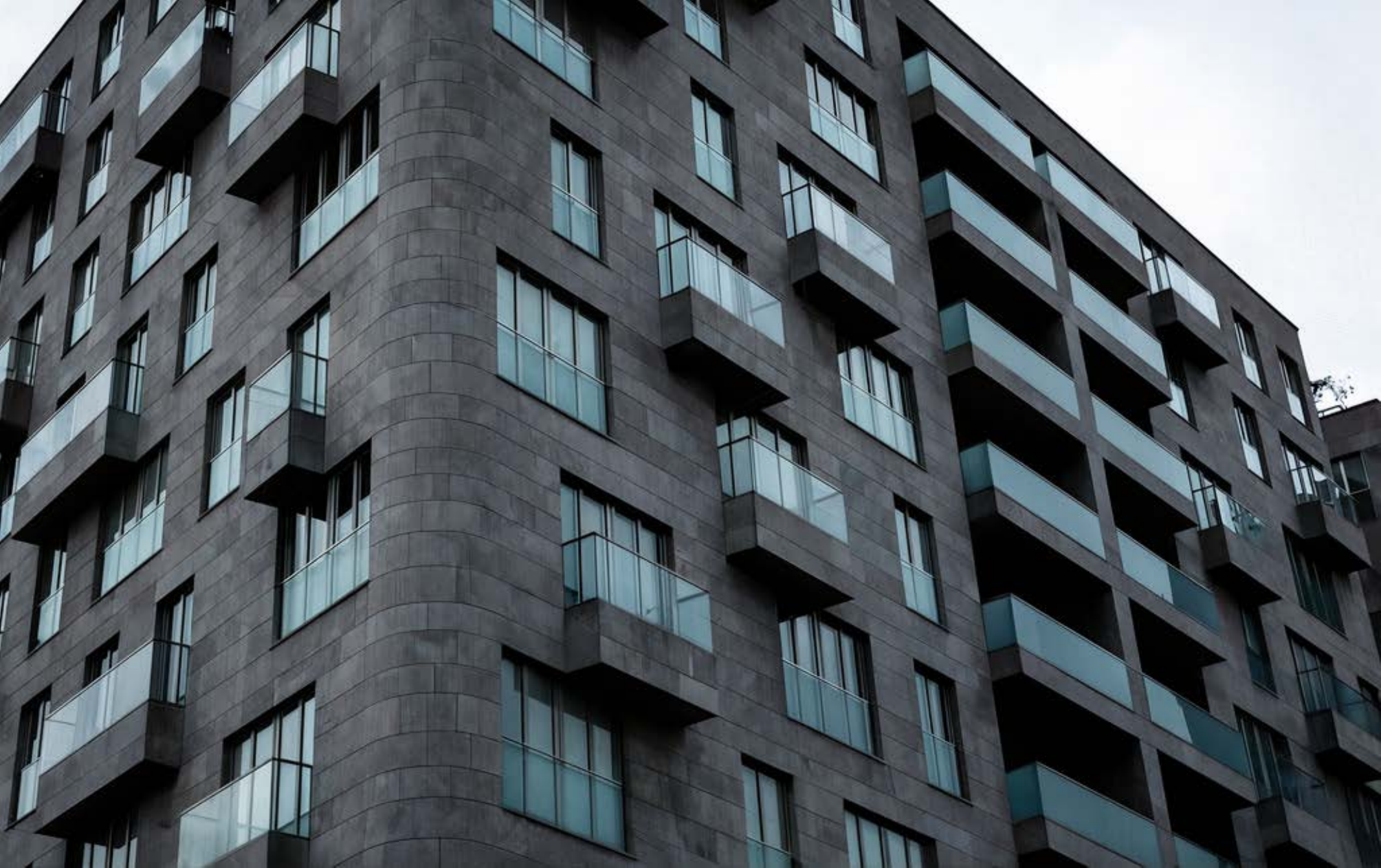} \\
\vspace{0.6em}
\small~SwinIR~\cite{SwinIR} 30.09 dB  & \small~Restormer~\cite{Restormer}, 30.65 dB & \small~\textbf{Ours}, \textbf{30.71 dB} \\

\vspace{0.4em}
\includegraphics[width=.33\textwidth,valign=t]{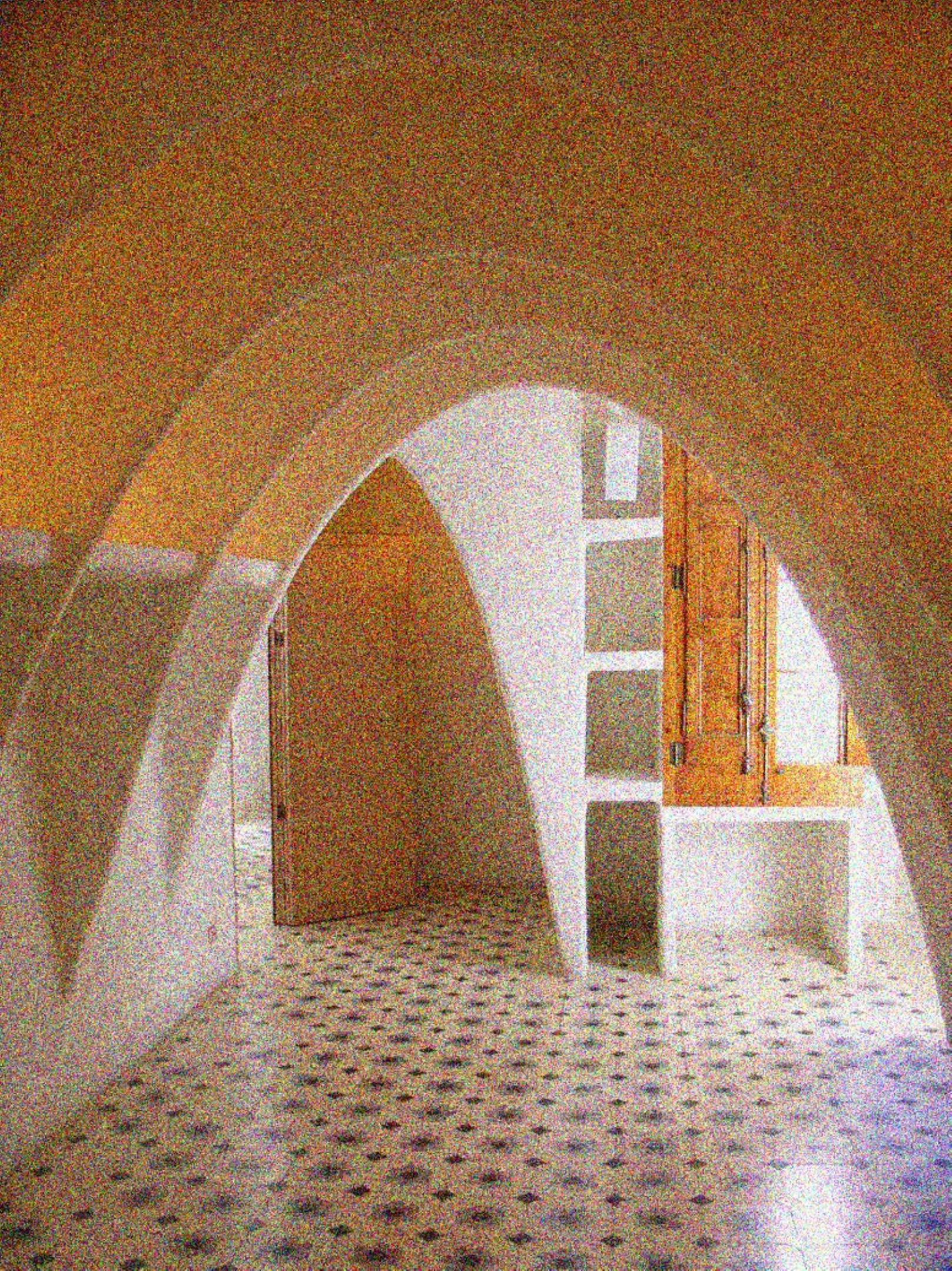}  &   
\includegraphics[width=.33\textwidth,valign=t]{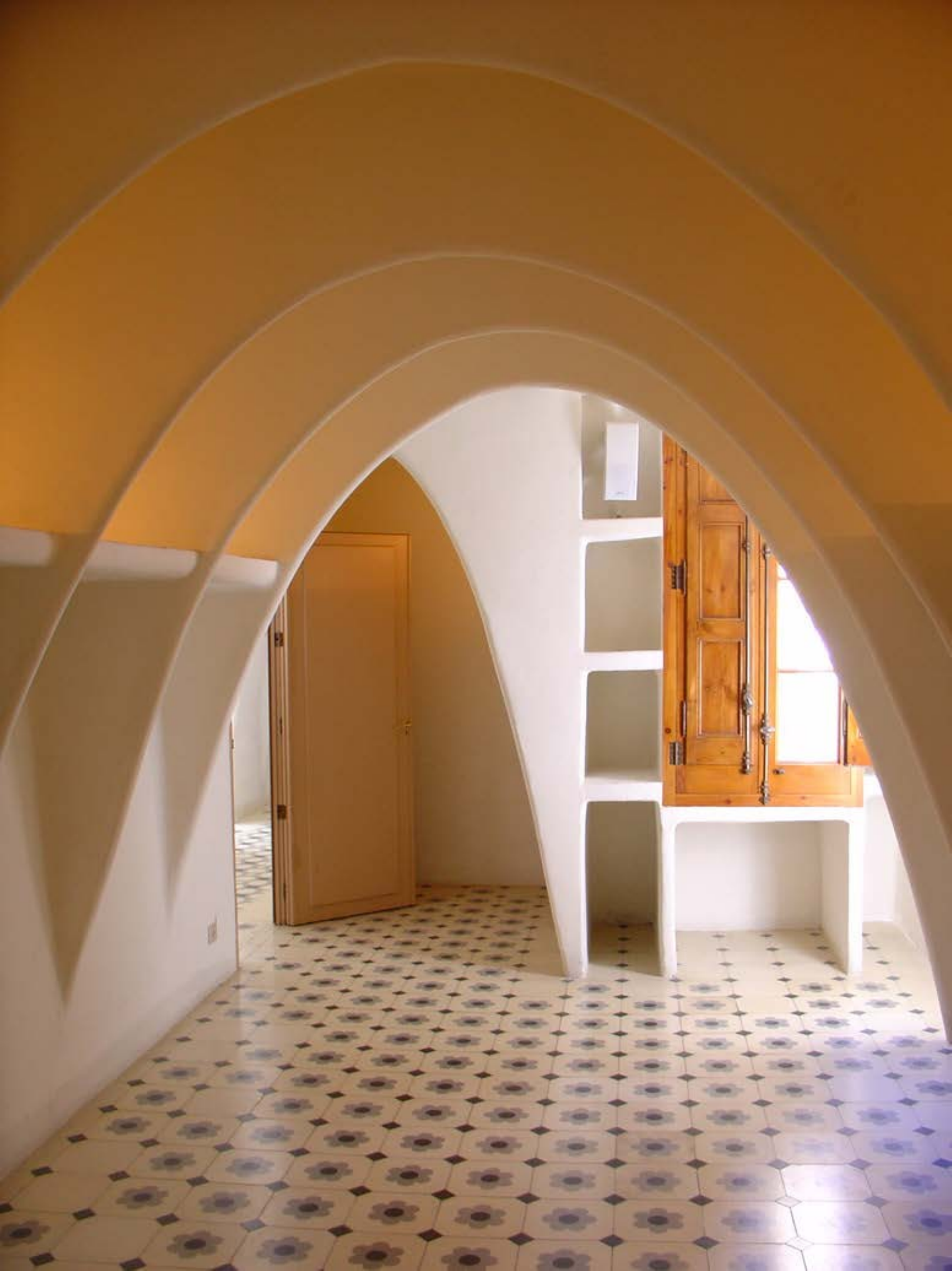} &   
\includegraphics[width=.33\textwidth,valign=t]{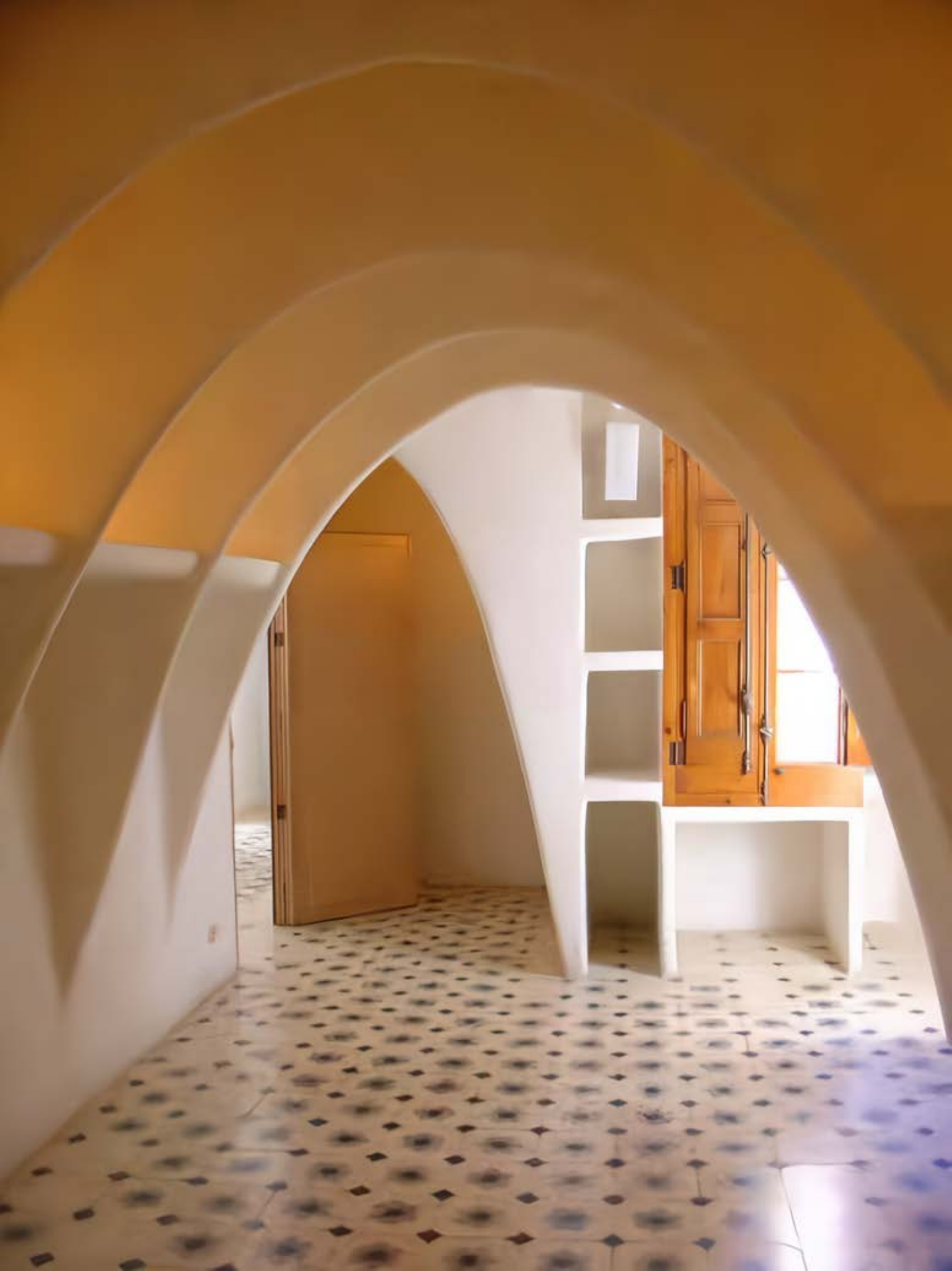} \\
\vspace{0.6em}
\small~Degraded  & \small~Label & \small~RNAN~\cite{RNAN}, 34.42 dB \\

\vspace{0.4em}
\includegraphics[width=.33\textwidth,valign=t]{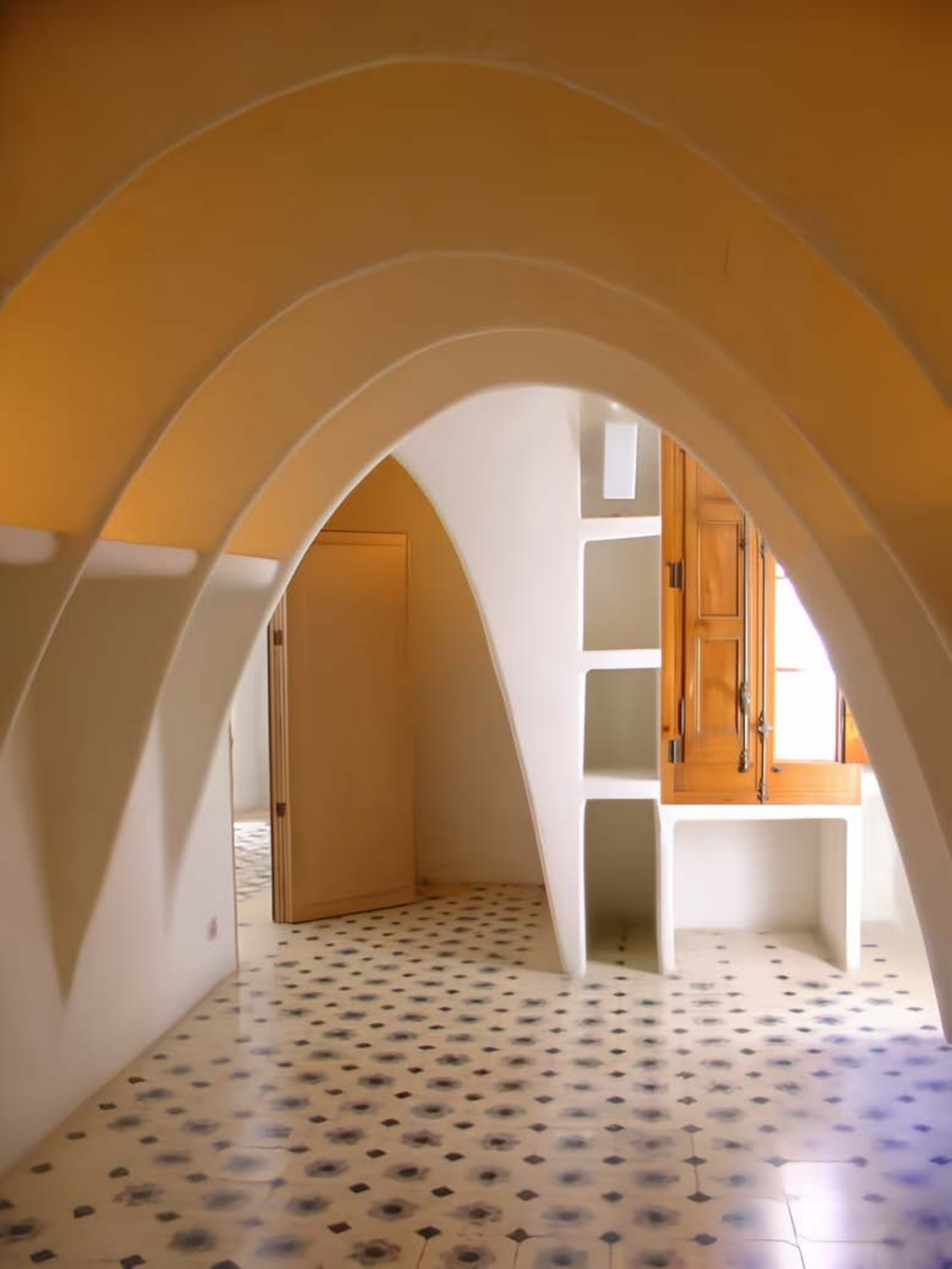}  &   
\includegraphics[width=.33\textwidth,valign=t]{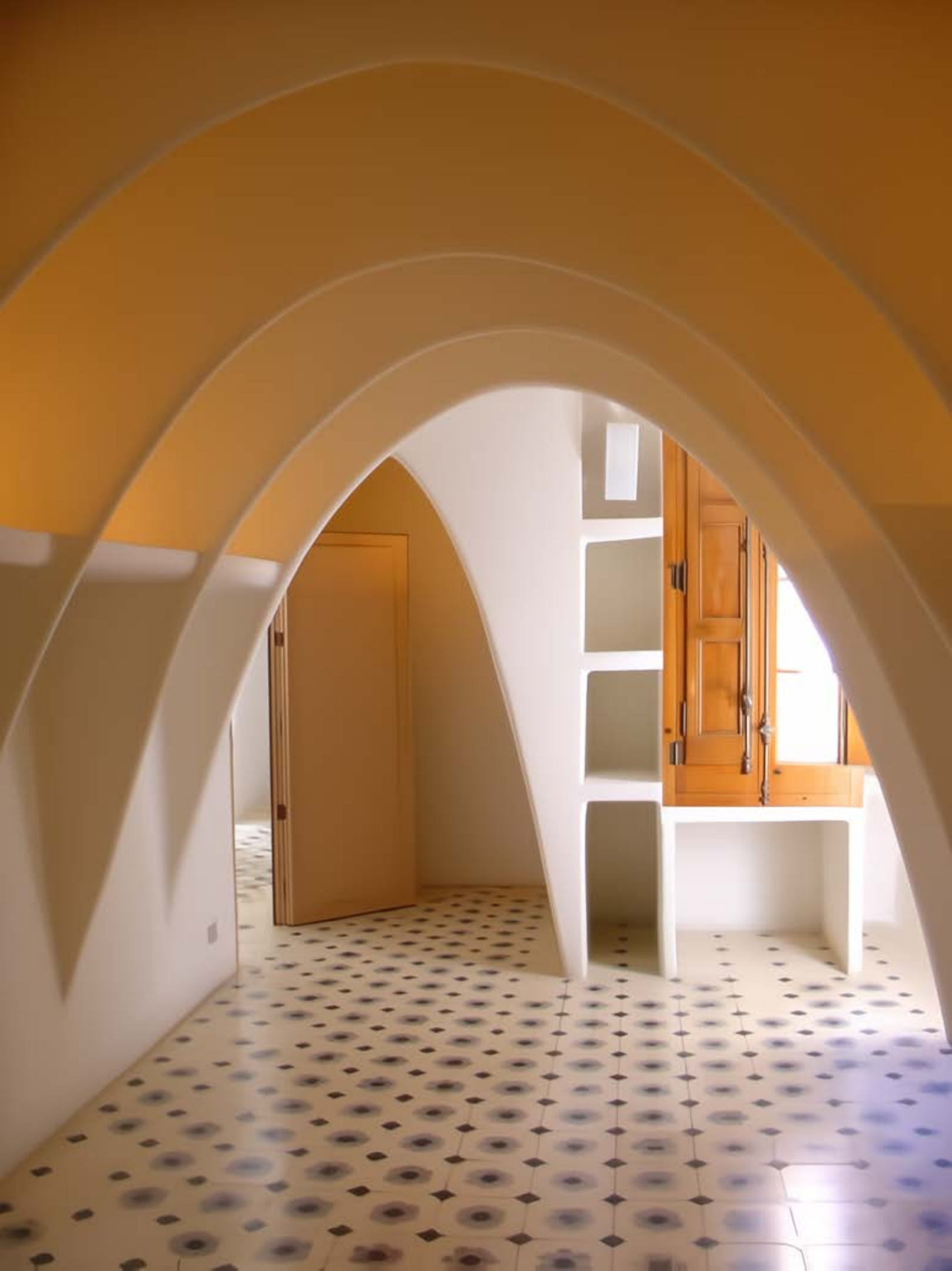} &   
\includegraphics[width=.33\textwidth,valign=t]{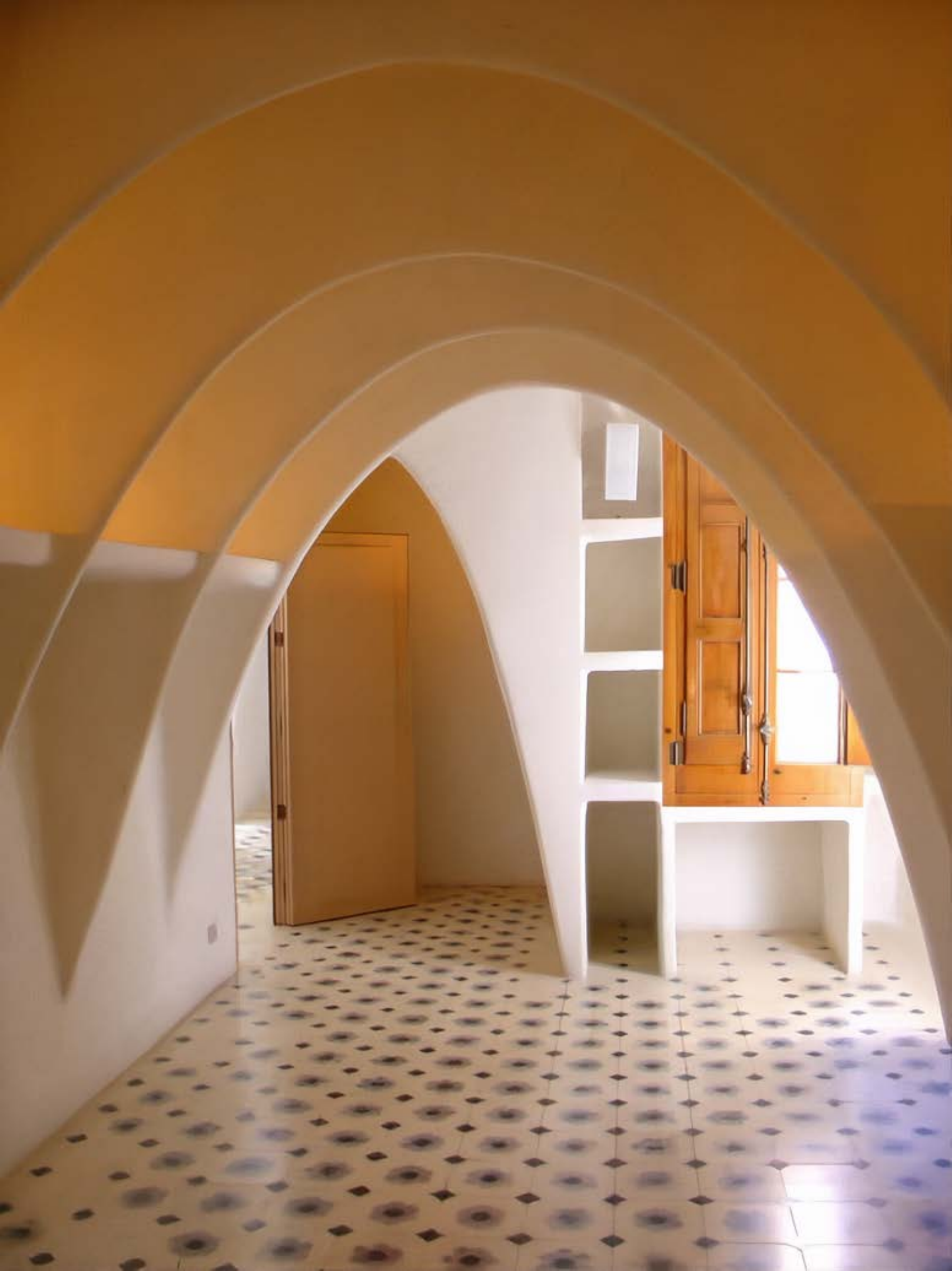} \\
\vspace{0.6em}
\small~SwinIR~\cite{SwinIR}, 35.12 dB  & \small~Restormer~\cite{Restormer}, 35.76 dB & \small~\textbf{Ours}, \textbf{35.81 dB} \\

\end{tabular}
}
\vspace{-5mm}
\caption{\small Gaussian color denoising results on Urban100~\cite{Urban100}.}
\label{fig:suppl_denoising_visual_results}
\end{center}
\vspace{-7mm}
\end{figure*}